\documentclass[onecolumn]{cohere}

\usepackage{pdfpages}
\usepackage[table]{colortbl}
\usepackage{wrapfig}
\setcitestyle{number,square}
\usepackage{xspace} 
\usepackage{lipsum}
\usepackage{blindtext}
\usepackage{array}
\usepackage{tabularx}
\usepackage{longtable}
\usepackage{xltabular}
\usepackage{placeins}
\usepackage{breqn} 
\usepackage{makecell} 
\usepackage{multirow} 

\usepackage[framemethod=TikZ]{mdframed}
\usepackage{tcolorbox}
\usepackage{multicol}
\usepackage{epsdice}
\usepackage{pifont}

\definecolor{ayad}{RGB}{148, 156, 229} 
\definecolor{ayadsymbol}{RGB}{76, 110, 230} 
\definecolor{lightblue}{RGB}{211, 227, 252} 
\definecolor{bgblue}{RGB}{247, 250, 255} 

\definecolor{ayac}{RGB}{255, 175, 71} 
\definecolor{lightyellow}{RGB}{250, 224, 189} 
\definecolor{bgyellow}{RGB}{255,251,246} 

\definecolor{ayae}{RGB}{196,178,188} 
\definecolor{ayaebackground}{RGB}{250, 245, 248}

\definecolor{ayaui}{RGB}{204, 232, 204} 

\newenvironment{card}[3]{
  \mdfsetup{
    frametitle={
      \tikz[baseline=(current bounding box.east),outer sep=0pt]
      \node[anchor=east,rectangle,fill=#2]{#1};
    },
    innertopmargin=7pt,
    innerbottommargin=7pt,
    innerleftmargin=7pt,
    innerrightmargin=7pt,
    linecolor=#2,
    linewidth=0.3pt,
    topline=true,
    backgroundcolor=#3,
    frametitleaboveskip=\dimexpr-\ht\strutbox\relax,
  }
  \begin{mdframed}[]\relax%
}{
  \end{mdframed}
}

\newtcolorbox{mybox}[2][]{
  colback=white, 
  colframe=lightblue,
  fonttitle=\bfseries,
  coltitle=black,  
  title=#2, 
  #1
}
\newtcolorbox{mybox2}[2][]{
  colback=white, 
  colframe=lightyellow,
  fonttitle=\bfseries,
  coltitle=black,  
  title=#2, 
  #1
}

\newtcolorbox{mybox3}[2][]{
  colback=white, 
  colframe=lightgray,
  fonttitle=\bfseries,
  coltitle=black,  
  title=#2, 
  #1
}

\DeclareSymbolFont{extraup}{U}{zavm}{m}{n}
\DeclareMathSymbol{\vardiamond}{\mathalpha}{extraup}{87}
\DeclareMathSymbol{\varspade}{\mathalpha}{extraup}{85}

\newcommand{\dia}{%
  \ooalign{%
    \scalebox{1.09}{\textcolor{ayadsymbol}{$\vardiamond$}}\cr
    \raisebox{.2ex}{\hspace{0.029em}\scalebox{1.1}{\textcolor{black}{$\diamondsuit$}}}\cr
  }%
}

\newcommand{\spade}{%
  \ooalign{%
    \textcolor{ayac}{\raisebox{0.21ex}{\hspace{.085em}\resizebox{0.75em}{0.6em}{$\spadesuit$}}}\cr 
    \raisebox{-0.18ex}{\scalebox{1.1}{\textcolor{black}{$\varspade$}}}\cr
  }%
}

\newcommand{\cmark}{\ding{51}}%
\newcommand{\xmark}{\ding{55}}%

\DeclareRobustCommand*{\specialdia}{\dia}
\DeclareRobustCommand*{\specialspade}{\spade}

\definecolor{Gray}{gray}{0.9}

\newcommand\invisiblesection[1]{%
  \refstepcounter{section}%
  \addcontentsline{toc}{section}{\protect\numberline{\thesection}#1}%
  \sectionmark{#1}}

\title{{\includegraphics[scale=0.2]{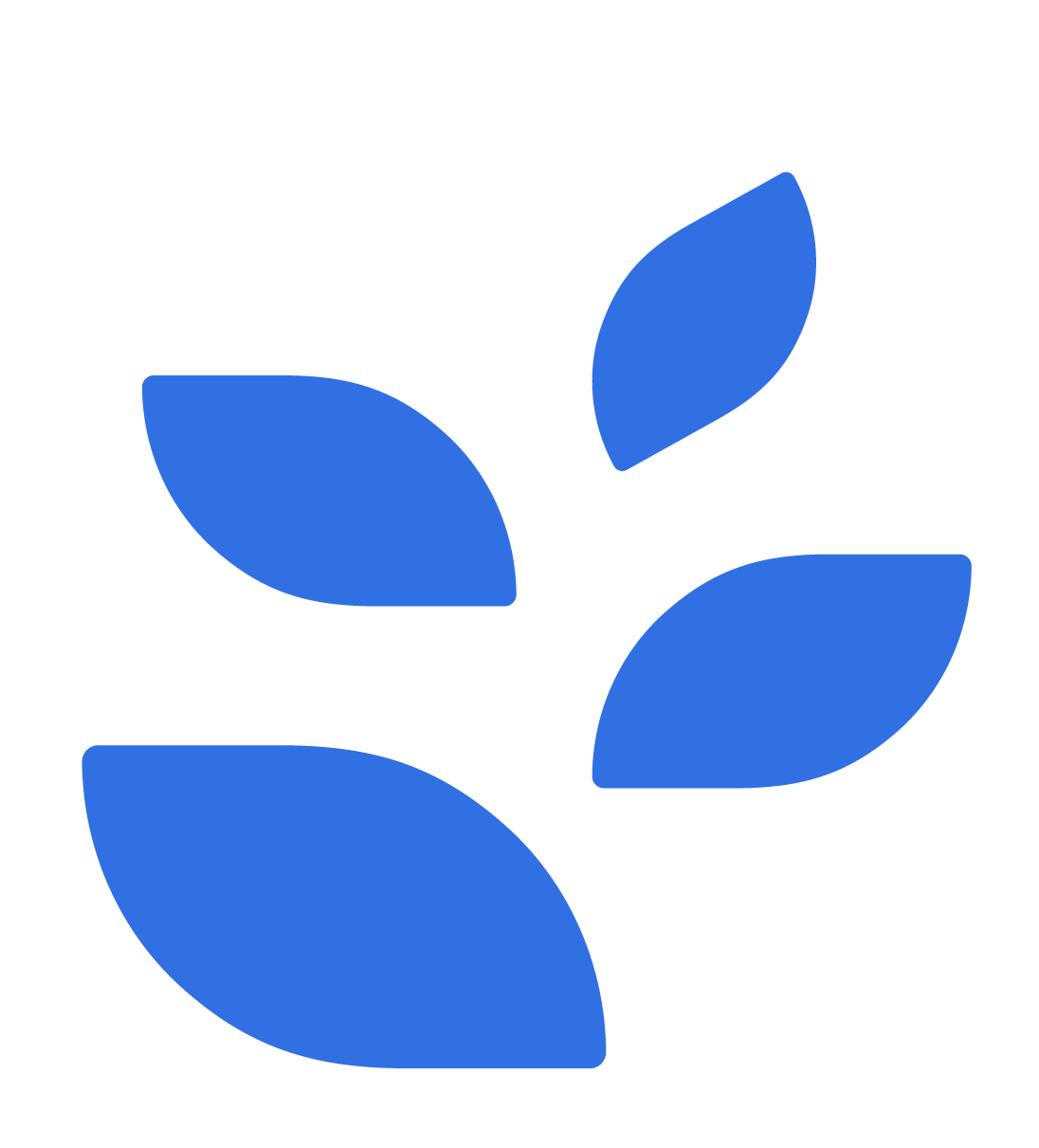}}Aya Dataset: An Open-Access Collection \\for Multilingual Instruction Tuning}

\renewcommand{\fasymbol}{\textcolor{ayadsymbol}{$\vardiamond$}}
\multiauthors
\author{
    name={Shivalika Singh\fa},
    affiliation={1},
}
\author{
    name={Freddie Vargus\fa},
    affiliation={1},
}
\author{
    name={Daniel D'souza\fa},
    affiliation={1},
}
\author{
    name={Börje F. Karlsson\fa},
    affiliation={2},
}
\author{
    name={Abinaya Mahendiran\fa},
    affiliation={1},
}
\author{
    name={Wei-Yin Ko\fa},
    affiliation={3},
}
\author{
    name={Herumb Shandilya\fa},
    affiliation={1},
}
\author{
    name={Jay Patel},
    affiliation={4},
}
\author{
    name={Deividas Mataciunas},
    affiliation={1},
}
\author{
    name={Laura O'Mahony},
    affiliation={5},
}
\author{
    name={Mike Zhang},
    affiliation={6},
}
\author{
    name={Ramith Hettiarachchi},
    affiliation={7},
}
\author{
    name={Joseph Wilson},
    affiliation={8},
}
\author{
    name={Marina Machado},
    affiliation={3},
}
\author{
    name={Luisa Souza Moura},
    affiliation={3},
}
\author{
    name={Dominik Krzemiński},
    affiliation={1},
}
\author{
    name={Hakimeh Fadaei},
    affiliation={1},
}
\author{
    name={Irem Ergün},
    affiliation={3},
}

\author{
    name={Ifeoma Okoh},
    affiliation={1},
}

\author{
    name={Aisha Alaagib},
    affiliation={1},
}
\author{
    name={Oshan Mudannayake},
    affiliation={1},
}
\author{
    name={Zaid Alyafeai},
    affiliation={9},
}
\author{
    name={Vu Minh Chien},
    affiliation={1},
}
\author{
    name={Sebastian Ruder},
    affiliation={3},
}
\author{
    name={Surya Guthikonda},
    affiliation={1},
}
\author{
    name={Emad A. Alghamdi},
    affiliation={10},
}
\author{
    name={Sebastian Gehrmann},
    affiliation={11},
}
\author{
    name={Niklas Muennighoff},
    affiliation={1},
}
\author{
    name={Max Bartolo},
    affiliation={3},
}
\author{
    name={Julia Kreutzer},
    affiliation={12},
}
\author{
    name={Ahmet Üstün},
    affiliation={12},
}
\author{
    name={Marzieh Fadaee},
    affiliation={12},
}
\author{
    name={Sara Hooker},
    affiliation={12},
}
\affiliations{
    \item[1] Cohere For AI Community
    \item[2] Beijing Academy of Artificial Intelligence 
    \item[3] Cohere
    \item[4] Binghamton University
    \item[5] University of Limerick
    \item[6] IT University of Copenhagen
    \item[7] MIT
    \item[8] University of Toronto
    \item[9] King Fahd University of Petroleum and Minerals 
    \item[10] King Abdulaziz University, ASAS.AI
    \item[11] Bloomberg LP
    \item[12] Cohere For AI
}
\corresponding[*]{Shivalika Singh \texttt{<shivalikasingh95@gmail.com>}, Marzieh Fadaee \texttt{<marzieh@cohere.com>}, Sara Hooker \texttt{<sarahooker@cohere.com>}}

\date{\today}
\abstract{Datasets are foundational to many breakthroughs in modern artificial intelligence. Many recent achievements in the space of natural language processing (NLP) can be attributed to the fine-tuning of pre-trained models on a diverse set of tasks that enables a large language model (LLM) to respond to instructions. Instruction fine-tuning (IFT) requires specifically constructed and annotated datasets. However, existing datasets are almost all in the English language. In this work, our primary goal is to bridge the language gap by building a human-curated instruction-following dataset spanning 65 languages. We worked with fluent speakers of languages from around the world to collect natural instances of instructions and completions. Furthermore, we create the most extensive multilingual collection to date, comprising 513 million instances through templating and translating existing datasets across 114 languages.
In total, we contribute four key resources: we develop and open-source the \textbf{\aya Annotation Platform}, the \textbf{\aya Dataset}, the \textbf{\aya Collection}, and the \textbf{\aya Evaluation Suite}. The \aya initiative also serves as a valuable case study in participatory research, involving collaborators from 119 countries. We see this as a valuable framework for future research collaborations that aim to bridge gaps in resources.
}

\newcommand{\aya}{\textbf{{Aya}}\xspace}
\newcommand{\Aya}{\textbf{{Aya}}\xspace}

\begin{document}


\section{Introduction}

Datasets are static representations of the world, far from the rich, ever-evolving environment we navigate as humans. Yet, these frozen snapshots in time are the foundation upon which progress in AI has been built.  Many recent breakthroughs in language modeling can be attributed to fine-tuning pre-trained models on a diverse set of tasks that enable a Large Language Model (LLM) to follow instructions~\citep{mccann2018natural,sanh2022multitask, wei2022finetuned, muennighoff-etal-2023-crosslingual,longpre2023flan}. Instruction fine-tuning (IFT) leverages the precept that Natural Language Processing (NLP) tasks can be described via natural language instructions, such as \textit{``What were the reviews like for the Barbie movie?"} 
or \textit{``Write a recipe from the following list of ingredients."} This process requires \emph{prompts} to be paired with expected \emph{completions}~\citep{ziegler2020finetuning,ouyang2022training} aiming to capture the variety of ways an LLM can be used in downstream tasks. Yet, the very act of curating data imparts a viewpoint about what distributions we want our model to represent and what is forgotten. So, \textit{what do these widely used datasets tell us about the assumptions underlying these breakthroughs?} 

\begin{figure}[tb!]
    \centering
\includegraphics[width=\linewidth]{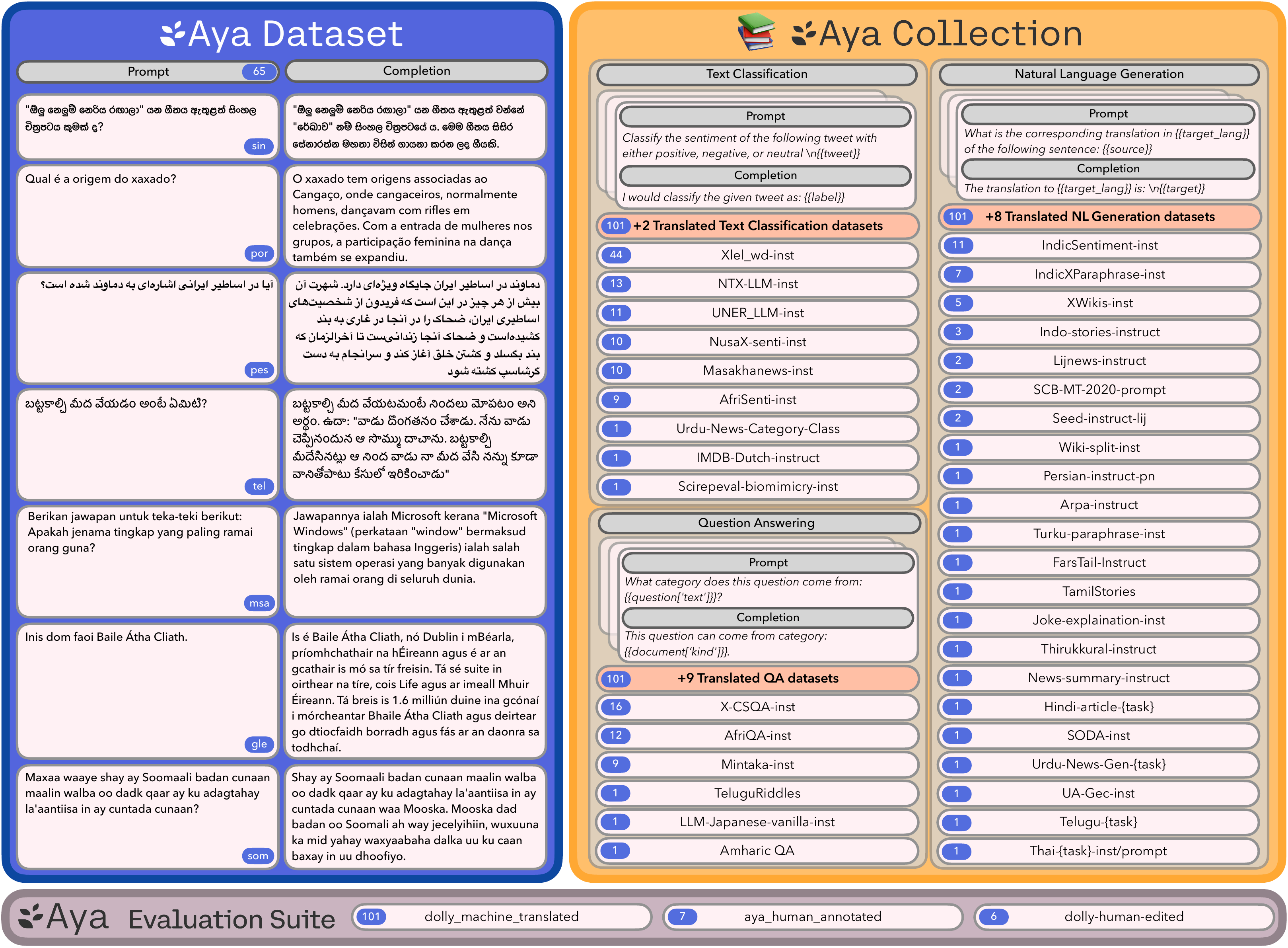}
    \caption{\textbf{Aya Dataset,  Aya Collection \& Aya Evaluation Suite.} On the left, we show examples of contributions in the \Aya Dataset. These are original human-curated prompt-completion pairs written by fluent speakers of 65 languages.
    On the right, we have the \aya Collection, an aggregation of 44 monolingual and multilingual templated instruction datasets and 19 translated datasets ranging over 114 languages and three main tasks: Text Classification, Natural Language Generation, and Question Answering. 
    The bottom block showcases the \aya Evaluation Suite for multilingual open-ended generation. This collection consists of original annotation and post-edits of translations covering several languages, and translation of high-quality and universal prompts into 101 languages.
    We indicate the number of languages in a dataset with the value in the blue ovals in the figure. 
    (Translated datasets have been visually merged due to space constraints).
    }\label{fig:aya_dataset}
\end{figure}

More than 7,000 languages\footnote{\url{https://www.ethnologue.com/}} are spoken around the world today, with a considerable number facing the challenges of being low-resourced, under-represented, or disappearing \citep{maxwell2006frontiers,Simons_2019,moran20204,secretariat2022international, 10.1371/journal.pone.0287850, ilhomovna2023you, marivate-etal-2020-investigating}. In contrast, the most widely used datasets and breakthroughs in NLP have coalesced around a few data-rich languages \citep{longpre2023data,alpaca,chung2022scaling,fan2021beyond,dodge2021,lucy2024aboutme}. IFT datasets are no exception; the creation of these datasets has almost entirely focused on English. Furthermore, the vast majority of the creators of these works originate from a few countries \citep{longpre2023data,zhang2022ai}.

The factors underlying the construction of the datasets impact how models perform for users around the world. Models perform better on the distribution they are trained to mimic \citep{kunchukuttan-etal-2021-large}. This often introduces known biases towards languages~\citep{schwartz2022towards, Kotek2023GenderBA, Khandelwal2023CasteistBN, vashishtha2023evaluating,khondaker2023gptaraeval} and dialects~\citep{jorgensen-etal-2015-challenges,blodgett-etal-2016-demographic,zampieri-etal-2017-findings,sun-etal-2023-dialect} not included during training and introduces critical security flaws \citep{yong2023lowresource, nasr2023scalable, Li2023PrivacyIL, Lukas2023AnalyzingLO, Deng2023MultilingualJC}. 

Datasets aren't simply raw materials that fuel breakthroughs but also make the poor \textit{poorer} and the rich \textit{richer} \citep{held2023material, durmus2023towards,Robinson2023ChatGPTMC}. Disparities in the access to technological resources predates the advent of LLMs \citep{garrette2013real}. However, as LLMs become more sophisticated and widely available, non-English languages will remain under-represented and will likely become more so. The imbalance between languages has created a growing divide in the cost of using this technology as marginalized languages require more tokens and incur higher latency for generations \citep{ji2023better,cui2023efficient}, consigning speakers of low-performing languages to lower quality technology \citep{held2023material, durmus2023towards,nicholas2023lost,ojo2023good}. Often, speakers of low-resource languages do not have the resources to improve NLP technology for their language, facing a \textit{low-resource double bind} with limited access to both compute and data~\citep{ahia-etal-2021-low-resource}.

\begin{table}[htbp]
\scriptsize
\begin{tabularx}{\textwidth}{>{\hsize=.3\hsize}X>{\centering\arraybackslash\hsize=.08\hsize}X>{\centering\arraybackslash\hsize=.06\hsize}X>{\centering\arraybackslash\hsize=.1\hsize}X>{\hsize=.4\hsize}X>{\centering\arraybackslash\hsize=.1\hsize}X}
       \toprule
       Dataset & \#Instances & \#Langs & \% English & Generation method & Permissive license  \\
       \midrule
       Llama2 IFT data \citep{touvron2023llama} & NA & 27 & 90\% & Human-annotations SFT datasets & \xmark \\
        Alpaca \citep{alpaca} & 52K & 1 & 100\% & Synthetic data generation  IFT datasets & $\approx$ \\
       P3 \citep{sanh2022multitask} & 12M & 1 & 100\% & Template generation given applied to English datasets & \cmark\\
       Flan 2022 \citep{longpre2023flan} & 15M & 60 & 100\% & Template generation applied to English datasets & \cmark \\
       xP3 \citep{muennighoff-etal-2023-crosslingual} & 81M & 46 & 39\% & Template generation applied to English datasets  & \cmark\\
       Sweinstruct \citep{holmstrom-doostmohammadi-2023-making}& 68K & 1 & 0\% & Machine translation English IFT datasets & $\approx$ \\
       Okapi \citep{dac2023okapi} & 158K & 26 & 45\% & Machine translation English IFT datasets & \cmark \\
       Bactrian-X \citep{li2023bactrian} & 3.4M & 52 & 2\% & Machine translation + synthetic data generation & $\approx$ \\
       \midrule
       \rowcolor{ayad} \aya Dataset & 204K & 65 & 2\% & Original IFT Human-annotations & \cmark \\
       \rowcolor{ayac} \aya Collection &  513M & 114 & 3.5\% & Template Generation and translating existing datasets  &  \cmark \\
       \bottomrule
   \end{tabularx}%
   \caption{Comparison of different instruction-tuning datasets. \cmark 
  represents permissive licenses that allow commercial use while $\approx$ represents restrictive licenses that do not allow commercial use. \xmark  represents non availability of license.}
   \label{tab:dataset_comparison}
\end{table}

\textbf{In this work, our goal is to reduce this linguistic inequality.} Efforts that aim to improve multilingual performance have often focused on improving data coverage~\citep{chen2023monolingual}. However, most of the limited effort to date has focused on multilingual pre-training~\citep{workshop2023bloom, wei2023polylm, lample2019cross} with even less work centered on imparting instruction following abilities. Approaches that have tried to translate English instruct-style datasets into other languages often suffer from translation biases~\citep{vanmassenhove-etal-2021-machine, hartung2023measuring, savoldi2021gender,muennighoff-etal-2023-crosslingual} or fail to reflect cultural context appropriately~\citep{wang-etal-2022-measuring, ji_ji_bouillon_seligman_2023, pudjiati2022post}. 
Automatic curation of multilingual datasets is a logical ---and sometimes necessary--- approach but often suffers from
noise and biases. This makes it difficult to validate the quality of the created datasets~\citep{kreutzer-etal-2022-quality, luccioni-viviano-2021-whats, ferrara2023should, caswell-etal-2020-language} or requires the curation of manual templates which often result in low instruction and completion diversity~\citep{muennighoff-etal-2023-crosslingual} critical for model performance~\citep{naik2023diversity, Chung_2023, li2023making, lahoti2023improving}. 

\textbf{In contrast, a key aspect of our work focused on harder-to-obtain human-curated data from fluent speakers of a language.} This curation process has received far less attention due to lack of access to fluent speakers, especially in low-resource languages~\citep{joshi-etal-2019-unsung}. We chose to close this gap by conducting a year-long participatory research initiative that involved working with fluent speakers of languages from around the world to collect human-curated instances of instructions and completions. 
By leveraging best practices from open-source and crowd-sourced science projects \citep{FRANZONI20141,BeckS2022,Lenart-Gansiniec2023}, we built a simple and intuitive user interface, the \aya Annotation Platform\footnote{This platform is accessible at: \url{https://aya.for.ai}} (\aya UI) which served as the central platform for contributors to join the \aya \footnote{The word \aya has its origins in the \texttt{Akan (Twi)} language and is translated as ``fern'' in English~\citep{willis1998adinkra}.} \footnote{\aya represents endurance, resourcefulness, and defiance -- like a fern growing in barren conditions.} project. In total, we had 2,997 collaborators spread across 119 countries around the world. Their collective efforts resulted in the \aya dataset which is the largest human-curated multilingual instruction-finetuned dataset to date, containing 204,114 high-quality annotations in 65 languages.

Additionally, we release and transform 44 pre-existing datasets into sets of instruction-completion pairs by crafting diverse templates manually, relying on fluent speakers for each language. We further expand this collection by translating datasets from English into 101 languages. We refer to this expanded collection of 513 million instances covering 114 languages in total as the \aya collection, which to date, is the most extensive collection of multilingual instruction-finetuning (IFT) data.

Overall, \aya contributes four key resources: \textbf{\aya Annotation Platform} (\aya UI); \textbf{\aya Dataset}; \textbf{\aya Collection}, and \textbf{\aya Evaluation Suite}. 
Figure~\ref{fig:aya_dataset} shows a visual representation of the \aya Dataset and Collection. Below, we briefly describe these core contributions:

\begin{enumerate}
    \item \colorbox{ayaui}{\textbf{\aya Annotation Platform}}(\aya UI): We built a robust annotation tool to facilitate the collection of high-quality multilingual data in an instruction-style format supporting 182
    languages, including dialects. Over eight months, we had a total of 2,997 
    registered users spanning 119 countries and 134 languages, including dialects. 
    \item \colorbox{ayad}{\textbf{\aya Dataset}}: We created the largest human-annotated multilingual instruction finetuning dataset to date, consisting of over 204K 
    instances that cover 65 languages. We include a data card \citep{Pushkarna2022} for the \aya Dataset in Appendix \ref{apx:aya_dataset_datacard}.
    \item \colorbox{ayac}{\textbf{\aya Collection}}: We collected instruction-style templates from fluent speakers and applied them to a curated list of 44 datasets, including tasks such as Text Classification, Text Generation, Machine Translation, Paraphrasing, and Open-domain Question Answering. 
    Some of these datasets also include equivalent multilingual versions produced through translation. We release 513M 
    instances that cover 114 languages.
    These contributions are made available as an open-source collection. We include a data card for the \aya Collection in Appendix \ref{apx:aya_collection_datacard}.
    \item \textbf{\colorbox{ayae}{\aya Evaluation Suite}}: 
    We curate and release a diverse evaluation suite for multilingual open-ended generation quality. It consists of 250 human-written prompts for each of 7 languages, 200 automatically translated but human-selected prompts for 101 languages (114 dialects), and human-edited prompts of the latter for 6 languages, and the English originals. The first set represents culturally-grounded and original prompts, while the translated and post-edited prompts are sourced from English Dolly~\citep{DatabricksBlog2023DollyV2} and selected for their cross-cultural relevance. We include a data card for the \aya Collection in Appendix \ref{apx:aya_evaluation_datacard}. 
\end{enumerate}

By fully open sourcing the \Aya Dataset, \Aya Collection and \aya Evaluation Suite with a permissive Apache 2.0 License\footnote{\url{https://www.apache.org/licenses/LICENSE-2.0}} as well as the code for our annotation platform, we hope to empower researchers and practitioners to further advance multilingual models and applications. All datasets are accessible for download.\footnote{\url{https://hf.co/datasets/CohereForAI/aya_dataset}}\footnote{\url{https://hf.co/datasets/CohereForAI/aya_collection}}\footnote{\url{https://hf.co/datasets/CohereForAI/aya_evaluation_suite}}

\textbf{Paper Organization $\:$} 
Section~\ref{sec:aya_annotation_platform_and_aya_dataset} discusses the design and development of the \aya Annotation Platform, as well as the preparation of the \aya Dataset, and Section~\ref{sec:analysis_of_aya_dataset} presents a detailed analysis of the \aya Dataset. Section~\ref{sec:aya_collection} and Section~\ref{sec:analysis_of_aya_collection} contain discussion and analysis of the \aya Collection. 
Section~\ref{sec:evaluationsuite} describes the details of the evaluation suite curated in this project.
In Section~\ref{sec:particpatory_approch_to_research}, we describe our approach to participatory research.  In Section~\ref{sec:related_work}, we review the existing literature, and in Section~\ref{sec:limitations} we discuss the limitations of our work. Section~\ref{sec:conclusion} concludes the paper.

\section[Aya Annotation Platform \& Aya Dataset]{\colorbox{ayaui}{\aya Annotation Platform} \& \colorbox{ayad}{\aya Dataset}}
\label{sec:aya_annotation_platform_and_aya_dataset}

\subsection[Aya Annotation Platform]{\aya Annotation Platform}

The goal of the \aya project is to facilitate annotations to a crowd-sourced dataset by individuals fluent in different languages.
Inputs from fluent speakers of each language ensure that the dataset is more likely to be organic, fluent, and representative of the speakers' cultures.
Including fluent and native speakers from various regions poses significant logistical challenges involving meticulous data selection, quality control measures, and custom annotation tools. We developed the \aya Annotation Platform to streamline the data collection process worldwide, accommodating a large number of decentralized contributors across multiple languages.

User Interfaces (UIs) play a pivotal role in the context of NLP data collection, serving as the primary point of interaction between human annotators and the data collection process. The \aya Annotation Platform\footnote{\url{https://aya.for.ai/}} had to accommodate users in 119 countries collecting data across 134 languages.
We designed the platform with a few key principles in mind, such as accessibility and ease of use for users who were unfamiliar with AI and machine learning. 
As part of our contribution, we fully open-source the code for our UI\footnote{\url{https://github.com/for-ai/aya-annotations-ui}}.

\textbf{Accessibility $\:$} As users worldwide use different devices and operating systems, we decided to support both mobile and desktop interfaces~\citep{muhammad-etal-2023-afrisenti}. 
Approximately 54\% of users accessed \aya UI via desktop browsers while 46\% utilized mobile browsers. We attribute the high fraction of mobile users to the skew towards mobile users in the Global South \citep{Avle2018ResearchOM}.  We supported Single Sign-On (SSO) capabilities to enable seamless tracking of user profiles and reward users with points for contributing data across multiple sessions. We initially only supported Discord sign-on but discovered that Discord is inaccessible or not widely used in certain countries.
Also, the necessity of a platform-specific account created an obstacle to user engagement with \aya. 
This prompted us to add Google sign-on as an alternative option.

\begin{figure}[tb]
  \centering
  \includegraphics[width=0.85\textwidth]{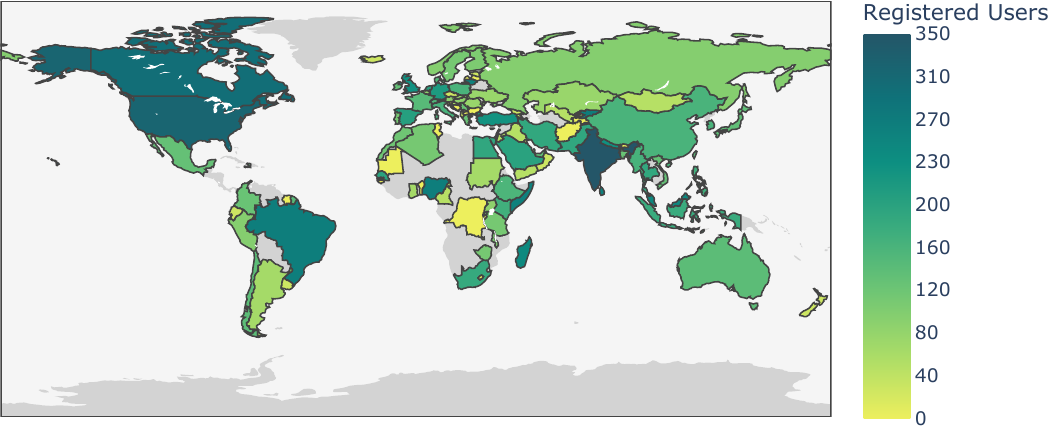}
  \caption{Geographical distribution of the users registered on the \aya platform.}\label{fig:registered_users_by_region}
\end{figure}

\textbf{Languages Supported $\:$} \aya project contributors could select the languages they are proficient in when signing up using the \aya UI. They could then make annotations in the language(s) they selected. Given the sheer number of languages we could collect annotations for, we chose to prioritize annotation support for the 101 languages available in the mT5 model \citep{mt5-2020}. 
We note that ultimately, some of these languages didn't receive enough contributions to include them in the final dataset. Conversely, we received substantial contributions from languages not initially part of the original list, like \texttt{Wolof}, leading to their inclusion; the final \aya Dataset covers 65 languages. Table~\ref{tab:language_codes} provides details of these languages.

\textbf{Contributors $\:$} We aimed to include individuals from diverse backgrounds---not limited to AI experts---enabling anyone proficient in a language to contribute. 
Our pool of contributors ultimately reflects this inclusive approach.
During the registration process, we request specific demographic details from each \aya UI user such as country of residence, languages of fluent communication, gender, age range, and familiar dialects. 
We display the onboarding form in Figure~\ref{fig:onboarding} in the Appendix.
The \aya community of contributors includes 2,997 registered users across 134 languages.

\begin{figure}[htb!]
     \centering
  \includegraphics[width=0.90\textwidth]{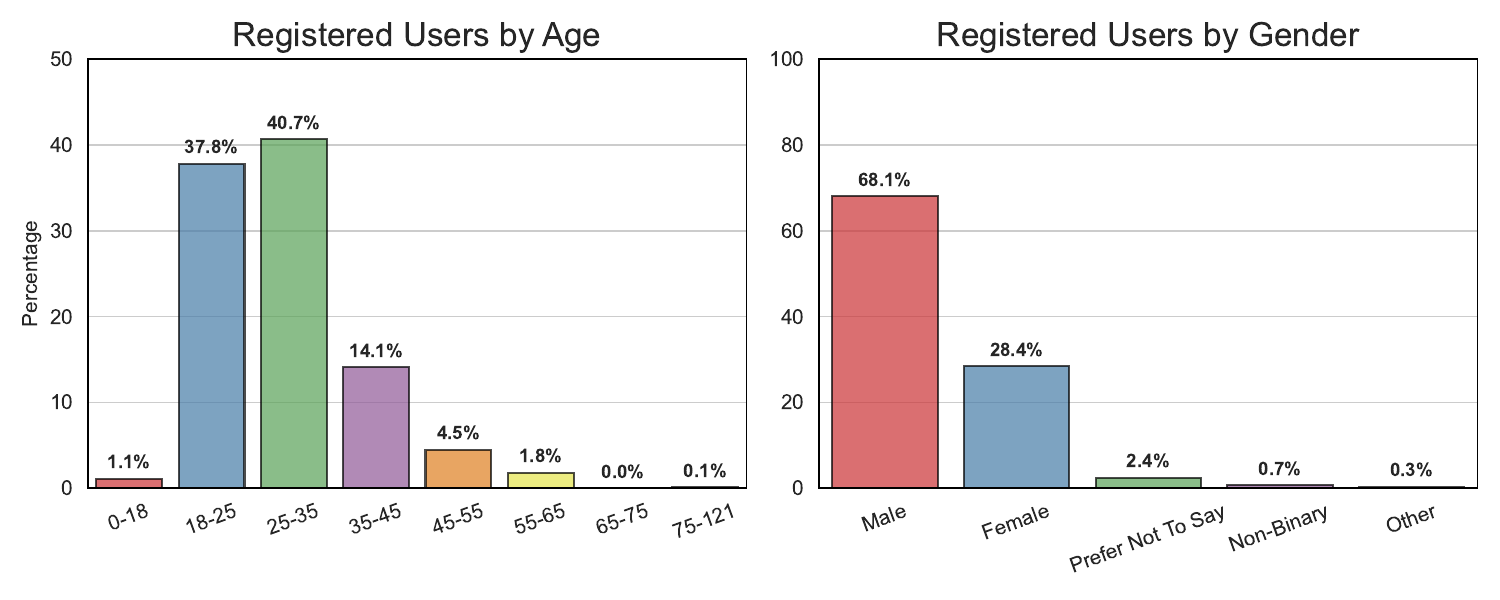}
  \caption{\textbf{Left:} Distribution of registered users on the \aya UI by age using specified values. \textbf{Right:} Distribution of registered users on the \aya UI by gender using specified values 
  }
\label{fig:registered_users_by_gender_age}
\end{figure}

\textbf{Demographics} Figure~\ref{fig:registered_users_by_gender_age} illustrates the demographics of registered \aya UI users by age and gender. Regarding the age profiles of users, more than two-thirds were aged between 18 and 35. Approximately 68.1\% of users identified themselves as male and 28.5\% as female. Overall, 6.6\% of users self-reported dialects. Within this group, 75\% specified one dialect, 20\% specific two dialects, and the remaining 5\% specified three or more dialects, with a maximum of six.

\begin{wrapfigure}{r}{0.45\textwidth}
  \centering
  \includegraphics[scale=0.65]{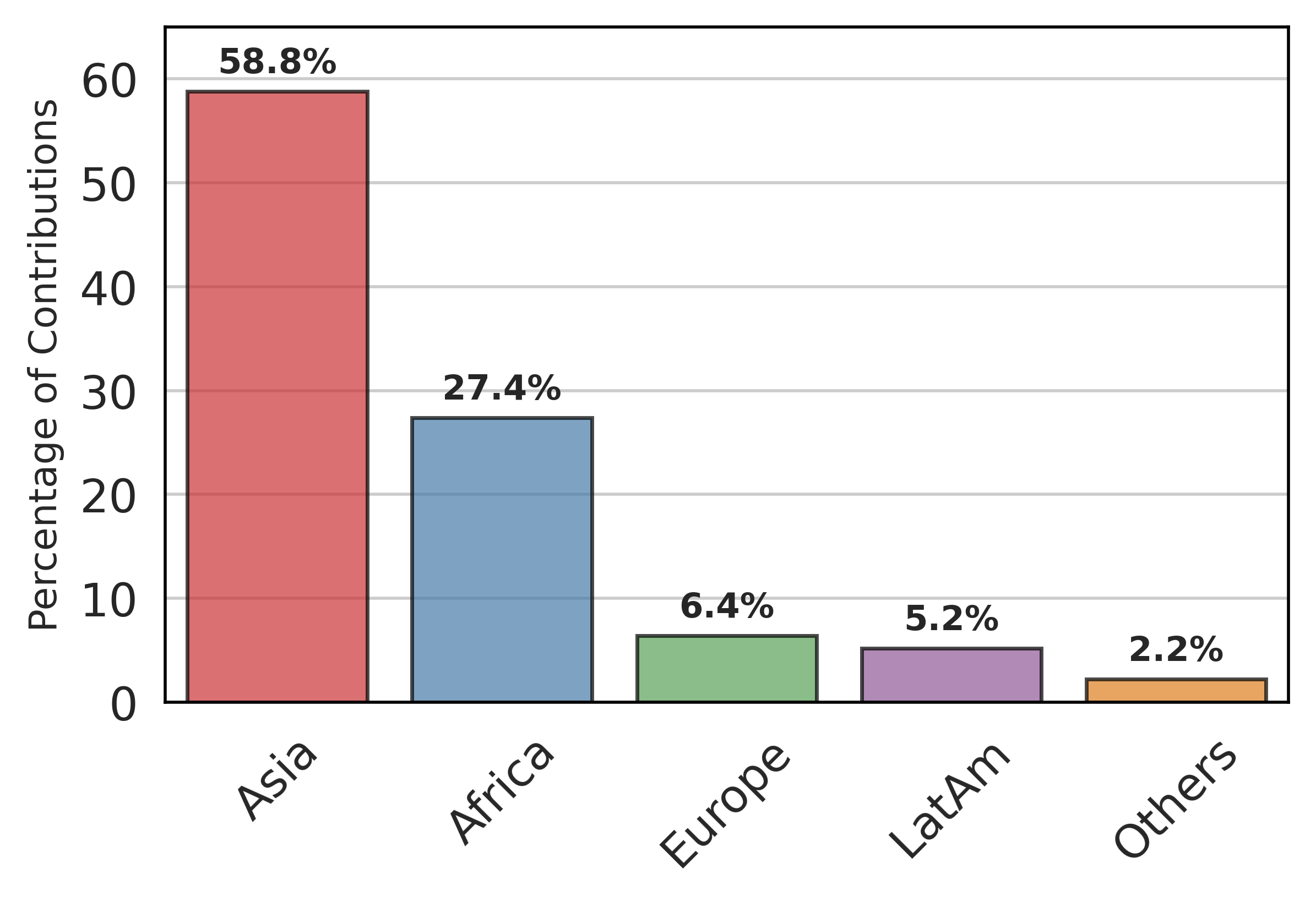}
  \caption{Distribution of total contributions across different regions.}
  \label{fig:contributions_by_region}
\end{wrapfigure}

During the development of \aya, registered users were geographically distributed across \textbf{119 countries} based on their residence. Certain countries like Afghanistan, Bulgaria, Kuwait, and Tajikistan had just one registered user. Figure~\ref{fig:registered_users_by_region} displays this global distribution, highlighting India with the highest number of registered users (346 out of 2,997).

\textbf{Geographic-Based Contribution Assessment $\:$}
We grouped the languages by the regions in which they either originate or are widely spoken. The language statistics by region for the original 101 languages we wanted to cover are as follows: 14 languages in Africa, 41 languages in Asia, 42 languages in Europe, and 4 languages in Latin America (See Appendix~\ref{apx:division_by_region} for more details and some exceptions of the distribution). As seen in Figure~\ref{fig:contributions_by_region},
more than half of all contributions for the \aya project came from Asia with 58.8\%, followed by the African region with 27.4\%. Europe, Latin America, and other regions account for the remaining 13.8\% of the contributions. 

We observe a large skew in terms of regional contributions, which deserves further research to understand why certain networks of contributors remained motivated for the entire project. These disparities in participation may be due to opportunity cost in time \citep{gerosa2021shifting,wu2007empirical}, cultural beliefs around sharing data \citep{huang2023acegpt}, or the belief that the language in question is not well served by the current technology \citep{nicholas2023lost}. 

\textbf{Acknowledgement of contributions $\:$} Recognition and transparency were maintained throughout the project through the use of a leaderboard\footnote{The \aya Leaderboard is accessible at: \url{https://aya.for.ai/leaderboard/}} to acknowledge contributions.
We implemented a scoring system where contributors earned a maximum of three points for each re-annotation, with one point awarded for rating the prompt and completion, one point for editing the prompt, and one point for editing the completion. Each original annotation was awarded with two points. We describe the different annotation tasks in the \aya UI in detail in Section~\ref{sec:annotation-tasks}.

The \aya Leaderboard is organized to display daily, weekly, and cumulative scores, providing a comprehensive overview of user contributions. The users have the flexibility to filter scores based on specific languages, allowing for a sense of community amongst contributors of a particular language. This design aimed to boost contributors' motivation to provide high-quality inputs for their chosen languages. Figure~\ref{fig:leaderboards} shows an example of the leaderboard.
We discuss further details on collaborating with the community in Section~\ref{sec:particpatory_approch_to_research}.

\subsection{Annotation tasks}\label{sec:annotation-tasks}

On the \aya Annotation Platform, contributors were able to contribute to three different tasks, following the find-fix-verify paradigm~\citep{DBLP:journals/cacm/BernsteinLMHAKC15}: Writing new examples from scratch (\textbf{original annotations}), editing existing examples to improve the quality and comprehensiveness (\textbf{re-annotations}), and giving feedback on the quality of existing contributions (\textbf{annotation feedback}). We describe each briefly below:

\subsubsection{Original Annotations}

This task facilitates the inclusion of human-generated organic content by allowing annotators to submit original prompt-completion pairs in their language. Existing multilingual models have been shown to produce generations influenced by Western culture~\citep{DBLP:journals/corr/abs-2111-06467,Naous2023HavingBA,lee-etal-2023-hate} reflecting the underlying representation bias \citep{mehrabi2021survey} of their training datasets. This task aims to encourage annotators to submit fresh samples that are representative of their language, culture, literature, history, and region.
The guidelines for contributors is available in Appendix~\ref{sec:annotation-guidelines}.

\begin{figure*}[ht]
    \centering
    \begin{subfigure}[b]{0.48\textwidth}
        \centering
        \includegraphics[width=0.90\linewidth]{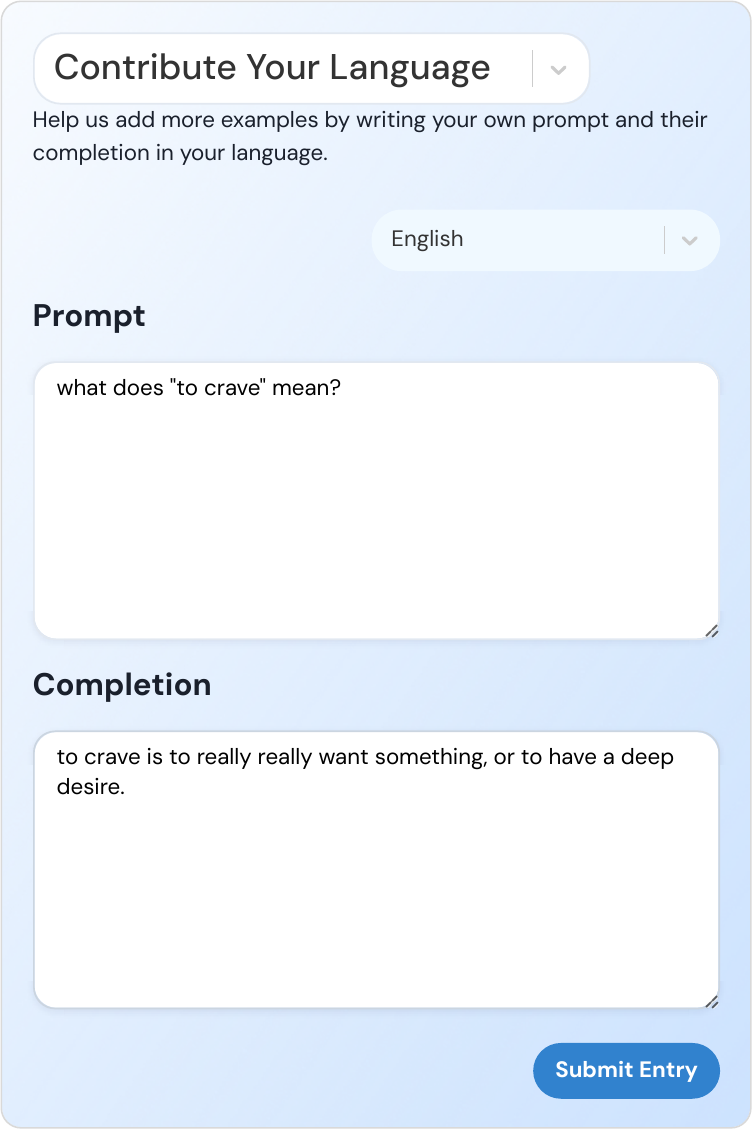}
        \caption{Example of an original annotation contribution. }
        \label{fig:task_examples_Task2}
    \end{subfigure}%
    \begin{subfigure}[b]{0.51\linewidth}
        \centering
        \includegraphics[width=\linewidth]{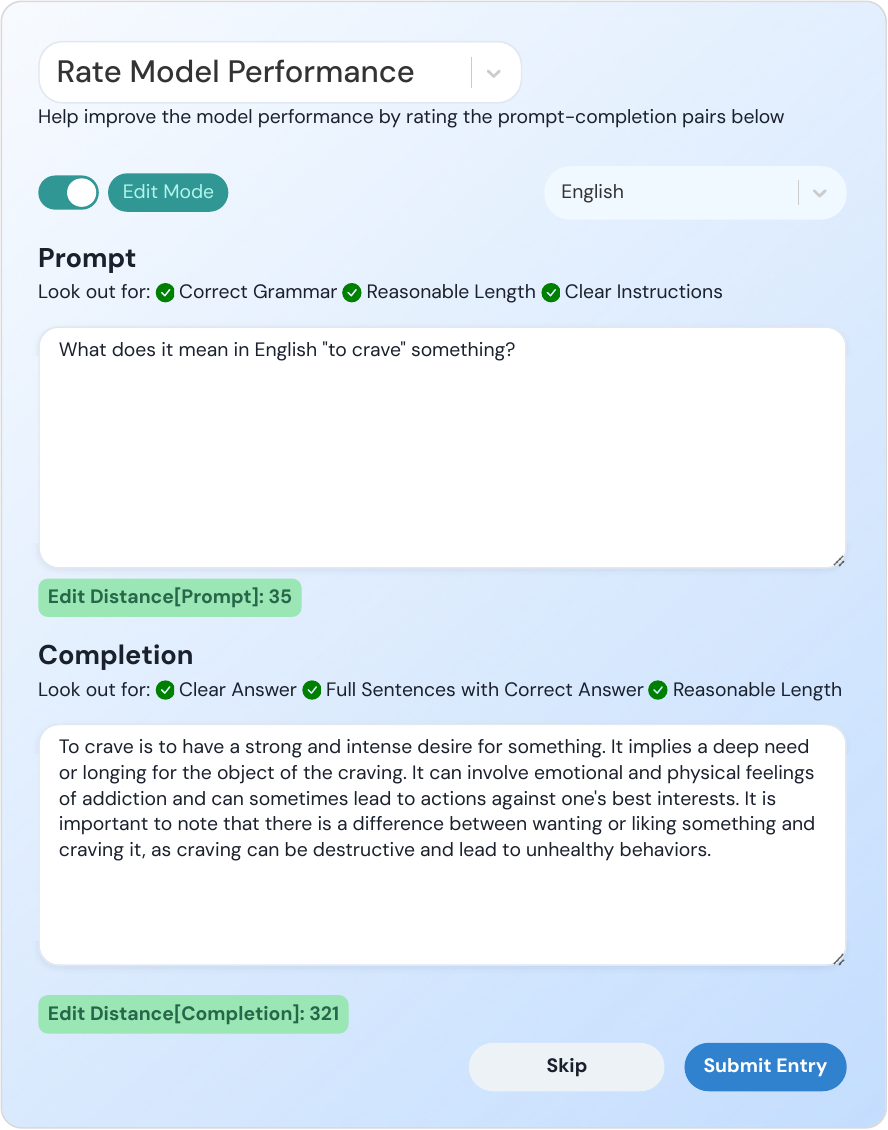}
        \caption{Example of a re-annotation contribution.}
        \label{fig:task_examples_Task1}
    \end{subfigure}
    \vspace{0.1cm}
    \caption{Demonstration of a sample original annotation and re-annotation contribution, in English. \textbf{(a)} exemplifies an original contribution input by an annotator from scratch. \textbf{(b)} shows a sample of re-annotation. Here, the annotator may have improved upon either a prompt and completion pair pulled from the database or a prompt and completion originally created by another contributor.}
    \label{fig:task_examples}
\end{figure*}

\subsubsection{Re-Annotations}
The purpose of this task is to facilitate the re-annotation or editing of prompt and completion pairs. The decision to add a re-annotation task partly stems from the need to help annotators understand the expected format of instruction-style datasets and to convey the variety of tasks in existing datasets, including question answering \citep{saad2023pdftriage, arefeen2023leancontext}, summarization \citep{stiennon2020learning, wu2021recursively}, paraphrasing \citep{witteveen2019paraphrasing, reimers2019sentence}, and translation \citep{nllbteam2022language, barrault2023seamlessm4t}. Editing examples from existing datasets not only helped familiarize annotators with the expected format but also allowed for human evaluation and rating of existing widely used instruction-style datasets.

In total, we collected datasets from 19 public data sources and translated them into 114 available languages, including dialects using the NLLB 3.3B parameter machine translation model~\citep{nllbteam2022language}. From each collection, we randomly chose 100 examples (per dataset and per language), creating our dataset for annotation, after which we had 1M translated prompt-completion pairs initially populated in the \aya UI as re-annotation tasks. These translated pairs served as a starting point for prompts and completions which annotators could improve. We release the raw translations as part of the \aya Collection, provide more details about the provenance of the translated datasets, and how they were selected in Section~\ref{sec:translation}.

In addition to translated examples, there are other available data sources suitable for re-annotation: original \aya pairs, pre-existing instruction-style datasets (e.g., xP3), and the transformation of datasets into an instruction-style format, i.e., templated datasets.
By re-annotating examples from different sources, we simultaneously enhance the quality of individual examples while obtaining a signal on the overall quality of the dataset in a specific language.
A demonstration of a re-annotation, where an annotator strengthens a given prompt/completion, is shown in Figure~\ref{fig:task_examples_Task1}.

\subsubsection{Annotation Feedback}

\begin{wrapfigure}{l}{0.55\textwidth}
  \centering
  \includegraphics[scale=0.30]{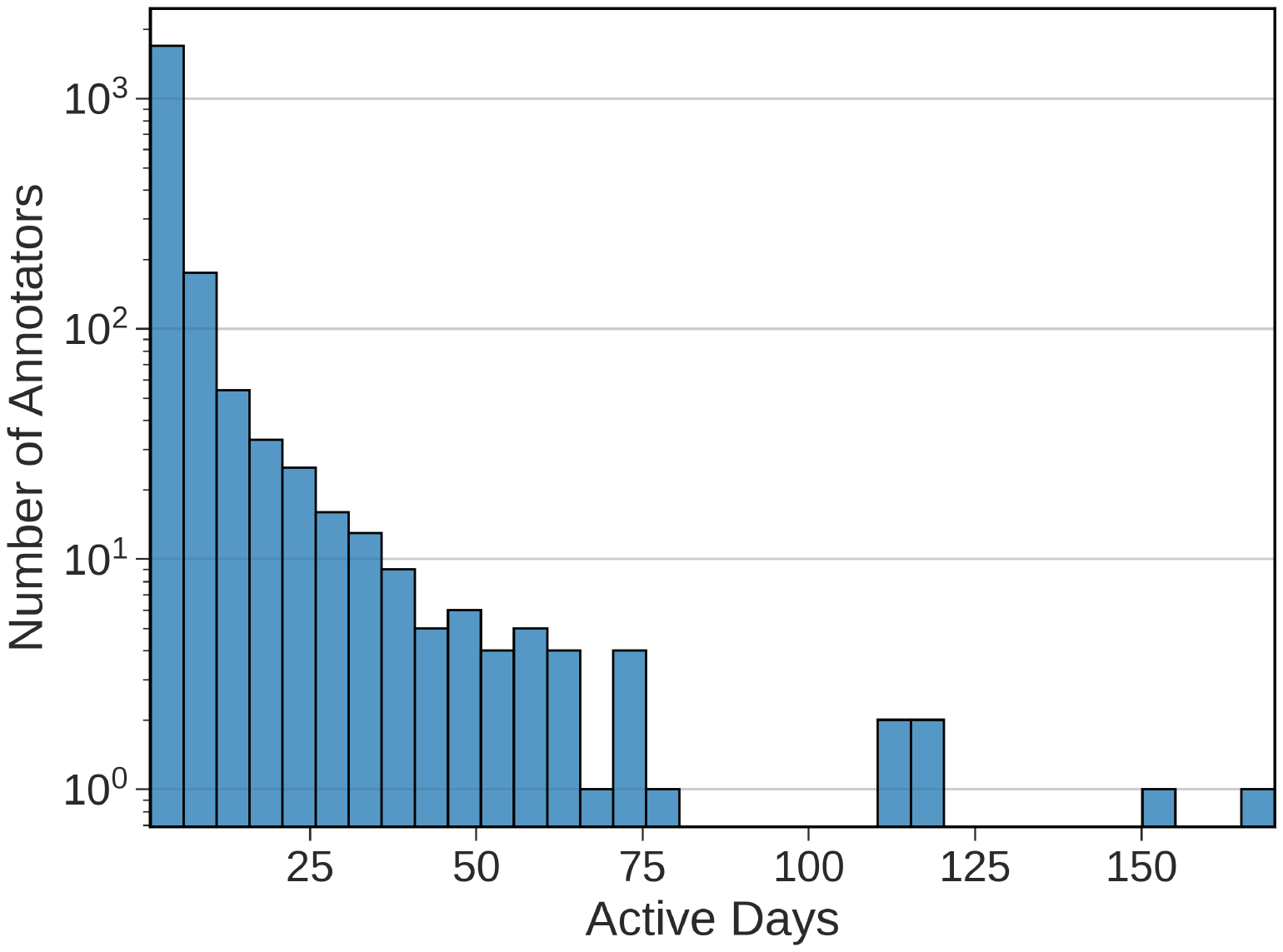}
  \caption{The distribution of annotators' engagement based on the number of days they actively contributed in \aya UI}
  \label{fig:user_activity}
\end{wrapfigure}

Data quality is critical to ensure that a model can represent a language well. Learning from noisy, low-quality datasets harms the overall model performance and the relatively high cost of encoding these noisy examples is a misuse of capacity \citep{hsueh2009data,dodge2021, luccioni-viviano-2021-whats,kreutzer-etal-2022-quality}. Prior work has shown that improvements to quality through data pruning or selection can have an significant impact on the downstream performance of a model \citep{longpre2023pretrainer,marion2023investigating,boubdir2023prompts,yang2023dataset}. In particular, for instruction-tuning datasets, a small subset of higher-quality instructions can greatly outperform a larger volume of lower-quality instructions \citep{alshikh2023becoming,zhou2023lima,chen2023alpagasus}. Given these findings, ensuring high quality contributions is of paramount importance. Ensuring consistent quality is particularly challenging in an open science initiative with a large number of contributors. We face two key challenges:

\textbf{Changes in the Annotator Pool.} During the year-long project, annotators joined and left the project at different points depending on their interests and availability. 
As a result, the window of contribution for each annotator was different. 
Only a small fraction of annotators participated for the entire duration of the year-long project. 
Annotators were active for an average of 1.3 sessions. 
Figure~\ref{fig:user_activity} presents a histogram depicting the distribution of user engagement based on the number of days they actively contributed. 
On average, \aya annotators spent five days contributing to the project. 
Annotators tended to be highly active shortly after joining, but their activity declined over time. There was a subgroup of annotators who maintained consistent activity over extended periods.

\textbf{Varying level of experience with AI.} An important goal of this project was to have a diverse pool of annotators and we thus did not limit the selection criteria to working knowledge of language models or AI in general.
As a result, there were different levels of understanding amongst the annotators what was meant by a \textit{prompt} and \textit{completion}. For example, we found at least one contributor with 3,684 contributions to three languages (English, Somali, Standard Arabic) who failed to structure their submissions as a prompt with a question. Instead, the contributor used an extract of text as the prompt and its continuation in the completion.
Prefacing such prompts with an instruction such as \textit{``Complete the following partial extract of text:"} would have been a more suitable format. 
While we routinely provided examples to contributors, there was a clear need for a systematic way to review and measure the quality of submissions.

\textbf{Validating the quality of contributions} We follow a peer-review approach where each annotator acts as a reviewer for the other annotators working on the same language. These reviews form the basis for a quality \aya score which is displayed on the leaderboard in the UI.
The quality score for an annotator is calculated by averaging the combined average ratings of their examples provided by other annotators who serve as reviewers. We provide more details about how annotations are reviewed in the Appendix Section \ref{sec:reviewing-annotators}.
All three tasks in the \aya UI are connected in a sequential pipeline where submissions from ``Original Annotations'' are reviewed in the ``Re-Annotations'' task, and the re-annotations are further reviewed as part of the ``Annotation Feedback'' task. This systematic approach allows for a robust evaluation and enhancement of the collected data.

\subsection[Criteria for Inclusion in Aya Dataset]{Criteria for Inclusion in \aya Dataset}\label{sec:selectioncritera}

The \aya Dataset includes all original annotations and a subset of all re-annotations. We only release re-annotations if there is a difference between the original and the edited version. To determine this subset, we compute the sum of edit distances \textbf{d} (Levenshtein distance \citep{levenshtein1966binary}) between the original and re-annotated prompts and completions on the character level and use an acceptance threshold of ($d \geq 5$). This ensures that we do not release duplicates of existing data. 

Only languages with at least 50 contributions were included in the final release of \aya Dataset. This threshold was picked as it represents a balance between achieving a reasonable level of data quality and considering the practical limitations of human resources for some languages. The goal is to include as many languages as possible without lowering the overall quality of the dataset.
Table~\ref{tab:language_codes} lists details of the languages included in the \aya Dataset. 

\begin{table}[htb!]
    \small
    \centering
    \begin{tabular}{l l r}
        \toprule
         & & \textbf{Count} \\
        \midrule
        Original Annotations & & 138,844 \\
        \midrule
        \multirow{3}{*}{Re-Annotations} 
            & xP3 datasets & 2859 \\
            & Translated datasets & 7757 \\
            & Templated datasets & 11013 \\ 
            & Original Annotations & 43641  \\

        \midrule
        \textbf{\aya Dataset Total} && \textbf{204,114} \\
        \bottomrule
    \end{tabular}
    \caption{\aya Dataset Statistics (number of pairs of prompts and completions obtained through various annotation tasks).}
    \label{tab:aya-stats}
\end{table}

\section[Analysis of the Aya Dataset]{Analysis of \colorbox{ayad}{\aya Dataset}}
\label{sec:analysis_of_aya_dataset}

\subsection{Statistics}

The \aya Dataset contains a total of 204,114 instances collected via the \aya Annotation Platform. Table~\ref{tab:aya-stats} provides the breakdown of original annotations and re-annotations in the final dataset. 
The dataset covers 65 languages: 22 high-resource, 12 mid-resource, and 31 low-resource languages (See Appendix~\ref{sec:lang-groups} for more details on our language mappings).

\subsection[Length of Aya Dataset]{Length of \aya Dataset}\label{sec:length_dataset}

One objective of this project was to collect fluid original human prompts and completions.
Table~\ref{tab:aya_dataset_examples} provides examples of prompts and completions from the \aya Dataset. During the data collection process, annotators were provided with examples and guidelines but were also trusted to explore their own creativity and cultural background to come up with new examples. As a result, it is meaningful to understand differences in aggregate statistics like length across datasets, language type and relationship with perceived quality.

\begin{figure}[t!]
  \centering
  \includegraphics[width=0.55\textwidth]{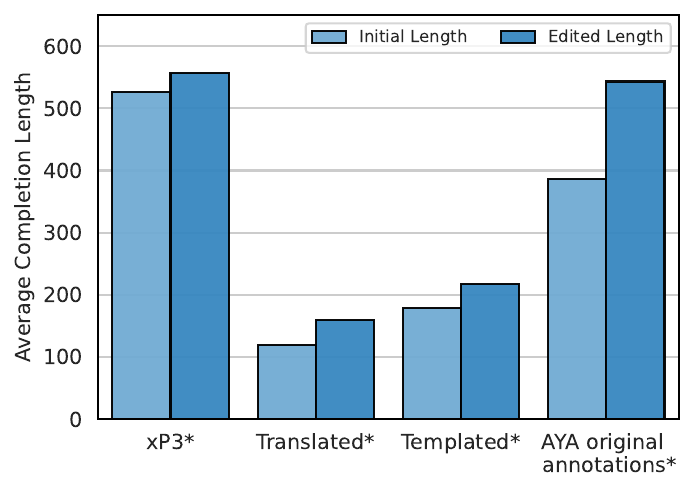} 
  \caption{Average Completion Length before and after re-annotation. Here (*) indicates the subset of all dataset categories (xP3, translated,  templated, and \aya original annotations) that were included in the \aya Dataset after re-annotation. Re-annotation improves average completion length across all datasets.
  }
  \label{fig:completion_length_comparison}
\end{figure}

\begin{wrapfigure}{r}{0.55\textwidth}
  \centering
  \includegraphics[width=0.5\textwidth]{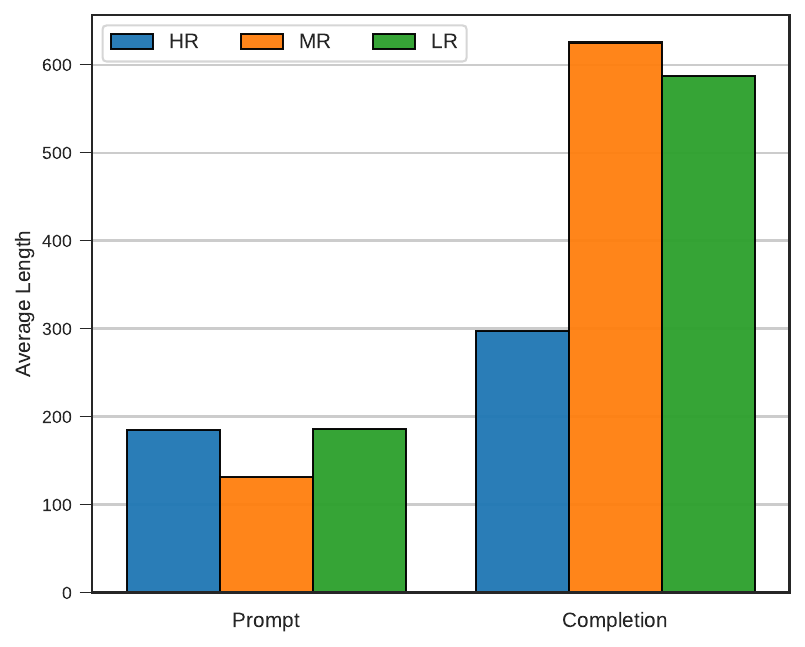} 
  \caption{Average prompt and completion length of instances in the \aya Dataset across different language categories (high (HR), mid (MR) and low (LR) resource languages, see Table~\ref{tab:language_codes}).}
\label{fig:Avg_prompt_completion_length_comparison_across_lang_groups}
\end{wrapfigure}

\textbf{Impact of Re-Annotation $\:$} When editing existing instances, we instructed the annotators to prioritize enhancing both the quality and richness of the prompts and completions. 
The average length of completions before and after edits are shown in Figure~\ref{fig:completion_length_comparison}.
We observe that across all data sources, the average length of completion increased after editing. On average, the length of completions after edits is 25\% longer than before edits. We observed the largest increase for \aya original annotations surfaced in the UI -- which were 40\% longer on average than the original length.

\textbf{Length difference across language groups $\:$} The average prompt and completion length (number of characters) observed across these different language groups is shown in Figure~\ref{fig:Avg_prompt_completion_length_comparison_across_lang_groups}. A distinct contrast exists in completion lengths between mid and low-resource languages when compared to high-resource languages.
Long completions and complete sentences are valuable in instruction-tuning datasets, particularly when training multilingual models to generate content in those languages.

\begin{figure}[t!]
  \centering
  \includegraphics[width=\textwidth]{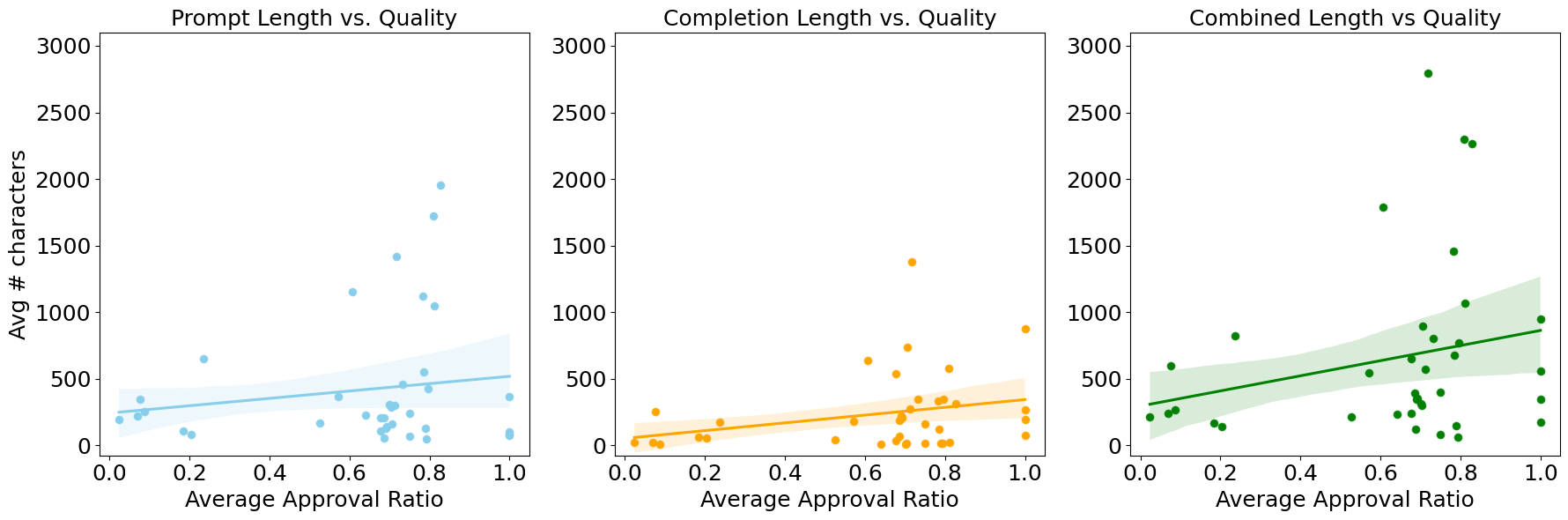}
  \caption{Relationship between Average Prompt and Completion Length in characters and the Average Approval Rate of the example. }
  \label{avg-length-vs-approval-ratio}
\end{figure}

\textbf{Length vs. Perceived Data Quality $\:$} Although longer completions can be valuable for training models to generate long and natural text, it does not necessarily imply higher quality.
Using annotators' feedback in the UI, we further investigate the impact of length on the perceived quality of the samples.
Figure~\ref{avg-length-vs-approval-ratio} showcases this analysis. We observe a positive correlation between how long the prompts and completions are and their resulting average approval ratio. 
Specifically, when we plot combined prompt and completion length against quality, we observe a correlation coefficient of 0.27. 
This finding emphasizes the importance of using longer prompts and completions and incorporating complete sentences to ensure a positive human experience when engaging with such a model.

\textbf{Length difference by language $\:$} Figure~\ref{fig:per_language_prompt_comp_length} in the Appendix illustrates the statistics per language. We observe an array of patterns that differed across languages.
For instance in \texttt{Japanese}, completions are on average 31\% shorter than prompts. 
On the other end, for \texttt{Urdu} and \texttt{Yoruba}, completions are notably long relative to prompts. On average, completions are 1258\% and 2516\% longer than the corresponding prompts for \texttt{Urdu} and \texttt{Yoruba}, respectively.
The average completion length in \texttt{Yoruba} is 1591\% longer than the average prompt length in \texttt{Japanese}. 
Figure~\ref{fig:Avg_prompt_comletion_length-Aya_dataset} provides the average length of the combination of prompts and completions per language.

\begin{figure}[htp!] 
  \centering
  \includegraphics[width=0.6\textwidth]{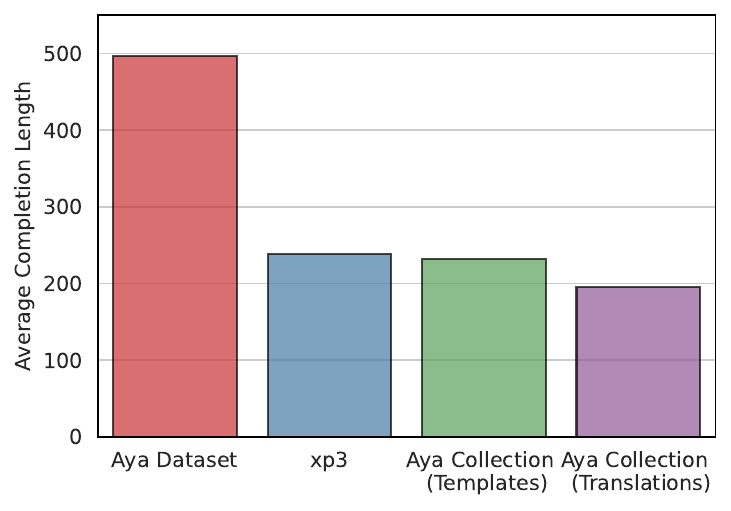}
  \caption{Comparison of completion lengths between \aya Dataset, \aya Collection, and xP3 (excluding the "code" split). 
  }
  \label{fig:completion-length-box-plot-xP3-vs-AYA}
\end{figure}

\textbf{Comparison in Completion length Across Datasets} The \aya Dataset has considerably longer completions on average when compared with other data collections as shown in Figure~\ref{fig:completion-length-box-plot-xP3-vs-AYA}.
This is particularly noteworthy given that the \aya Dataset is human-curated.
Given the presence of longer completions in the training data for many low-resource languages, we expect that models trained on the \aya Dataset will generate longer and more natural responses.

\begin{table}[t!]
    \centering
    \begin{tabular}{c}
     \includegraphics[
     clip,
     trim=4.5cm 6cm 2cm 6cm, 
     width=0.8\textwidth]{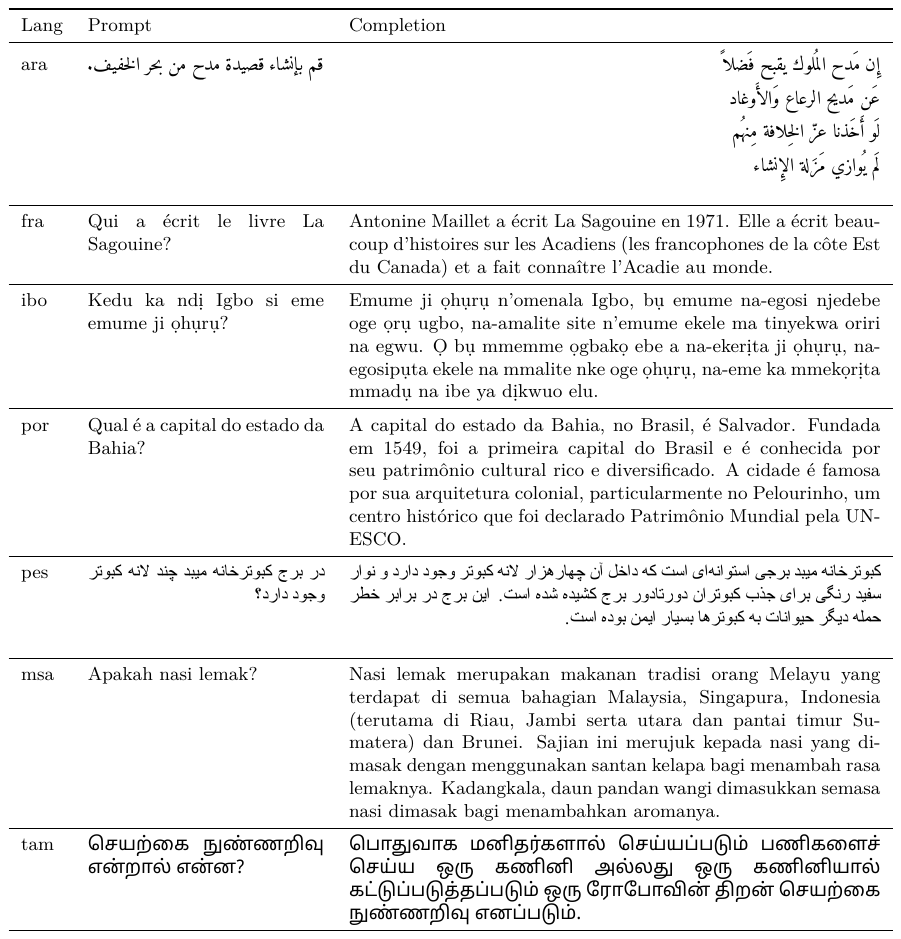}\\
    \end{tabular}
    \caption{Examples of prompt and completions in the \aya Dataset.}
    \label{tab:aya_dataset_examples}
\end{table}

\subsection{Annotator Skew}
\label{sec:anskew}
A feature of participatory research projects is the challenge of establishing and maintaining a balanced number of annotations across groups of annotators.
In the \aya project, the number of annotators per language varied due to numerous factors. 
As a result, the distribution of annotators is not uniform across languages. 
Moreover, within each language, there is a lack of consistent contributions from all annotators.
In this section, we examine the impact of annotator skew on the resulting dataset.

\begin{figure}[htb!]
\centering 
  \includegraphics[scale=0.63]{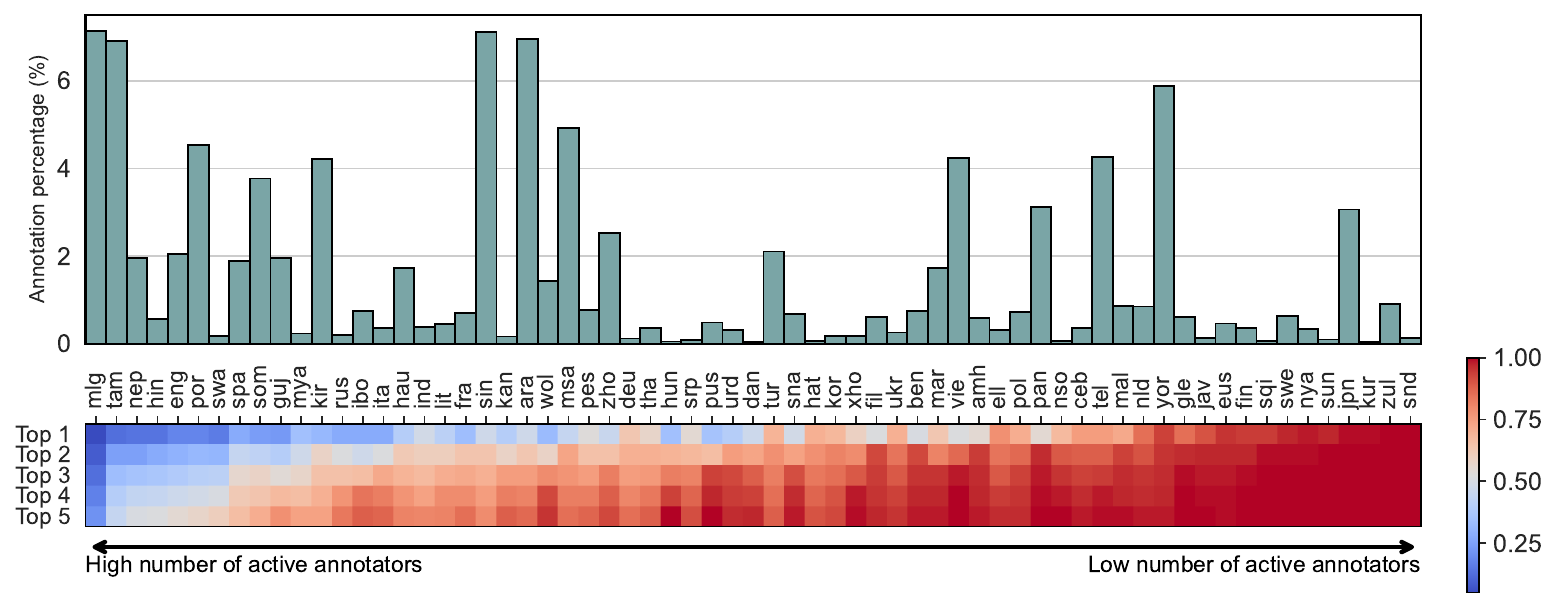}
\caption{\textbf{Top}: Ratio of all annotations per language with respect to the whole dataset. \textbf{Bottom:} Ratio of annotations done by the top-k most active contributors ($k=1,...,5$).  Languages annotations follow their respective ISO codes from Table~\ref{tab:language_codes}.}
\label{fig:skew} 
\end{figure}

\textbf{Annotator Skew Across Languages.} Annotators were encouraged to contribute to any language in which they could comfortably read and write and were asked to focus most of their efforts on languages other than \texttt{English}. Although a significant number of participants registered for many languages, the engagement level of annotators was not equal, which resulted in considerable differences in the number of contributions across languages.
Figure~\ref{fig:skew} (top) provides an overview of the percentage of each language present in the final compilation. The highest number of contributions is for \texttt{Malagasy} with 14,597 instances, and the lowest is 79 for \texttt{Kurdish}.

\textbf{Annotator Skew Within a Language.} The final contributions for each language in the \aya Dataset are not evenly distributed among annotators.
The median number of annotators per language is 15 (mean is 24.75) with one language having only a single active annotator (\texttt{Sindhi}) and some having over 80 annotators (\texttt{English} and \texttt{Portuguese}). 
Note that annotators made contributions at varying rates, and there is no direct correlation between the number of annotators and the ultimate count of language contributions. A limited pool of annotators for some languages implies that most instances in that language originate from a smaller group of individuals.
Figure~\ref{fig:skew} (bottom) illustrates the proportion of instances in a language originating from the most active annotators.
We observe a skewed pattern where for 12 languages, the 5 most active annotators contributed all examples.
There is an uneven distribution of contributions for many languages because those languages had a smaller number of voluntary annotators throughout the entire project despite rigorous outreach.
Additionally, we did not establish a specific quota for annotators to meet; everyone contributed as they desired, resulting in varying levels of activity among annotators.

The most extreme cases are \texttt{Zulu} and \texttt{Sindhi}, where one annotator in each language volunteered for all contributions in Annotation and Re-annotation tasks. Thus, in Figure~\ref{fig:skew} their top-1 contributor ratio is 1.0 and does not change when moving to top-2 or further.
The languages with the least skewed distributions are \texttt{Malagasy, Tamil, Nepali, Hindi, English} and \texttt{Portuguese}. The language \texttt{English} also had the highest number of unique annotators with 130 individuals out of which 95 annotators contributed to \texttt{English} as their second language for annotation purposes. 
Given the uneven distribution of annotators per language, it is important to acknowledge that individual annotator quality has a disproportionate influence on some languages.

\subsection{The impact of introducing the \aya Score}
As part of our collaborative annotation effort in \aya, we emphasized the importance of quality as well as long completions that contain clear responses to the instructions specified in the prompt during the project.
To encourage high-quality examples from the annotators, we introduced the \aya Score (Section \ref{sec:highquality}) halfway through the project to focus on the quality, in addition to the quantity, of contributions. 

\begin{figure}[ht!]
    \centering
    \begin{subfigure}[b]{.45\textwidth}
    \centering
    {\includegraphics[width=\linewidth]{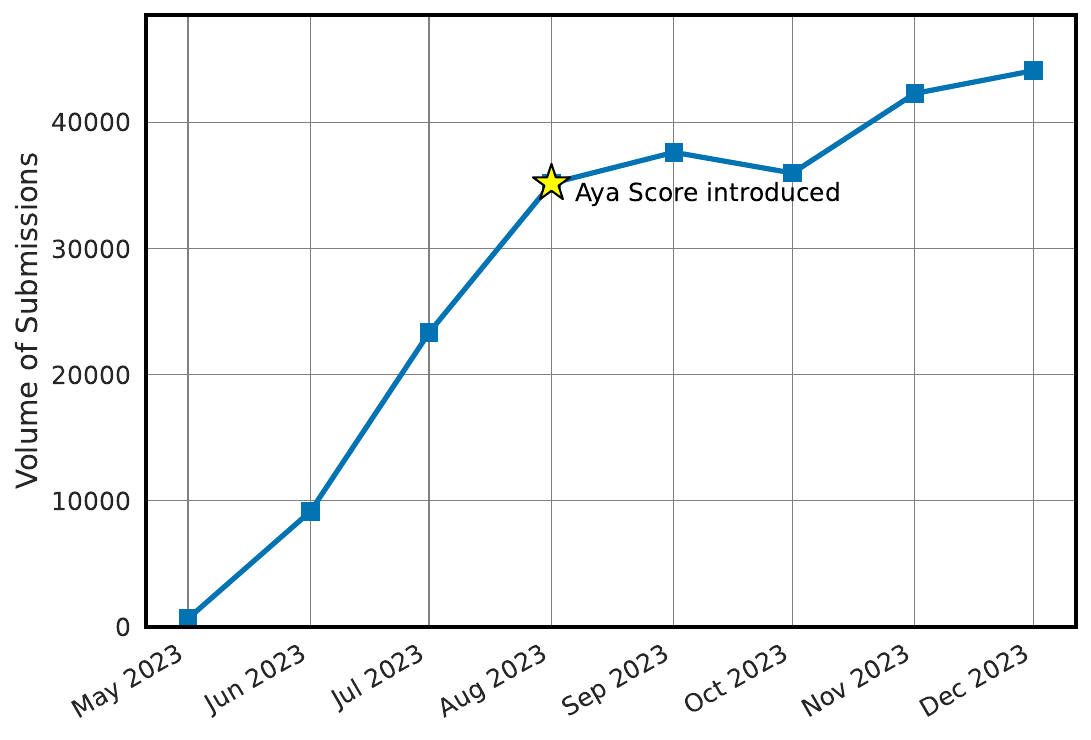}}
    \caption{Volume of submissions over time}
    \end{subfigure}
    \begin{subfigure}[b]{.45\textwidth}
    \centering
    {\includegraphics[width=\linewidth]{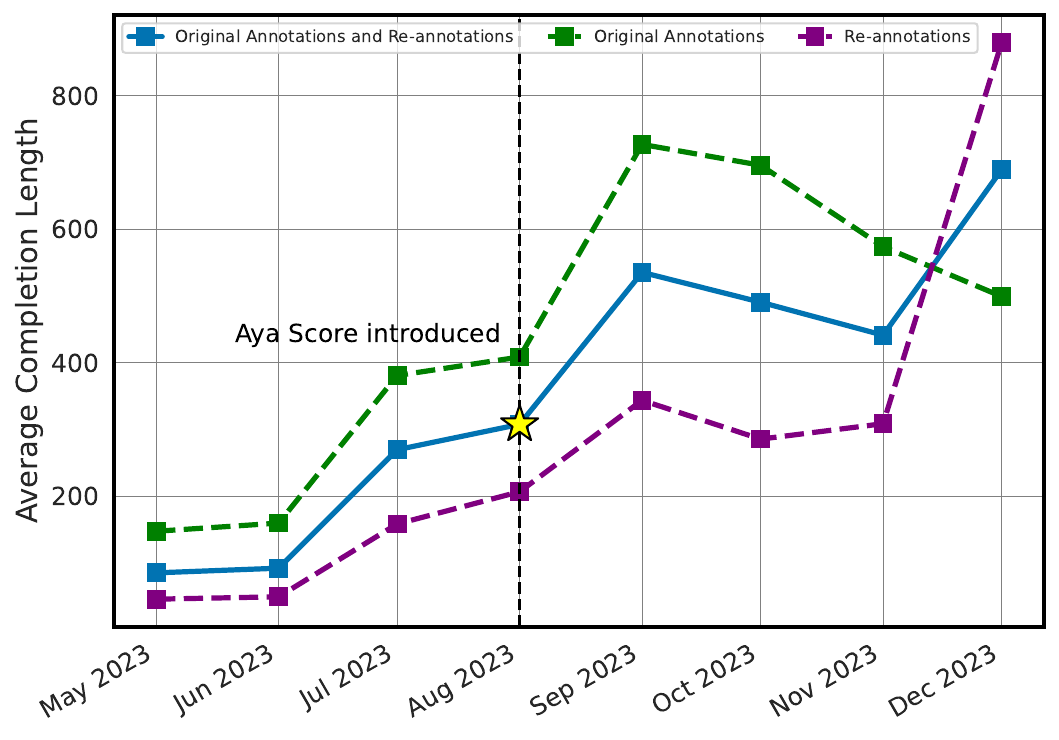}}
    \caption{Average completion length over time}
    \end{subfigure}
    \caption{The volume of original annotations and re-annotations increases after the introduction of \aya Score. We also observe a marked uptick in the completion length of all submitted annotations with the introduction of the \aya Score.
    }
    \label{fig:impact_of_aya_score}
\end{figure}

The \aya Score encouraged participants to incorporate more edits during annotation, with one specific guideline urging them to transform short answers into full sentences or paragraphs.
Figure~\ref{fig:impact_of_aya_score}(right) shows the change in the completion lengths over time. We observe that after introduction of the \aya score, there is a marked uptick in the completion length of all submitted annotations.

\section[Aya Collection]{\colorbox{ayac}{Aya Collection}}
\label{sec:aya_collection}

We introduce the \aya Collection, a comprehensive, large corpus of datasets that can be used by researchers around the world to train multilingual models. Our goal is only to include datasets with permissive licensing for manipulation and redistribution.\footnote{\url{https://en.wikipedia.org/wiki/Permissive_software_license}} Where possible, we report the license associated with each dataset within the \aya Collection. 

The \aya Collection consists of three different sources of data:
\begin{enumerate}
    \item \textbf{Templated data:} We collaborated with fluent speakers to create templates that allowed for the automatic expansion of existing datasets into various languages.
    \item \textbf{Translated data:}  We translated a hand-selected subset of 19 datasets into 101 languages (114 dialects) using the NLLB 3.3B parameter machine translation model~\citep{nllbteam2022language}. The full list of datasets translated is listed in Appendix Table \ref{tab:aya_collection_translated}.
    \item \textbf{\aya Dataset:} We release the \aya Dataset described in Section \ref{sec:analysis_of_aya_dataset} as a subset of the overall collection. This is the only dataset in the collection that is human-annotated in its entirety.
\end{enumerate}

\textbf{Dataset Selection Criteria $\:$} The templated and translated datasets in the \aya Collection were selectively hand picked to achieve a mix of different task types.
Our criteria prioritized datasets with high-quality natural and complete sentences, suitable for creating pairs of prompts and completions. 
Datasets that could potentially yield single-word answers were deliberately excluded.
Finally, to create a high-quality collection, we examined all datasets and excluded those identified as unclean or noisy, primarily attributable to their automatic creation processes.

\subsection{Templating Existing Datasets}
\label{sec:templated}
We explored the automatic expansion of existing datasets in various languages with human-written \emph{prompt templates}, following previous works \citep{naturalinstructions,bach2022promptsource,wei2022finetuned,wang-etal-2022-super}.
Unlike prior works that still either use English prompts in a multilingual dataset or rely on automatic translation to generate multilingual prompts, to our knowledge, \aya Collection is the first broad effort to involve fluent speakers in creating prompts unique to their language to expand existing datasets for instruction tuning.

We used the \texttt{PromptSource} framework \citep{bach2022promptsource} to template these datasets. 
We asked \aya community members to submit instructions and create templates for datasets in the languages they were proficient in. 
Our process includes: 1) Templating datasets with instructions in the same language as the original dataset; 2) If the dataset is not in English, annotating instructions in English.
Our input prompts can be monolingual or code-mixed, depending on whether we apply templates in the same language or in English to the dataset of a particular language. 
Note that code-mixed input prompts here refer to a \textit{structured} mixing of English instructions with non-English monolingual data \citep{lin-etal-2022-shot}, which is different from the typical sociolinguistic definition of code-mixing (or code-switching) of languages in natural conversational utterances \citep{winata-etal-2023-decades,yong2023prompting,dougruoz2023representativeness, srivastava2021challenges}. 

We examined the suggested templates and subsequently converted each dataset into an instruction-style format. We release these datasets under the \aya Collection. We list the details of all datasets we apply templates to in Appendix Table \ref{tab:list_of_templated_datasets_aya_collection}.

\subsection{Automatic Translation}
\label{sec:translation}

Research has demonstrated that training models with translated data can yield significant benefits \citep{aharoni2019massively, ZhiruiZhang18, TangY21}. 
We experiment with improving coverage of low-resource languages by selectively translating high-quality datasets from various existing collections.

\textbf{Setup} We selectively pick 19 high-quality 
IFT datasets from xP3~\citep{muennighoff-etal-2023-crosslingual}, the Flan Collection~\citep{longpre2023flan}, Dolly \citep{DatabricksBlog2023DollyV2}, along with additional sources such as SODA \citep{kim2022soda} and Mintaka \citep{sen-etal-2022-mintaka}. Datasets were prioritized for translation based on the richness of task diversity and length of completions. The complete list of these datasets is given in Appendix \ref{tab:aya_collection_translated}. These translations are available and open-source as part of the \aya Collection.
We process datasets for translation using the No Language Left Behind (NLLB)~\citep{nllbteam2022language} machine translation model, which is capable of single-sentence translations between 200 different languages and dialects in various scripts. 
For best performance, we use the largest NLLB model with 3.3B parameters.  

\begin{wrapfigure}{r}{0.35\textwidth}
    \centering
\includegraphics[width=0.28\textwidth]{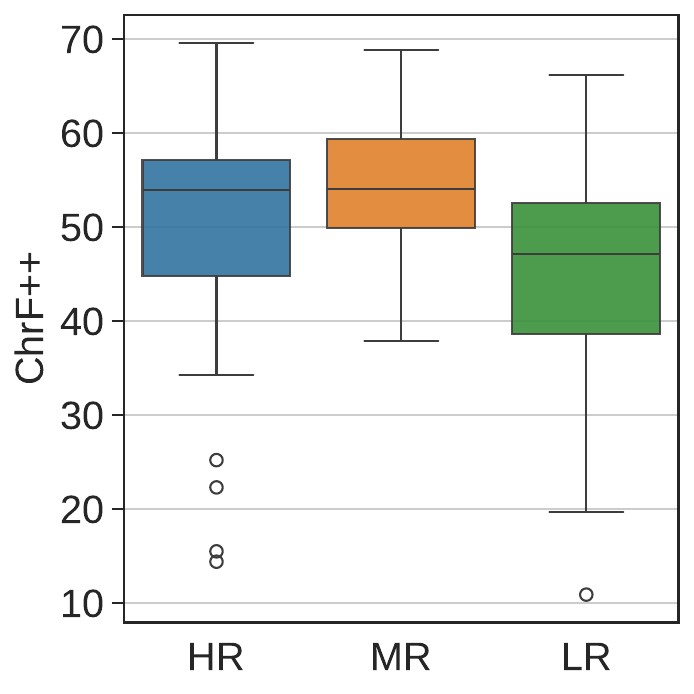}
    \caption{ChrF++ scores for the NLLB translation model, averaged across resourcefulness buckets.}
    \label{fig:nllb_quality_box}
\end{wrapfigure}

\textbf{Translation Quality $\:$} 
Appendix Section~\ref{app:nllb_quality} lists NLLB translation quality for each of the languages of interest, as reported in~\citep{nllbteam2022language}. Figure~\ref{fig:nllb_quality_box} shows the translation quality across languages grouped by their resourcefulness. The mean ChrF++ score on FLORES is 48.17 (min: 10.9, max: 69.6) for translations out of English, with a few outliers for HR and LR. We interpret this optimistically as strong enough to sufficiently serve our translation needs.
However, upon inspection of translation outputs for fine-tuning data, we encounter significant translation errors with Standard Arabic in Latin script and Minangkabau in Arabic script, so we exclude them from our translated dataset. 
In total, 19 public datasets 
were translated into 101 languages (114 dialects).
Details of these datasets can be found in Appendix Table~\ref{tab:aya_collection_translated}.  

In addition to releasing the translated datasets used as a basis for re-annotation, we also translated Dolly~\citep{DatabricksBlog2023DollyV2}. Dolly is a 15k instruction dataset Databricks collected by relying on its employees as annotators \citep{DatabricksBlog2023DollyV2}. Annotators were instructed to curate prompt and completion pairs in each of eight different instruction categories. 
In contrast to the mentioned NLP datasets, Dolly was purposefully designed to align language models with human expectations. It stands out as a high-quality, manually curated dataset covering a range of topics including brainstorming, classification, closed question answering, generation, information extraction, open question answering, and summarization.
The addition of the translated Dolly datasets is a valuable resource for languages that experience a scarcity of conversational instruction fine-tuning datasets.

The list of datasets, along with the number of languages, templates, and other statistics, can be found in Appendix Table~\ref{tab:aya_collection_translated}.

\section[Analysis of the Aya Collection]{Analysis of \colorbox{ayac}{\aya Collection}}
\label{sec:analysis_of_aya_collection}

\begin{table}[hbt!]
\small
    \centering
    \begin{tabular}{ll}
    \toprule
    \textbf{Main Task Type} & \textbf{Fine-grained Task Type} \\
    \midrule
    Question Answering              &   --- \\ \midrule
    Natural Language Generation     &   Summarization \\
                                    &   Translation \\
                                    &   Paraphrasing \\
                                    &   Dialogue \\
                                    &   Text Simplification \\ \midrule
    Text Classification             &   Sentiment Analysis \\
                                    &   Information Extraction \\
                                    &   Named Entity Recognition \\
                                    &   Event Linking \\
                                    &   Natural Language Inference \\
                                    &   Document Representation \\
                                    \bottomrule
    \end{tabular}
    \caption{Task Taxonomy of NLP tasks in the \aya Collection.}
    \label{tab:task_taxonomy}
\end{table}

\subsection{Statistics}

\textbf{Overview $\:$} The \aya Collection consists of existing NLP datasets that are templated to include instructions as well as datasets already in instruction format submitted by the \aya community. 
Table~\ref{tab:list_of_templated_datasets_aya_collection} shows the detailed list of datasets. 
The full list of templates is available in Section~\ref{tab:aya_collection_templates}.
The final \aya Collection consists of 44 multilingual and non-English templated datasets and 19 translated datasets, with 513M individual instances. 
Overall, the collection covers 114 languages\footnote{We release the \aya Dataset as part of the \aya Collection, bringing the total number of languages in the collection to 115. However, for the sake of clarity, when referencing the \aya Collection statistics in this paper, we exclude the \aya Dataset.}.

\textbf{Tasks Covered Across Templated and Translated Datasets $\:$} 
We aim to include datasets from various tasks in the collection while ensuring that they follow our selection criteria.
Table~\ref{tab:task_taxonomy} illustrates our task coverage in the \aya Collection, drawing inspiration from xP3 and the Flan Collection. 
We have a total of three main task types: Question Answering (QA), Natural Language Generation (NLG), and Text Classification (TC). Within these larger umbrella tasks, we define several finer-grained task types based on the datasets, resulting in a total of 11 finer-grained task types. These finer-grained task types are determined by the frequency of datasets in the \aya Collection encapsulating that task. 

\begin{figure}[htb!]
  \centering
  \includegraphics[scale=0.45]{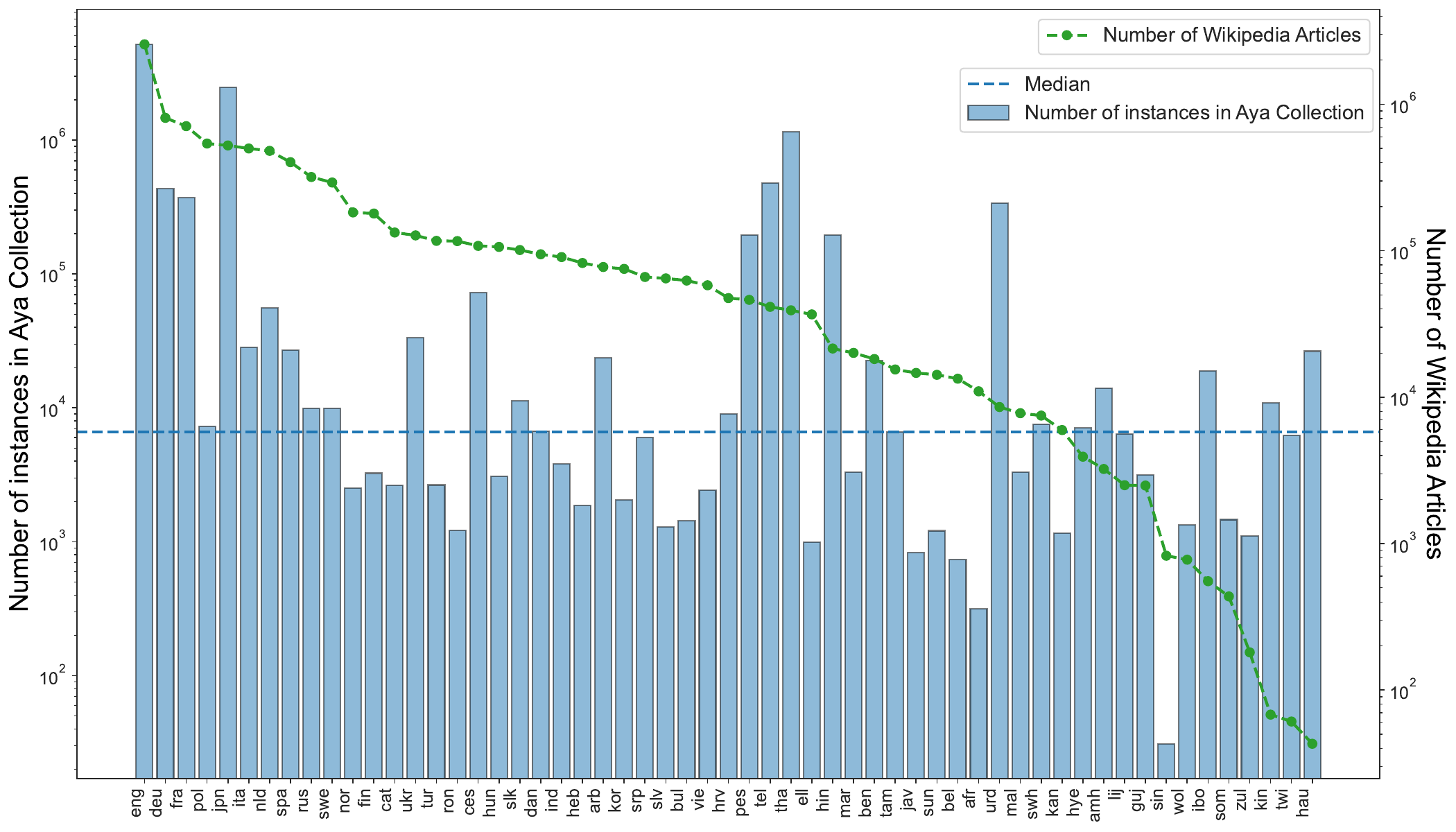}
  \caption{Number of prompt/completion pairs in each language in the \aya Collection (templated). Many languages with limited digital presence, as indicated by a low number of Wikipedia pages, are well-represented in the templated portion of the \aya Collection. Note that absolute  Both axes are in log-scale. }
  \label{fig:Aya_collection_lang}
\end{figure}

For QA, we decided to keep only the main task type, as the intended goal of question-answering tasks is clear: \textit{Answer a proposed question}. 
The type of the question can be different: open-ended, close-ended, multiple-choice, single response.
For NLG, finer-grained task types include Summarization, Translation, Paraphrasing, Dialogue (Generation), and Text Simplification.
For TC, we include the following finer-grained task types: Sentiment Analysis, Information Extraction, Named Entity Recognition, Event Linking, Natural Language Inference, and Scientific Document Representation. 
Finally, we label the task categories of each dataset in the \aya Collection in Table~\ref{tab:templated_taxonomy} and Table~\ref{tab:translated_taxonomy}. 
If we are not able to find a fine-grained task type for the dataset, we keep the main task type.

\textbf{Language Balance $\:$} One of the objectives of templating (and translating) existing datasets is to broaden the available resources for languages that have limited digital data. 
To examine if our final collection adheres to a similar distribution pattern, we use the number of Wikipedia pages in each language as a proxy for the online presence of its fluent speakers.
Figure~\ref{fig:Aya_collection_lang} showcases that although the number of instances for languages varies in the \aya Collection (templated subset), it does not disadvantage languages with fewer Wikipedia pages. 
The distribution still ensures a reasonable coverage across all languages.
It is imperative to emphasize that our analysis does not involve a direct comparison of absolute values, given the disparate units of measurement involved. Instead, we examine the \textit{patterns} of data scarcity for various languages in our collection versus Wikipedia.
Including the translated datasets in the \aya Collection further reduces disparities between languages and contributes to creating a more balanced collection.

\begin{figure}[htb!]
  \centering
  \includegraphics[scale=0.55]{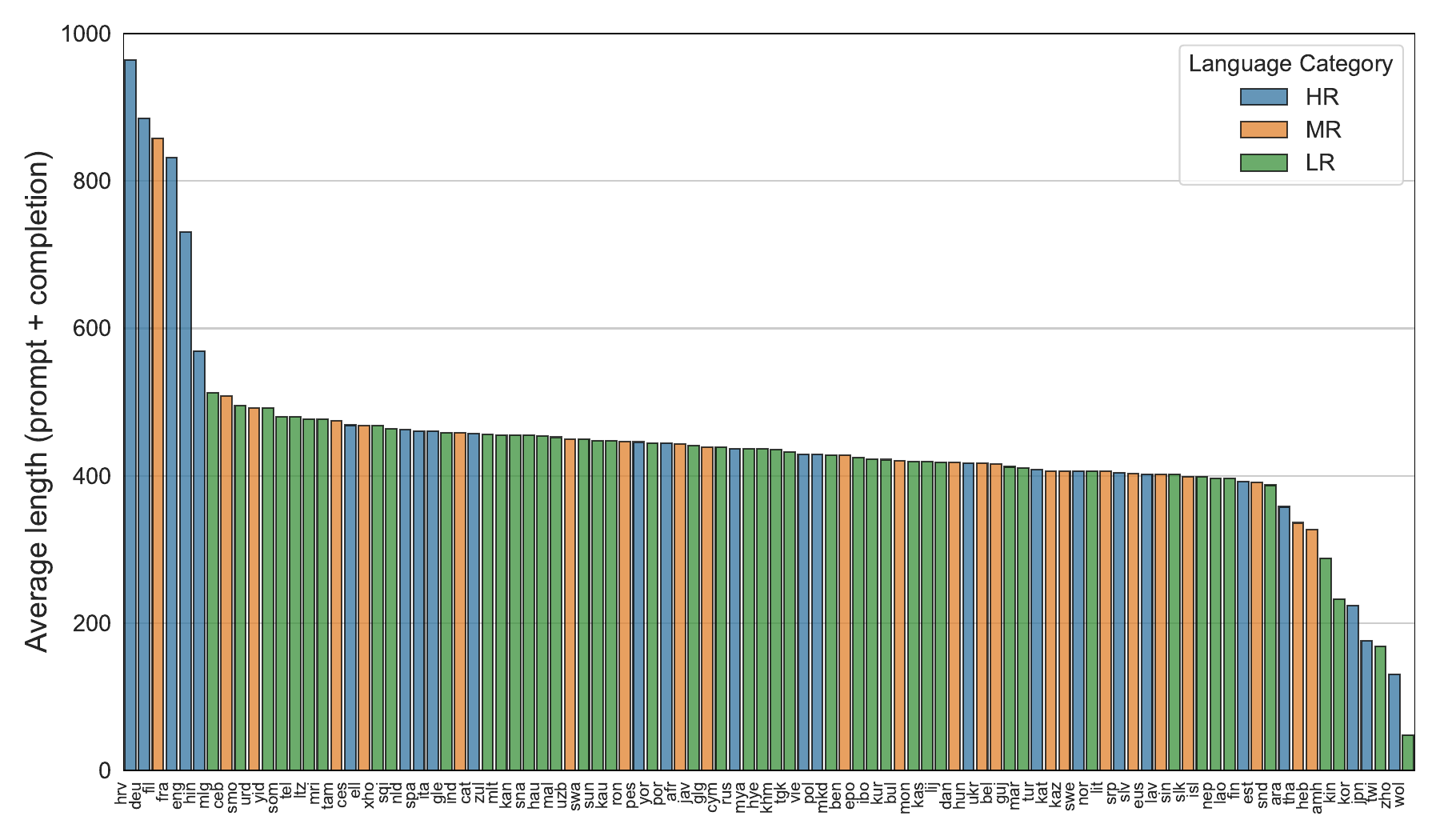}
  \caption{The average length of prompts and completions for high (HR), medium (MR) and low-resource (LR) languages in \aya Collection.} \label{fig:Aya_collection_lang_length}
\end{figure}

\textbf{Prompt and Completion Lengths $\:$} Figure~\ref{fig:Aya_collection_lang_length} shows the distribution of length across languages.
No discernible pattern is observed when examining lengths for high-resource languages compared to low-resource languages. 
Low-resource languages appear at both ends of the distribution, occupying both the head and tail.
In the \aya Collection some low-resource languages (e.g., \texttt{Somali} and \texttt{Amharic}) have longer average completions length than medium or even high-resource languages.
The dedication of individual participants in identifying datasets in their own language and templating them has made a significant difference for many languages.

\subsection{Quality Assessment of All Different Data Sources}

As previously stated, binary feedback on the quality of the prompt-completion pairings was collected from the annotators.
We define the average approval ratio per dataset which serves as a valuable metric for assessing the quality of datasets across various languages and diverse data sources.
We compute the average approval ratio as $\mathcal{T_{+}} / \mathcal{T} $, where $\mathcal{T_{+}}$ represents the total number of thumbs up, and $\mathcal{T}$ represents the total number of votes per dataset.
An average approval ratio of 1.0 would indicate that every annotation was perceived to be of good quality and all prompts and completions had received a thumbs up. An average approval ratio of 0.0 would indicate that every annotation was perceived to be of poor quality, and all prompts and completions had received a thumbs down. We constrained our quality analysis to the 40 datasets in our pool for which we had at least 20 instances of feedback.

\begin{figure}[htb!]
  \centering
  \includegraphics[width=0.58\textwidth]{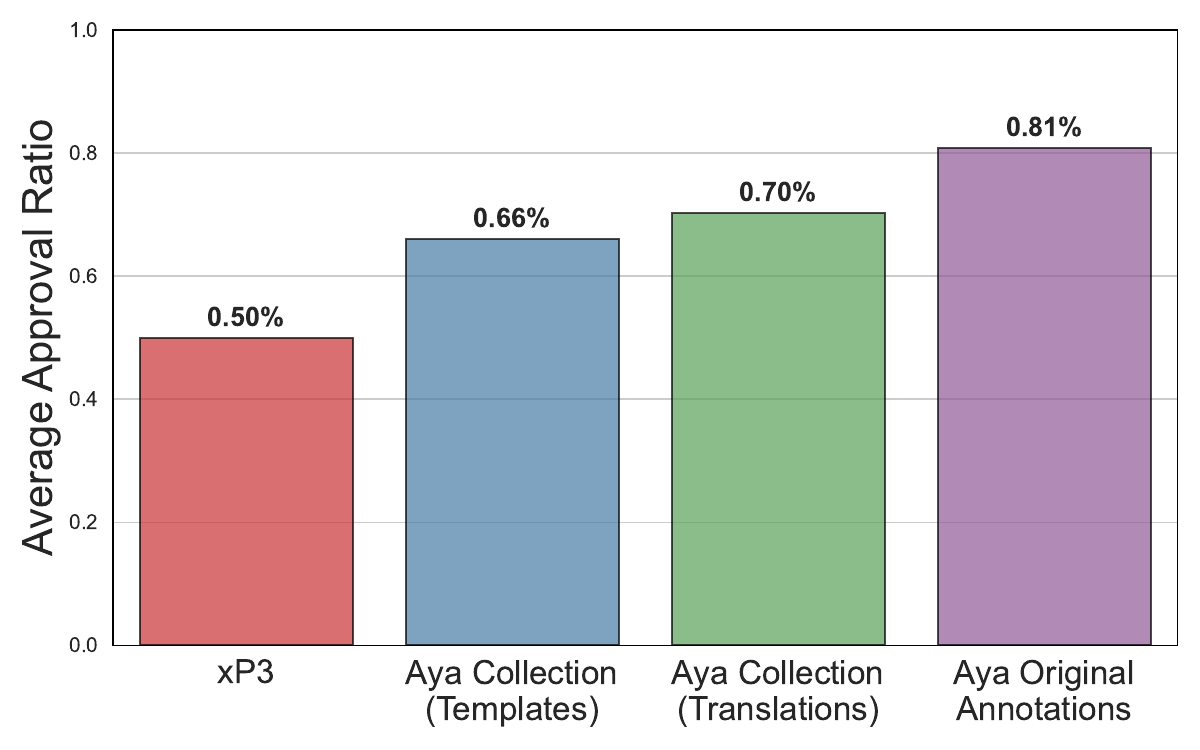}
  \caption{Average approval ratio per dataset group, constrained to datasets receiving at least 20 votes.}
  \label{avg-approval-ratio-per-dataset-group}
\end{figure}

Overall, we observe that the majority of datasets were of above average (0.5) quality based on their approval ratio, with all translated data as well as Original Annotations being above average. However, across all the datasets within each group ---xP3, Templated, Translated, and \aya original annotations--- \aya original annotations were perceived to be of the highest quality, with an average approval ratio of approximately 0.81, compared to the lowest quality dataset, xP3, which had an average approval ratio of approximately 0.50. 
This aligns with our intuition that carefully curated datasets lead to high-quality annotations as perceived by human annotators.
Figure~\ref{avg-approval-ratio-per-dataset-group} provides a summary of the results for each group. Figure~\ref{avg-approval-ratio-per-dataset} in the Appendix provides approval ratios per datasets in each group. 

\section{\colorbox{ayae}{\aya Evaluation Suite}}
\label{sec:evaluationsuite}

Lastly, as part of the \aya project we curate and release an evaluation suite tailored for multilingual models.
This set is a valuable contribution in tackling the scarcity of multilingual data, a challenge that becomes even more apparent when considering evaluation sets.
While there are several test sets available for evaluating multilingual models~\citep{conneau2018xnli,ponti2020xcopa,lin-etal-2022-shot}, they focus primarily on discriminative tasks.
To evaluate multilingual models' generations, the literature includes task-specific evaluation sets such as Translation~\citep{goyal2021flores101}, Summarization~\citep{hasan-etal-2021-xl} and Question Answering~\citep{tydiqa}.
However there is currently a gap in evaluating \textit{open-ended generation} capabilities of LLMs within a multilingual context. 
We aim to address this gap by curating a multilingual evaluation set tailored for assessing the open-ended generation capabilities of LLMs, such as brainstorming, planning, and other unstructured, long-form responses.

To strike a balance between language coverage and the quality that comes with human attention, we create an evaluation suite that includes \textbf{(1)} human-curated examples in a limited set of languages, \textbf{(2)} automatically translations of handpicked examples into a more extensive number of languages, and \textbf{(3)} human-post-edited translations into a small number of languages. 
We consider two primary sources of data: original annotations from \aya dataset (comprising new examples culturally curated for different languages) and Dolly prompts (high-quality, human-written examples carefully selected to have a universal reach).
The subsets comprising the \aya evaluation suite are:

\textbf{\textsc{aya-human-annotated} test set $\:$} For ease of future adoption, we have partitioned the \aya dataset into training and testing splits. The test set of the \aya Dataset contains 1,750 of the total instances (250 instances from 7 languages), selected at random from original annotations. Our goal is to achieve a balanced representation of languages in the test set and ensure a sufficient number of examples per language. To guarantee enough data remains for training, we focused on languages with at least 2000 original annotations.  In order to ensure linguistic diversity, we included languages that were varied in terms of high, mid, or low-resourcedness, as well as script and language families. For those reasons, the test set consists of \texttt{English} (high-resource, Latin script, Indo-European), \texttt{Portuguese} (mid-resource, Latin script, Indo-European), \texttt{Simplified Chinese} (high-resource, Han, Sino-Tibetan), \texttt{Standard Arabic} (high-resource, Arabic script, Afro-Asiatic), \texttt{Telugu} (low-resource, Telugu script, Dravidian), \texttt{Turkish}  (mid-resource, Latin script, Turkic), and \texttt{Yoruba} (low-resource, Latin script, Atlantic-Congo). See Table~\ref{tab:language_codes} for more details. 

\textbf{\textsc{dolly-machine-translated} test set $\:$} We separate a curated subset of 200 Dolly prompts~\citep{DatabricksBlog2023DollyV2} to serve as an additional translated evaluation set. 
Our aim with this selection was to exclude any culturally or geographically specific prompts and completions. 
Hence, two reviewers inspected a set of initially 500 English prompts that were uniformly sampled based on the task categories in Dolly. The reviewers excluded prompts that rely on geographic knowledge such as \textit{``Looking at cities in Australia that are on the east coast and the west coast of the country, which coast are the cities of Fremantle, Sydney, Brisbane, Perth, Cairns, Townsville, Newcastle located on?''}, or prompts such as \textit{``Why is NFL football called football when players use their hands mainly?''} that rely on overly specific cultural references. When two reviewers disagreed, a third reviewer was asked to break the tie. We kept prompts such as \textit{``Is art useless?''} or \textit{``Write a short paragraph about why you should not have both a pet cat and a pet bird.''} and questions that refer to geographic specific knowledge where the supporting evidence was provided in the prompt itself e.g., \textit{``Given a reference text about Minister for Food, Agriculture and Fisheries of Denmark, when was the position created and was was it named?''}. 
Although not perfect, the intention behind this selection was to gather a test set that allows us to evaluate the fluency and quality of responses in various languages while avoiding model assessment on prompts tied to specific cultural or geographic references that might have language-dependent validity. We automatically translate the prompts with NLLB into 101 languages and their dialects that are captured by NLLB.
Including the original \texttt{English} prompts this dataset covers 115 dialects.

\textbf{\textsc{dolly-human-edited} test set $\:$} The automatic translation process may introduce errors in the prompts that render them nonsensical. For example, the prompt \textit{``Which is a species of fish? Bleak or Weary''} requires domain expertise to choose the right translation of the fish names rather than literal translations of the adjectives (as e.g. in the NLLB Translation into \texttt{Spanish}: \textit{``Desanimado o cansado.''} (=\textit{``discouraged or tired''})). If the prompt does not make any sense, there is no clear expectation and measurement of what a good and correct completion should look like. 
To confidently interpret evaluation results, it is imperative to establish a reliable set of prompts for evaluation. 
To enhance the reliability of testing on these prompts, we therefore enlist professional human annotators to post-edit the examples (e.g. for the example above \textit{``Alburno o Cansado''} (=\textit{``[Fish name] or Tired''}).
We post-edit the prompts for a subset of six languages: \texttt{Arabic}, \texttt{Hindi}, \texttt{Spanish}, \texttt{French}, \texttt{Serbian} and \texttt{Russian}. Appendix~\ref{apx:postedit_setup} describes the post-editing process and effort in more detail.
The example above illustrates that some prompts, even when translated correctly, might still not transfer well into other languages---which is the main difference between a translated English-centric set like this and an evaluation set originally written in each target language like \textsc{aya-human-annotated}.

We open-source the \textbf{\textsc{dolly-machine-translated} test set} to be an additional resource for researchers, although warn that the expressiveness of a translated evaluation set is limited by the quality of the translation model (and human post-edit) and may adversely impact an estimate of ability in languages where translations are not adequate~\citep{nogara2023toxic}. Ultimately, this is a compromise between having evaluation coverage in a more complete set of languages (101 languages and 114 dialects in total) versus having human-annotated evaluation sets. \textbf{If using the automatically translated test set, we recommend it be paired and reported with the professionally post-edited \textsc{dolly-human-edited} for 6 languages, or the \textsc{aya-human-annotated} set which also only covers 7 languages but is entirely created by proficient target language speakers.}

\section{A Participatory Approach to Research}
\label{sec:particpatory_approch_to_research}

Recent breakthroughs in NLP have predominantly come from narrow collaborations that involve researchers from a handful of institutions and regions of the world~\citep{Nakamura2023}. This reliance on small, specialized collaboration networks has been shown to hinder innovation~\citep{Park2023PapersAP}. Dataset creation as a process has often been undervalued, with minimization of the value of creators' contributions~\citep{andressaddressing, peng2021mitigating, hanley2020ethical}. Under such conditions, the richness and diversity of the data are often compromised, as it reflects a limited perspective that aligns with the interests of those who wield greater power in these transactions. Data is not, as metaphors such as \textit{`data mining'} \citep{Puschmann2014}, or \textit{`data is the new oil'} \citep{Stark2019, Awati2015}, might suggest, a natural resource waiting to be exploited. Whenever we engage with data, we are also engaging with the connections that data has to the people who produce, prepare, and distribute it \citep{Seaver2021, Pinel2020,crawford2021a}. Participatory approaches in AI design and research are one way to address gaps in access to resources needed for research: through collaborative partnerships with language speakers and local communities. 

\aya is an example of a participatory research project \citep{Birhane_2022,corbett2023,Delgado2023ThePT}. Here, the research is the result of a broad cross-institutional, global collaboration. This type of cross-sectional work facilitates the collection of vital linguistic data and community engagement, which is crucial for developing effective language technologies \citep{joshi-etal-2019-unsung,forall-nekoto-etal-2020-participatory}.
We describe below some of the guiding principles we followed throughout the year-long \aya project.

\textbf{Fluid Ownership and Growth $\:$} Our open science framework allowed us to challenge the norms of how computer science usually proceeds \citep{WittOS, SabouOS}. Traditional research approaches often involve rigid hierarchies; typically, research is conducted within academic institutions or corporate labs where roles are clearly defined, and collaboration is mostly synchronous, relying on in-person meetings or real-time communication. In contrast, \aya took a decentralized and democratic approach to collaboration, supporting fluid leadership and flexible role adoption. This empowered members to take initiative and lead in areas where they had passion or expertise, regardless of their position in academia, or when they became involved in the project. For example, members became Language Ambassadors at many different points during the year-long project, and mentorship roles evolved naturally with more experienced researchers providing guidance to those more junior (see Appendix~\ref{sec:repr_communities} for more details of different roles in the project).

\textbf{Organizational Structure $\:$} The communication channels and organizational structure of \aya were designed to facilitate rich collaboration that could evolve with the interests of participating researchers over the year-long project. For example, most communication between independent researchers involved within \aya was asynchronous over Discord, which allowed researchers in different time zones to participate in discussions. Monthly meetings were open for anyone to attend and were recorded for asynchronous viewing. We describe the structure of meetings and communication more thoroughly in Appendix~\ref{apx:platforms} and~\ref{apx:meetings}.

\textbf{Inclusion and Access $\:$} The open nature of the \hyperlink{https://aya.for.ai/}{\aya UI} allowed us to bypass the gate-keeping mechanisms of academic science that often marginalize non-English speakers and people without formal academic credentials \citep{West2020}. Expertise in the command of a spoken or written language is clearly distinct from expertise in machine learning. The inclusion of such a wide range of volunteers gave us more representative data in a wide variety of languages and also gave volunteers a glimpse into the often obscure world of machine learning.

\textbf{Who Participated in \aya $\:$} The motivations of contributors were not based on financial remuneration but on ideals of community, identity, and social justice. Participants saw their roles as Language Ambassadors and contributors as crucial to ensuring the inclusion of their languages in the ongoing transition to a digital, information-driven economy. The Language Ambassador for \texttt{Malagasy}, a language-driven to the risk of extinction by colonial French rule in Madagascar \citep{Spolsky2017}, is planning hackathons in 2024 to use the \aya Dataset to create voice-to-text apps that will help non-literate speakers of \texttt{Malagasy} participate in the modern economy. In \texttt{Telugu}, a traditional genre of poetry known as Sathakam is an integral part of the educational system. However, chatbots that can translate text into \texttt{Telugu} have little to no understanding of the Sathakam form. The \texttt{Telugu} Language Ambassador told a newspaper in Toronto that ``in \aya, we made sure to include as many Sathakams as we could find''~\citep{Castaldo2023}.

These motivations are not peripheral to the strength of the final \aya Dataset but are key factors in the data’s provenance \citep{Loukissas2019}. These qualitative dimensions remind us that language is, for the people who use it every day, an intimately social phenomenon. Beyond the symbolic notation that connects tokens to referents in the real world, we find a robust network of social relations that are necessary for languages to flourish \citep{Sidnell2012, Goodwin2017, Agha2006}. The social interactions between contributors, ML researchers, and social scientists in the \aya project were crucial to its success. Contributors shared playlists of their favorite songs from their home country, recipes from their childhood, and snapshots of the views from their home offices. They debated subtle nuances of how they wanted their language represented in the dataset and pushed back on some of the assumptions made by project coordinators on what constituted a distinct language as opposed to a regional dialect (see Section~\ref{sec:limitations}).  More than one contributor sat down with their grandparents to contribute to a language that spanned three generations of use. 

The realities of the conditions under which many people work and live were present every day. For example, Zoom meetings were cut short for some volunteers due to power outages in their countries or lack of access to a stable internet connection. \texttt{Burmese}, a language spoken in Myanmar, started out strong in the project with a group of 35 motivated volunteers but saw a sudden pause in contributions as civil war broke out in the country resulting in the withdrawal of the volunteers from the project \citep{Reuters2023a}. The Language Ambassador for \texttt{Armenian} also had to drop out of the project because of a conflict in that country \citep{Reuters2023}. In some countries, postal services only functioned a few days per month because of ongoing warfare, creating challenges for organizers when mailing out \aya gifts to thank committed volunteers. Ultimately, organizers were not able to send gifts to thank volunteers who participated from Somalia, Yemen and Palestine. For Somalia and Yemen, both Canada Post, DHL and Fedex where not able to support shipments. For Palestine, the cost of shipment proved to be prohibitively expensive -- with an estimated shipping cost of $294$ US dollars per t-shirt. These geo-political realities shaped both our contributors' experience as well as the progress of the project.  

Including these factors in our post-mortem analysis of the project is crucial to understanding both the motivation of people willing to volunteer for open-science projects, and also to understanding the data itself: its breadth, its provenance, its shortcomings, and its living history.

\begin{center}
\scriptsize
\begin{longtable}{lllllcc}
\toprule
ISO Code & Language & Script & Family & Subgrouping & Resources & Included \\
\midrule 
ace & Achinese & Arabic/Latin  & Austronesian   & Malayo-Polynesian & Low & \spade \\
afr & Afrikaans & Latin & Indo-European  & Germanic  & Mid & \spade \\
amh & Amharic & Ge'ez & Afro-Asiatic & Semitic & Low & {\dia} \spade \\
ara & Arabic & Arabic & Afro-Asiatic   & Semitic                  & High           & {\dia} \spade \\
aze & Azerbaijani & Arabic/Latin & Turkic & Common Turkic & Low & \spade \\ 
ban & Balinese                  & Latin         & Austronesian   & Malayo-Polynesian        & Low           & \spade \\
bbc & Toba Batak                & Latin         & Austronesian   & Malayo-Polynesian        & Low           & \spade \\
bel & Belarusian                & Cyrillic      & Indo-European  & Balto-Slavic             & Mid           & \spade \\
bem & Bemba                     & Latin         & Niger-Congo    & Atlantic-Congo           & Low           & \spade \\
ben & Bengali                   & Bengali       & Indo-European  & Indo-Aryan               & Mid           & {\dia} \spade \\
bjn & Banjar                 & Arabic/Latin  & Austronesian   & Malayo-Polynesian        & Low           & \spade \\
bul & Bulgarian                 & Cyrillic      & Indo-European  & Balto-Slavic             & Mid           & \spade \\
cat & Catalan                   & Latin         & Indo-European  & Italic                   & High           & \spade \\
ceb & Cebuano                   & Latin         & Austronesian   & Malayo-Polynesian        & Mid           & {\dia} \spade \\ 
ces & Czech & Latin & Indo-European & Balto-Slavic & High & \spade \\ 
cym & Welsh & Latin & Indo-European & Celtic & Low & \spade \\ 
dan & Danish                    & Latin         & Indo-European  & Germanic                 & Mid           & {\dia} \spade \\
deu & German                    & Latin         & Indo-European  & Germanic                 & High          & {\dia} \spade \\ 
ell & Greek                     & Greek         & Indo-European  & Graeco-Phrygian          & Mid           & {\dia} \spade \\ 
eng & English                   & Latin         & Indo-European  & Germanic                 & High          & {\dia} \spade \\ 
epo & Esperanto & Latin & Constructed & Esperantic & Low & \spade \\ 
est & Estonian & Latin & Uralic & Finnic & Med & \spade \\ 
eus & Basque                    & Latin         & Basque  & - & High & {\dia} \spade \\ 
fil & Filipino                  & Latin         & Austronesian   & Malayo-Polynesian        & Mid           & {\dia} \spade \\
fin & Finnish                   & Latin         & Uralic         & Finnic                   & Mid           & {\dia} \spade \\
fon & Fon                       & Latin         & Niger-Congo    & Atlantic-Congo           & Low           & \spade \\
fra & French                    & Latin         & Indo-European  & Italic                   & High          & {\dia} \spade \\ 
gla & Scottish Gaelic & Latin & Indo-European & Celtic & Low & \spade \\ 
gle & Irish                     & Latin         & Indo-European  & Celtic                   & Low           & {\dia} \spade \\ 
glg & Galician & Latin & Indo-European & Italic & Med & \spade  \\ 
guj & Gujarati                  & Gujarati      & Indo-European  & Indo-Aryan               & Low           & {\dia} \spade \\ 
hat & Haitian Creole            & Latin         & Indo-European  & Italic                   & Low           & {\dia} \spade \\ 
hau & Hausa                     & Latin         & Afro-Asiatic   & Chadic                   & Low           & {\dia} \spade \\ 
heb & Hebrew                    & Hebrew        & Afro-Asiatic   & Semitic                  & Mid           & \spade \\ 
hin & Hindi                     & Devanagari    & Indo-European  & Indo-Aryan               & High           & {\dia} \spade \\ 
hrv & Croatian                  & Latin         & Indo-European  & Balto-Slavic.            & High           & \spade \\
hun & Hungarian                 & Latin         & Uralic         & -  & High           & {\dia} \spade \\ 
hye & Armenian                  & Armenian      & Indo-European  &  Armenic      & Low           & \spade \\
ibo & Igbo                      & Latin         & Atlantic-Congo & Benue-Congo              & Low           & {\dia} \spade \\
ind & Indonesian                & Latin         & Austronesian   & Malayo-Polynesian        & Mid           & {\dia} \spade \\ 
isl & Icelandic & Latin & Indo-European & Germanic & Low & \spade \\ 
ita & Italian                   & Latin         & Indo-European  & Italic                   & High           & {\dia} \spade \\ 
jav & Javanese                  & Latin         & Austronesian   & Malayo-Polynesian        & Low           & {\dia} \spade \\ 
jpn & Japanese                  & Japanese      & Japonic        & Japanesic                & High          & {\dia} \spade \\ 
kan & Kannada                   & Kannada       & Dravidian      & South Dravidian          & Low           & {\dia} \spade \\
kas & Kashmiri & Arabic & Indo-European & Indo-Aryan & Low & \spade \\ 
kat & Georgian & Georgian & Kartvelian & Georgian-Zan & Mid & \spade \\ 
kau & Kanuri & Arabic/Latin & Saharan & Western Saharan & Low & \spade \\ 
kaz & Kazakh & Cyrillic & Turkic & Common Turkic & Mid & \spade \\ 
khm & Khmer & Khmer & Austroasiatic & Khmeric & Low & \spade \\ 
kin & Kinyarwanda               & Latin         & Niger-Congo    & Atlantic-Congo           & Low           & \spade \\
kir & Kyrgyz                    & Cyrillic      & Turkic         & Common Turkic            & Low           & {\dia} \spade \\
kor & Korean                    & Hangul        & Koreanic       & Korean                   & Mid           & {\dia} \spade \\ 
kur & Kurdish           & Latin         & Indo-European  & Iranian                  & Low           & {\dia} \spade \\
lao & Lao & Lao & Tai-Kadai & Kam-Tai & Low & \spade \\ 
lav & Latvian & Latin & Indo-European & Balto-Slavic & Mid & \spade \\ 
lij & Ligurian                 & Latin         & Indo-European  & Italic                   & Low           & \spade \\
lit & Lithuanian                & Latin         & Indo-European  & Balto-Slavic             & Mid           & {\dia} \spade \\
ltz & Luxembourgish & Latin & Indo-European & Germanic & Low & \spade \\ 
mad & Madurese                  & Latin         & Austronesian   & Malayo-Polynesian        & Low           & \spade \\
mal & Malayalam & Malayalam & Dravidian & South Dravidian          & Low           & {\dia} \spade \\ 
man & Manipuri & Bengali & Sino-Tibetan & Kuki-Chin-Naga & Low & \spade \\
mar & Marathi                   & Devanagari    & Indo-European  & Indo-Aryan               & Low           & {\dia} \spade \\
min & Minangkabau               & Latin         & Austronesian   & Malayo-Polynesian        & Low           & \spade \\
mkd & Macedonian & Cyrillic & Indo-European & Balto-Slavic & Low & \spade \\ 
mlg & Malagasy          & Latin         & Austronesian   & Malayo-Polynesian        & Low           & {\dia} \spade \\
mlt & Maltese & Latin & Afro-Asiatic & Semitic & Low & \spade \\ 
mon & Mongolian & Cyrillic & Mongolic-Khitan & Mongolic & Low & \spade \\ 
mri & Maori & Latin & Austronesian & Malayo-Polynesian & Low & \spade \\ 
msa & Malay            & Latin         & Austronesian   & Malayo-Polynesian        & Mid           & {\dia} \spade \\ 
mya & Burmese                   & Myanmar       & Sino-Tibetan   & Burmo-Qiangic            & Low           & {\dia} \spade \\
nep & Nepali                    & Devanagari    & Indo-European  & Indo-Aryan               & Low           & {\dia} \spade \\
nij & Ngaju & Latin & Austronesian & Malayo-Polynesian & Low & \spade \\
nld & Dutch                     & Latin         & Indo-European  & Germanic                 & High           & {\dia} \spade \\ 
nor & Norwegian                 & Latin         & Indo-European  & Germanic                 & Low           & \spade \\
nso & Northern Sotho & Latin & Atlantic-Congo & Benue-Congo & Low & {\dia} \spade \\
nya & Chichewa & Latin         & Atlantic-Congo & Benue-Congo              & Low           & {\dia} \\
pan & Punjabi                   & Gurmukhi      & Indo-European  & Indo-Aryan               & Low           & {\dia} \spade \\ 
pes & Persian           & Arabic        & Indo-European  & Iranian                  & High           & {\dia} \spade \\ 
pol & Polish                    & Latin         & Indo-European  & Balto-Slavic             & High           & {\dia} \spade \\ 
por & Portuguese                & Latin         & Indo-European  & Italic                   & High           & {\dia} \spade \\ 
pus & Pashto & Arabic & Indo-European  & Iranian & Low & {\dia} \spade \\ 
ron & Romanian & Latin & Indo-European & Italic & Mid & \spade\\ 
rus & Russian                   & Cyrillic      & Indo-European & Balto-Slavic & High & {\dia} \spade \\ 
sin & Sinhala & Sinhala & Indo-European  & Indo-Aryan               & Low           & {\dia} \spade \\
slk & Slovak                    & Latin         & Indo-European  & Balto-Slavic             & Mid           & \spade \\
slv & Slovenian                 & Latin         & Indo-European  & Balto-Slavic             & Mid           & \spade \\
smo & Samoan & Latin & Austronesian & Malayo-Polynesian & Low & \spade \\ 
sna & Shona                     & Latin         & Indo-European  & Indo-Aryan               & Low           & {\dia} \spade \\ 
snd & Sindhi                    & Arabic        & Indo-European  & Indo-Aryan               & Low           & {\dia} \spade \\ 
som & Somali                    & Latin         & Afro-Asiatic   & Cushitic                 & Low           & {\dia} \spade \\ 
sot & Southern Sotho            & Latin         & Atlantic-Congo & Benue-Congo              & Low           & \spade \\
spa & Spanish                   & Latin         & Indo-European  & Italic                   & High          & {\dia} \spade \\ 
sqi & Albanian                  & Latin         & Indo-European  & Albanian                 & Low           & {\dia} \spade \\
srp & Serbian                   & Cyrillic      & Indo-European  & Balto-Slavic             & High           & {\dia} \spade \\ 
sun & Sundanese                 & Latin         & Austronesian   & Malayo-Polynesian        & Low           & {\dia} \spade \\ 
swa & Swahili                   & Latin         & Atlantic-Congo & Benue-Congo              & Low           & {\dia} \spade \\ 
swe & Swedish                   & Latin         & Indo-European  & Germanic                 & High           & {\dia} \spade \\ 
tam & Tamil                     & Tamil         & Dravidian      & South Dravidian          & Mid           & {\dia} \spade \\ 
taq & Tamasheq & Latin/Tifinagh & Afro-Asiatic & Berber & Low & \spade \\ 
tel & Telugu                    & Telugu        & Dravidian      & South Dravidian          & Low           & {\dia} \spade \\ 
tgk & Tajik & Cyrillic & Indo-European & Iranian & Low & \spade \\ 
tha & Thai                      & Thai          & Tai-Kadai      & Kam-Tai                  & Mid           & {\dia} \spade \\
tur & Turkish                   & Latin         & Turkic         & Common Turkic            & High           & {\dia} \spade \\ 
twi & Twi                       & Latin         & Niger-Congo    & Atlantic-Congo           & Low           & \spade \\ 
ukr & Ukrainian                 & Cyrillic      & Indo-European  & Balto-Slavic             & Mid           & {\dia} \spade \\ 
urd & Urdu                      & Arabic        & Indo-European  & Indo-Aryan               & Mid           & {\dia} \spade \\ 
uzb & Uzbek & Latin & Turkic & Common Turkic & Med & \spade  \\  
vie & Vietnamese                & Latin         & Austroasiatic  & Vietic                   & High           & {\dia} \spade \\ 
wol & Wolof                     & Latin         & Atlantic-Congo & North-Central Atlantic   & Low           & {\dia} \spade \\
xho & Xhosa                     & Latin         & Atlantic-Congo & Benue-Congo              & Low           & {\dia} \spade \\
yid & Yiddish & Hebrew & Indo-European & Germanic & Low & \spade \\ 
yor & Yorùbá                    & Latin         & Atlantic-Congo & Benue-Congo              & Low           & {\dia} \spade \\ 
zho & Chinese       & Han           & Sino-Tibetan   & Sinitic                  & High          & {\dia} \spade \\
zul & Zulu                      & Latin         & Atlantic-Congo & Benue-Congo              & Low           & {\dia} \spade \\

\bottomrule
\caption{65 languages in the \aya Dataset and 114 languages in the \aya Collection, each language's corresponding script, family, subgrouping, and if it is classified as ``lower-'', ``mid-'' or ``higher''-resourced according to the taxonomy classes by \citep{joshi-etal-2020-state} (low: [0, 1, 2], mid: [3], high: [4, 5]). The language is either included in the \aya Dataset (\specialdia), \aya Collection (\specialspade), or both. Note that \texttt{Ngaju} (nij) and \texttt{Toba Batak} (bbc) are not listed in~\citep{joshi-etal-2020-state}.}
\label{tab:language_codes}
\end{longtable}
\end{center}

\section{Related Work}
\label{sec:related_work}

\subsection{Multilingual datasets}

Low-resource languages have long been a challenge in NLP, with limited data impacting task performance \citep{kunchukuttan-etal-2021-large}. To address this, researchers have explored techniques like data augmentation \citep{sennrich-etal-2016-neural,dhole2021nl}, transfer learning \citep{zoph-etal-2016-transfer}, repeating~\citep{luukkonen2023fingpt,muennighoff2023scaling}, and multilingual models \citep{10.1145/3406095, muennighoff-etal-2023-crosslingual, yong-etal-2023-bloom}, achieving promising results in areas like machine translation. Here, we focus on efforts that are centered on multilingual dataset creation.

Several works have created large-scale multilingual corpora. These are often unstructured texts, ideal for large-scale unsupervised pre-training~\citep{AbadjiOrtizSuarezRomaryetal.2021,OrtizSuarezSagotRomary2019,workshop2023bloom,scao2022language,laurenccon2022bigscience,kudugunta2023madlad, Whitehouse2023LLM}. Another group of multilingual datasets is focused on machine translation \citep{Lucia2010Dataset, fan2021beyond}. They consist of parallel texts in two or more languages, enabling models to learn the mappings between them. Ideally, machine translation datasets encompass diverse domains and language pairs, from commonly spoken languages to resource-scarce ones, promoting inclusivity and linguistic diversity. One of the most extensive collections of parallel corpora is available at the OPUS project website\footnote{\url{https://opus.nlpl.eu}} \citep{TIEDEMANN12}.
Large capacity models for language understanding may obtain strong performance on high-resource languages while greatly improving low-resource languages~\citep{Goyal2021Larger}. In \cite{Whitehouse2023LLM}, the effectiveness of LLM-powered data augmentation in cross-lingual commonsense reasoning was demonstrated. An improved performance was shown when smaller cross-lingual models were finetuned with data generated by LLMs. Some recently released datasets focus on specialized language domains such as law~\citep{Niklaus2023MultiLegalPile}, education~\citep{zhang2023m3exam}, or healthcare~\citep{wang2023clinicalgpt}.

These corpora often suffer from inadequate data quality and require extensive cleaning~\citep{abadji-etal-2022-towards,kreutzer-etal-2022-quality}. Task-specific datasets, such as XCOPA~\citep{ponti2020xcopa} or XNLI~\citep{conneau2018xnli}, are smaller in scale but offer higher quality data targeted at a specific model capability such as cross-lingual understanding and transfer learning. This type of data is crucial for evaluating and enhancing the performance of models in diverse linguistic contexts. 

No Language Left Behind \citep{nllbteam2022language} open-sourced bitext, mined bitext, and data generated using back-translation in 200+ languages specifically for text-to-text translation. While Seamless4MT \citep{barrault2023seamlessm4t} released the metadata of SeamlessAlign, an open multimodal translation dataset, there are relatively fewer works for data creation/curation in low-resource languages. \cite{cahyawijaya2023nusacrowd} introduced NusaCrowd, a standardized collection of 137 datasets covering 19 Indonesian local languages in text, speech, and image modalities.
Our work differs from previous datasets as we create a large-scale instruction-tuning dataset spanning hundreds of different tasks, yet retain high-quality by involving human annotation and rigorous quality control across the entire data creation process.

\subsection{Instruction-tuning datasets}

Instruction-tuning datasets are collections of human-curated instructions and response pairs, templatized NLP tasks, or synthetic instructions generated by a language model. 
There are a growing number of NLP meta-datasets such as Natural instructions \citep{naturalinstructions}, SuperNatural Instructions\citep{wang2022super}, Flan 2021~\citep{wei2022finetuned}, Flan 2022 \citep{longpre2023flan}, Public Pool of Prompts (P3)~\citep{sanh2022multitask}, Unnatural Instructions~\citep{honovich-etal-2023-unnatural}, OPT-IML \citep{iyer2022opt}, inter alia \citep{khashabi2020unifiedqa,ye2021crossfit,min2021metaicl} that collate numerous instruction finetuned datasets together. Some work focuses on specific applications such as dialogue~\citep{kopf2023openassistant}, structured knowledge grounding~\citep{xie2022unifiedskg}, or chain-of-thought reasoning~\citep{wei2022chain,kim2023cot}. 
Manual efforts include Open Assistant \citep{kopf2023openassistant} crowd-sourcing volunteers who wrote both instructions and responses, Databricks employees creating 15k examples in Dolly \citep{DatabricksBlog2023DollyV2}, and LIMA \citep{zhou2023lima} which is a collection of 1,000 author-curated IFT examples.

Synthetic instruction-tuning datasets comprise instructions sampled from a language model, such as the Self-Instruct dataset \citep{wang2022self} generated by GPT-3 \citep{brown2020language}, the Alpaca dataset \citep{alpaca} generated by GPT-3.5, and the Guanaco dataset \citep{joseph_cheung_2023}. Increasingly, the synthetic generation of instruction-finetuned datasets is more sophisticated. \citep{xu2023wizardlm} propose a novel Evol-Instruct framework to obtain complex
and difficult instructions gradually. \citep{luo2023wizardcoder} and \citep{gunasekar2023textbooks} further expand this idea to promote reasoning, code generation, and algorithmic skills. InstructionWild \citep{instructionwild} and ShareGPT\footnote{\url{https://sharegpt.com/}} are collections of user-shared conversations with ChatGPT. 

\subsection{Multilingual Instruction-Tuning Datasets} 

Despite ever-larger collections of IFT datasets, prior work has been largely English-centric. Most approaches to extend instruction finetuned datasets outside of English have relied on \textbf{1)} translating English datasets into other languages \citep{holmstrom-doostmohammadi-2023-making, li2023bactrian, winata-etal-2023-nusax}, \textbf{2)} template based dataset creation \citep{yu2023large, gupta2023targen} or \textbf{3)}  human curating instruction datasets in languages outside of English \citep{muennighoff-etal-2023-crosslingual, li2023m3it, wang2022supernaturalinstructions}. 
There have been some notable exceptions with large proportions of non-English data~\citep{joseph_cheung_2023, kopf2023openassistant, lai2023okapi, li2023bactrian, longpre2023flan, muennighoff2023octopack, muennighoff-etal-2023-crosslingual, zhuo2024astraios, oig2023}. 

\textbf{Template-Based Datasets.} The most relevant effort is recent work by \citep{muennighoff-etal-2023-crosslingual} releasing Crosslingual Public Pool of Prompts (xP3). xP3 expands the P3 taxonomy and adds 28 new multilingual datasets. However, their datasets usually use the same template in different languages, thus limiting task diversity. For example, a random batch from their dataset may include the same sample in different languages multiple times. Their xP3 corpus has task instructions exclusively in English. In \citep{muennighoff-etal-2023-crosslingual}, the experiments with matching the task instruction to the respective language of the sample via machine translation (xP3mt) showed slightly improved performance for non-English task instructions at inference. Our work is distinct in that our human-curated constructed dataset is unique for each of the 65 languages. Such diversity has been emphasized as a key ingredient for instruction tuning~\citep{longpre2023flan}. Further, we create non-English task instructions via human annotators, ensuring these are of high-quality, which is another pillar of a good performance~\citep{zhou2023lima}.

\textbf{Machine Translated Datasets.} Machine-translated prompts often lack variability and the cultural nuance inherent in natively written text. However, they are still useful for expanding the language coverage of the training data and can help bridge the resource gap for languages with limited training data \citep{urbizu-etal-2023-enough, lin-etal-2022-shot}. They can also adapt already-trained instruction-tuned language models to follow instructions in new languages \citep{yong-etal-2023-bloom}. Furthermore, LLMs trained on designed prompts have also been shown to be successful at tasks like EAE (Event Argument Extraction) from multilingual data in a zero-shot setup \citep{huang2022multilingual}. \citep{zhang2023chinese} constructed high-quality Chinese instructions from existing English instruction datasets. They first translated the English instructions into Chinese and then used a human verification process to determine whether these translations were usable; the verified dataset set consists of around 200k Chinese instruction-tuning samples. \citep{li2023bactrian} constructed instruction data for 52 popular languages using Google Translate to translate English prompts and completions from Alpaca \citep{alpaca} (52K) and Dolly \citep{DatabricksBlog2023DollyV2} (15K) dataset, then used this data to finetune LLaMA \citep{touvron2023llama} using the LoRA \citep{hu2021lora} technology. \citep{zhang2023bayling} prompted LLMs to translate a task request, which was overlaid with the more granular user-based corrects. This process naturally connects different languages as well as human preferences with LLMs, leveraging LLaMA \citep{touvron2023llama} for foundational support and employing automatic construction of interactive translation instructions for instructional tuning, thereby enhancing the model's multilingual capability and alignment with diverse linguistic needs.

\textbf{Human-Curated Multilingual Examples.} Most relevant to our work on the \aya dataset are other datasets that have been curated by humans, often in English. Databricks collected a 15k instruction dataset \texttt{databricks-dolly-15k} by relying on its employees as annotators \citep{DatabricksBlog2023DollyV2}.  Annotators were instructed to curate prompt / response pairs in each of eight different instruction categories. \citep{köpf2023openassistant} released the OpenAssistant corpus with over 10,000 dialogues from more than 13,500 international annotators. While this dataset contains multilingual annotations, this was not an explicit goal of the initiative. In contrast to our corpus which only has 2.05\% contributions in English, 42.8\% of the OpenAssistant project remains in English~\citep{kopf2023openassistant}. 

\subsection{Participatory Research in Machine Learning}

\begin{quote}
    \textit{If you want to go fast go alone; if you want to go far, go together.} \textbf{--- African Proverb}
\end{quote}

Prior participatory research initiatives have centered around regions or specific tasks like translation or character recognition. For example, \citep{Clanuwat2018DeepLF} tackles the problem of reading and understanding \textit{Kuzushiji}, a cursive style of \texttt{Japanese} writing no longer in common use. Another example of culturally diverse data collection is \citep{liu-etal-2021-visually}, which recruited native speakers from five languages (\texttt{Indonesian, Swahili, Tamil, Turkish}, and \texttt{Mandarin Chinese}) that are typologically, genealogically, and geographically diverse, to provide images of concepts that are representative of their cultures. Then, they recruited native-speaking professional linguists to write captions for these images. However, this dataset is small (less than 8,000 data points) and thus limited to evaluation only. It is worth noting that these works are solely focused on the image domain, unlike our work, which concentrates on text.

More relevant to our work are participatory data creation initiatives focused on NLP. \citep{guevara-rukoz-etal-2020-crowdsourcing} presents a study focusing on the creation of a crowd-sourced corpus for Latin American Spanish dialects to address the scarcity of resources for these languages. \citep{forall-nekoto-etal-2020-participatory} focuses on the task of Machine Translation (MT), and curates a dataset in 30 under-represented African languages according to a participatory research framework. Our work is very much in the spirit of these prior efforts, with differences in terms of global rather than regional focus. In contrast to these works, which have a specific regional focus, \aya collaborators came from multiple continents covering a diverse range of languages.

Several works have explored the organizational structures required to facilitate the development of research communities around under-represented languages. \citep{siminyu2021} details work on the AI4D - African Language Program, which aimed to enhance language resources for African languages. The outcome included creating over nine open-source African language datasets and establishing baseline models, demonstrating the program's significant impact on language technology for African languages. \citep{Azunre2021NLPFG} describes the establishment of NLP Ghana, with its collaborative open-source community. \citep{strassel-tracey-2016-lorelei} discusses the challenges of developing resources for low-resource languages under the LORELEI (Low Resource Languages for Emergent Incidents) program. They focus on the pressing need for digital resources in these languages, particularly in critical situations such as mitigating the effects of natural disasters. 

Open science community initiatives like \aya yield significant advancements in language modeling. Related efforts (in terms of compute and resources required) can be found in the BigScience Workshop \citep{akiki2022bigscience}, which began in 2021. The BigScience project was initiated to address the limitations in LLM development, emphasizing open science and inclusive collaboration. Leveraging open science principles, it united a global network of researchers working to collaboratively and ethically enhance machine learning. Their work culminated in key developments like the BLOOM model \citep{workshop2023bloom} and ROOTS corpus \citep{laurenccon2022bigscience}. These achievements underscore the value of community-driven, ethical, and diverse research programs for large-scale language technologies. Following Big Science, there have been other recent efforts on open science in language modeling~\citep{olmo20247b,dolma}.

\section{Limitations of our work}
\label{sec:limitations}

\begin{enumerate}

\item \textbf {Language and dialect coverage}: The \aya Dataset and \aya Collection cover 65 and 114 languages respectively---significantly more than existing multilingual datasets. However, this is still only a tiny fraction of the world's linguistic diversity. Of the world's approximately 7,000 languages, only half of them are captured in any sort of written form \citep{ADDA20168}. Of this half, only a few hundred are included on the internet in machine readable corpora \citep{ADDA20168}. This means that 93\% of the world’s languages are still not being used to train LLMs. It is also notoriously difficult to determine the dividing line between different languages and different dialects of the same language \citep{vanrooy}. Geo-cultural variation within a language often gives rise to new dialects or creoles over time \citep{zampieri2020natural, wolfram1997issues, brown2020language, lent-etal-2022-creole, blaschke-etal-2023-survey} and, as such, dialects can serve an important function in establishing and maintaining cultural identity \citep{falck2012dialects}. Many different dialects that are generally recognized as belonging to a single parent language are not represented in the dataset. For example, in the case of \texttt{Malay}, one of the largest Southeast Asian languages in the dataset, there are no contributions for regional dialects that are widely spoken in certain states of Malaysia. Contributions by volunteers who wished to self-identify as speaking a particular dialect were tagged as such in the data to allow for limited analysis of the use of regional dialects in annotations. Lastly, socio-linguistic data show that multilingual speakers often `code-switch' between languages or dialects depending on context \citep{myers2017code}, but in this project, we kept the languages isolated to make them easier to classify and to be used downstream for language-specific applications. \aya also does not cover programming languages. There has been prior work on covering diverse programming languages~\citep{li2023starcoder,allal2023santacoder} and we leave further explorations in this direction to future work.

\item \textbf {Uneven distribution of contributions}: As explored in Section~\ref{sec:anskew}, despite the large number of participants, the activity of annotators was skewed, with a `long tail' of annotators only contributing one or two annotations. Relatively few contributors accounted for the most annotations (see~Figure~\ref{fig:skew} - bottom).
Similarly, there is a huge gap between languages with the highest number of contributions and ones with the lowest number of contributions. Consequently, this suggests potential unevenness in dataset distributions across different languages and a lack of annotator diversity within some languages dominated by one or two frequent contributors.

\item \textbf {Cultural or personal bias}: Another limitation is the presence of annotations with particular cultural biases. Some languages in our dataset have limited representation, with only a few annotators responsible for annotating the bulk of their dataset. This might mean that data for a particular language is dominated by annotations that represent the opinions or priorities of a particular contributor or could represent a narrow selection of cultural viewpoints. For example, annotations in French might contain many examples about the history of France, its food, songs, and other cultural practices, but not contain much information about the cultural heritage of French-speaking communities in Québec, Togo, or Senegal \citep{doi:10.1146/annurev-anthro-092611-145804}. This bias is particularly problematic given the skewed distribution of the most active annotators. There is also a potential bias in the availability of particular kinds of content. For example, it is easier to find online text from news sites for many African languages than it is to find text from other domains. Accordingly, these datasets will be skewed towards the grammar and lexicon used in news reports instead of the kind of natural language people use in everyday life \citep{hovy2021five}. 

\item \textbf{Gendered pronouns}: Many of the languages in the \aya Dataset only contain pronouns that are explicitly gendered (e.g., Arabic) or that lack gender-neutral third-person pronouns for gender-neutral reference (e.g. Estonian). This means that in responding to prompts that might not specify a gender, care needs to be taken to ensure that responses remain neutral as to the gender of any assumed participants \citep{ghosh2023chatgpt}. For example, if a response requires reference to \textit{``a teacher''} in French, the annotator would need to include references to both \textit{``un/e enseignant/e''}. While care was taken to ensure neutral responses for new annotations, gendered annotations in existing datasets might not have been flagged, as they are not, strictly speaking, incorrect. Instead, they merely presuppose a gendered reading where one might not be implied \citep{hardmeier2018pronoun}. 

\item \textbf{Formality distinctions}: Many of the languages in the \aya Dataset also require the speaker or annotator to make situational choices as to the formality of the pronoun used in response to a particular prompt. Languages such as \texttt{Japanese, Persian, Indonesian, Javanese, Yoruba, French, Spanish,} and \texttt{German} include different levels of honorifics that are used in formal or informal settings, or used between community members who differ in status (determined by a variety of factors such as age, profession, seniority, or ethnicity) \citep{BrownGilman+1968+252+275}. In \texttt{Yoruba}, for example, the pronoun that roughly translates as \textit{"they"} can either be used as a singular honorific or as a third-person plural pronoun \citep{Yusuf2022}.
We deferred to the individual annotators in crafting their responses, allowing them to rely on the norms of their particular speech community to determine how to respond. Often, these decisions hinged on the content being discussed, or on how formally the prompt was crafted in the original data set. When in doubt, annotators were asked to imagine what kind of `voice' they would expect an LLM to have when answering a given prompt \citep{Wilson2023}. 

This means our released dataset contains many languages that have varying levels of standardization and differing style guidelines. Standardization is often deeply intertwined with power and identity, and the manner of speech may be connected to aspects of identity like age, education level, tribal affiliation, and religion. The lack of standardization is also largely due to regional and cultural differences across the same language, exemplified by \texttt{Portuguese} in the dataset: European Portuguese diverges from Brazilian Portuguese not only in formality but also in grammar, spelling, and vocabulary. Often, standards are projected by others to ensure adherence to cultural values~\citep{bourdieu1987makes,de2015big,haugen1959planning,rickford2012african}.

\item \textbf{Toxic or offensive speech}: The \aya Annotation Platform did not contain specific flags for toxic, harmful, or offensive speech, so it is possible that malicious users could submit unsafe data. We believe this is of relatively low risk because of the high rate of human-verified annotations and peer-review, making it unlikely that toxic prompts or completions made it into the final dataset.  However, there is no guarantee that every entry was audited. While data poisoning has rarely been observed as a viable threat in practice, it has been demonstrated to be of concern for instruction-tuning with very few examples \citep{xu2023instructions,wan2023poisoning} and for pre-training under realistic conditions \citep{carlini2023poisoning}. During the eight months of crowd-sourced annotating, there were no reported cases of hateful or toxic speech in the existing datasets nor were there any instances of offensive speech reported in the peer-reviewing phase of new annotations. 

We also note that data that might be offensive to one annotator might not be offensive to another, for instance, the completion of a prompt that asks for a definition of the word ``woke'' \citep{Castaldo2023a}. Prompts written on partisan political topics, or the inclusion of political advertisements or campaign messages could cause offense depending on the political proclivities of the annotator. In short, we tried to mitigate offensive speech by relying heavily on human annotation and peer review, but there is no guarantee that all such data points were removed from the corpus. 

\item \textbf{Accounting for mislabeled data}: The \aya Annotation Platform did not contain any components that enabled re-labeling the assigned language of annotations. This may result in prompts and completions that appear under a particular language, but were submitted incorrectly and would need to be re-categorized into a different language.
Additionally, while we trusted annotators were able to follow directions and had a high rate of manual auditing, some examples likely made it into the \aya Dataset that were not in instruction-style format or were free-form texts.



\end{enumerate}

\section{Conclusion}
\label{sec:conclusion}
Open participatory research continues to be under-resourced and undervalued, particularly when that work focuses on data creation ~\citep{Sambasivan2021datawork}. \aya involved participants from many different countries, different ages, and different levels of familiarity with the field of natural language processing. We see continued opportunity for computational linguists and machine-learning engineers to collaborate with social scientists such as sociolinguists, anthropologists, sociologists, and media studies scholars. As new norms in open science emerge \citep{KRISHNA202061,BowserOS}, collaborations like these can help ensure that projects in NLP are motivated by an understanding of what language means to the people who use it every day.

With \aya, we hope to change the way data is created for multilingual NLP research. In line with this view, we release the \colorbox{ayad}{\aya Dataset} which is the first human-curated open-source, multilingual instruction-style dataset consisting of 204,114 prompt-completion pairs covering 65 languages.
This dataset was built with the help of our open-science community of 2,997 collaborators from 119 countries over a period of eight months.

We also release the \colorbox{ayac}{\aya Collection}, which consists of 44 instruction-style datasets. These were prepared by transforming existing NLP datasets into prompt-completion pairs that can be leveraged for instruction tuning.
Furthermore, we translate several high-quality datasets into 101 languages, thereby expanding coverage, particularly for many low-resource languages.
This collection consists of 513M prompt and completion pairs covering 114 languages in total and is the largest multilingual instruction-finetuning collection today.
Additionally, we release \colorbox{ayae}{\aya Evaluation Suite}, consisting of human-curated examples in 13 languages and translation of carefully selected prompts in 101 languages.
Finally, we are also open-sourcing the \colorbox{ayaui}{\aya Annotation Platform} so that communities can continue to use the platform to support the process of multilingual data collection. We hope these communities continue to grow and develop, and to connect speakers of low-resource languages around the world.

\bibliography{main,anthology}

\begin{thebibliography}{376}
\providecommand{\natexlab}[1]{#1}
\providecommand{\url}[1]{\texttt{#1}}
\expandafter\ifx\csname urlstyle\endcsname\relax
  \providecommand{\doi}[1]{doi: #1}\else
  \providecommand{\doi}{doi: \begingroup \urlstyle{rm}\Url}\fi

\bibitem[Abadji et~al.(2021)Abadji, Su{\'a}rez, Romary, and Sagot]{AbadjiOrtizSuarezRomaryetal.2021}
Julien Abadji, Pedro Javier~Ortiz Su{\'a}rez, Laurent Romary, and Beno{\^i}t Sagot.
\newblock Ungoliant: An optimized pipeline for the generation of a very large-scale multilingual web corpus.
\newblock In Harald L{\"u}ngen, Marc Kupietz, Piotr Bański, Adrien Barbaresi, Simon Clematide, and Ines Pisetta (eds.), \emph{CMLC 2021-9th Workshop on Challenges in the Management of Large Corpora}, Proceedings of the Workshop on Challenges in the Management of Large Corpora (CMLC-9) 2021. Limerick, 12 July 2021 (Online-Event), pp.\  1 -- 9, Mannheim, 2021. Leibniz-Institut f{\"u}r Deutsche Sprache.
\newblock \doi{10.14618/ids-pub-10468}.
\newblock URL \url{https://nbn-resolving.org/urn:nbn:de:bsz:mh39-104688}.

\bibitem[Abadji et~al.(2022)Abadji, Ortiz~Suarez, Romary, and Sagot]{abadji-etal-2022-towards}
Julien Abadji, Pedro Ortiz~Suarez, Laurent Romary, and Beno{\^\i}t Sagot.
\newblock Towards a cleaner document-oriented multilingual crawled corpus.
\newblock In \emph{Proceedings of the Thirteenth Language Resources and Evaluation Conference}, pp.\  4344--4355, Marseille, France, June 2022. European Language Resources Association.
\newblock URL \url{https://aclanthology.org/2022.lrec-1.463}.

\bibitem[Abedissa et~al.(2023)Abedissa, Usbeck, and Assabie]{abedissa2023amqa}
Tilahun Abedissa, Ricardo Usbeck, and Yaregal Assabie.
\newblock {AmQA: Amharic Question Answering Dataset}.
\newblock \emph{arXiv preprint arXiv:2303.03290}, 2023.

\bibitem[Adda et~al.(2016)Adda, Stüker, Adda-Decker, Ambouroue, Besacier, Blachon, Bonneau-Maynard, Godard, Hamlaoui, Idiatov, Kouarata, Lamel, Makasso, Rialland, {Van de Velde}, Yvon, and Zerbian]{ADDA20168}
Gilles Adda, Sebastian Stüker, Martine Adda-Decker, Odette Ambouroue, Laurent Besacier, David Blachon, Hélène Bonneau-Maynard, Pierre Godard, Fatima Hamlaoui, Dmitry Idiatov, Guy-Noël Kouarata, Lori Lamel, Emmanuel-Moselly Makasso, Annie Rialland, Mark {Van de Velde}, François Yvon, and Sabine Zerbian.
\newblock Breaking the unwritten language barrier: The bulb project.
\newblock \emph{Procedia Computer Science}, 81:\penalty0 8--14, 2016.
\newblock ISSN 1877-0509.
\newblock \doi{https://doi.org/10.1016/j.procs.2016.04.023}.
\newblock URL \url{https://www.sciencedirect.com/science/article/pii/S1877050916300370}.
\newblock SLTU-2016 5th Workshop on Spoken Language Technologies for Under-resourced languages 09-12 May 2016 Yogyakarta, Indonesia.

\bibitem[Adelani et~al.(2023)Adelani, Masiak, Azime, Alabi, Tonja, Mwase, Ogundepo, Dossou, Oladipo, Nixdorf, Emezue, sana~al azzawi, Sibanda, David, Ndolela, Mukiibi, Ajayi, Moteu, Odhiambo, Owodunni, Obiefuna, Mohamed, Muhammad, Ababu, Salahudeen, Yigezu, Gwadabe, Abdulmumin, Taye, Awoyomi, Shode, Adelani, Abdulganiyu, Omotayo, Adeeko, Afolabi, Aremu, Samuel, Siro, Kimotho, Ogbu, Mbonu, Chukwuneke, Fanijo, Ojo, Awosan, Kebede, Sakayo, Nyatsine, Sidume, Yousuf, Oduwole, Tshinu, Kimanuka, Diko, Nxakama, Nigusse, Johar, Mohamed, Hassan, Mehamed, Ngabire, Jules, Ssenkungu, and Stenetorp]{adelani2023masakhanews}
David~Ifeoluwa Adelani, Marek Masiak, Israel~Abebe Azime, Jesujoba Alabi, Atnafu~Lambebo Tonja, Christine Mwase, Odunayo Ogundepo, Bonaventure F.~P. Dossou, Akintunde Oladipo, Doreen Nixdorf, Chris~Chinenye Emezue, sana~al azzawi, Blessing Sibanda, Davis David, Lolwethu Ndolela, Jonathan Mukiibi, Tunde Ajayi, Tatiana Moteu, Brian Odhiambo, Abraham Owodunni, Nnaemeka Obiefuna, Muhidin Mohamed, Shamsuddeen~Hassan Muhammad, Teshome~Mulugeta Ababu, Saheed~Abdullahi Salahudeen, Mesay~Gemeda Yigezu, Tajuddeen Gwadabe, Idris Abdulmumin, Mahlet Taye, Oluwabusayo Awoyomi, Iyanuoluwa Shode, Tolulope Adelani, Habiba Abdulganiyu, Abdul-Hakeem Omotayo, Adetola Adeeko, Abeeb Afolabi, Anuoluwapo Aremu, Olanrewaju Samuel, Clemencia Siro, Wangari Kimotho, Onyekachi Ogbu, Chinedu Mbonu, Chiamaka Chukwuneke, Samuel Fanijo, Jessica Ojo, Oyinkansola Awosan, Tadesse Kebede, Toadoum~Sari Sakayo, Pamela Nyatsine, Freedmore Sidume, Oreen Yousuf, Mardiyyah Oduwole, Tshinu Tshinu, Ussen Kimanuka, Thina Diko, Siyanda Nxakama, Sinodos
  Nigusse, Abdulmejid Johar, Shafie Mohamed, Fuad~Mire Hassan, Moges~Ahmed Mehamed, Evrard Ngabire, Jules Jules, Ivan Ssenkungu, and Pontus Stenetorp.
\newblock {MasakhaNEWS: News Topic Classification for African Languages}, 2023.

\bibitem[Agha(2006)]{Agha2006}
Asif Agha.
\newblock \emph{Language and Social Relations}.
\newblock Studies in the Social and Cultural Foundations of Language. Cambridge University Press, 2006.

\bibitem[Aharoni et~al.(2019)Aharoni, Johnson, and Firat]{aharoni2019massively}
Roee Aharoni, Melvin Johnson, and Orhan Firat.
\newblock Massively multilingual neural machine translation, 2019.

\bibitem[Ahia et~al.(2021)Ahia, Kreutzer, and Hooker]{ahia-etal-2021-low-resource}
Orevaoghene Ahia, Julia Kreutzer, and Sara Hooker.
\newblock The low-resource double bind: An empirical study of pruning for low-resource machine translation.
\newblock In \emph{Findings of the Association for Computational Linguistics: EMNLP 2021}, pp.\  3316--3333, Punta Cana, Dominican Republic, November 2021. Association for Computational Linguistics.
\newblock \doi{10.18653/v1/2021.findings-emnlp.282}.
\newblock URL \url{https://aclanthology.org/2021.findings-emnlp.282}.

\bibitem[AhmadMustafa(2023{\natexlab{a}})]{Urdu-Instruct-News-Article-Generation}
AhmadMustafa.
\newblock {Urdu-Instruct-News-Article-Generation}.
\newblock \url{https://huggingface.co/datasets/AhmadMustafa/Urdu-Instruct-News-Article-Generation}, 2023{\natexlab{a}}.
\newblock Accessed: 2023-11-28.

\bibitem[AhmadMustafa(2023{\natexlab{b}})]{Urdu-Instruct-News-Category-Classification}
AhmadMustafa.
\newblock {Urdu-Instruct-News-Category-Classification}.
\newblock \url{https://huggingface.co/datasets/AhmadMustafa/Urdu-Instruct-News-Category-Classification}, 2023{\natexlab{b}}.
\newblock Accessed: 2023-11-28.

\bibitem[AhmadMustafa(2023{\natexlab{c}})]{Urdu-Instruct-News-Headline-Generation}
AhmadMustafa.
\newblock {Urdu-Instruct-News-Headline-Generation}.
\newblock \url{https://huggingface.co/datasets/AhmadMustafa/Urdu-Instruct-News-Headline-Generation}, 2023{\natexlab{c}}.
\newblock Accessed: 2023-11-28.

\bibitem[{AI Tamil Nadu}(2023{\natexlab{a}})]{aitamilnaduTamilStories}
{AI Tamil Nadu}.
\newblock Tamil stories.
\newblock \url{https://huggingface.co/datasets/aitamilnadu/tamil_stories}, 2023{\natexlab{a}}.
\newblock Accessed: 2023-12-15.

\bibitem[{AI Tamil Nadu}(2023{\natexlab{b}})]{aitamilnaduThirukkuralInstuct}
{AI Tamil Nadu}.
\newblock {Thirukkural Instruct}.
\newblock \url{https://huggingface.co/datasets/aitamilnadu/thirukkural_instruct}, 2023{\natexlab{b}}.
\newblock Accessed: 2023-11-30.

\bibitem[Akiki et~al.(2022)Akiki, Pistilli, Mieskes, Gallé, Wolf, Ilić, and Jernite]{akiki2022bigscience}
Christopher Akiki, Giada Pistilli, Margot Mieskes, Matthias Gallé, Thomas Wolf, Suzana Ilić, and Yacine Jernite.
\newblock Bigscience: A case study in the social construction of a multilingual large language model, 2022.

\bibitem[Allal et~al.(2023)Allal, Li, Kocetkov, Mou, Akiki, Ferrandis, Muennighoff, Mishra, Gu, Dey, et~al.]{allal2023santacoder}
Loubna~Ben Allal, Raymond Li, Denis Kocetkov, Chenghao Mou, Christopher Akiki, Carlos~Munoz Ferrandis, Niklas Muennighoff, Mayank Mishra, Alex Gu, Manan Dey, et~al.
\newblock Santacoder: don't reach for the stars!
\newblock \emph{arXiv preprint arXiv:2301.03988}, 2023.

\bibitem[AlShikh et~al.(2023)AlShikh, Daaboul, Goddard, Imel, Kamble, Kulkarni, and Russak]{alshikh2023becoming}
Waseem AlShikh, Manhal Daaboul, Kirk Goddard, Brock Imel, Kiran Kamble, Parikshith Kulkarni, and Melisa Russak.
\newblock Becoming self-instruct: introducing early stopping criteria for minimal instruct tuning.
\newblock \emph{arXiv preprint arXiv:2307.03692}, 2023.

\bibitem[Amirkhani et~al.(2023)Amirkhani, AzariJafari, Faridan-Jahromi, Kouhkan, Pourjafari, and Amirak]{amirkhani2023farstail}
Hossein Amirkhani, Mohammad AzariJafari, Soroush Faridan-Jahromi, Zeinab Kouhkan, Zohreh Pourjafari, and Azadeh Amirak.
\newblock Farstail: {A} {P}ersian natural language inference dataset.
\newblock \emph{Soft Computing}, 2023.
\newblock \doi{10.1007/s00500-023-08959-3}.

\bibitem[Andress et~al.(2020)Andress, Hall, Davis, Levine, Cripps, and Guinn]{andressaddressing}
Lauri Andress, Tristen Hall, Sheila Davis, Judith Levine, Kimberly Cripps, and Dominique Guinn.
\newblock Addressing power dynamics in community-engaged research partnerships.
\newblock \emph{Journal of Patient-Reported Outcomes 4: 24}, 2020.

\bibitem[Arefeen et~al.(2023)Arefeen, Debnath, and Chakradhar]{arefeen2023leancontext}
Md~Adnan Arefeen, Biplob Debnath, and Srimat Chakradhar.
\newblock Leancontext: Cost-efficient domain-specific question answering using llms.
\newblock \emph{arXiv preprint arXiv:2309.00841}, 2023.

\bibitem[Artetxe et~al.(2019)Artetxe, Ruder, and Yogatama]{Artetxe:etal:2019}
Mikel Artetxe, Sebastian Ruder, and Dani Yogatama.
\newblock On the cross-lingual transferability of monolingual representations.
\newblock \emph{CoRR}, abs/1910.11856, 2019.

\bibitem[Avle et~al.(2018)Avle, Quartey, and Hutchful]{Avle2018ResearchOM}
Seyram Avle, Emmanuel Quartey, and David Hutchful.
\newblock Research on mobile phone data in the global south.
\newblock \emph{The Oxford Handbook of Networked Communication}, 2018.
\newblock URL \url{https://api.semanticscholar.org/CorpusID:168167342}.

\bibitem[Awati \& Shum(2015)Awati and Shum]{Awati2015}
Kailash Awati and Simon~Buckingham Shum.
\newblock Big data metaphors we live by.
\newblock \emph{Towards Data Science}, 2015.
\newblock URL \url{https://towardsdatascience.com/big-data-metaphors-we-live-by-98d3fa44ebf8}.

\bibitem[Azunre et~al.(2021)Azunre, Osei, Addo, Adu-Gyamfi, Moore, Adabankah, Opoku, Asare-Nyarko, Nyarko, Amoaba, Appiah, Akwerh, Lawson, Budu, Debrah, Boateng, Ofori, Buabeng-Munkoh, Adjei, Ampomah, Otoo., Borkor, Mensah, Mensah, Marcel, Amponsah, and Hayfron-Acquah]{Azunre2021NLPFG}
Paul Azunre, Salomey Osei, Salomey~Afua Addo, Lawrence~Asamoah Adu-Gyamfi, Stephen Moore, Bernard Adabankah, Bernard Opoku, Clara Asare-Nyarko, Samuel Nyarko, Cynthia Amoaba, Esther~Dansoa Appiah, Felix Akwerh, Richard Nii~Lante Lawson, Joel Budu, Emmanuel Debrah, Nana~Adowaa Boateng, Wisdom Ofori, Edwin Buabeng-Munkoh, Franklin Adjei, Isaac. K.~E. Ampomah, Joseph Otoo., Reindorf~Nartey Borkor, Standylove~Birago Mensah, Lucien Mensah, Mark~Amoako Marcel, Anokye~Acheampong Amponsah, and James~Ben Hayfron-Acquah.
\newblock Nlp for ghanaian languages.
\newblock \emph{ArXiv}, abs/2103.15475, 2021.
\newblock URL \url{https://api.semanticscholar.org/CorpusID:232404908}.

\bibitem[Bach et~al.(2022)Bach, Sanh, Yong, Webson, Raffel, Nayak, Sharma, Kim, Bari, Fevry, et~al.]{bach2022promptsource}
Stephen~H Bach, Victor Sanh, Zheng-Xin Yong, Albert Webson, Colin Raffel, Nihal~V Nayak, Abheesht Sharma, Taewoon Kim, M~Saiful Bari, Thibault Fevry, et~al.
\newblock Promptsource: An integrated development environment and repository for natural language prompts.
\newblock \emph{arXiv preprint arXiv:2202.01279}, 2022.

\bibitem[Barrault et~al.(2023)Barrault, Chung, Meglioli, Dale, Dong, Duquenne, Elsahar, Gong, Heffernan, Hoffman, et~al.]{barrault2023seamlessm4t}
Lo{\"\i}c Barrault, Yu-An Chung, Mariano~Cora Meglioli, David Dale, Ning Dong, Paul-Ambroise Duquenne, Hady Elsahar, Hongyu Gong, Kevin Heffernan, John Hoffman, et~al.
\newblock Seamlessm4t-massively multilingual \& multimodal machine translation.
\newblock \emph{arXiv preprint arXiv:2308.11596}, 2023.

\bibitem[Bartolo et~al.(2020)Bartolo, Roberts, Welbl, Riedel, and Stenetorp]{bartolo2020beat}
Max Bartolo, Alastair Roberts, Johannes Welbl, Sebastian Riedel, and Pontus Stenetorp.
\newblock Beat the ai: Investigating adversarial human annotation for reading comprehension.
\newblock \emph{Transactions of the Association for Computational Linguistics}, 8:\penalty0 662--678, 2020.
\newblock \doi{10.1162/tacl\_a\_00338}.
\newblock URL \url{https://doi.org/10.1162/tacl_a_00338}.

\bibitem[Bastanfard et~al.(2023)Bastanfard, Shahabipour, and Amirkhani]{Bastanfard2023}
Azam Bastanfard, Mohammad Shahabipour, and Dariush Amirkhani.
\newblock Crowdsourcing of labeling image objects: an online gamification application for data collection.
\newblock \emph{Multimedia Tools and Applications}, Aug 2023.
\newblock ISSN 1573-7721.
\newblock \doi{10.1007/s11042-023-16325-6}.
\newblock URL \url{https://doi.org/10.1007/s11042-023-16325-6}.

\bibitem[Beck et~al.(2022)Beck, Bergenholtz, Bogers, Brasseur, Conradsen, Marco, Distel, Dobusch, Dörler, Effert, Fecher, Filiou, Frederiksen, Gillier, Grimpe, Gruber, Haeussler, Heigl, Hoisl, Hyslop, Kokshagina, LaFlamme, Lawson, Lifshitz-Assaf, Lukas, Nordberg, Norn, Poetz, Ponti, Pruschak, Priego, Radziwon, Rafner, Romanova, Ruser, Sauermann, Shah, Sherson, Suess-Reyes, Tucci, Tuertscher, Vedel, Velden, Verganti, Wareham, Wiggins, and Xu]{BeckS2022}
Susanne Beck, Carsten Bergenholtz, Marcel Bogers, Tiare-Maria Brasseur, Marie~Louise Conradsen, Diletta~Di Marco, Andreas~P. Distel, Leonhard Dobusch, Daniel Dörler, Agnes Effert, Benedikt Fecher, Despoina Filiou, Lars Frederiksen, Thomas Gillier, Christoph Grimpe, Marc Gruber, Carolin Haeussler, Florian Heigl, Karin Hoisl, Katie Hyslop, Olga Kokshagina, Marcel LaFlamme, Cornelia Lawson, Hila Lifshitz-Assaf, Wolfgang Lukas, Markus Nordberg, Maria~Theresa Norn, Marion Poetz, Marisa Ponti, Gernot Pruschak, Laia~Pujol Priego, Agnieszka Radziwon, Janet Rafner, Gergana Romanova, Alexander Ruser, Henry Sauermann, Sonali~K. Shah, Jacob~F. Sherson, Julia Suess-Reyes, Christopher~L. Tucci, Philipp Tuertscher, Jane~Bjørn Vedel, Theresa Velden, Roberto Verganti, Jonathan Wareham, Andrea Wiggins, and Sunny~Mosangzi Xu.
\newblock The open innovation in science research field: a collaborative conceptualisation approach.
\newblock \emph{Industry and Innovation}, 29\penalty0 (2):\penalty0 136--185, 2022.
\newblock \doi{10.1080/13662716.2020.1792274}.
\newblock URL \url{https://doi.org/10.1080/13662716.2020.1792274}.

\bibitem[Berant et~al.(2013)Berant, Chou, Frostig, and Liang]{berant-etal-2013-semantic}
Jonathan Berant, Andrew Chou, Roy Frostig, and Percy Liang.
\newblock Semantic parsing on {F}reebase from question-answer pairs.
\newblock In \emph{Proceedings of the 2013 Conference on Empirical Methods in Natural Language Processing}, pp.\  1533--1544, Seattle, Washington, USA, October 2013. Association for Computational Linguistics.
\newblock URL \url{https://aclanthology.org/D13-1160}.

\bibitem[Bernstein et~al.(2015)Bernstein, Little, Miller, Hartmann, Ackerman, Karger, Crowell, and Panovich]{DBLP:journals/cacm/BernsteinLMHAKC15}
Michael~S. Bernstein, Greg Little, Robert~C. Miller, Bj{\"{o}}rn Hartmann, Mark~S. Ackerman, David~R. Karger, David Crowell, and Katrina Panovich.
\newblock Soylent: a word processor with a crowd inside.
\newblock \emph{Commun. {ACM}}, 58\penalty0 (8):\penalty0 85--94, 2015.
\newblock \doi{10.1145/2791285}.
\newblock URL \url{https://doi.org/10.1145/2791285}.

\bibitem[Bird(2022)]{bird-2022-local}
Steven Bird.
\newblock Local languages, third spaces, and other high-resource scenarios.
\newblock In \emph{Proceedings of the 60th Annual Meeting of the Association for Computational Linguistics (Volume 1: Long Papers)}, pp.\  7817--7829, Dublin, Ireland, May 2022. Association for Computational Linguistics.
\newblock \doi{10.18653/v1/2022.acl-long.539}.
\newblock URL \url{https://aclanthology.org/2022.acl-long.539}.

\bibitem[Birhane et~al.(2022)Birhane, Isaac, Prabhakaran, Diaz, Elish, Gabriel, and Mohamed]{Birhane_2022}
Abeba Birhane, William Isaac, Vinodkumar Prabhakaran, Mark Diaz, Madeleine~Clare Elish, Iason Gabriel, and Shakir Mohamed.
\newblock Power to the people? opportunities and challenges for participatory ai.
\newblock In \emph{Equity and Access in Algorithms, Mechanisms, and Optimization}, EAAMO ’22. ACM, October 2022.
\newblock \doi{10.1145/3551624.3555290}.
\newblock URL \url{http://dx.doi.org/10.1145/3551624.3555290}.

\bibitem[Bisk et~al.(2020)Bisk, Zellers, Bras, Gao, and Choi]{Bisk2020}
Yonatan Bisk, Rowan Zellers, Ronan~Le Bras, Jianfeng Gao, and Yejin Choi.
\newblock Piqa: Reasoning about physical commonsense in natural language.
\newblock In \emph{Thirty-Fourth AAAI Conference on Artificial Intelligence}, 2020.

\bibitem[Blaschke et~al.(2023)Blaschke, Schuetze, and Plank]{blaschke-etal-2023-survey}
Verena Blaschke, Hinrich Schuetze, and Barbara Plank.
\newblock A survey of corpora for {G}ermanic low-resource languages and dialects.
\newblock In \emph{Proceedings of the 24th Nordic Conference on Computational Linguistics (NoDaLiDa)}, pp.\  392--414, T{\'o}rshavn, Faroe Islands, May 2023. University of Tartu Library.
\newblock URL \url{https://aclanthology.org/2023.nodalida-1.41}.

\bibitem[Blodgett et~al.(2016)Blodgett, Green, and O{'}Connor]{blodgett-etal-2016-demographic}
Su~Lin Blodgett, Lisa Green, and Brendan O{'}Connor.
\newblock Demographic dialectal variation in social media: A case study of {A}frican-{A}merican {E}nglish.
\newblock In \emph{Proceedings of the 2016 Conference on Empirical Methods in Natural Language Processing}, pp.\  1119--1130, Austin, Texas, November 2016. Association for Computational Linguistics.
\newblock \doi{10.18653/v1/D16-1120}.
\newblock URL \url{https://aclanthology.org/D16-1120}.

\bibitem[Botha et~al.(2018)Botha, Faruqui, Alex, Baldridge, and Das]{botha-etal-2018-learning}
Jan~A. Botha, Manaal Faruqui, John Alex, Jason Baldridge, and Dipanjan Das.
\newblock Learning to split and rephrase from {W}ikipedia edit history.
\newblock In \emph{Proceedings of the 2018 Conference on Empirical Methods in Natural Language Processing}, pp.\  732--737, Brussels, Belgium, October-November 2018. Association for Computational Linguistics.
\newblock \doi{10.18653/v1/D18-1080}.
\newblock URL \url{https://aclanthology.org/D18-1080}.

\bibitem[Boubdir et~al.(2023)Boubdir, Kim, Ermis, Fadaee, and Hooker]{boubdir2023prompts}
Meriem Boubdir, Edward Kim, Beyza Ermis, Marzieh Fadaee, and Sara Hooker.
\newblock Which prompts make the difference? data prioritization for efficient human llm evaluation, 2023.

\bibitem[Bourdieu(1987)]{bourdieu1987makes}
Pierre Bourdieu.
\newblock What makes a social class? on the theoretical and practical existence of groups.
\newblock \emph{Berkeley journal of sociology}, 32:\penalty0 1--17, 1987.

\bibitem[Bowser et~al.(2020)Bowser, Cooper, De~Sherbinin, Wiggins, Brenton, Chuang, Faustman, Haklay, and Meloche]{BowserOS}
Anne Bowser, Caren Cooper, Alex De~Sherbinin, Andrea Wiggins, Peter Brenton, Tyng-Ruey Chuang, Elaine Faustman, Mordechai Haklay, and Metis Meloche.
\newblock Still in need of norms: the state of the data in citizen science.
\newblock \emph{Citizen Science: Theory and Practice}, 5\penalty0 (1), 2020.

\bibitem[Brown \& Gilman(1968)Brown and Gilman]{BrownGilman+1968+252+275}
Roger Brown and Albert Gilman.
\newblock \emph{THE PRONOUNS OF POWER AND SOLIDARITY}, pp.\  252--275.
\newblock De Gruyter Mouton, Berlin, Boston, 1968.
\newblock ISBN 9783110805376.
\newblock \doi{doi:10.1515/9783110805376.252}.
\newblock URL \url{https://doi.org/10.1515/9783110805376.252}.

\bibitem[Brown et~al.(2020)Brown, Mann, Ryder, Subbiah, Kaplan, Dhariwal, Neelakantan, Shyam, Sastry, Askell, Agarwal, Herbert-Voss, Krueger, Henighan, Child, Ramesh, Ziegler, Wu, Winter, Hesse, Chen, Sigler, Litwin, Gray, Chess, Clark, Berner, McCandlish, Radford, Sutskever, and Amodei]{brown2020language}
Tom~B. Brown, Benjamin Mann, Nick Ryder, Melanie Subbiah, Jared Kaplan, Prafulla Dhariwal, Arvind Neelakantan, Pranav Shyam, Girish Sastry, Amanda Askell, Sandhini Agarwal, Ariel Herbert-Voss, Gretchen Krueger, Tom Henighan, Rewon Child, Aditya Ramesh, Daniel~M. Ziegler, Jeffrey Wu, Clemens Winter, Christopher Hesse, Mark Chen, Eric Sigler, Mateusz Litwin, Scott Gray, Benjamin Chess, Jack Clark, Christopher Berner, Sam McCandlish, Alec Radford, Ilya Sutskever, and Dario Amodei.
\newblock Language models are few-shot learners, 2020.

\bibitem[Cahyawijaya et~al.(2023)Cahyawijaya, Lovenia, Aji, Winata, Wilie, Koto, Mahendra, Wibisono, Romadhony, Vincentio, et~al.]{cahyawijaya2023nusacrowd}
Samuel Cahyawijaya, Holy Lovenia, Alham~Fikri Aji, Genta Winata, Bryan Wilie, Fajri Koto, Rahmad Mahendra, Christian Wibisono, Ade Romadhony, Karissa Vincentio, et~al.
\newblock Nusacrowd: Open source initiative for indonesian nlp resources.
\newblock In \emph{Findings of the Association for Computational Linguistics: ACL 2023}, pp.\  13745--13818, 2023.

\bibitem[Carlini et~al.(2023)Carlini, Jagielski, Choquette-Choo, Paleka, Pearce, Anderson, Terzis, Thomas, and Tramèr]{carlini2023poisoning}
Nicholas Carlini, Matthew Jagielski, Christopher~A. Choquette-Choo, Daniel Paleka, Will Pearce, Hyrum Anderson, Andreas Terzis, Kurt Thomas, and Florian Tramèr.
\newblock Poisoning web-scale training datasets is practical, 2023.

\bibitem[Castaldo(2023)]{Castaldo2023}
Joe Castaldo.
\newblock {``AI chatbots fall short in dozens of languages. A non-profit project aims to fix that"}.
\newblock \emph{Globe \& Mail}, 2023.

\bibitem[Castaldo(September 16, 2023)]{Castaldo2023a}
Joe Castaldo.
\newblock Meet the gig workers making ai machines more accurate, capable and powerful.
\newblock \emph{The Globe and Mail}, September 16, 2023.

\bibitem[Caswell et~al.(2020)Caswell, Breiner, van Esch, and Bapna]{caswell-etal-2020-language}
Isaac Caswell, Theresa Breiner, Daan van Esch, and Ankur Bapna.
\newblock Language {ID} in the wild: Unexpected challenges on the path to a thousand-language web text corpus.
\newblock In \emph{Proceedings of the 28th International Conference on Computational Linguistics}, pp.\  6588--6608, Barcelona, Spain (Online), December 2020. International Committee on Computational Linguistics.
\newblock \doi{10.18653/v1/2020.coling-main.579}.
\newblock URL \url{https://aclanthology.org/2020.coling-main.579}.

\bibitem[Chen et~al.(2023{\natexlab{a}})Chen, Li, Yan, Wang, Gunaratna, Yadav, Tang, Srinivasan, Zhou, Huang, et~al.]{chen2023alpagasus}
Lichang Chen, Shiyang Li, Jun Yan, Hai Wang, Kalpa Gunaratna, Vikas Yadav, Zheng Tang, Vijay Srinivasan, Tianyi Zhou, Heng Huang, et~al.
\newblock Alpagasus: Training a better alpaca with fewer data.
\newblock \emph{arXiv preprint arXiv:2307.08701}, 2023{\natexlab{a}}.

\bibitem[Chen et~al.(2023{\natexlab{b}})Chen, Ji, Bogoychev, Haddow, and Heafield]{chen2023monolingual}
Pinzhen Chen, Shaoxiong Ji, Nikolay Bogoychev, Barry Haddow, and Kenneth Heafield.
\newblock Monolingual or multilingual instruction tuning: Which makes a better alpaca, 2023{\natexlab{b}}.

\bibitem[Chen et~al.(2023{\natexlab{c}})Chen, Chen, and Karlsson]{chen2023dataset}
Sanxing Chen, Yongqiang Chen, and B{\"o}rje~F. Karlsson.
\newblock Dataset and baseline system for multi-lingual extraction and normalization of temporal and numerical expressions.
\newblock \emph{arXiv preprint arXiv:2303.18103}, 2023{\natexlab{c}}.

\bibitem[Chung et~al.(2022)Chung, Hou, Longpre, Zoph, Tay, Fedus, Li, Wang, Dehghani, Brahma, et~al.]{chung2022scaling}
Hyung~Won Chung, Le~Hou, Shayne Longpre, Barret Zoph, Yi~Tay, William Fedus, Yunxuan Li, Xuezhi Wang, Mostafa Dehghani, Siddhartha Brahma, et~al.
\newblock Scaling instruction-finetuned language models.
\newblock \emph{arXiv preprint arXiv:2210.11416}, 2022.

\bibitem[Chung et~al.(2023)Chung, Kamar, and Amershi]{Chung_2023}
John Chung, Ece Kamar, and Saleema Amershi.
\newblock Increasing diversity while maintaining accuracy: Text data generation with large language models and human interventions.
\newblock In \emph{Proceedings of the 61st Annual Meeting of the Association for Computational Linguistics (Volume 1: Long Papers)}. Association for Computational Linguistics, 2023.
\newblock \doi{10.18653/v1/2023.acl-long.34}.
\newblock URL \url{http://dx.doi.org/10.18653/v1/2023.acl-long.34}.

\bibitem[Clanuwat et~al.(2018)Clanuwat, Bober-Irizar, Kitamoto, Lamb, Yamamoto, and Ha]{Clanuwat2018DeepLF}
Tarin Clanuwat, Mikel Bober-Irizar, Asanobu Kitamoto, Alex Lamb, Kazuaki Yamamoto, and David Ha.
\newblock Deep learning for classical japanese literature.
\newblock \emph{ArXiv}, abs/1812.01718, 2018.
\newblock URL \url{https://api.semanticscholar.org/CorpusID:54458639}.

\bibitem[Clark et~al.(2019)Clark, Lee, Chang, Kwiatkowski, Collins, and Toutanova]{clark2019boolq}
Christopher Clark, Kenton Lee, Ming-Wei Chang, Tom Kwiatkowski, Michael Collins, and Kristina Toutanova.
\newblock Boolq: Exploring the surprising difficulty of natural yes/no questions.
\newblock In \emph{NAACL}, 2019.

\bibitem[Clark et~al.(2020)Clark, Choi, Collins, Garrette, Kwiatkowski, Nikolaev, and Palomaki]{tydiqa}
Jonathan~H. Clark, Eunsol Choi, Michael Collins, Dan Garrette, Tom Kwiatkowski, Vitaly Nikolaev, and Jennimaria Palomaki.
\newblock Tydi qa: A benchmark for information-seeking question answering in typologically diverse languages.
\newblock \emph{Transactions of the Association for Computational Linguistics}, 2020.

\bibitem[Clark et~al.(2018)Clark, Cowhey, Etzioni, Khot, Sabharwal, Schoenick, and Tafjord]{allenai:arc}
Peter Clark, Isaac Cowhey, Oren Etzioni, Tushar Khot, Ashish Sabharwal, Carissa Schoenick, and Oyvind Tafjord.
\newblock Think you have solved question answering? try arc, the ai2 reasoning challenge.
\newblock \emph{arXiv:1803.05457v1}, 2018.

\bibitem[Conneau et~al.(2018)Conneau, Lample, Rinott, Williams, Bowman, Schwenk, and Stoyanov]{conneau2018xnli}
Alexis Conneau, Guillaume Lample, Ruty Rinott, Adina Williams, Samuel~R Bowman, Holger Schwenk, and Veselin Stoyanov.
\newblock Xnli: Evaluating cross-lingual sentence representations.
\newblock \emph{arXiv preprint arXiv:1809.05053}, 2018.

\bibitem[Conover et~al.(2023)Conover, Hayes, Mathur, Xie, Wan, Shah, Ghodsi, Wendell, Zaharia, and Xin]{DatabricksBlog2023DollyV2}
Mike Conover, Matt Hayes, Ankit Mathur, Jianwei Xie, Jun Wan, Sam Shah, Ali Ghodsi, Patrick Wendell, Matei Zaharia, and Reynold Xin.
\newblock {Free Dolly: Introducing the World's First Truly Open Instruction-Tuned LLM}, 2023.
\newblock URL \url{https://www.databricks.com/blog/2023/04/12/dolly-first-open-commercially-viable-instruction-tuned-llm}.

\bibitem[ConseggioLigure(2023{\natexlab{a}})]{lijnews-instruct-ita-lij}
ConseggioLigure.
\newblock Lij news instruct ita-lij.
\newblock \url{https://huggingface.co/datasets/ConseggioLigure/lijnews-instruct-ita-lij}, 2023{\natexlab{a}}.
\newblock Accessed: 2023-11-28.

\bibitem[ConseggioLigure(2023{\natexlab{b}})]{lijnews-instruct-lij-ita}
ConseggioLigure.
\newblock Lij news instruct lij-ita.
\newblock \url{https://huggingface.co/datasets/ConseggioLigure/lijnews-instruct-lij-ita}, 2023{\natexlab{b}}.
\newblock Accessed: 2023-11-28.

\bibitem[ConseggioLigure(2023{\natexlab{c}})]{seed-instruct-eng-lij}
ConseggioLigure.
\newblock Seed instruct eng-lij.
\newblock \url{https://huggingface.co/datasets/ConseggioLigure/seed-instruct-eng-lij}, 2023{\natexlab{c}}.
\newblock Accessed: 2023-11-28.

\bibitem[ConseggioLigure(2023{\natexlab{d}})]{seed-instruct-lij-eng}
ConseggioLigure.
\newblock Seed instruct lij-eng.
\newblock \url{https://huggingface.co/datasets/ConseggioLigure/seed-instruct-lij-eng}, 2023{\natexlab{d}}.
\newblock Accessed: 2023-11-28.

\bibitem[Corbett et~al.(2023)Corbett, Denton, and Erete]{corbett2023}
Eric Corbett, Emily Denton, and Sheena Erete.
\newblock Power and public participation in ai.
\newblock In \emph{Proceedings of the 3rd ACM Conference on Equity and Access in Algorithms, Mechanisms, and Optimization}, EAAMO '23, New York, NY, USA, 2023. Association for Computing Machinery.
\newblock ISBN 9798400703812.
\newblock \doi{10.1145/3617694.3623228}.
\newblock URL \url{https://doi.org/10.1145/3617694.3623228}.

\bibitem[Crawford(2021)]{crawford2021a}
Kate Crawford.
\newblock \emph{Atlas of AI: Power, Politics, and the Planetary Costs of Artificial Intelligence}.
\newblock Yale University Press, New Haven, Connecticut, 2021.

\bibitem[Cui et~al.(2019)Cui, Liu, Che, Xiao, Chen, Ma, Wang, and Hu]{cui-etal-2019-span}
Yiming Cui, Ting Liu, Wanxiang Che, Li~Xiao, Zhipeng Chen, Wentao Ma, Shijin Wang, and Guoping Hu.
\newblock A span-extraction dataset for {C}hinese machine reading comprehension.
\newblock In \emph{Proceedings of the 2019 Conference on Empirical Methods in Natural Language Processing and the 9th International Joint Conference on Natural Language Processing (EMNLP-IJCNLP)}, pp.\  5883--5889, Hong Kong, China, November 2019. Association for Computational Linguistics.
\newblock \doi{10.18653/v1/D19-1600}.
\newblock URL \url{https://aclanthology.org/D19-1600}.

\bibitem[Cui et~al.(2023)Cui, Yang, and Yao]{cui2023efficient}
Yiming Cui, Ziqing Yang, and Xin Yao.
\newblock Efficient and effective text encoding for chinese llama and alpaca, 2023.

\bibitem[Dabre et~al.(2020)Dabre, Chu, and Kunchukuttan]{10.1145/3406095}
Raj Dabre, Chenhui Chu, and Anoop Kunchukuttan.
\newblock A survey of multilingual neural machine translation.
\newblock \emph{ACM Comput. Surv.}, 53\penalty0 (5), sep 2020.
\newblock ISSN 0360-0300.
\newblock \doi{10.1145/3406095}.
\newblock URL \url{https://doi.org/10.1145/3406095}.

\bibitem[Dac~Lai et~al.(2023)Dac~Lai, Van~Nguyen, Ngo, Nguyen, Dernoncourt, Rossi, and Nguyen]{dac2023okapi}
Viet Dac~Lai, Chien Van~Nguyen, Nghia~Trung Ngo, Thuat Nguyen, Franck Dernoncourt, Ryan~A Rossi, and Thien~Huu Nguyen.
\newblock Okapi: Instruction-tuned large language models in multiple languages with reinforcement learning from human feedback.
\newblock \emph{arXiv e-prints}, pp.\  arXiv--2307, 2023.

\bibitem[Dasigi et~al.(2019)Dasigi, Liu, Marasovic, Smith, and Gardner]{allenai:quoref}
Pradeep Dasigi, Nelson~F. Liu, Ana Marasovic, Noah~A. Smith, and Matt Gardner.
\newblock Quoref: A reading comprehension dataset with questions requiring coreferential reasoning.
\newblock \emph{arXiv:1908.05803v2}, 2019.

\bibitem[de~Franga et~al.(2015)de~Franga, Vivacqua, and Campos]{deFranga2015}
Flavio~A. de~Franga, Adriana~S. Vivacqua, and Maria Luiza~M. Campos.
\newblock Designing a gamification mechanism to encourage contributions in a crowdsourcing system.
\newblock In \emph{2015 IEEE 19th International Conference on Computer Supported Cooperative Work in Design (CSCWD)}, pp.\  462--466, 2015.
\newblock \doi{10.1109/CSCWD.2015.7231003}.

\bibitem[De~Mauro et~al.(2015)De~Mauro, Greco, and Grimaldi]{de2015big}
Andrea De~Mauro, Marco Greco, and Michele Grimaldi.
\newblock What is big data? a consensual definition and a review of key research topics.
\newblock In \emph{AIP conference proceedings}, volume 1644, pp.\  97--104. American Institute of Physics, 2015.

\bibitem[Delgado et~al.(2023)Delgado, Yang, Madaio, and Yang]{Delgado2023ThePT}
Fernando Delgado, Stephen Yang, Michael Madaio, and Qian Yang.
\newblock The participatory turn in ai design: Theoretical foundations and the current state of practice.
\newblock \emph{Proceedings of the 3rd ACM Conference on Equity and Access in Algorithms, Mechanisms, and Optimization}, 2023.
\newblock URL \url{https://api.semanticscholar.org/CorpusID:263605822}.

\bibitem[Deng et~al.(2023)Deng, Zhang, Pan, and Bing]{Deng2023MultilingualJC}
Yue Deng, Wenxuan Zhang, Sinno~Jialin Pan, and Lidong Bing.
\newblock Multilingual jailbreak challenges in large language models.
\newblock \emph{ArXiv}, abs/2310.06474, 2023.
\newblock URL \url{https://api.semanticscholar.org/CorpusID:263831094}.

\bibitem[desik98(2023)]{desik98TeluguRiddles}
desik98.
\newblock {Telugu Riddles}.
\newblock \url{https://huggingface.co/datasets/desik98/TeluguRiddles}, 2023.
\newblock Accessed: 2023-11-30.

\bibitem[Dhole et~al.(2021)Dhole, Gangal, Gehrmann, Gupta, Li, Mahamood, Mahendiran, Mille, Shrivastava, Tan, et~al.]{dhole2021nl}
Kaustubh~D Dhole, Varun Gangal, Sebastian Gehrmann, Aadesh Gupta, Zhenhao Li, Saad Mahamood, Abinaya Mahendiran, Simon Mille, Ashish Shrivastava, Samson Tan, et~al.
\newblock Nl-augmenter: A framework for task-sensitive natural language augmentation.
\newblock \emph{arXiv preprint arXiv:2112.02721}, 2021.

\bibitem[Doddapaneni et~al.(2023)Doddapaneni, Aralikatte, Ramesh, Goyal, Khapra, Kunchukuttan, and Kumar]{doddapaneni-etal-2023-towards}
Sumanth Doddapaneni, Rahul Aralikatte, Gowtham Ramesh, Shreya Goyal, Mitesh~M. Khapra, Anoop Kunchukuttan, and Pratyush Kumar.
\newblock Towards leaving no {I}ndic language behind: Building monolingual corpora, benchmark and models for {I}ndic languages.
\newblock In \emph{Proceedings of the 61st Annual Meeting of the Association for Computational Linguistics (Volume 1: Long Papers)}, pp.\  12402--12426, Toronto, Canada, July 2023. Association for Computational Linguistics.
\newblock URL \url{https://aclanthology.org/2023.acl-long.693}.

\bibitem[Dodge et~al.(2021)Dodge, Sap, Marasovic, Agnew, Ilharco, Groeneveld, and Gardner]{dodge2021}
Jesse Dodge, Maarten Sap, Ana Marasovic, William Agnew, Gabriel Ilharco, Dirk Groeneveld, and Matt Gardner.
\newblock Documenting the english colossal clean crawled corpus.
\newblock \emph{CoRR}, abs/2104.08758, 2021.
\newblock URL \url{https://arxiv.org/abs/2104.08758}.

\bibitem[Do{\u{g}}ru{\"o}z et~al.(2023)Do{\u{g}}ru{\"o}z, Sitaram, and Yong]{dougruoz2023representativeness}
A~Seza Do{\u{g}}ru{\"o}z, Sunayana Sitaram, and Zheng-Xin Yong.
\newblock Representativeness as a forgotten lesson for multilingual and code-switched data collection and preparation.
\newblock \emph{arXiv preprint arXiv:2310.20470}, 2023.

\bibitem[Dolan \& Brockett(2005)Dolan and Brockett]{dolan2005automatically}
Bill Dolan and Chris Brockett.
\newblock Automatically constructing a corpus of sentential paraphrases.
\newblock In \emph{Third International Workshop on Paraphrasing (IWP2005)}. Asia Federation of Natural Language Processing, January 2005.
\newblock URL \url{https://www.microsoft.com/en-us/research/publication/automatically-constructing-a-corpus-of-sentential-paraphrases/}.

\bibitem[Durmus et~al.(2023)Durmus, Nyugen, Liao, Schiefer, Askell, Bakhtin, Chen, Hatfield-Dodds, Hernandez, Joseph, et~al.]{durmus2023towards}
Esin Durmus, Karina Nyugen, Thomas~I Liao, Nicholas Schiefer, Amanda Askell, Anton Bakhtin, Carol Chen, Zac Hatfield-Dodds, Danny Hernandez, Nicholas Joseph, et~al.
\newblock Towards measuring the representation of subjective global opinions in language models.
\newblock \emph{arXiv preprint arXiv:2306.16388}, 2023.

\bibitem[el2e10(2023{\natexlab{a}})]{aya-indicsentiment}
el2e10.
\newblock {Aya Indic Sentiment}.
\newblock \url{https://huggingface.co/datasets/el2e10/aya-indicsentiment}, 2023{\natexlab{a}}.
\newblock Accessed: 2023-11-28.

\bibitem[el2e10(2023{\natexlab{b}})]{aya-parapharse}
el2e10.
\newblock {Aya Paraphrase}.
\newblock \url{https://huggingface.co/datasets/el2e10/aya-paraphrase}, 2023{\natexlab{b}}.
\newblock Accessed: 2023-11-28.

\bibitem[Fabbri et~al.(2019)Fabbri, Li, She, Li, and Radev]{alex2019multinews}
Alexander~R. Fabbri, Irene Li, Tianwei She, Suyi Li, and Dragomir~R. Radev.
\newblock Multi-news: a large-scale multi-document summarization dataset and abstractive hierarchical model, 2019.

\bibitem[Falck et~al.(2012)Falck, Heblich, Lameli, and S{\"u}dekum]{falck2012dialects}
Oliver Falck, Stephan Heblich, Alfred Lameli, and Jens S{\"u}dekum.
\newblock Dialects, cultural identity, and economic exchange.
\newblock \emph{Journal of urban economics}, 72\penalty0 (2-3):\penalty0 225--239, 2012.

\bibitem[Fan et~al.(2021)Fan, Bhosale, Schwenk, Ma, El-Kishky, Goyal, Baines, Celebi, Wenzek, Chaudhary, et~al.]{fan2021beyond}
Angela Fan, Shruti Bhosale, Holger Schwenk, Zhiyi Ma, Ahmed El-Kishky, Siddharth Goyal, Mandeep Baines, Onur Celebi, Guillaume Wenzek, Vishrav Chaudhary, et~al.
\newblock Beyond english-centric multilingual machine translation.
\newblock \emph{Journal of Machine Learning Research}, 22\penalty0 (107):\penalty0 1--48, 2021.

\bibitem[Farahani et~al.(2021)Farahani, Gharachorloo, and Manthouri]{pnSummary}
Mehrdad Farahani, Mohammad Gharachorloo, and Mohammad Manthouri.
\newblock Leveraging parsbert and pretrained mt5 for persian abstractive text summarization.
\newblock In \emph{2021 26th International Computer Conference, Computer Society of Iran (CSICC)}, pp.\  1--6. IEEE, 2021.

\bibitem[Ferrara(2023)]{ferrara2023should}
Emilio Ferrara.
\newblock Should chatgpt be biased? challenges and risks of bias in large language models.
\newblock \emph{arXiv preprint arXiv:2304.03738}, 2023.

\bibitem[$\forall$ et~al.(2020)$\forall$, Nekoto, Marivate, Matsila, Fasubaa, Fagbohungbe, Akinola, Muhammad, Kabongo~Kabenamualu, Osei, Sackey, Niyongabo, Macharm, Ogayo, Ahia, Berhe, Adeyemi, Mokgesi-Selinga, Okegbemi, Martinus, Tajudeen, Degila, Ogueji, Siminyu, Kreutzer, Webster, Ali, Abbott, Orife, Ezeani, Dangana, Kamper, Elsahar, Duru, Kioko, Espoir, van Biljon, Whitenack, Onyefuluchi, Emezue, Dossou, Sibanda, Bassey, Olabiyi, Ramkilowan, {\"O}ktem, Akinfaderin, and Bashir]{forall-nekoto-etal-2020-participatory}
{}~$\forall$, Wilhelmina Nekoto, Vukosi Marivate, Tshinondiwa Matsila, Timi Fasubaa, Taiwo Fagbohungbe, Solomon~Oluwole Akinola, Shamsuddeen Muhammad, Salomon Kabongo~Kabenamualu, Salomey Osei, Freshia Sackey, Rubungo~Andre Niyongabo, Ricky Macharm, Perez Ogayo, Orevaoghene Ahia, Musie~Meressa Berhe, Mofetoluwa Adeyemi, Masabata Mokgesi-Selinga, Lawrence Okegbemi, Laura Martinus, Kolawole Tajudeen, Kevin Degila, Kelechi Ogueji, Kathleen Siminyu, Julia Kreutzer, Jason Webster, Jamiil~Toure Ali, Jade Abbott, Iroro Orife, Ignatius Ezeani, Idris~Abdulkadir Dangana, Herman Kamper, Hady Elsahar, Goodness Duru, Ghollah Kioko, Murhabazi Espoir, Elan van Biljon, Daniel Whitenack, Christopher Onyefuluchi, Chris~Chinenye Emezue, Bonaventure F.~P. Dossou, Blessing Sibanda, Blessing Bassey, Ayodele Olabiyi, Arshath Ramkilowan, Alp {\"O}ktem, Adewale Akinfaderin, and Abdallah Bashir.
\newblock Participatory research for low-resourced machine translation: A case study in {A}frican languages.
\newblock In Trevor Cohn, Yulan He, and Yang Liu (eds.), \emph{Findings of the Association for Computational Linguistics: EMNLP 2020}, Online, November 2020. Association for Computational Linguistics.
\newblock \doi{10.18653/v1/2020.findings-emnlp.195}.
\newblock URL \url{https://aclanthology.org/2020.findings-emnlp.195}.

\bibitem[Franzoni \& Sauermann(2014)Franzoni and Sauermann]{FRANZONI20141}
Chiara Franzoni and Henry Sauermann.
\newblock Crowd science: The organization of scientific research in open collaborative projects.
\newblock \emph{Research Policy}, 43\penalty0 (1):\penalty0 1--20, 2014.
\newblock ISSN 0048-7333.
\newblock \doi{https://doi.org/10.1016/j.respol.2013.07.005}.
\newblock URL \url{https://www.sciencedirect.com/science/article/pii/S0048733313001212}.

\bibitem[ganeshjcs(2023{\natexlab{a}})]{hindi-article-summarization}
ganeshjcs.
\newblock {Hindi Article Summarization}.
\newblock \url{https://huggingface.co/datasets/ganeshjcs/hindi-article-summarization}, 2023{\natexlab{a}}.
\newblock Accessed: 2023-11-28.

\bibitem[ganeshjcs(2023{\natexlab{b}})]{hindi-headline-article-generation}
ganeshjcs.
\newblock {Hindi Headline Article Generation}.
\newblock \url{https://huggingface.co/datasets/ganeshjcs/hindi-headline-article-generation}, 2023{\natexlab{b}}.
\newblock Accessed: 2023-11-28.

\bibitem[Gao \& Liu(2023)Gao and Liu]{10.1371/journal.pone.0287850}
Ya~Gao and WenQi Liu.
\newblock Measures to sustain endangered languages: A bilingual competition model with sliding mode control.
\newblock \emph{PLOS ONE}, 18\penalty0 (6):\penalty0 1--16, 06 2023.
\newblock \doi{10.1371/journal.pone.0287850}.
\newblock URL \url{https://doi.org/10.1371/journal.pone.0287850}.

\bibitem[Garrette et~al.(2013)Garrette, Mielens, and Baldridge]{garrette2013real}
Dan Garrette, Jason Mielens, and Jason Baldridge.
\newblock Real-world semi-supervised learning of pos-taggers for low-resource languages.
\newblock In \emph{Proceedings of the 51st Annual Meeting of the Association for Computational Linguistics (Volume 1: Long Papers)}, pp.\  583--592, 2013.

\bibitem[Gerosa et~al.(2021)Gerosa, Wiese, Trinkenreich, Link, Robles, Treude, Steinmacher, and Sarma]{gerosa2021shifting}
Marco Gerosa, Igor Wiese, Bianca Trinkenreich, Georg Link, Gregorio Robles, Christoph Treude, Igor Steinmacher, and Anita Sarma.
\newblock The shifting sands of motivation: Revisiting what drives contributors in open source.
\newblock In \emph{2021 IEEE/ACM 43rd International Conference on Software Engineering (ICSE)}, pp.\  1046--1058. IEEE, 2021.

\bibitem[Ghosh \& Caliskan(2023)Ghosh and Caliskan]{ghosh2023chatgpt}
Sourojit Ghosh and Aylin Caliskan.
\newblock Chatgpt perpetuates gender bias in machine translation and ignores non-gendered pronouns: Findings across bengali and five other low-resource languages, 2023.

\bibitem[Gliwa et~al.(2019)Gliwa, Mochol, Biesek, and Wawer]{gliwa-etal-2019-samsum}
Bogdan Gliwa, Iwona Mochol, Maciej Biesek, and Aleksander Wawer.
\newblock {SAMS}um corpus: A human-annotated dialogue dataset for abstractive summarization.
\newblock In \emph{Proceedings of the 2nd Workshop on New Frontiers in Summarization}, pp.\  70--79, Hong Kong, China, November 2019. Association for Computational Linguistics.
\newblock \doi{10.18653/v1/D19-5409}.
\newblock URL \url{https://aclanthology.org/D19-5409}.

\bibitem[Goodwin(2017)]{Goodwin2017}
Charles Goodwin.
\newblock \emph{Co-Operative Action}.
\newblock Learning in Doing: Social, Cognitive and Computational Perspectives. Cambridge University Press, 2017.

\bibitem[Goyal et~al.(2021{\natexlab{a}})Goyal, Du, Ott, Anantharaman, and Conneau]{Goyal2021Larger}
Naman Goyal, Jingfei Du, Myle Ott, Giri Anantharaman, and Alexis Conneau.
\newblock Larger-{Scale} {Transformers} for {Multilingual} {Masked} {Language} {Modeling}.
\newblock In \emph{Proceedings of the 6th {Workshop} on {Representation} {Learning} for {NLP} ({RepL4NLP}-2021)}. Association for Computational Linguistics, 2021{\natexlab{a}}.

\bibitem[Goyal et~al.(2021{\natexlab{b}})Goyal, Gao, Chaudhary, Chen, Wenzek, Ju, Krishnan, Ranzato, Guzman, and Fan]{goyal2021flores101}
Naman Goyal, Cynthia Gao, Vishrav Chaudhary, Peng-Jen Chen, Guillaume Wenzek, Da~Ju, Sanjana Krishnan, Marc'Aurelio Ranzato, Francisco Guzman, and Angela Fan.
\newblock The flores-101 evaluation benchmark for low-resource and multilingual machine translation, 2021{\natexlab{b}}.

\bibitem[Graff et~al.(2003)Graff, Kong, Chen, and Maeda]{graff2003english}
David Graff, Junbo Kong, Ke~Chen, and Kazuaki Maeda.
\newblock English gigaword.
\newblock \emph{Linguistic Data Consortium, Philadelphia}, 4\penalty0 (1):\penalty0 34, 2003.

\bibitem[Grano et~al.(2017)Grano, Di~Sorbo, Mercaldo, Visaggio, Canfora, and Panichella]{ZurichOpenRepositoryandArchive:dataset}
Giovanni Grano, Andrea Di~Sorbo, Francesco Mercaldo, Corrado~A Visaggio, Gerardo Canfora, and Sebastiano Panichella.
\newblock Android apps and user feedback: a dataset for software evolution and quality improvement.
\newblock In \emph{Proceedings of the 2nd ACM SIGSOFT international workshop on app market analytics}, pp.\  8--11, 2017.

\bibitem[Groeneveld et~al.(2024)Groeneveld, Beltagy, Walsh, Bhagia, Kinney, Tafjord, Jha, Ivison, Magnusson, Wang, Arora, Atkinson, Authur, Chandu, Cohan, Dumas, Elazar, Gu, Hessel, Khot, Merrill, Morrison, Muennighoff, Naik, Nam, Peters, Pyatkin, Ravichander, Schwenk, Shah, Smith, Subramani, Wortsman, Dasigi, Lambert, Richardson, Dodge, Lo, Soldaini, Smith, and Hajishirzi]{olmo20247b}
Dirk Groeneveld, Iz~Beltagy, Pete Walsh, Akshita Bhagia, Rodney Kinney, Oyvind Tafjord, Ananya~Harsh Jha, Hamish Ivison, Ian Magnusson, Yizhong Wang, Shane Arora, David Atkinson, Russell Authur, Khyathi Chandu, Arman Cohan, Jennifer Dumas, Yanai Elazar, Yuling Gu, Jack Hessel, Tushar Khot, William Merrill, Jacob Morrison, Niklas Muennighoff, Aakanksha Naik, Crystal Nam, Matthew~E. Peters, Valentina Pyatkin, Abhilasha Ravichander, Dustin Schwenk, Saurabh Shah, Will Smith, Nishant Subramani, Mitchell Wortsman, Pradeep Dasigi, Nathan Lambert, Kyle Richardson, Jesse Dodge, Kyle Lo, Luca Soldaini, Noah~A. Smith, and Hannaneh Hajishirzi.
\newblock {OLMo: Accelerating the Science of Language Models}.
\newblock \emph{arXiv preprint}, 2024.

\bibitem[Gu et~al.(2022)Gu, Dalvi, and Clark]{gu-etal-2022-dream}
Yuling Gu, Bhavana Dalvi, and Peter Clark.
\newblock {DREAM}: Improving situational {QA} by first elaborating the situation.
\newblock In \emph{Proceedings of the 2022 Conference of the North American Chapter of the Association for Computational Linguistics: Human Language Technologies}, pp.\  1115--1127, Seattle, United States, July 2022. Association for Computational Linguistics.
\newblock \doi{10.18653/v1/2022.naacl-main.82}.
\newblock URL \url{https://aclanthology.org/2022.naacl-main.82}.

\bibitem[Guevara-Rukoz et~al.(2020)Guevara-Rukoz, Demirsahin, He, Chu, Sarin, Pipatsrisawat, Gutkin, Butryna, and Kjartansson]{guevara-rukoz-etal-2020-crowdsourcing}
Adriana Guevara-Rukoz, Isin Demirsahin, Fei He, Shan-Hui~Cathy Chu, Supheakmungkol Sarin, Knot Pipatsrisawat, Alexander Gutkin, Alena Butryna, and Oddur Kjartansson.
\newblock Crowdsourcing {L}atin {A}merican {S}panish for low-resource text-to-speech.
\newblock In \emph{Proceedings of the Twelfth Language Resources and Evaluation Conference}, pp.\  6504--6513, Marseille, France, May 2020. European Language Resources Association.
\newblock ISBN 979-10-95546-34-4.
\newblock URL \url{https://aclanthology.org/2020.lrec-1.801}.

\bibitem[Gulli(2005)]{gulli2005ag}
Antonio Gulli.
\newblock {AG’s Corpus of News Articles}.
\newblock \emph{Dipartimento di Informatica, University of Pisa, Nov}, 2005.
\newblock URL \url{http://www.di.unipi.it/~gulli/AG_corpus_of_news_articles.html}.

\bibitem[Gunasekar et~al.(2023)Gunasekar, Zhang, Aneja, Mendes, Giorno, Gopi, Javaheripi, Kauffmann, de~Rosa, Saarikivi, Salim, Shah, Behl, Wang, Bubeck, Eldan, Kalai, Lee, and Li]{gunasekar2023textbooks}
Suriya Gunasekar, Yi~Zhang, Jyoti Aneja, Caio César~Teodoro Mendes, Allie~Del Giorno, Sivakanth Gopi, Mojan Javaheripi, Piero Kauffmann, Gustavo de~Rosa, Olli Saarikivi, Adil Salim, Shital Shah, Harkirat~Singh Behl, Xin Wang, Sébastien Bubeck, Ronen Eldan, Adam~Tauman Kalai, Yin~Tat Lee, and Yuanzhi Li.
\newblock Textbooks are all you need, 2023.

\bibitem[Gupta et~al.(2023)Gupta, Scaria, Anantheswaran, Verma, Parmar, Sawant, Baral, and Mishra]{gupta2023targen}
Himanshu Gupta, Kevin Scaria, Ujjwala Anantheswaran, Shreyas Verma, Mihir Parmar, Saurabh~Arjun Sawant, Chitta Baral, and Swaroop Mishra.
\newblock Targen: Targeted data generation with large language models, 2023.

\bibitem[Hanley et~al.(2020)Hanley, Khandelwal, Averbuch-Elor, Snavely, and Nissenbaum]{hanley2020ethical}
Margot Hanley, Apoorv Khandelwal, Hadar Averbuch-Elor, Noah Snavely, and Helen Nissenbaum.
\newblock An ethical highlighter for people-centric dataset creation.
\newblock \emph{arXiv preprint arXiv:2011.13583}, 2020.

\bibitem[Hardmeier \& Guillou(2018)Hardmeier and Guillou]{hardmeier2018pronoun}
Christian Hardmeier and Liane Guillou.
\newblock Pronoun translation in english-french machine translation: An analysis of error types, 2018.

\bibitem[Hartung et~al.(2023)Hartung, Herygers, Kurlekar, Zakaria, Volkan, Gröttrup, and Georges]{hartung2023measuring}
Kai Hartung, Aaricia Herygers, Shubham Kurlekar, Khabbab Zakaria, Taylan Volkan, Sören Gröttrup, and Munir Georges.
\newblock Measuring sentiment bias in machine translation, 2023.

\bibitem[Hasan et~al.(2021)Hasan, Bhattacharjee, Islam, Mubasshir, Li, Kang, Rahman, and Shahriyar]{hasan-etal-2021-xl}
Tahmid Hasan, Abhik Bhattacharjee, Md.~Saiful Islam, Kazi Mubasshir, Yuan-Fang Li, Yong-Bin Kang, M.~Sohel Rahman, and Rifat Shahriyar.
\newblock {XL}-sum: Large-scale multilingual abstractive summarization for 44 languages.
\newblock In \emph{Findings of the Association for Computational Linguistics: ACL-IJCNLP 2021}, pp.\  4693--4703, Online, August 2021. Association for Computational Linguistics.
\newblock \doi{10.18653/v1/2021.findings-acl.413}.
\newblock URL \url{https://aclanthology.org/2021.findings-acl.413}.

\bibitem[Haugen(1959)]{haugen1959planning}
Einar Haugen.
\newblock Planning for a standard language in modern norway.
\newblock \emph{Anthropological linguistics}, pp.\  8--21, 1959.

\bibitem[Held et~al.(2023)Held, Harris, Best, and Yang]{held2023material}
William Held, Camille Harris, Michael Best, and Diyi Yang.
\newblock A material lens on coloniality in nlp, 2023.

\bibitem[Hermann et~al.(2015)Hermann, Kocisky, Grefenstette, Espeholt, Kay, Suleyman, and Blunsom]{hermann2015teaching}
Karl~Moritz Hermann, Tomas Kocisky, Edward Grefenstette, Lasse Espeholt, Will Kay, Mustafa Suleyman, and Phil Blunsom.
\newblock Teaching machines to read and comprehend.
\newblock In \emph{Advances in neural information processing systems}, pp.\  1693--1701, 2015.

\bibitem[hghader1(2023)]{hghader1FarsTailInstructLLM}
hghader1.
\newblock {FarsTail-Instruct-LLM}.
\newblock \url{https://huggingface.co/datasets/hghader1/FarsTail-Instruct-LLM}, 2023.
\newblock Accessed: 2023-11-28.

\bibitem[Holmstr{\"o}m \& Doostmohammadi(2023)Holmstr{\"o}m and Doostmohammadi]{holmstrom-doostmohammadi-2023-making}
Oskar Holmstr{\"o}m and Ehsan Doostmohammadi.
\newblock Making instruction finetuning accessible to non-{E}nglish languages: A case study on {S}wedish models.
\newblock In \emph{Proceedings of the 24th Nordic Conference on Computational Linguistics (NoDaLiDa)}, pp.\  634--642, T{\'o}rshavn, Faroe Islands, May 2023. University of Tartu Library.
\newblock URL \url{https://aclanthology.org/2023.nodalida-1.62}.

\bibitem[Honovich et~al.(2023)Honovich, Scialom, Levy, and Schick]{honovich-etal-2023-unnatural}
Or~Honovich, Thomas Scialom, Omer Levy, and Timo Schick.
\newblock Unnatural instructions: Tuning language models with (almost) no human labor.
\newblock In \emph{Proceedings of the 61st Annual Meeting of the Association for Computational Linguistics (Volume 1: Long Papers)}, pp.\  14409--14428, Toronto, Canada, July 2023. Association for Computational Linguistics.
\newblock URL \url{https://aclanthology.org/2023.acl-long.806}.

\bibitem[Hovy \& Prabhumoye(2021)Hovy and Prabhumoye]{hovy2021five}
Dirk Hovy and Shrimai Prabhumoye.
\newblock Five sources of bias in natural language processing.
\newblock \emph{Language and Linguistics Compass}, 15\penalty0 (8):\penalty0 e12432, 2021.

\bibitem[Hovy et~al.(2001)Hovy, Gerber, Hermjakob, Lin, and Ravichandran]{hovy-etal-2001-toward}
Eduard Hovy, Laurie Gerber, Ulf Hermjakob, Chin-Yew Lin, and Deepak Ravichandran.
\newblock Toward semantics-based answer pinpointing.
\newblock In \emph{Proceedings of the First International Conference on Human Language Technology Research}, 2001.
\newblock URL \url{https://aclanthology.org/H01-1069}.

\bibitem[Hsueh et~al.(2009)Hsueh, Melville, and Sindhwani]{hsueh2009data}
Pei-Yun Hsueh, Prem Melville, and Vikas Sindhwani.
\newblock Data quality from crowdsourcing: a study of annotation selection criteria.
\newblock In \emph{Proceedings of the NAACL HLT 2009 workshop on active learning for natural language processing}, pp.\  27--35, 2009.

\bibitem[Hu et~al.(2021)Hu, Shen, Wallis, Allen-Zhu, Li, Wang, Wang, and Chen]{hu2021lora}
Edward~J. Hu, Yelong Shen, Phillip Wallis, Zeyuan Allen-Zhu, Yuanzhi Li, Shean Wang, Lu~Wang, and Weizhu Chen.
\newblock Lora: Low-rank adaptation of large language models, 2021.

\bibitem[Huang et~al.(2023)Huang, Yu, Zhu, Sun, Cheng, Song, Chen, Alharthi, An, Liu, et~al.]{huang2023acegpt}
Huang Huang, Fei Yu, Jianqing Zhu, Xuening Sun, Hao Cheng, Dingjie Song, Zhihong Chen, Abdulmohsen Alharthi, Bang An, Ziche Liu, et~al.
\newblock Acegpt, localizing large language models in arabic.
\newblock \emph{arXiv preprint arXiv:2309.12053}, 2023.

\bibitem[Huang et~al.(2022)Huang, Hsu, Natarajan, Chang, and Peng]{huang2022multilingual}
Kuan-Hao Huang, I-Hung Hsu, Premkumar Natarajan, Kai-Wei Chang, and Nanyun Peng.
\newblock Multilingual generative language models for zero-shot cross-lingual event argument extraction, 2022.

\bibitem[Huang et~al.(2019)Huang, Le~Bras, Bhagavatula, and Choi]{huang-etal-2019-cosmos}
Lifu Huang, Ronan Le~Bras, Chandra Bhagavatula, and Yejin Choi.
\newblock Cosmos {QA}: Machine reading comprehension with contextual commonsense reasoning.
\newblock In \emph{Proceedings of the 2019 Conference on Empirical Methods in Natural Language Processing and the 9th International Joint Conference on Natural Language Processing (EMNLP-IJCNLP)}, pp.\  2391--2401, Hong Kong, China, November 2019. Association for Computational Linguistics.
\newblock \doi{10.18653/v1/D19-1243}.
\newblock URL \url{https://aclanthology.org/D19-1243}.

\bibitem[Hussain et~al.(2021)Hussain, Mughal, Ali, Hassan, and Daudpota]{hussain-etal-2021-urdunews}
Khalid Hussain, Nimra Mughal, Irfan Ali, Saif Hassan, and Sher~Muhammad Daudpota.
\newblock {Urdu News Dataset 1M}.
\newblock Technical report, Mendeley Data, V3, 2021.

\bibitem[Hämäläinen(2021)]{H_m_l_inen_2021}
Mika Hämäläinen.
\newblock \emph{Endangered Languages are not Low-Resourced!}, pp.\  1–11.
\newblock University of Helsinki, March 2021.
\newblock \doi{10.31885/9789515150257.1}.
\newblock URL \url{http://dx.doi.org/10.31885/9789515150257.1}.

\bibitem[Iftitahu(2023{\natexlab{a}})]{indonesian_instruct_stories}
Iftitahu.
\newblock {Indonesian Instruct Stories}.
\newblock \url{https://huggingface.co/datasets/Iftitahu/indonesian_instruct_stories}, 2023{\natexlab{a}}.
\newblock Accessed: 2023-11-28.

\bibitem[Iftitahu(2023{\natexlab{b}})]{javanese_instruct_stories}
Iftitahu.
\newblock {Javanese Instruct Stories}.
\newblock \url{https://huggingface.co/datasets/Iftitahu/javanese_instruct_stories}, 2023{\natexlab{b}}.
\newblock Accessed: 2023-11-28.

\bibitem[Iftitahu(2023{\natexlab{c}})]{sundanese_instruct_stories}
Iftitahu.
\newblock {Sudanese Instruct Stories}.
\newblock \url{https://huggingface.co/datasets/Iftitahu/sundanese_instruct_stories}, 2023{\natexlab{c}}.
\newblock Accessed: 2023-11-28.

\bibitem[Ilhomovna \& Yuldasheva(2023)Ilhomovna and Yuldasheva]{ilhomovna2023you}
Amirova~Gulruh Ilhomovna and S~Yuldasheva.
\newblock You have got to know your language to understand your culture.
\newblock In \emph{Interdiscipline Innovation and Scientific Research Conference}, volume~1, pp.\  103--106, 2023.

\bibitem[Iyer et~al.(2012)Iyer, Dandekar, and Csernai]{qqp}
Shankar Iyer, Nikhil Dandekar, and Korn{\"{a}}l Csernai.
\newblock Quora question pairs, 2012.

\bibitem[Iyer et~al.(2022)Iyer, Lin, Pasunuru, Mihaylov, Simig, Yu, Shuster, Wang, Liu, Koura, et~al.]{iyer2022opt}
Srinivasan Iyer, Xi~Victoria Lin, Ramakanth Pasunuru, Todor Mihaylov, Daniel Simig, Ping Yu, Kurt Shuster, Tianlu Wang, Qing Liu, Punit~Singh Koura, et~al.
\newblock Opt-iml: Scaling language model instruction meta learning through the lens of generalization.
\newblock \emph{arXiv preprint arXiv:2212.12017}, 2022.

\bibitem[Ji et~al.(2023{\natexlab{a}})Ji, Ji, Bouillon, and Seligman]{ji_ji_bouillon_seligman_2023}
Meng Ji, Meng Ji, Pierrette Bouillon, and Mark Seligman.
\newblock \emph{Cultural and Linguistic Bias of Neural Machine Translation Technology}, pp.\  100–128.
\newblock Studies in Natural Language Processing. Cambridge University Press, 2023{\natexlab{a}}.
\newblock \doi{10.1017/9781108938976.005}.

\bibitem[Ji et~al.(2023{\natexlab{b}})Ji, Gong, Deng, Peng, Niu, Ma, and Li]{ji2023better}
Yunjie Ji, Yan Gong, Yong Deng, Yiping Peng, Qiang Niu, Baochang Ma, and Xiangang Li.
\newblock Towards better instruction following language models for chinese: Investigating the impact of training data and evaluation, 2023{\natexlab{b}}.

\bibitem[jjzha(2023)]{imdb-dutch-instruct}
jjzha.
\newblock {IMDB Dutch Instruct}.
\newblock \url{https://huggingface.co/datasets/jjzha/imdb-dutch-instruct}, 2023.
\newblock Accessed: 2023-11-28.

\bibitem[J{\o}rgensen et~al.(2015)J{\o}rgensen, Hovy, and S{\o}gaard]{jorgensen-etal-2015-challenges}
Anna J{\o}rgensen, Dirk Hovy, and Anders S{\o}gaard.
\newblock Challenges of studying and processing dialects in social media.
\newblock In \emph{Proceedings of the Workshop on Noisy User-generated Text}, pp.\  9--18, Beijing, China, July 2015. Association for Computational Linguistics.
\newblock \doi{10.18653/v1/W15-4302}.
\newblock URL \url{https://aclanthology.org/W15-4302}.

\bibitem[{Joseph Cheung}(2023)]{joseph_cheung_2023}
{Joseph Cheung}.
\newblock { GuanacoDataset (Revision 8cf0d29) }, 2023.
\newblock URL \url{https://huggingface.co/datasets/JosephusCheung/GuanacoDataset}.

\bibitem[{Joshi} et~al.(2017){Joshi}, {Choi}, {Weld}, and {Zettlemoyer}]{2017arXivtriviaqa}
Mandar {Joshi}, Eunsol {Choi}, Daniel {Weld}, and Luke {Zettlemoyer}.
\newblock {triviaqa: A Large Scale Distantly Supervised Challenge Dataset for Reading Comprehension}.
\newblock \emph{arXiv e-prints}, art. arXiv:1705.03551, 2017.

\bibitem[Joshi et~al.(2019)Joshi, Barnes, Santy, Khanuja, Shah, Srinivasan, Bhattamishra, Sitaram, Choudhury, and Bali]{joshi-etal-2019-unsung}
Pratik Joshi, Christain Barnes, Sebastin Santy, Simran Khanuja, Sanket Shah, Anirudh Srinivasan, Satwik Bhattamishra, Sunayana Sitaram, Monojit Choudhury, and Kalika Bali.
\newblock Unsung challenges of building and deploying language technologies for low resource language communities.
\newblock In \emph{Proceedings of the 16th International Conference on Natural Language Processing}, pp.\  211--219, International Institute of Information Technology, Hyderabad, India, December 2019. NLP Association of India.
\newblock URL \url{https://aclanthology.org/2019.icon-1.25}.

\bibitem[Joshi et~al.(2020)Joshi, Santy, Budhiraja, Bali, and Choudhury]{joshi-etal-2020-state}
Pratik Joshi, Sebastin Santy, Amar Budhiraja, Kalika Bali, and Monojit Choudhury.
\newblock The state and fate of linguistic diversity and inclusion in the {NLP} world.
\newblock In \emph{Proceedings of the 58th Annual Meeting of the Association for Computational Linguistics}, pp.\  6282--6293, Online, July 2020. Association for Computational Linguistics.
\newblock \doi{10.18653/v1/2020.acl-main.560}.
\newblock URL \url{https://aclanthology.org/2020.acl-main.560}.

\bibitem[jxmorris12 et~al.(2023)jxmorris12, thomwolf, lhoestq, and lewtun]{ag_news}
jxmorris12, thomwolf, lhoestq, and lewtun.
\newblock ag_news.
\newblock \url{https://huggingface.co/datasets/ag_news}, 2023.
\newblock Accessed: 2023-11-28.

\bibitem[Kanerva et~al.(2021)Kanerva, Ginter, Chang, Rastas, Skantsi, Kilpel{\"a}inen, Kupari, Saarni, Sev{\'o}n, and Tarkka]{kanerva-etal-2021-finnish}
Jenna Kanerva, Filip Ginter, Li-Hsin Chang, Iiro Rastas, Valtteri Skantsi, Jemina Kilpel{\"a}inen, Hanna-Mari Kupari, Jenna Saarni, Maija Sev{\'o}n, and Otto Tarkka.
\newblock {F}innish paraphrase corpus.
\newblock In \emph{Proceedings of the 23rd Nordic Conference on Computational Linguistics (NoDaLiDa)}, pp.\  288--298, Reykjavik, Iceland (Online), May 31--2 June 2021. Link{\"o}ping University Electronic Press, Sweden.
\newblock URL \url{https://aclanthology.org/2021.nodalida-main.29}.

\bibitem[Khandelwal et~al.(2023)Khandelwal, Tonneau, Bean, Kirk, and Hale]{Khandelwal2023CasteistBN}
Khyati Khandelwal, Manuel Tonneau, Andrew~M. Bean, Hannah~Rose Kirk, and Scott~A. Hale.
\newblock Casteist but not racist? quantifying disparities in large language model bias between india and the west.
\newblock \emph{ArXiv}, abs/2309.08573, 2023.
\newblock URL \url{https://api.semanticscholar.org/CorpusID:262013517}.

\bibitem[Khashabi et~al.(2018)Khashabi, Chaturvedi, Roth, Upadhyay, and Roth]{MultiRC2018}
Daniel Khashabi, Snigdha Chaturvedi, Michael Roth, Shyam Upadhyay, and Dan Roth.
\newblock Looking beyond the surface:a challenge set for reading comprehension over multiple sentences.
\newblock In \emph{Proceedings of North American Chapter of the Association for Computational Linguistics (NAACL)}, 2018.

\bibitem[Khashabi et~al.(2020)Khashabi, Min, Khot, Sabharwal, Tafjord, Clark, and Hajishirzi]{khashabi2020unifiedqa}
Daniel Khashabi, Sewon Min, Tushar Khot, Ashish Sabharwal, Oyvind Tafjord, Peter Clark, and Hannaneh Hajishirzi.
\newblock Unifiedqa: Crossing format boundaries with a single qa system.
\newblock \emph{arXiv preprint arXiv:2005.00700}, 2020.

\bibitem[Khondaker et~al.(2023)Khondaker, Waheed, Nagoudi, and Abdul-Mageed]{khondaker2023gptaraeval}
Md~Tawkat~Islam Khondaker, Abdul Waheed, El~Moatez~Billah Nagoudi, and Muhammad Abdul-Mageed.
\newblock Gptaraeval: A comprehensive evaluation of chatgpt on arabic nlp, 2023.

\bibitem[Khot et~al.(2020)Khot, Clark, Guerquin, Jansen, and Sabharwal]{allenai:qasc}
Tushar Khot, Peter Clark, Michal Guerquin, Peter Jansen, and Ashish Sabharwal.
\newblock Qasc: A dataset for question answering via sentence composition.
\newblock \emph{arXiv:1910.11473v2}, 2020.

\bibitem[Kim et~al.(2022)Kim, Hessel, Jiang, West, Lu, Yu, Zhou, Bras, Alikhani, Kim, Sap, and Choi]{kim2022soda}
Hyunwoo Kim, Jack Hessel, Liwei Jiang, Peter West, Ximing Lu, Youngjae Yu, Pei Zhou, Ronan~Le Bras, Malihe Alikhani, Gunhee Kim, Maarten Sap, and Yejin Choi.
\newblock Soda: Million-scale dialogue distillation with social commonsense contextualization.
\newblock \emph{ArXiv}, abs/2212.10465, 2022.

\bibitem[Kim et~al.(2021)Kim, Maddela, Kriz, Xu, and Callison-Burch]{kim-etal-2021-bisect}
Joongwon Kim, Mounica Maddela, Reno Kriz, Wei Xu, and Chris Callison-Burch.
\newblock {B}i{SECT}: Learning to split and rephrase sentences with bitexts.
\newblock In \emph{Proceedings of the 2021 Conference on Empirical Methods in Natural Language Processing}, pp.\  6193--6209, Online and Punta Cana, Dominican Republic, November 2021. Association for Computational Linguistics.
\newblock \doi{10.18653/v1/2021.emnlp-main.500}.
\newblock URL \url{https://aclanthology.org/2021.emnlp-main.500}.

\bibitem[Kim et~al.(2023)Kim, Joo, Kim, Jang, Ye, Shin, and Seo]{kim2023cot}
Seungone Kim, Se~June Joo, Doyoung Kim, Joel Jang, Seonghyeon Ye, Jamin Shin, and Minjoon Seo.
\newblock The cot collection: Improving zero-shot and few-shot learning of language models via chain-of-thought fine-tuning.
\newblock \emph{arXiv preprint arXiv:2305.14045}, 2023.

\bibitem[K{\"o}pf et~al.(2023)K{\"o}pf, Kilcher, von R{\"u}tte, Anagnostidis, Tam, Stevens, Barhoum, Duc, Stanley, Nagyfi, et~al.]{kopf2023openassistant}
Andreas K{\"o}pf, Yannic Kilcher, Dimitri von R{\"u}tte, Sotiris Anagnostidis, Zhi-Rui Tam, Keith Stevens, Abdullah Barhoum, Nguyen~Minh Duc, Oliver Stanley, Rich{\'a}rd Nagyfi, et~al.
\newblock Openassistant conversations--democratizing large language model alignment.
\newblock \emph{arXiv preprint arXiv:2304.07327}, 2023.

\bibitem[Kotek et~al.(2023)Kotek, Dockum, and Sun]{Kotek2023GenderBA}
Hadas Kotek, Rikker Dockum, and David~Q. Sun.
\newblock Gender bias and stereotypes in large language models.
\newblock \emph{Proceedings of The ACM Collective Intelligence Conference}, 2023.
\newblock URL \url{https://api.semanticscholar.org/CorpusID:261276445}.

\bibitem[Kreutzer et~al.(2022)Kreutzer, Caswell, Wang, Wahab, van Esch, Ulzii-Orshikh, Tapo, Subramani, Sokolov, Sikasote, Setyawan, Sarin, Samb, Sagot, Rivera, Rios, Papadimitriou, Osei, Suarez, Orife, Ogueji, Rubungo, Nguyen, M{\"u}ller, M{\"u}ller, Muhammad, Muhammad, Mnyakeni, Mirzakhalov, Matangira, Leong, Lawson, Kudugunta, Jernite, Jenny, Firat, Dossou, Dlamini, de~Silva, {\c{C}}abuk~Ball{\i}, Biderman, Battisti, Baruwa, Bapna, Baljekar, Azime, Awokoya, Ataman, Ahia, Ahia, Agrawal, and Adeyemi]{kreutzer-etal-2022-quality}
Julia Kreutzer, Isaac Caswell, Lisa Wang, Ahsan Wahab, Daan van Esch, Nasanbayar Ulzii-Orshikh, Allahsera Tapo, Nishant Subramani, Artem Sokolov, Claytone Sikasote, Monang Setyawan, Supheakmungkol Sarin, Sokhar Samb, Beno{\^\i}t Sagot, Clara Rivera, Annette Rios, Isabel Papadimitriou, Salomey Osei, Pedro~Ortiz Suarez, Iroro Orife, Kelechi Ogueji, Andre~Niyongabo Rubungo, Toan~Q. Nguyen, Mathias M{\"u}ller, Andr{\'e} M{\"u}ller, Shamsuddeen~Hassan Muhammad, Nanda Muhammad, Ayanda Mnyakeni, Jamshidbek Mirzakhalov, Tapiwanashe Matangira, Colin Leong, Nze Lawson, Sneha Kudugunta, Yacine Jernite, Mathias Jenny, Orhan Firat, Bonaventure F.~P. Dossou, Sakhile Dlamini, Nisansa de~Silva, Sakine {\c{C}}abuk~Ball{\i}, Stella Biderman, Alessia Battisti, Ahmed Baruwa, Ankur Bapna, Pallavi Baljekar, Israel~Abebe Azime, Ayodele Awokoya, Duygu Ataman, Orevaoghene Ahia, Oghenefego Ahia, Sweta Agrawal, and Mofetoluwa Adeyemi.
\newblock Quality at a glance: An audit of web-crawled multilingual datasets.
\newblock \emph{Transactions of the Association for Computational Linguistics}, 10:\penalty0 50--72, 2022.
\newblock \doi{10.1162/tacl_a_00447}.
\newblock URL \url{https://aclanthology.org/2022.tacl-1.4}.

\bibitem[Krishna(2020)]{KRISHNA202061}
Venni~V. Krishna.
\newblock Open science and its enemies: Challenges for a sustainable science–society social contract.
\newblock \emph{Journal of Open Innovation: Technology, Market, and Complexity}, 6\penalty0 (3):\penalty0 61, 2020.
\newblock ISSN 2199-8531.
\newblock \doi{https://doi.org/10.3390/joitmc6030061}.

\bibitem[Kudugunta et~al.(2023)Kudugunta, Caswell, Zhang, Garcia, Choquette-Choo, Lee, Xin, Kusupati, Stella, Bapna, et~al.]{kudugunta2023madlad}
Sneha Kudugunta, Isaac Caswell, Biao Zhang, Xavier Garcia, Christopher~A Choquette-Choo, Katherine Lee, Derrick Xin, Aditya Kusupati, Romi Stella, Ankur Bapna, et~al.
\newblock Madlad-400: A multilingual and document-level large audited dataset.
\newblock \emph{arXiv preprint arXiv:2309.04662}, 2023.

\bibitem[Kunchukuttan et~al.(2021)Kunchukuttan, Jain, and Kejriwal]{kunchukuttan-etal-2021-large}
Anoop Kunchukuttan, Siddharth Jain, and Rahul Kejriwal.
\newblock A large-scale evaluation of neural machine transliteration for {I}ndic languages.
\newblock In \emph{Proceedings of the 16th Conference of the European Chapter of the Association for Computational Linguistics: Main Volume}, pp.\  3469--3475, Online, April 2021. Association for Computational Linguistics.
\newblock \doi{10.18653/v1/2021.eacl-main.303}.
\newblock URL \url{https://aclanthology.org/2021.eacl-main.303}.

\bibitem[Kwiatkowski et~al.(2019)Kwiatkowski, Palomaki, Redfield, Collins, Parikh, Alberti, Epstein, Polosukhin, Devlin, Lee, Toutanova, Jones, Kelcey, Chang, Dai, Uszkoreit, Le, and Petrov]{kwiatkowski-etal-2019-natural}
Tom Kwiatkowski, Jennimaria Palomaki, Olivia Redfield, Michael Collins, Ankur Parikh, Chris Alberti, Danielle Epstein, Illia Polosukhin, Jacob Devlin, Kenton Lee, Kristina Toutanova, Llion Jones, Matthew Kelcey, Ming-Wei Chang, Andrew~M. Dai, Jakob Uszkoreit, Quoc Le, and Slav Petrov.
\newblock Natural questions: A benchmark for question answering research.
\newblock \emph{Transactions of the Association for Computational Linguistics}, 7:\penalty0 452--466, 2019.
\newblock \doi{10.1162/tacl_a_00276}.
\newblock URL \url{https://aclanthology.org/Q19-1026}.

\bibitem[Köpf et~al.(2023)Köpf, Kilcher, von Rütte, Anagnostidis, Tam, Stevens, Barhoum, Duc, Stanley, Nagyfi, ES, Suri, Glushkov, Dantuluri, Maguire, Schuhmann, Nguyen, and Mattick]{köpf2023openassistant}
Andreas Köpf, Yannic Kilcher, Dimitri von Rütte, Sotiris Anagnostidis, Zhi-Rui Tam, Keith Stevens, Abdullah Barhoum, Nguyen~Minh Duc, Oliver Stanley, Richárd Nagyfi, Shahul ES, Sameer Suri, David Glushkov, Arnav Dantuluri, Andrew Maguire, Christoph Schuhmann, Huu Nguyen, and Alexander Mattick.
\newblock Openassistant conversations -- democratizing large language model alignment, 2023.

\bibitem[Ladhak et~al.(2020)Ladhak, Durmus, Cardie, and McKeown]{ladhak-etal-2020-wikilingua}
Faisal Ladhak, Esin Durmus, Claire Cardie, and Kathleen McKeown.
\newblock {W}iki{L}ingua: A new benchmark dataset for cross-lingual abstractive summarization.
\newblock In \emph{Findings of the Association for Computational Linguistics: EMNLP 2020}, pp.\  4034--4048, Online, November 2020. Association for Computational Linguistics.
\newblock \doi{10.18653/v1/2020.findings-emnlp.360}.
\newblock URL \url{https://aclanthology.org/2020.findings-emnlp.360}.

\bibitem[Lahoti et~al.(2023)Lahoti, Blumm, Ma, Kotikalapudi, Potluri, Tan, Srinivasan, Packer, Beirami, Beutel, and Chen]{lahoti2023improving}
Preethi Lahoti, Nicholas Blumm, Xiao Ma, Raghavendra Kotikalapudi, Sahitya Potluri, Qijun Tan, Hansa Srinivasan, Ben Packer, Ahmad Beirami, Alex Beutel, and Jilin Chen.
\newblock Improving diversity of demographic representation in large language models via collective-critiques and self-voting, 2023.

\bibitem[Lai et~al.(2017)Lai, Xie, Liu, Yang, and Hovy]{lai-etal-2017-race}
Guokun Lai, Qizhe Xie, Hanxiao Liu, Yiming Yang, and Eduard Hovy.
\newblock {RACE}: Large-scale {R}e{A}ding comprehension dataset from examinations.
\newblock In \emph{Proceedings of the 2017 Conference on Empirical Methods in Natural Language Processing}, pp.\  785--794, Copenhagen, Denmark, September 2017. Association for Computational Linguistics.
\newblock \doi{10.18653/v1/D17-1082}.
\newblock URL \url{https://aclanthology.org/D17-1082}.

\bibitem[Lai et~al.(2023)Lai, Van~Nguyen, Ngo, Nguyen, Dernoncourt, Rossi, and Nguyen]{lai2023okapi}
Viet~Dac Lai, Chien Van~Nguyen, Nghia~Trung Ngo, Thuat Nguyen, Franck Dernoncourt, Ryan~A Rossi, and Thien~Huu Nguyen.
\newblock Okapi: Instruction-tuned large language models in multiple languages with reinforcement learning from human feedback.
\newblock \emph{arXiv preprint arXiv:2307.16039}, 2023.

\bibitem[Lample \& Conneau(2019)Lample and Conneau]{lample2019cross}
Guillaume Lample and Alexis Conneau.
\newblock Cross-lingual language model pretraining.
\newblock \emph{arXiv preprint arXiv:1901.07291}, 2019.

\bibitem[Lauren{\c{c}}on et~al.(2022)Lauren{\c{c}}on, Saulnier, Wang, Akiki, Villanova~del Moral, Le~Scao, Von~Werra, Mou, Gonz{\'a}lez~Ponferrada, Nguyen, et~al.]{laurenccon2022bigscience}
Hugo Lauren{\c{c}}on, Lucile Saulnier, Thomas Wang, Christopher Akiki, Albert Villanova~del Moral, Teven Le~Scao, Leandro Von~Werra, Chenghao Mou, Eduardo Gonz{\'a}lez~Ponferrada, Huu Nguyen, et~al.
\newblock The bigscience roots corpus: A 1.6 tb composite multilingual dataset.
\newblock \emph{Advances in Neural Information Processing Systems}, 35:\penalty0 31809--31826, 2022.

\bibitem[Lebret et~al.(2016)Lebret, Grangier, and Auli]{DBLP:journals/corr/LebretGA16}
R{\'{e}}mi Lebret, David Grangier, and Michael Auli.
\newblock Generating text from structured data with application to the biography domain.
\newblock \emph{CoRR}, abs/1603.07771, 2016.
\newblock URL \url{http://arxiv.org/abs/1603.07771}.

\bibitem[Lee et~al.(2023)Lee, Jung, and Oh]{lee-etal-2023-hate}
Nayeon Lee, Chani Jung, and Alice Oh.
\newblock Hate speech classifiers are culturally insensitive.
\newblock In \emph{Proceedings of the First Workshop on Cross-Cultural Considerations in NLP (C3NLP)}, pp.\  35--46, Dubrovnik, Croatia, May 2023. Association for Computational Linguistics.
\newblock URL \url{https://aclanthology.org/2023.c3nlp-1.5}.

\bibitem[Lehmann et~al.(2014)Lehmann, Isele, Jakob, Jentzsch, Kontokostas, Mendes, Hellmann, Morsey, Van~Kleef, Auer, and Bizer]{dbpedia}
Jens Lehmann, Robert Isele, Max Jakob, Anja Jentzsch, Dimitris Kontokostas, Pablo Mendes, Sebastian Hellmann, Mohamed Morsey, Patrick Van~Kleef, Sören Auer, and Christian Bizer.
\newblock Dbpedia - a large-scale, multilingual knowledge base extracted from wikipedia.
\newblock \emph{Semantic Web Journal}, 6, 01 2014.
\newblock \doi{10.3233/SW-140134}.

\bibitem[Lenart-Gansiniec et~al.(2023)Lenart-Gansiniec, Czakon, Su{\l}kowski, and Pocek]{Lenart-Gansiniec2023}
Regina Lenart-Gansiniec, Wojciech Czakon, {\L}ukasz Su{\l}kowski, and Jasna Pocek.
\newblock Understanding crowdsourcing in science.
\newblock \emph{Review of Managerial Science}, 17\penalty0 (8):\penalty0 2797--2830, Nov 2023.
\newblock ISSN 1863-6691.
\newblock \doi{10.1007/s11846-022-00602-z}.
\newblock URL \url{https://doi.org/10.1007/s11846-022-00602-z}.

\bibitem[Lent et~al.(2022)Lent, Ogueji, de~Lhoneux, Ahia, and S{\o}gaard]{lent-etal-2022-creole}
Heather Lent, Kelechi Ogueji, Miryam de~Lhoneux, Orevaoghene Ahia, and Anders S{\o}gaard.
\newblock What a creole wants, what a creole needs.
\newblock In \emph{Proceedings of the Thirteenth Language Resources and Evaluation Conference}, pp.\  6439--6449, Marseille, France, June 2022. European Language Resources Association.
\newblock URL \url{https://aclanthology.org/2022.lrec-1.691}.

\bibitem[Levenshtein et~al.(1966)]{levenshtein1966binary}
Vladimir~I Levenshtein et~al.
\newblock Binary codes capable of correcting deletions, insertions, and reversals.
\newblock In \emph{Soviet physics doklady}, volume~10, pp.\  707--710. Soviet Union, 1966.

\bibitem[Lewis et~al.(2020)Lewis, Oğuz, Rinott, Riedel, and Schwenk]{lewis2020mlqa}
Patrick Lewis, Barlas Oğuz, Ruty Rinott, Sebastian Riedel, and Holger Schwenk.
\newblock Mlqa: Evaluating cross-lingual extractive question answering, 2020.

\bibitem[Li et~al.(2023{\natexlab{a}})Li, Koto, Wu, Aji, and Baldwin]{li2023bactrian}
Haonan Li, Fajri Koto, Minghao Wu, Alham~Fikri Aji, and Timothy Baldwin.
\newblock Bactrian-x: A multilingual replicable instruction-following model with low-rank adaptation.
\newblock \emph{arXiv preprint arXiv:2305.15011}, 2023{\natexlab{a}}.

\bibitem[Li et~al.(2023{\natexlab{b}})Li, Chen, Luo, Kang, Zhang, Hu, Chan, and Song]{Li2023PrivacyIL}
Haoran Li, Yulin Chen, Jinglong Luo, Yan Kang, Xiaojin Zhang, Qi~Hu, Chunkit Chan, and Yangqiu Song.
\newblock Privacy in large language models: Attacks, defenses and future directions.
\newblock \emph{ArXiv}, abs/2310.10383, 2023{\natexlab{b}}.
\newblock URL \url{https://api.semanticscholar.org/CorpusID:264145758}.

\bibitem[Li et~al.(2023{\natexlab{c}})Li, Yin, Li, Chen, Wang, Ren, Li, Yang, Xu, Sun, Kong, and Liu]{li2023m3it}
Lei Li, Yuwei Yin, Shicheng Li, Liang Chen, Peiyi Wang, Shuhuai Ren, Mukai Li, Yazheng Yang, Jingjing Xu, Xu~Sun, Lingpeng Kong, and Qi~Liu.
\newblock M$^3$it: A large-scale dataset towards multi-modal multilingual instruction tuning, 2023{\natexlab{c}}.

\bibitem[Li et~al.(2023{\natexlab{d}})Li, Allal, Zi, Muennighoff, Kocetkov, Mou, Marone, Akiki, Li, Chim, et~al.]{li2023starcoder}
Raymond Li, Loubna~Ben Allal, Yangtian Zi, Niklas Muennighoff, Denis Kocetkov, Chenghao Mou, Marc Marone, Christopher Akiki, Jia Li, Jenny Chim, et~al.
\newblock Starcoder: may the source be with you!
\newblock \emph{arXiv preprint arXiv:2305.06161}, 2023{\natexlab{d}}.

\bibitem[Li \& Roth(2002)Li and Roth]{li-roth-2002-learning}
Xin Li and Dan Roth.
\newblock Learning question classifiers.
\newblock In \emph{{COLING} 2002: The 19th International Conference on Computational Linguistics}, 2002.
\newblock URL \url{https://aclanthology.org/C02-1150}.

\bibitem[Li et~al.(2023{\natexlab{e}})Li, Lin, Zhang, Fu, Chen, Lou, and Chen]{li2023making}
Yifei Li, Zeqi Lin, Shizhuo Zhang, Qiang Fu, Bei Chen, Jian-Guang Lou, and Weizhu Chen.
\newblock Making large language models better reasoners with step-aware verifier, 2023{\natexlab{e}}.

\bibitem[Li et~al.(2022)Li, Zhang, Zhao, Shen, Liu, Mao, and Zhang]{li2022csl}
Yudong Li, Yuqing Zhang, Zhe Zhao, Linlin Shen, Weijie Liu, Weiquan Mao, and Hui Zhang.
\newblock Csl: A large-scale chinese scientific literature dataset, 2022.

\bibitem[Lignos et~al.(2022)Lignos, Holley, Palen-Michel, and S{\"a}lev{\"a}]{lignos-etal-2022-toward}
Constantine Lignos, Nolan Holley, Chester Palen-Michel, and Jonne S{\"a}lev{\"a}.
\newblock Toward more meaningful resources for lower-resourced languages.
\newblock In \emph{Findings of the Association for Computational Linguistics: ACL 2022}, pp.\  523--532, Dublin, Ireland, May 2022. Association for Computational Linguistics.
\newblock \doi{10.18653/v1/2022.findings-acl.44}.
\newblock URL \url{https://aclanthology.org/2022.findings-acl.44}.

\bibitem[Lin et~al.(2020)Lin, Zhou, Shen, Zhou, Bhagavatula, Choi, and Ren]{lin-etal-2020-commongen}
Bill~Yuchen Lin, Wangchunshu Zhou, Ming Shen, Pei Zhou, Chandra Bhagavatula, Yejin Choi, and Xiang Ren.
\newblock {C}ommon{G}en: A constrained text generation challenge for generative commonsense reasoning.
\newblock In \emph{Findings of the Association for Computational Linguistics: EMNLP 2020}, pp.\  1823--1840, Online, November 2020. Association for Computational Linguistics.
\newblock \doi{10.18653/v1/2020.findings-emnlp.165}.
\newblock URL \url{https://aclanthology.org/2020.findings-emnlp.165}.

\bibitem[Lin et~al.(2021)Lin, Lee, Qiao, and Ren]{lin-etal-2021-common}
Bill~Yuchen Lin, Seyeon Lee, Xiaoyang Qiao, and Xiang Ren.
\newblock Common sense beyond {E}nglish: Evaluating and improving multilingual language models for commonsense reasoning.
\newblock In \emph{Proceedings of the 59th Annual Meeting of the Association for Computational Linguistics and the 11th International Joint Conference on Natural Language Processing (Volume 1: Long Papers)}, pp.\  1274--1287, Online, August 2021. Association for Computational Linguistics.
\newblock \doi{10.18653/v1/2021.acl-long.102}.
\newblock URL \url{https://aclanthology.org/2021.acl-long.102}.

\bibitem[Lin et~al.(2019)Lin, Tafjord, Clark, and Gardner]{Lin2019ReasoningOP}
Kevin Lin, Oyvind Tafjord, Peter Clark, and Matt Gardner.
\newblock Reasoning over paragraph effects in situations.
\newblock In \emph{EMNLP 2019 MRQA Workshop}, pp.\ ~58, 2019.

\bibitem[Lin et~al.(2022)Lin, Mihaylov, Artetxe, Wang, Chen, Simig, Ott, Goyal, Bhosale, Du, Pasunuru, Shleifer, Koura, Chaudhary, O{'}Horo, Wang, Zettlemoyer, Kozareva, Diab, Stoyanov, and Li]{lin-etal-2022-shot}
Xi~Victoria Lin, Todor Mihaylov, Mikel Artetxe, Tianlu Wang, Shuohui Chen, Daniel Simig, Myle Ott, Naman Goyal, Shruti Bhosale, Jingfei Du, Ramakanth Pasunuru, Sam Shleifer, Punit~Singh Koura, Vishrav Chaudhary, Brian O{'}Horo, Jeff Wang, Luke Zettlemoyer, Zornitsa Kozareva, Mona Diab, Veselin Stoyanov, and Xian Li.
\newblock Few-shot learning with multilingual generative language models.
\newblock In \emph{Proceedings of the 2022 Conference on Empirical Methods in Natural Language Processing}, pp.\  9019--9052, Abu Dhabi, United Arab Emirates, December 2022. Association for Computational Linguistics.
\newblock URL \url{https://aclanthology.org/2022.emnlp-main.616}.

\bibitem[Liu et~al.(2021)Liu, Bugliarello, Ponti, Reddy, Collier, and Elliott]{liu-etal-2021-visually}
Fangyu Liu, Emanuele Bugliarello, Edoardo~Maria Ponti, Siva Reddy, Nigel Collier, and Desmond Elliott.
\newblock Visually grounded reasoning across languages and cultures.
\newblock In \emph{Proceedings of the 2021 Conference on Empirical Methods in Natural Language Processing}, pp.\  10467--10485, Online and Punta Cana, Dominican Republic, November 2021. Association for Computational Linguistics.
\newblock \doi{10.18653/v1/2021.emnlp-main.818}.
\newblock URL \url{https://aclanthology.org/2021.emnlp-main.818}.

\bibitem[Longpre et~al.(2023{\natexlab{a}})Longpre, Hou, Vu, Webson, Chung, Tay, Zhou, Le, Zoph, Wei, and Roberts]{longpre2023flan}
Shayne Longpre, Le~Hou, Tu~Vu, Albert Webson, Hyung~Won Chung, Yi~Tay, Denny Zhou, Quoc~V. Le, Barret Zoph, Jason Wei, and Adam Roberts.
\newblock The flan collection: Designing data and methods for effective instruction tuning.
\newblock In \emph{Proceedings of the 40th International Conference on Machine Learning}, ICML'23. JMLR.org, 2023{\natexlab{a}}.

\bibitem[Longpre et~al.(2023{\natexlab{b}})Longpre, Mahari, Chen, Obeng-Marnu, Sileo, Brannon, Muennighoff, Khazam, Kabbara, Perisetla, Wu, Shippole, Bollacker, Wu, Villa, Pentland, and Hooker]{longpre2023data}
Shayne Longpre, Robert Mahari, Anthony Chen, Naana Obeng-Marnu, Damien Sileo, William Brannon, Niklas Muennighoff, Nathan Khazam, Jad Kabbara, Kartik Perisetla, Xinyi Wu, Enrico Shippole, Kurt Bollacker, Tongshuang Wu, Luis Villa, Sandy Pentland, and Sara Hooker.
\newblock The data provenance initiative: A large scale audit of dataset licensing \& attribution in ai, 2023{\natexlab{b}}.

\bibitem[Longpre et~al.(2023{\natexlab{c}})Longpre, Yauney, Reif, Lee, Roberts, Zoph, Zhou, Wei, Robinson, Mimno, et~al.]{longpre2023pretrainer}
Shayne Longpre, Gregory Yauney, Emily Reif, Katherine Lee, Adam Roberts, Barret Zoph, Denny Zhou, Jason Wei, Kevin Robinson, David Mimno, et~al.
\newblock A pretrainer's guide to training data: Measuring the effects of data age, domain coverage, quality, \& toxicity.
\newblock \emph{arXiv preprint arXiv:2305.13169}, 2023{\natexlab{c}}.

\bibitem[Loukissas(2019)]{Loukissas2019}
Yanni~Alexander Loukissas.
\newblock \emph{All Data Are Local: Thinking Critically in a Data-Driven Society}.
\newblock MIT Press, Cambridge, Massachusetts, 2019.

\bibitem[Lowphansirikul et~al.(2022)Lowphansirikul, Polpanumas, Rutherford, and Nutanong]{lowphansirikul2020scb}
Lalita Lowphansirikul, Charin Polpanumas, Attapol~T Rutherford, and Sarana Nutanong.
\newblock {A large English--Thai parallel corpus from the web and machine-generated text}.
\newblock \emph{Language Resources and Evaluation}, 56\penalty0 (2):\penalty0 477--499, 2022.

\bibitem[Luccioni \& Viviano(2021)Luccioni and Viviano]{luccioni-viviano-2021-whats}
Alexandra Luccioni and Joseph Viviano.
\newblock What{'}s in the box? an analysis of undesirable content in the {C}ommon {C}rawl corpus.
\newblock In \emph{Proceedings of the 59th Annual Meeting of the Association for Computational Linguistics and the 11th International Joint Conference on Natural Language Processing (Volume 2: Short Papers)}, pp.\  182--189, Online, August 2021. Association for Computational Linguistics.
\newblock \doi{10.18653/v1/2021.acl-short.24}.
\newblock URL \url{https://aclanthology.org/2021.acl-short.24}.

\bibitem[{Lucia Specia} et~al.(2010){Lucia Specia}, {Nicola Cancedda}, and {Marc Dymetman}]{Lucia2010Dataset}
{Lucia Specia}, {Nicola Cancedda}, and {Marc Dymetman}.
\newblock A {Dataset} for {Assessing} {Machine} {Translation} {Evaluation} {Metrics}.
\newblock \emph{International Conference on Language Resources and Evaluation}, 2010.

\bibitem[Lucy et~al.(2024)Lucy, Gururangan, Soldaini, Strubell, Bamman, Klein, and Dodge]{lucy2024aboutme}
Li~Lucy, Suchin Gururangan, Luca Soldaini, Emma Strubell, David Bamman, Lauren Klein, and Jesse Dodge.
\newblock Aboutme: {U}sing self-descriptions in webpages to document the effects of english pretraining data filters, 2024.

\bibitem[Lukas et~al.(2023)Lukas, Salem, Sim, Tople, Wutschitz, and Zanella-B'eguelin]{Lukas2023AnalyzingLO}
Nils Lukas, A.~Salem, Robert Sim, Shruti Tople, Lukas Wutschitz, and Santiago Zanella-B'eguelin.
\newblock Analyzing leakage of personally identifiable information in language models.
\newblock \emph{2023 IEEE Symposium on Security and Privacy (SP)}, pp.\  346--363, 2023.
\newblock URL \url{https://api.semanticscholar.org/CorpusID:256459554}.

\bibitem[Luo et~al.(2023)Luo, Xu, Zhao, Sun, Geng, Hu, Tao, Ma, Lin, and Jiang]{luo2023wizardcoder}
Ziyang Luo, Can Xu, Pu~Zhao, Qingfeng Sun, Xiubo Geng, Wenxiang Hu, Chongyang Tao, Jing Ma, Qingwei Lin, and Daxin Jiang.
\newblock Wizardcoder: Empowering code large language models with evol-instruct, 2023.

\bibitem[Luukkonen et~al.(2023)Luukkonen, Komulainen, Luoma, Eskelinen, Kanerva, Kupari, Ginter, Laippala, Muennighoff, Piktus, et~al.]{luukkonen2023fingpt}
Risto Luukkonen, Ville Komulainen, Jouni Luoma, Anni Eskelinen, Jenna Kanerva, Hanna-Mari Kupari, Filip Ginter, Veronika Laippala, Niklas Muennighoff, Aleksandra Piktus, et~al.
\newblock Fingpt: Large generative models for a small language.
\newblock \emph{arXiv preprint arXiv:2311.05640}, 2023.

\bibitem[Maas et~al.(2011)Maas, Daly, Pham, Huang, Ng, and Potts]{maas-EtAl:2011:ACL-HLT2011}
Andrew~L. Maas, Raymond~E. Daly, Peter~T. Pham, Dan Huang, Andrew~Y. Ng, and Christopher Potts.
\newblock Learning word vectors for sentiment analysis.
\newblock In \emph{Proceedings of the 49th Annual Meeting of the Association for Computational Linguistics: Human Language Technologies}, pp.\  142--150, Portland, Oregon, USA, June 2011. Association for Computational Linguistics.
\newblock URL \url{http://www.aclweb.org/anthology/P11-1015}.

\bibitem[Maillard et~al.(2023)Maillard, Gao, Kalbassi, Sadagopan, Goswami, Koehn, Fan, and Guzmán]{seed-23}
Jean Maillard, Cynthia Gao, Elahe Kalbassi, Kaushik~Ram Sadagopan, Vedanuj Goswami, Philipp Koehn, Angela Fan, and Francisco Guzmán.
\newblock Small data, big impact: Leveraging minimal data for effective machine translation.
\newblock In \emph{Proceedings of the 61st Annual Meeting of the Association for Computational Linguistics (Volume 1: Long Papers)}, pp.\  2740--2756, Toronto, Canada, 2023. Association for Computational Linguistics.
\newblock URL \url{https://aclanthology.org/2023.acl-long.154}.

\bibitem[Marion et~al.(2023)Marion, Üstün, Pozzobon, Wang, Fadaee, and Hooker]{marion2023investigating}
Max Marion, Ahmet Üstün, Luiza Pozzobon, Alex Wang, Marzieh Fadaee, and Sara Hooker.
\newblock When less is more: Investigating data pruning for pretraining llms at scale, 2023.

\bibitem[Marivate et~al.(2020)Marivate, Sefara, Chabalala, Makhaya, Mokgonyane, Mokoena, and Modupe]{marivate-etal-2020-investigating}
Vukosi Marivate, Tshephisho Sefara, Vongani Chabalala, Keamogetswe Makhaya, Tumisho Mokgonyane, Rethabile Mokoena, and Abiodun Modupe.
\newblock Investigating an approach for low resource language dataset creation, curation and classification: Setswana and sepedi.
\newblock In \emph{Proceedings of the first workshop on Resources for African Indigenous Languages}, pp.\  15--20, Marseille, France, May 2020. European Language Resources Association (ELRA).
\newblock ISBN 979-10-95546-60-3.
\newblock URL \url{https://aclanthology.org/2020.rail-1.3}.

\bibitem[maxbartolo(2023{\natexlab{a}})]{adversarial_qa_dbert}
maxbartolo.
\newblock {AdversarialQA D(BERT)}.
\newblock \url{https://huggingface.co/datasets/adversarial_qa/viewer/dbert}, 2023{\natexlab{a}}.
\newblock Accessed: 2023-11-28.

\bibitem[maxbartolo(2023{\natexlab{b}})]{adversarial_qa_dbidaf}
maxbartolo.
\newblock {AdversarialQA D(BiDAF)}.
\newblock \url{https://huggingface.co/datasets/adversarial_qa/viewer/dbidaf}, 2023{\natexlab{b}}.
\newblock Accessed: 2023-11-28.

\bibitem[maxbartolo(2023{\natexlab{c}})]{adversarial_qa_droberta}
maxbartolo.
\newblock {AdversarialQA D(RoBERTa)}.
\newblock \url{https://huggingface.co/datasets/adversarial_qa/viewer/droberta}, 2023{\natexlab{c}}.
\newblock Accessed: 2023-11-28.

\bibitem[Maxwell \& Hughes(2006)Maxwell and Hughes]{maxwell2006frontiers}
Mike Maxwell and Baden Hughes.
\newblock Frontiers in linguistic annotation for lower-density languages.
\newblock In \emph{Proceedings of the workshop on frontiers in linguistically annotated corpora 2006}, pp.\  29--37, 2006.

\bibitem[Mayhew et~al.(2023)Mayhew, Blevins, Liu, Šuppa, Gonen, Imperial, Karlsson, Lin, Ljubešić, Miranda, Plank, Riabi, and Pinter]{mayhew2023universal}
Stephen Mayhew, Terra Blevins, Shuheng Liu, Marek Šuppa, Hila Gonen, Joseph~Marvin Imperial, Börje~F. Karlsson, Peiqin Lin, Nikola Ljubešić, LJ~Miranda, Barbara Plank, Arij Riabi, and Yuval Pinter.
\newblock {Universal NER: A Gold-Standard Multilingual Named Entity Recognition Benchmark}.
\newblock \emph{arXiv preprint arXiv:2311.09122}, 2023.

\bibitem[McCann et~al.(2018)McCann, Keskar, Xiong, and Socher]{mccann2018natural}
Bryan McCann, Nitish~Shirish Keskar, Caiming Xiong, and Richard Socher.
\newblock The natural language decathlon: Multitask learning as question answering.
\newblock \emph{arXiv preprint arXiv:1806.08730}, 2018.

\bibitem[Mehrabi et~al.(2021)Mehrabi, Morstatter, Saxena, Lerman, and Galstyan]{mehrabi2021survey}
Ninareh Mehrabi, Fred Morstatter, Nripsuta Saxena, Kristina Lerman, and Aram Galstyan.
\newblock A survey on bias and fairness in machine learning.
\newblock \emph{ACM computing surveys (CSUR)}, 54\penalty0 (6):\penalty0 1--35, 2021.

\bibitem[Mihaylov et~al.(2018)Mihaylov, Clark, Khot, and Sabharwal]{OpenBookQA2018}
Todor Mihaylov, Peter Clark, Tushar Khot, and Ashish Sabharwal.
\newblock Can a suit of armor conduct electricity? a new dataset for open book question answering.
\newblock In \emph{EMNLP}, 2018.

\bibitem[Min et~al.(2021)Min, Lewis, Zettlemoyer, and Hajishirzi]{min2021metaicl}
Sewon Min, Mike Lewis, Luke Zettlemoyer, and Hannaneh Hajishirzi.
\newblock Metaicl: Learning to learn in context.
\newblock \emph{arXiv preprint arXiv:2110.15943}, 2021.

\bibitem[Mishra et~al.(2022)Mishra, Khashabi, Baral, and Hajishirzi]{naturalinstructions}
Swaroop Mishra, Daniel Khashabi, Chitta Baral, and Hannaneh Hajishirzi.
\newblock Cross-task generalization via natural language crowdsourcing instructions.
\newblock In \emph{ACL}, 2022.

\bibitem[Moran \& Chiarcos(2020)Moran and Chiarcos]{moran20204}
Steven Moran and Christian Chiarcos.
\newblock 4 linguistic linked open data and under-resourced languages: From collection to application.
\newblock \emph{Development of Linguistic Linked Open Data Resources for Collaborative Data-Intensive Research in the Language Sciences}, pp.\ ~39, 2020.

\bibitem[Morschheuser et~al.(2017)Morschheuser, Hamari, Koivisto, and Maedche]{Morschheuser2017}
Benedikt Morschheuser, Juho Hamari, Jonna Koivisto, and Alexander Maedche.
\newblock Gamified crowdsourcing: Conceptualization, literature review, and future agenda.
\newblock \emph{International Journal of Human-Computer Studies}, 106:\penalty0 26--43, 2017.
\newblock ISSN 1071-5819.
\newblock \doi{https://doi.org/10.1016/j.ijhcs.2017.04.005}.
\newblock URL \url{https://www.sciencedirect.com/science/article/pii/S1071581917300642}.

\bibitem[Muennighoff et~al.(2023{\natexlab{a}})Muennighoff, Liu, Zebaze, Zheng, Hui, Zhuo, Singh, Tang, von Werra, and Longpre]{muennighoff2023octopack}
Niklas Muennighoff, Qian Liu, Armel Zebaze, Qinkai Zheng, Binyuan Hui, Terry~Yue Zhuo, Swayam Singh, Xiangru Tang, Leandro von Werra, and Shayne Longpre.
\newblock Octopack: Instruction tuning code large language models.
\newblock \emph{arXiv preprint arXiv:2308.07124}, 2023{\natexlab{a}}.

\bibitem[Muennighoff et~al.(2023{\natexlab{b}})Muennighoff, Rush, Barak, Scao, Piktus, Tazi, Pyysalo, Wolf, and Raffel]{muennighoff2023scaling}
Niklas Muennighoff, Alexander~M Rush, Boaz Barak, Teven~Le Scao, Aleksandra Piktus, Nouamane Tazi, Sampo Pyysalo, Thomas Wolf, and Colin Raffel.
\newblock Scaling data-constrained language models.
\newblock \emph{arXiv preprint arXiv:2305.16264}, 2023{\natexlab{b}}.

\bibitem[Muennighoff et~al.(2023{\natexlab{c}})Muennighoff, Wang, Sutawika, Roberts, Biderman, Le~Scao, Bari, Shen, Yong, Schoelkopf, Tang, Radev, Aji, Almubarak, Albanie, Alyafeai, Webson, Raff, and Raffel]{muennighoff-etal-2023-crosslingual}
Niklas Muennighoff, Thomas Wang, Lintang Sutawika, Adam Roberts, Stella Biderman, Teven Le~Scao, M~Saiful Bari, Sheng Shen, Zheng~Xin Yong, Hailey Schoelkopf, Xiangru Tang, Dragomir Radev, Alham~Fikri Aji, Khalid Almubarak, Samuel Albanie, Zaid Alyafeai, Albert Webson, Edward Raff, and Colin Raffel.
\newblock Crosslingual generalization through multitask finetuning.
\newblock In \emph{Proceedings of the 61st Annual Meeting of the Association for Computational Linguistics (Volume 1: Long Papers)}, pp.\  15991--16111, Toronto, Canada, July 2023{\natexlab{c}}. Association for Computational Linguistics.
\newblock URL \url{https://aclanthology.org/2023.acl-long.891}.

\bibitem[Muhammad et~al.(2023)Muhammad, Abdulmumin, Ayele, Ousidhoum, Adelani, Yimam, Ahmad, Beloucif, Mohammad, Ruder, Hourrane, Jorge, Brazdil, Ali, David, Osei, Shehu-Bello, Lawan, Gwadabe, Rutunda, Belay, Messelle, Balcha, Chala, Gebremichael, Opoku, and Arthur]{muhammad-etal-2023-afrisenti}
Shamsuddeen Muhammad, Idris Abdulmumin, Abinew Ayele, Nedjma Ousidhoum, David Adelani, Seid Yimam, Ibrahim Ahmad, Meriem Beloucif, Saif Mohammad, Sebastian Ruder, Oumaima Hourrane, Alipio Jorge, Pavel Brazdil, Felermino Ali, Davis David, Salomey Osei, Bello Shehu-Bello, Falalu Lawan, Tajuddeen Gwadabe, Samuel Rutunda, Tadesse Belay, Wendimu Messelle, Hailu Balcha, Sisay Chala, Hagos Gebremichael, Bernard Opoku, and Stephen Arthur.
\newblock {A}fri{S}enti: A {T}witter sentiment analysis benchmark for {A}frican languages.
\newblock In Houda Bouamor, Juan Pino, and Kalika Bali (eds.), \emph{Proceedings of the 2023 Conference on Empirical Methods in Natural Language Processing}, pp.\  13968--13981, Singapore, December 2023. Association for Computational Linguistics.
\newblock \doi{10.18653/v1/2023.emnlp-main.862}.
\newblock URL \url{https://aclanthology.org/2023.emnlp-main.862}.

\bibitem[Myers-Scotton(2017)]{myers2017code}
Carol Myers-Scotton.
\newblock Code-switching.
\newblock \emph{The handbook of sociolinguistics}, pp.\  217--237, 2017.

\bibitem[Naik et~al.(2023)Naik, Chandrasekaran, Yuksekgonul, Palangi, and Nushi]{naik2023diversity}
Ranjita Naik, Varun Chandrasekaran, Mert Yuksekgonul, Hamid Palangi, and Besmira Nushi.
\newblock Diversity of thought improves reasoning abilities of large language models, 2023.

\bibitem[Nakamura et~al.(2023)Nakamura, Soares, Pillar, Diniz-Filho, and Duarte]{Nakamura2023}
Gabriel Nakamura, Bruno Soares, Valério Pillar, José Diniz-Filho, and Leandro Duarte.
\newblock Three pathways to better recognize the expertise of global south researchers.
\newblock \emph{npj Biodiversity}, 08 2023.
\newblock \doi{10.1038/s44185-023-00021-7}.

\bibitem[Nallapati et~al.(2016)Nallapati, Zhou, dos santos, Gulcehre, and Xiang]{nallapati2016abstractive}
Ramesh Nallapati, Bowen Zhou, Cicero~Nogueira dos santos, Caglar Gulcehre, and Bing Xiang.
\newblock Abstractive text summarization using sequence-to-sequence rnns and beyond, 2016.

\bibitem[Naous et~al.(2023)Naous, Ryan, and Xu]{Naous2023HavingBA}
Tarek Naous, Michael~Joseph Ryan, and Wei Xu.
\newblock Having beer after prayer? measuring cultural bias in large language models.
\newblock \emph{ArXiv}, abs/2305.14456, 2023.
\newblock URL \url{https://api.semanticscholar.org/CorpusID:258865272}.

\bibitem[Narayan et~al.(2018)Narayan, Cohen, and Lapata]{Narayan2018DontGM}
Shashi Narayan, Shay~B. Cohen, and Mirella Lapata.
\newblock Don't give me the details, just the summary! topic-aware convolutional neural networks for extreme summarization.
\newblock \emph{ArXiv}, abs/1808.08745, 2018.

\bibitem[Nasr et~al.(2023)Nasr, Carlini, Hayase, Jagielski, Cooper, Ippolito, Choquette-Choo, Wallace, Tramèr, and Lee]{nasr2023scalable}
Milad Nasr, Nicholas Carlini, Jonathan Hayase, Matthew Jagielski, A.~Feder Cooper, Daphne Ippolito, Christopher~A. Choquette-Choo, Eric Wallace, Florian Tramèr, and Katherine Lee.
\newblock Scalable extraction of training data from (production) language models, 2023.

\bibitem[Nguyen et~al.(2023)Nguyen, Suri, Tsui, and Schuhmann]{oig2023}
Huu Nguyen, Sameer Suri, Ken Tsui, and Christoph Schuhmann.
\newblock The open instruction generalist (oig) dataset.
\newblock \url{https://laion.ai/blog/oig-dataset/}, 2023.

\bibitem[Ni et~al.(2023)Ni, Xue, Jain, Shah, Zheng, and You]{instructionwild}
Jinjie Ni, Fuzhao Xue, Kabir Jain, Mahir~Hitesh Shah, Zangwei Zheng, and Yang You.
\newblock Instruction in the wild: A user-based instruction dataset.
\newblock \url{https://github.com/XueFuzhao/InstructionWild}, 2023.

\bibitem[Nicholas \& Bhatia(2023)Nicholas and Bhatia]{nicholas2023lost}
Gabriel Nicholas and Aliya Bhatia.
\newblock Lost in translation: Large language models in non-english content analysis, 2023.

\bibitem[Niklaus et~al.(2023)Niklaus, Matoshi, St{\"u}rmer, Chalkidis, and Ho]{Niklaus2023MultiLegalPile}
Joel Niklaus, Veton Matoshi, Matthias St{\"u}rmer, Ilias Chalkidis, and Daniel~E Ho.
\newblock Multilegalpile: A 689gb multilingual legal corpus.
\newblock \emph{arXiv preprint arXiv:2306.02069}, 2023.

\bibitem[NLLB-Team et~al.(2022)NLLB-Team, Costa-jussà, Cross, Çelebi, Elbayad, Heafield, Heffernan, Kalbassi, Lam, Licht, Maillard, Sun, Wang, Wenzek, Youngblood, Akula, Barrault, Gonzalez, Hansanti, Hoffman, Jarrett, Sadagopan, Rowe, Spruit, Tran, Andrews, Ayan, Bhosale, Edunov, Fan, Gao, Goswami, Guzmán, Koehn, Mourachko, Ropers, Saleem, Schwenk, and Wang]{nllbteam2022language}
NLLB-Team, Marta~R. Costa-jussà, James Cross, Onur Çelebi, Maha Elbayad, Kenneth Heafield, Kevin Heffernan, Elahe Kalbassi, Janice Lam, Daniel Licht, Jean Maillard, Anna Sun, Skyler Wang, Guillaume Wenzek, Al~Youngblood, Bapi Akula, Loic Barrault, Gabriel~Mejia Gonzalez, Prangthip Hansanti, John Hoffman, Semarley Jarrett, Kaushik~Ram Sadagopan, Dirk Rowe, Shannon Spruit, Chau Tran, Pierre Andrews, Necip~Fazil Ayan, Shruti Bhosale, Sergey Edunov, Angela Fan, Cynthia Gao, Vedanuj Goswami, Francisco Guzmán, Philipp Koehn, Alexandre Mourachko, Christophe Ropers, Safiyyah Saleem, Holger Schwenk, and Jeff Wang.
\newblock No language left behind: Scaling human-centered machine translation, 2022.

\bibitem[Nogara et~al.(2023)Nogara, Pierri, Cresci, Luceri, Törnberg, and Giordano]{nogara2023toxic}
Gianluca Nogara, Francesco Pierri, Stefano Cresci, Luca Luceri, Petter Törnberg, and Silvia Giordano.
\newblock Toxic bias: Perspective api misreads german as more toxic, 2023.

\bibitem[Ogundepo et~al.(2023)Ogundepo, Gwadabe, Rivera, Clark, Ruder, Adelani, Dossou, Diop, Sikasote, Hacheme, Buzaaba, Ezeani, Mabuya, Osei, Emezue, Kahira, Muhammad, Oladipo, Owodunni, Tonja, Shode, Asai, Aremu, Awokoya, Opoku, Chukwuneke, Mwase, Siro, Arthur, Ajayi, Otiende, Rubungo, Sinkala, Ajisafe, Onwuegbuzia, Lawan, Ahmad, Alabi, Mbonu, Adeyemi, Phiri, Ahia, Iro, and Adhiambo]{ogundepo-etal-2023-cross}
Odunayo Ogundepo, Tajuddeen Gwadabe, Clara Rivera, Jonathan Clark, Sebastian Ruder, David Adelani, Bonaventure Dossou, Abdou Diop, Claytone Sikasote, Gilles Hacheme, Happy Buzaaba, Ignatius Ezeani, Rooweither Mabuya, Salomey Osei, Chris Emezue, Albert Kahira, Shamsuddeen Muhammad, Akintunde Oladipo, Abraham Owodunni, Atnafu Tonja, Iyanuoluwa Shode, Akari Asai, Anuoluwapo Aremu, Ayodele Awokoya, Bernard Opoku, Chiamaka Chukwuneke, Christine Mwase, Clemencia Siro, Stephen Arthur, Tunde Ajayi, Verrah Otiende, Andre Rubungo, Boyd Sinkala, Daniel Ajisafe, Emeka Onwuegbuzia, Falalu Lawan, Ibrahim Ahmad, Jesujoba Alabi, Chinedu Mbonu, Mofetoluwa Adeyemi, Mofya Phiri, Orevaoghene Ahia, Ruqayya Iro, and Sonia Adhiambo.
\newblock Afriqa: Cross-lingual open-retrieval question answering for african languages.
\newblock In Houda Bouamor, Juan Pino, and Kalika Bali (eds.), \emph{Findings of the Association for Computational Linguistics: EMNLP 2023}, pp.\  14957--14972, Singapore, December 2023. Association for Computational Linguistics.
\newblock \doi{10.18653/v1/2023.findings-emnlp.997}.
\newblock URL \url{https://aclanthology.org/2023.findings-emnlp.997}.

\bibitem[Ojo et~al.(2023)Ojo, Ogueji, Stenetorp, and Adelani]{ojo2023good}
Jessica Ojo, Kelechi Ogueji, Pontus Stenetorp, and David~I. Adelani.
\newblock How good are large language models on african languages?, 2023.

\bibitem[{Ortiz Su{'a}rez} et~al.(2019){Ortiz Su{'a}rez}, Sagot, and Romary]{OrtizSuarezSagotRomary2019}
Pedro~Javier {Ortiz Su{'a}rez}, Benoit Sagot, and Laurent Romary.
\newblock Asynchronous pipelines for processing huge corpora on medium to low resource infrastructures.
\newblock In Piotr Bański, Adrien Barbaresi, Hanno Biber, Evelyn Breiteneder, Simon Clematide, Marc Kupietz, Harald L{"u}ngen, and Caroline Iliadi (eds.), \emph{7th Workshop on the Challenges in the Management of Large Corpora (CMLC-7)}, Proceedings of the Workshop on Challenges in the Management of Large Corpora (CMLC-7) 2019. Cardiff, 22nd July 2019, pp.\  9 -- 16, Mannheim, 2019. Leibniz-Institut f{"u}r Deutsche Sprache.
\newblock \doi{10.14618/ids-pub-9021}.
\newblock URL \url{http://nbn-resolving.de/urn:nbn:de:bsz:mh39-90215}.

\bibitem[osyvokon(2023)]{ua_gec_instruction_tuning}
osyvokon.
\newblock { UA-GEC instruction tuning }.
\newblock \url{https://huggingface.co/datasets/osyvokon/ua_gec_instruction_tuning}, 2023.
\newblock Accessed: 2023-11-28.

\bibitem[Ouyang et~al.(2022)Ouyang, Wu, Jiang, Almeida, Wainwright, Mishkin, Zhang, Agarwal, Slama, Ray, Schulman, Hilton, Kelton, Miller, Simens, Askell, Welinder, Christiano, Leike, and Lowe]{ouyang2022training}
Long Ouyang, Jeff Wu, Xu~Jiang, Diogo Almeida, Carroll~L. Wainwright, Pamela Mishkin, Chong Zhang, Sandhini Agarwal, Katarina Slama, Alex Ray, John Schulman, Jacob Hilton, Fraser Kelton, Luke Miller, Maddie Simens, Amanda Askell, Peter Welinder, Paul Christiano, Jan Leike, and Ryan Lowe.
\newblock Training language models to follow instructions with human feedback, 2022.

\bibitem[Pang \& Lee(2005)Pang and Lee]{Pang+Lee:05a}
Bo~Pang and Lillian Lee.
\newblock Seeing stars: Exploiting class relationships for sentiment categorization with respect to rating scales.
\newblock In \emph{Proceedings of the ACL}, 2005.

\bibitem[Paperno et~al.(2016)Paperno, Kruszewski, Lazaridou, Pham, Bernardi, Pezzelle, Baroni, Boleda, and Fernández]{paperno2016lambada}
Denis Paperno, Germán Kruszewski, Angeliki Lazaridou, Quan~Ngoc Pham, Raffaella Bernardi, Sandro Pezzelle, Marco Baroni, Gemma Boleda, and Raquel Fernández.
\newblock The lambada dataset: Word prediction requiring a broad discourse context, 2016.

\bibitem[Park et~al.(2023)Park, Leahey, and Funk]{Park2023PapersAP}
Michael Park, Erin Leahey, and Russell~J. Funk.
\newblock Papers and patents are becoming less disruptive over time.
\newblock \emph{Nature}, 613:\penalty0 138--144, 2023.
\newblock URL \url{https://api.semanticscholar.org/CorpusID:255466666}.

\bibitem[Peng et~al.(2021)Peng, Mathur, and Narayanan]{peng2021mitigating}
Kenny Peng, Arunesh Mathur, and Arvind Narayanan.
\newblock Mitigating dataset harms requires stewardship: Lessons from 1000 papers.
\newblock \emph{arXiv preprint arXiv:2108.02922}, 2021.

\bibitem[Perez-Beltrachini \& Lapata(2021)Perez-Beltrachini and Lapata]{clads-emnlp}
Laura Perez-Beltrachini and Mirella Lapata.
\newblock Models and datasets for cross-lingual summarisation.
\newblock In \emph{Proceedings of The 2021 Conference on Empirical Methods in Natural Language Processing}, Punta Cana, Dominican Republic, 2021.

\bibitem[Petroni et~al.(2021)Petroni, Piktus, Fan, Lewis, Yazdani, Cao, Thorne, Jernite, Karpukhin, Maillard, Plachouras, Rockt{\"{a}}schel, and Riedel]{kilt_tasks}
Fabio Petroni, Aleksandra Piktus, Angela Fan, Patrick S.~H. Lewis, Majid Yazdani, Nicola~De Cao, James Thorne, Yacine Jernite, Vladimir Karpukhin, Jean Maillard, Vassilis Plachouras, Tim Rockt{\"{a}}schel, and Sebastian Riedel.
\newblock {KILT:} a benchmark for knowledge intensive language tasks.
\newblock In Kristina Toutanova, Anna Rumshisky, Luke Zettlemoyer, Dilek Hakkani{-}T{\"{u}}r, Iz~Beltagy, Steven Bethard, Ryan Cotterell, Tanmoy Chakraborty, and Yichao Zhou (eds.), \emph{Proceedings of the 2021 Conference of the North American Chapter of the Association for Computational Linguistics: Human Language Technologies, {NAACL-HLT} 2021, Online, June 6-11, 2021}, pp.\  2523--2544. Association for Computational Linguistics, 2021.
\newblock URL \url{https://www.aclweb.org/anthology/2021.naacl-main.200/}.

\bibitem[Petty(2023)]{Reuters2023a}
Martin Petty.
\newblock {Explainer: Why is Myanmar's military holding an election?}, 2023.
\newblock URL \url{https://www.reuters.com/world/asia-pacific/why-is-myanmars-military-holding-an-election-2023-03-29/}.
\newblock Accessed on Jan. 17, 2024.

\bibitem[Pilehvar \& Camacho-Collados(2019)Pilehvar and Camacho-Collados]{pilehvar2019wic}
Mohammad~Taher Pilehvar and Jose Camacho-Collados.
\newblock Wic: the word-in-context dataset for evaluating context-sensitive meaning representations, 2019.

\bibitem[Pinel~C(2020)]{Pinel2020}
McKevitt~C Pinel~C, Prainsack~B.
\newblock Caring for data: Value creation in a data-intensive research laboratory.
\newblock \emph{Social Studies of Science}, 50\penalty0 (2):\penalty0 175--197, April 2020.

\bibitem[Ponti et~al.(2020)Ponti, Glava{\v{s}}, Majewska, Liu, Vuli{\'c}, and Korhonen]{ponti2020xcopa}
Edoardo~Maria Ponti, Goran Glava{\v{s}}, Olga Majewska, Qianchu Liu, Ivan Vuli{\'c}, and Anna Korhonen.
\newblock Xcopa: A multilingual dataset for causal commonsense reasoning.
\newblock \emph{arXiv preprint arXiv:2005.00333}, 2020.

\bibitem[Popovi{\'c}(2015)]{popovic-2015-chrf}
Maja Popovi{\'c}.
\newblock chr{F}: character n-gram {F}-score for automatic {MT} evaluation.
\newblock In \emph{Proceedings of the Tenth Workshop on Statistical Machine Translation}, pp.\  392--395, Lisbon, Portugal, September 2015. Association for Computational Linguistics.
\newblock \doi{10.18653/v1/W15-3049}.
\newblock URL \url{https://aclanthology.org/W15-3049}.

\bibitem[Post(2018)]{post-2018-call}
Matt Post.
\newblock A call for clarity in reporting {BLEU} scores.
\newblock In \emph{Proceedings of the Third Conference on Machine Translation: Research Papers}, pp.\  186--191, Brussels, Belgium, October 2018. Association for Computational Linguistics.
\newblock \doi{10.18653/v1/W18-6319}.
\newblock URL \url{https://aclanthology.org/W18-6319}.

\bibitem[Pratapa et~al.(2022)Pratapa, Gupta, and Mitamura]{pratapa-etal-2022-multilingual}
Adithya Pratapa, Rishubh Gupta, and Teruko Mitamura.
\newblock Multilingual event linking to {W}ikidata.
\newblock In \emph{Proceedings of the Workshop on Multilingual Information Access (MIA)}, pp.\  37--58, Seattle, USA, July 2022. Association for Computational Linguistics.
\newblock \doi{10.18653/v1/2022.mia-1.5}.
\newblock URL \url{https://aclanthology.org/2022.mia-1.5}.

\bibitem[Pudjiati et~al.(2022)Pudjiati, Lustyantie, Iskandar, and Fitria]{pudjiati2022post}
Danti Pudjiati, Ninuk Lustyantie, Ifan Iskandar, and Tira~Nur Fitria.
\newblock Post-editing of machine translation: Creating a better translation of cultural specific terms.
\newblock \emph{Language Circle: Journal of Language and Literature}, 17\penalty0 (1):\penalty0 61--73, 2022.

\bibitem[Puschmann \& Burgess(2014)Puschmann and Burgess]{Puschmann2014}
Cornelius Puschmann and Jean Burgess.
\newblock Big data, big questions| metaphors of big data.
\newblock \emph{International Journal of Communication}, 8\penalty0 (0), 2014.

\bibitem[Pushkarna et~al.(2022)Pushkarna, Zaldivar, and Kjartansson]{Pushkarna2022}
Mahima Pushkarna, Andrew Zaldivar, and Oddur Kjartansson.
\newblock Data cards: Purposeful and transparent dataset documentation for responsible ai.
\newblock In \emph{Proceedings of the 2022 ACM Conference on Fairness, Accountability, and Transparency}, FAccT '22, pp.\  1776–1826, New York, NY, USA, 2022. Association for Computing Machinery.
\newblock ISBN 9781450393522.
\newblock \doi{10.1145/3531146.3533231}.
\newblock URL \url{https://doi.org/10.1145/3531146.3533231}.

\bibitem[PyThaiNLP(2023{\natexlab{a}})]{pythainlpScbMt2020en2thprompt}
PyThaiNLP.
\newblock scb_mt_2020_en2th_prompt.
\newblock \url{https://huggingface.co/datasets/pythainlp/scb_mt_2020_en2th_prompt}, 2023{\natexlab{a}}.
\newblock Accessed: 2023-11-29.

\bibitem[PyThaiNLP(2023{\natexlab{b}})]{pythainlpScbMt2020th2enprompt}
PyThaiNLP.
\newblock scb_mt_2020_th2en_prompt.
\newblock \url{https://huggingface.co/datasets/pythainlp/scb_mt_2020_th2en_prompt}, 2023{\natexlab{b}}.
\newblock Accessed: 2023-11-29.

\bibitem[PyThaiNLP(2023{\natexlab{c}})]{pythainlpThaiPosPrompt}
PyThaiNLP.
\newblock {Thai-Pos-prompt}.
\newblock \url{https://huggingface.co/datasets/pythainlp/Thai-Pos-prompt}, 2023{\natexlab{c}}.
\newblock Accessed: 2023-11-29.

\bibitem[PyThaiNLP(2023{\natexlab{d}})]{pythainlpThaiUsembassyEn2thPrompt}
PyThaiNLP.
\newblock thai_usembassy_en2th_prompt.
\newblock \url{https://huggingface.co/datasets/pythainlp/thai_usembassy_en2th_prompt}, 2023{\natexlab{d}}.
\newblock Accessed: 2023-11-29.

\bibitem[PyThaiNLP(2023{\natexlab{e}})]{pythainlpThaiUsembassyth2enPrompt}
PyThaiNLP.
\newblock thai_usembassy_th2en_prompt.
\newblock \url{https://huggingface.co/datasets/pythainlp/thai_usembassy_th2en_prompt}, 2023{\natexlab{e}}.
\newblock Accessed: 2023-11-29.

\bibitem[PyThaiNLP(2023{\natexlab{f}})]{thai-wiktionary-prompt}
PyThaiNLP.
\newblock thai-wiktionary-prompt.
\newblock \url{https://huggingface.co/datasets/pythainlp/thai-wiktionary-prompt}, 2023{\natexlab{f}}.
\newblock Accessed: 2023-11-29.

\bibitem[Raganato et~al.(2020)Raganato, Pasini, Camacho-Collados, and Pilehvar]{raganato-etal-2020-xl}
Alessandro Raganato, Tommaso Pasini, Jose Camacho-Collados, and Mohammad~Taher Pilehvar.
\newblock {XL}-{W}i{C}: A multilingual benchmark for evaluating semantic contextualization.
\newblock In \emph{Proceedings of the 2020 Conference on Empirical Methods in Natural Language Processing (EMNLP)}, pp.\  7193--7206, Online, November 2020. Association for Computational Linguistics.
\newblock \doi{10.18653/v1/2020.emnlp-main.584}.
\newblock URL \url{https://aclanthology.org/2020.emnlp-main.584}.

\bibitem[Rajani et~al.(2019)Rajani, McCann, Xiong, and Socher]{rajani2019explain}
Nazneen~Fatema Rajani, Bryan McCann, Caiming Xiong, and Richard Socher.
\newblock Explain yourself! leveraging language models for commonsense reasoning.
\newblock In \emph{Proceedings of the 2019 Conference of the Association for Computational Linguistics (ACL2019)}, 2019.
\newblock URL \url{https://arxiv.org/abs/1906.02361}.

\bibitem[{Rajpurkar} et~al.(2016){Rajpurkar}, {Zhang}, {Lopyrev}, and {Liang}]{2016arXiv160605250R}
Pranav {Rajpurkar}, Jian {Zhang}, Konstantin {Lopyrev}, and Percy {Liang}.
\newblock {SQuAD: 100,000+ Questions for Machine Comprehension of Text}.
\newblock \emph{arXiv e-prints}, art. arXiv:1606.05250, 2016.

\bibitem[Reddy et~al.(2019)Reddy, Chen, and Manning]{reddy-etal-2019-coqa}
Siva Reddy, Danqi Chen, and Christopher~D. Manning.
\newblock {C}o{QA}: A conversational question answering challenge.
\newblock \emph{Transactions of the Association for Computational Linguistics}, 7:\penalty0 249--266, 2019.
\newblock \doi{10.1162/tacl_a_00266}.
\newblock URL \url{https://aclanthology.org/Q19-1016}.

\bibitem[Reimers \& Gurevych(2019)Reimers and Gurevych]{reimers2019sentence}
Nils Reimers and Iryna Gurevych.
\newblock Sentence-bert: Sentence embeddings using siamese bert-networks.
\newblock \emph{arXiv preprint arXiv:1908.10084}, 2019.

\bibitem[Reuters(2023)]{Reuters2023}
Reuters.
\newblock {Explainer: What is happening between Armenia and Azerbaijan over Nagorno-Karabakh?}, 2023.
\newblock URL \url{https://www.reuters.com/world/what-is-happening-between-armenia-azerbaijan-over-nagorno-karabakh-2023-09-19/}.
\newblock Accessed on Jan. 17, 2024.

\bibitem[Rickford et~al.(2012)Rickford, Sweetland, Rickford, and Grano]{rickford2012african}
John~R Rickford, Julie Sweetland, Angela~E Rickford, and Thomas Grano.
\newblock \emph{African American, Creole, and other vernacular Englishes in education: A bibliographic resource}.
\newblock Routledge, 2012.

\bibitem[Robinson et~al.(2023)Robinson, Ogayo, Mortensen, and Neubig]{Robinson2023ChatGPTMC}
Nathaniel~R. Robinson, Perez Ogayo, David~R. Mortensen, and Graham Neubig.
\newblock Chatgpt mt: Competitive for high- (but not low-) resource languages.
\newblock \emph{ArXiv}, abs/2309.07423, 2023.
\newblock URL \url{https://api.semanticscholar.org/CorpusID:261824661}.

\bibitem[Rogers et~al.(2020)Rogers, Kovaleva, Downey, and Rumshisky]{DBLP:conf/aaai/RogersKDR20}
Anna Rogers, Olga Kovaleva, Matthew Downey, and Anna Rumshisky.
\newblock Getting closer to {AI} complete question answering: {A} set of prerequisite real tasks.
\newblock In \emph{The Thirty-Fourth {AAAI} Conference on Artificial Intelligence, {AAAI} 2020, The Thirty-Second Innovative Applications of Artificial Intelligence Conference, {IAAI} 2020, The Tenth {AAAI} Symposium on Educational Advances in Artificial Intelligence, {EAAI} 2020, New York, NY, USA, February 7-12, 2020}, pp.\  8722--8731. {AAAI} Press, 2020.
\newblock URL \url{https://aaai.org/ojs/index.php/AAAI/article/view/6398}.

\bibitem[Rooy(2021)]{vanrooy}
Raf~Van Rooy.
\newblock \emph{Language or Dialect? The History of a Conceptual Pair}.
\newblock Oxford University Press, 2021.

\bibitem[Rush et~al.(2015)Rush, Chopra, and Weston]{Rush_2015}
Alexander~M. Rush, Sumit Chopra, and Jason Weston.
\newblock A neural attention model for abstractive sentence summarization.
\newblock \emph{Proceedings of the 2015 Conference on Empirical Methods in Natural Language Processing}, 2015.
\newblock \doi{10.18653/v1/d15-1044}.
\newblock URL \url{http://dx.doi.org/10.18653/v1/D15-1044}.

\bibitem[Saad-Falcon et~al.(2023)Saad-Falcon, Barrow, Siu, Nenkova, Rossi, and Dernoncourt]{saad2023pdftriage}
Jon Saad-Falcon, Joe Barrow, Alexa Siu, Ani Nenkova, Ryan~A Rossi, and Franck Dernoncourt.
\newblock Pdftriage: Question answering over long, structured documents.
\newblock \emph{arXiv preprint arXiv:2309.08872}, 2023.

\bibitem[Sabou et~al.(2012)Sabou, Bontcheva, and Scharl]{SabouOS}
Marta Sabou, Kalina Bontcheva, and Arno Scharl.
\newblock Crowdsourcing research opportunities: Lessons from natural language processing.
\newblock In \emph{Proceedings of the 12th International Conference on Knowledge Management and Knowledge Technologies}, i-KNOW '12, New York, NY, USA, 2012. Association for Computing Machinery.
\newblock ISBN 9781450312424.
\newblock \doi{10.1145/2362456.2362479}.
\newblock URL \url{https://doi.org/10.1145/2362456.2362479}.

\bibitem[Saha et~al.(2018)Saha, Aralikatte, Khapra, and Sankaranarayanan]{DuoRC}
Amrita Saha, Rahul Aralikatte, Mitesh~M. Khapra, and Karthik Sankaranarayanan.
\newblock {DuoRC: Towards Complex Language Understanding with Paraphrased Reading Comprehension}.
\newblock In \emph{Meeting of the Association for Computational Linguistics (ACL)}, 2018.

\bibitem[Sambasivan et~al.(2021)Sambasivan, Kapania, Highfill, Akrong, Paritosh, and Aroyo]{Sambasivan2021datawork}
Nithya Sambasivan, Shivani Kapania, Hannah Highfill, Diana Akrong, Praveen Paritosh, and Lora~M Aroyo.
\newblock “everyone wants to do the model work, not the data work”: Data cascades in high-stakes ai.
\newblock In \emph{Proceedings of the 2021 CHI Conference on Human Factors in Computing Systems}, CHI '21, New York, NY, USA, 2021. Association for Computing Machinery.
\newblock ISBN 9781450380966.
\newblock \doi{10.1145/3411764.3445518}.
\newblock URL \url{https://doi.org/10.1145/3411764.3445518}.

\bibitem[Sanh et~al.(2022)Sanh, Webson, Raffel, Bach, Sutawika, Alyafeai, Chaffin, Stiegler, Raja, Dey, Bari, Xu, Thakker, Sharma, Szczechla, Kim, Chhablani, Nayak, Datta, Chang, Jiang, Wang, Manica, Shen, Yong, Pandey, Bawden, Wang, Neeraj, Rozen, Sharma, Santilli, Fevry, Fries, Teehan, Scao, Biderman, Gao, Wolf, and Rush]{sanh2022multitask}
Victor Sanh, Albert Webson, Colin Raffel, Stephen Bach, Lintang Sutawika, Zaid Alyafeai, Antoine Chaffin, Arnaud Stiegler, Arun Raja, Manan Dey, M~Saiful Bari, Canwen Xu, Urmish Thakker, Shanya~Sharma Sharma, Eliza Szczechla, Taewoon Kim, Gunjan Chhablani, Nihal Nayak, Debajyoti Datta, Jonathan Chang, Mike Tian-Jian Jiang, Han Wang, Matteo Manica, Sheng Shen, Zheng~Xin Yong, Harshit Pandey, Rachel Bawden, Thomas Wang, Trishala Neeraj, Jos Rozen, Abheesht Sharma, Andrea Santilli, Thibault Fevry, Jason~Alan Fries, Ryan Teehan, Teven~Le Scao, Stella Biderman, Leo Gao, Thomas Wolf, and Alexander~M Rush.
\newblock Multitask prompted training enables zero-shot task generalization.
\newblock In \emph{International Conference on Learning Representations}, 2022.
\newblock URL \url{https://openreview.net/forum?id=9Vrb9D0WI4}.

\bibitem[Sap et~al.(2019)Sap, Rashkin, Chen, LeBras, and Choi]{sap2019socialiqa}
Maarten Sap, Hannah Rashkin, Derek Chen, Ronan LeBras, and Yejin Choi.
\newblock Socialiqa: Commonsense reasoning about social interactions, 2019.

\bibitem[Savoldi et~al.(2021)Savoldi, Gaido, Bentivogli, Negri, and Turchi]{savoldi2021gender}
Beatrice Savoldi, Marco Gaido, Luisa Bentivogli, Matteo Negri, and Marco Turchi.
\newblock Gender bias in machine translation, 2021.

\bibitem[Scao et~al.(2022{\natexlab{a}})Scao, Fan, Akiki, Pavlick, Ili{\'c}, Hesslow, Castagn{\'e}, Luccioni, Yvon, Gall{\'e}, et~al.]{workshop2023bloom}
Teven~Le Scao, Angela Fan, Christopher Akiki, Ellie Pavlick, Suzana Ili{\'c}, Daniel Hesslow, Roman Castagn{\'e}, Alexandra~Sasha Luccioni, Fran{\c{c}}ois Yvon, Matthias Gall{\'e}, et~al.
\newblock Bloom: A 176b-parameter open-access multilingual language model.
\newblock \emph{arXiv preprint arXiv:2211.05100}, 2022{\natexlab{a}}.

\bibitem[Scao et~al.(2022{\natexlab{b}})Scao, Wang, Hesslow, Saulnier, Bekman, Bari, Bideman, Elsahar, Muennighoff, Phang, et~al.]{scao2022language}
Teven~Le Scao, Thomas Wang, Daniel Hesslow, Lucile Saulnier, Stas Bekman, M~Saiful Bari, Stella Bideman, Hady Elsahar, Niklas Muennighoff, Jason Phang, et~al.
\newblock What language model to train if you have one million gpu hours?
\newblock \emph{arXiv preprint arXiv:2210.15424}, 2022{\natexlab{b}}.

\bibitem[Schwartz et~al.(2022)Schwartz, Vassilev, Greene, Perine, Burt, Hall, et~al.]{schwartz2022towards}
Reva Schwartz, Apostol Vassilev, Kristen Greene, Lori Perine, Andrew Burt, Patrick Hall, et~al.
\newblock Towards a standard for identifying and managing bias in artificial intelligence.
\newblock \emph{NIST special publication}, 1270\penalty0 (10.6028), 2022.

\bibitem[Seaver(2021)]{Seaver2021}
Nick Seaver.
\newblock Care and scale: Decorrelative ethics in algorithmic recommendation.
\newblock \emph{Cultural Anthropology}, 36\penalty0 (3):\penalty0 509--537, 2021.

\bibitem[Secretariat(2022)]{secretariat2022international}
UN~Secretariat.
\newblock International decade of indigenous languages, 2022--2032: Global action plan: note/by the secretariat, 2022.

\bibitem[See et~al.(2017)See, Liu, and Manning]{DBLP:journals/corr/SeeLM17}
Abigail See, Peter~J. Liu, and Christopher~D. Manning.
\newblock Get to the point: Summarization with pointer-generator networks.
\newblock \emph{CoRR}, abs/1704.04368, 2017.
\newblock URL \url{http://arxiv.org/abs/1704.04368}.

\bibitem[Sen et~al.(2022)Sen, Aji, and Saffari]{sen-etal-2022-mintaka}
Priyanka Sen, Alham~Fikri Aji, and Amir Saffari.
\newblock Mintaka: A complex, natural, and multilingual dataset for end-to-end question answering.
\newblock In \emph{Proceedings of the 29th International Conference on Computational Linguistics}, pp.\  1604--1619, Gyeongju, Republic of Korea, October 2022. International Committee on Computational Linguistics.
\newblock URL \url{https://aclanthology.org/2022.coling-1.138}.

\bibitem[Sennrich et~al.(2016)Sennrich, Haddow, and Birch]{sennrich-etal-2016-neural}
Rico Sennrich, Barry Haddow, and Alexandra Birch.
\newblock Neural machine translation of rare words with subword units.
\newblock In \emph{Proceedings of the 54th Annual Meeting of the Association for Computational Linguistics (Volume 1: Long Papers)}, pp.\  1715--1725, Berlin, Germany, August 2016. Association for Computational Linguistics.
\newblock \doi{10.18653/v1/P16-1162}.
\newblock URL \url{https://aclanthology.org/P16-1162}.

\bibitem[Shafagh(2023{\natexlab{a}})]{ShafaghAyaPersianInstructionPnSummary}
Shafagh.
\newblock {Aya Persian Instruction pn Summary}.
\newblock \url{https://huggingface.co/datasets/Shafagh/aya_persian_instruction_pn-summary}, 2023{\natexlab{a}}.
\newblock Accessed: 2023-11-28.

\bibitem[Shafagh(2023{\natexlab{b}})]{ShafaghAyaPersianInstructionPnSummaryTitle}
Shafagh.
\newblock {Aya Persian Instruction pn Summary Title}.
\newblock \url{https://huggingface.co/datasets/Shafagh/aya_persian_instruction_pn-summary-title}, 2023{\natexlab{b}}.
\newblock Accessed: 2023-11-28.

\bibitem[Shao et~al.(2019)Shao, Liu, Lai, Tseng, and Tsai]{shao2019drcd}
Chih~Chieh Shao, Trois Liu, Yuting Lai, Yiying Tseng, and Sam Tsai.
\newblock Drcd: a chinese machine reading comprehension dataset, 2019.

\bibitem[Sidnell \& Enfield(2012)Sidnell and Enfield]{Sidnell2012}
Jack Sidnell and N.~J. Enfield.
\newblock Language diversity and social action: A third locus of linguistic relativity.
\newblock \emph{Current Anthropology}, 53\penalty0 (3):\penalty0 302--333, 2012.

\bibitem[Siminyu et~al.(2021)Siminyu, Kalipe, Orlic, Abbott, Marivate, Freshia, Sibal, Neupane, Adelani, Taylor, Ali, Degila, Balogoun, Diop, David, Fourati, Haddad, and Naski]{siminyu2021}
Kathleen Siminyu, Godson Kalipe, Davor Orlic, Jade~Z. Abbott, Vukosi Marivate, Sackey Freshia, Prateek Sibal, Bhanu Neupane, David~Ifeoluwa Adelani, Amelia~V. Taylor, Jamiil~Toure Ali, Kevin Degila, Momboladji Balogoun, Thierno~Ibrahima Diop, Davis David, Chayma Fourati, Hatem Haddad, and Malek Naski.
\newblock {AI4D} - african language program.
\newblock In Kathleen Siminyu, Julia Kreutzer, Hady Elsahar, Vukosi Marivate, Nishant Subramani, Jade~Z. Abbott, and Bernardt Duvenhage (eds.), \emph{2nd AfricaNLP Workshop Proceedings, AfricaNLP@EACL 2021, Virtual Event, April 19, 2021}, 2021.
\newblock URL \url{https://arxiv.org/abs/2104.02516}.

\bibitem[Simons(2019)]{Simons_2019}
Gary~F. Simons.
\newblock Two centuries of spreading language loss.
\newblock \emph{Proceedings of the Linguistic Society of America}, 4\penalty0 (1):\penalty0 27:1–12, Mar. 2019.
\newblock \doi{10.3765/plsa.v4i1.4532}.
\newblock URL \url{https://journals.linguisticsociety.org/proceedings/index.php/PLSA/article/view/4532}.

\bibitem[Singh et~al.(2022)Singh, D'Arcy, Cohan, Downey, and Feldman]{Singh2022SciRepEvalAM}
Amanpreet Singh, Mike D'Arcy, Arman Cohan, Doug Downey, and Sergey Feldman.
\newblock {SciRepEval: A Multi-Format Benchmark for Scientific Document Representations}.
\newblock \emph{ArXiv}, abs/2211.13308, 2022.

\bibitem[Soldaini et~al.(2024)Soldaini, Kinney, Bhagia, Schwenk, Atkinson, Authur, Bogin, Chandu, Dumas, Elazar, Hofmann, Jha, Kumar, Lucy, Lyu, Magnusson, Morrison, Muennighoff, Naik, Nam, Peters, Ravichander, Richardson, Shen, Strubell, Subramani, Tafjord, Walsh, Hajishirzi, Smith, Zettlemoyer, Beltagy, Groeneveld, Dodge, and Lo]{dolma}
Luca Soldaini, Rodney Kinney, Akshita Bhagia, Dustin Schwenk, David Atkinson, Russell Authur, Ben Bogin, Khyathi Chandu, Jennifer Dumas, Yanai Elazar, Valentin Hofmann, Ananya~Harsh Jha, Sachin Kumar, Li~Lucy, Xinxi Lyu, Ian Magnusson, Jacob Morrison, Niklas Muennighoff, Aakanksha Naik, Crystal Nam, Matthew~E. Peters, Abhilasha Ravichander, Kyle Richardson, Zejiang Shen, Emma Strubell, Nishant Subramani, Oyvind Tafjord, Evan~Pete Walsh, Hannaneh Hajishirzi, Noah~A. Smith, Luke Zettlemoyer, Iz~Beltagy, Dirk Groeneveld, Jesse Dodge, and Kyle Lo.
\newblock {Dolma: An Open Corpus of Three Trillion Tokens for Language Model Pretraining Research}.
\newblock \emph{arXiv preprint}, 2024.

\bibitem[Specia \& Farzindar(2010)Specia and Farzindar]{specia-farzindar-2010-estimating}
Lucia Specia and Atefeh Farzindar.
\newblock Estimating machine translation post-editing effort with {HTER}.
\newblock In \emph{Proceedings of the Second Joint EM+/CNGL Workshop: Bringing MT to the User: Research on Integrating MT in the Translation Industry}, pp.\  33--43, Denver, Colorado, USA, November 4 2010. Association for Machine Translation in the Americas.
\newblock URL \url{https://aclanthology.org/2010.jec-1.5}.

\bibitem[Spolsky(2018)]{Spolsky2017}
Bernard Spolsky.
\newblock Language policy in french colonies and after independence.
\newblock \emph{Current Issues in Language Planning}, 19\penalty0 (3):\penalty0 231--315, 2018.
\newblock \doi{10.1080/14664208.2018.1444948}.

\bibitem[Srivastava \& Singh(2021)Srivastava and Singh]{srivastava2021challenges}
Vivek Srivastava and Mayank Singh.
\newblock Challenges and considerations with code-mixed nlp for multilingual societies, 2021.

\bibitem[Stark \& Hoffmann(2019)Stark and Hoffmann]{Stark2019}
Luke Stark and Anna~Lauren Hoffmann.
\newblock Data is the new what? popular metaphors \& professional ethics in emerging data culture.
\newblock \emph{Journal of Cultural Analytics}, 4\penalty0 (1), 2019.

\bibitem[Stiennon et~al.(2020)Stiennon, Ouyang, Wu, Ziegler, Lowe, Voss, Radford, Amodei, and Christiano]{stiennon2020learning}
Nisan Stiennon, Long Ouyang, Jeffrey Wu, Daniel Ziegler, Ryan Lowe, Chelsea Voss, Alec Radford, Dario Amodei, and Paul~F Christiano.
\newblock Learning to summarize with human feedback.
\newblock \emph{Advances in Neural Information Processing Systems}, 33:\penalty0 3008--3021, 2020.

\bibitem[Strassel \& Tracey(2016)Strassel and Tracey]{strassel-tracey-2016-lorelei}
Stephanie Strassel and Jennifer Tracey.
\newblock {LORELEI} language packs: Data, tools, and resources for technology development in low resource languages.
\newblock In \emph{Proceedings of the Tenth International Conference on Language Resources and Evaluation ({LREC}'16)}, pp.\  3273--3280, Portoro{\v{z}}, Slovenia, May 2016. European Language Resources Association (ELRA).
\newblock URL \url{https://aclanthology.org/L16-1521}.

\bibitem[Sun et~al.(2023)Sun, Sellam, Clark, Vu, Dozat, Garrette, Siddhant, Eisenstein, and Gehrmann]{sun-etal-2023-dialect}
Jiao Sun, Thibault Sellam, Elizabeth Clark, Tu~Vu, Timothy Dozat, Dan Garrette, Aditya Siddhant, Jacob Eisenstein, and Sebastian Gehrmann.
\newblock Dialect-robust evaluation of generated text.
\newblock In \emph{Proceedings of the 61st Annual Meeting of the Association for Computational Linguistics (Volume 1: Long Papers)}, pp.\  6010--6028, Toronto, Canada, July 2023. Association for Computational Linguistics.
\newblock URL \url{https://aclanthology.org/2023.acl-long.331}.

\bibitem[SuryaKrishna02(2023{\natexlab{a}})]{SuryaKrishna02AyaTeluguFoodRecipes}
SuryaKrishna02.
\newblock {Aya Telugu Food Recipes}.
\newblock \url{https://huggingface.co/datasets/SuryaKrishna02/aya-telugu-food-recipes}, 2023{\natexlab{a}}.
\newblock Accessed: 2023-11-28.

\bibitem[SuryaKrishna02(2023{\natexlab{b}})]{SuryaKrishna02AyaTeluguJokes}
SuryaKrishna02.
\newblock {Aya Telugu Jokes}.
\newblock \url{https://huggingface.co/datasets/SuryaKrishna02/aya-telugu-jokes}, 2023{\natexlab{b}}.
\newblock Accessed: 2023-11-28.

\bibitem[SuryaKrishna02(2023{\natexlab{c}})]{SuryaKrishna02AyaTeluguNewsArticle}
SuryaKrishna02.
\newblock {Aya Telugu News Articles}.
\newblock \url{https://huggingface.co/datasets/SuryaKrishna02/aya-telugu-news-articles}, 2023{\natexlab{c}}.
\newblock Accessed: 2023-11-28.

\bibitem[SuryaKrishna02(2023{\natexlab{d}})]{SuryaKrishna02AyaTeluguParaphrase}
SuryaKrishna02.
\newblock {Aya Telugu Paraphrase}.
\newblock \url{https://huggingface.co/datasets/SuryaKrishna02/aya-telugu-paraphrase}, 2023{\natexlab{d}}.
\newblock Accessed: 2023-11-28.

\bibitem[SuryaKrishna02(2023{\natexlab{e}})]{SuryaKrishna02AyaTeluguPoems}
SuryaKrishna02.
\newblock {Aya Telugu Poems}.
\newblock \url{https://huggingface.co/datasets/SuryaKrishna02/aya-telugu-poems}, 2023{\natexlab{e}}.
\newblock Accessed: 2023-11-28.

\bibitem[Suzuki et~al.(2023)Suzuki, Hirano, and Sakaji]{Suzuki2023-llmvanilla}
Masahiro Suzuki, Masanori Hirano, and Hiroki Sakaji.
\newblock From base to conversational: Japanese instruction dataset and tuning large language models.
\newblock \emph{arXiv preprint arXiv:2309.03412}, 2023.

\bibitem[syntaxshill(2023)]{syntaxshillArpaAya}
syntaxshill.
\newblock Arpa aya.
\newblock \url{https://huggingface.co/datasets/syntaxshill/arpa-aya}, 2023.
\newblock Accessed: 2023-11-28.

\bibitem[Syvokon et~al.(2023)Syvokon, Nahorna, Kuchmiichuk, and Osidach]{syvokon-etal-2023-ua}
Oleksiy Syvokon, Olena Nahorna, Pavlo Kuchmiichuk, and Nastasiia Osidach.
\newblock {UA}-{GEC}: Grammatical error correction and fluency corpus for the {U}krainian language.
\newblock In \emph{Proceedings of the Second Ukrainian Natural Language Processing Workshop (UNLP)}, pp.\  96--102, Dubrovnik, Croatia, May 2023. Association for Computational Linguistics.
\newblock URL \url{https://aclanthology.org/2023.unlp-1.12}.

\bibitem[Tafjord et~al.(2019{\natexlab{a}})Tafjord, Clark, Gardner, Yih, and Sabharwal]{quarel_v1}
Oyvind Tafjord, Peter Clark, Matt Gardner, Wen-tau Yih, and Ashish Sabharwal.
\newblock Quarel: A dataset and models for answering questions about qualitative relationships.
\newblock In \emph{Proceedings of the AAAI Conference on Artificial Intelligence}, volume~33, pp.\  7063--7071, 2019{\natexlab{a}}.

\bibitem[Tafjord et~al.(2019{\natexlab{b}})Tafjord, Gardner, Lin, and Clark]{quartz}
Oyvind Tafjord, Matt Gardner, Kevin Lin, and Peter Clark.
\newblock Quartz: An open-domain dataset of qualitative relationship questions.
\newblock \emph{arXiv preprint arXiv:1909.03553}, 2019{\natexlab{b}}.

\bibitem[TahmidH(2023)]{TahmidHAnnotatedNewsSummary}
TahmidH.
\newblock {Annotated News Summary}.
\newblock \url{https://huggingface.co/datasets/TahmidH/annotated_news_summary}, 2023.
\newblock Accessed: 2023-11-28.

\bibitem[Tandon et~al.(2019)Tandon, Mishra, Sakaguchi, Bosselut, and Clark]{wiqa}
Niket Tandon, Bhavana~Dalvi Mishra, Keisuke Sakaguchi, Antoine Bosselut, and Peter Clark.
\newblock Wiqa: A dataset for "what if..." reasoning over procedural text.
\newblock \emph{arXiv:1909.04739v1}, 2019.

\bibitem[Tang et~al.(2021)Tang, Tran, Li, Chen, Goyal, Chaudhary, Gu, and Fan]{TangY21}
Yuqing Tang, Chau Tran, Xian Li, Peng-Jen Chen, Naman Goyal, Vishrav Chaudhary, Jiatao Gu, and Angela Fan.
\newblock Multilingual translation from denoising pre-training.
\newblock \emph{Findings of the Association for Computational Linguistics: ACL-IJCNLP}, 2021.
\newblock \doi{10.18653/v1/2021.findings-acl.304}.

\bibitem[Taori et~al.(2023)Taori, Gulrajani, Zhang, Dubois, Li, Guestrin, Liang, and Hashimoto]{alpaca}
Rohan Taori, Ishaan Gulrajani, Tianyi Zhang, Yann Dubois, Xuechen Li, Carlos Guestrin, Percy Liang, and Tatsunori~B. Hashimoto.
\newblock Stanford alpaca: An instruction-following llama model.
\newblock \url{https://github.com/tatsu-lab/stanford_alpaca}, 2023.

\bibitem[{Tellarin.ai}(2023{\natexlab{a}})]{llm-japanese-dataset-vanilla-aya-format}
{Tellarin.ai}.
\newblock {LLM Japanese Dataset Vanilla Aya Format}.
\newblock \url{https://huggingface.co/datasets/tellarin-ai/llm-japanese-dataset-vanilla-aya-format}, 2023{\natexlab{a}}.
\newblock Accessed: 2023-11-28.

\bibitem[{Tellarin.ai}(2023{\natexlab{b}})]{ntx_llm_inst}
{Tellarin.ai}.
\newblock {NTX LLM Instructions}.
\newblock \url{https://huggingface.co/datasets/tellarin-ai/ntx_llm_instructions}, 2023{\natexlab{b}}.
\newblock Accessed: 2023-11-28.

\bibitem[theblackcat102(2023)]{theblackcat102JokeExplaination}
theblackcat102.
\newblock Joke explaination.
\newblock \url{https://huggingface.co/datasets/theblackcat102/joke_explaination}, 2023.
\newblock Accessed: 2023-11-29.

\bibitem[Tiedemann(2012)]{TIEDEMANN12}
Jörg Tiedemann.
\newblock Parallel data, tools and interfaces in opus.
\newblock In Nicoletta Calzolari~(Conference Chair), Khalid Choukri, Thierry Declerck, Mehmet~Ugur Dogan, Bente Maegaard, Joseph Mariani, Jan Odijk, and Stelios Piperidis (eds.), \emph{Proceedings of the Eight International Conference on Language Resources and Evaluation (LREC'12)}, Istanbul, Turkey, may 2012. European Language Resources Association (ELRA).
\newblock ISBN 978-2-9517408-7-7.

\bibitem[Touvron et~al.(2023)Touvron, Martin, Stone, Albert, Almahairi, Babaei, Bashlykov, Batra, Bhargava, Bhosale, Bikel, Blecher, Ferrer, Chen, Cucurull, Esiobu, Fernandes, Fu, Fu, Fuller, Gao, Goswami, Goyal, Hartshorn, Hosseini, Hou, Inan, Kardas, Kerkez, Khabsa, Kloumann, Korenev, Koura, Lachaux, Lavril, Lee, Liskovich, Lu, Mao, Martinet, Mihaylov, Mishra, Molybog, Nie, Poulton, Reizenstein, Rungta, Saladi, Schelten, Silva, Smith, Subramanian, Tan, Tang, Taylor, Williams, Kuan, Xu, Yan, Zarov, Zhang, Fan, Kambadur, Narang, Rodriguez, Stojnic, Edunov, and Scialom]{touvron2023llama}
Hugo Touvron, Louis Martin, Kevin Stone, Peter Albert, Amjad Almahairi, Yasmine Babaei, Nikolay Bashlykov, Soumya Batra, Prajjwal Bhargava, Shruti Bhosale, Dan Bikel, Lukas Blecher, Cristian~Canton Ferrer, Moya Chen, Guillem Cucurull, David Esiobu, Jude Fernandes, Jeremy Fu, Wenyin Fu, Brian Fuller, Cynthia Gao, Vedanuj Goswami, Naman Goyal, Anthony Hartshorn, Saghar Hosseini, Rui Hou, Hakan Inan, Marcin Kardas, Viktor Kerkez, Madian Khabsa, Isabel Kloumann, Artem Korenev, Punit~Singh Koura, Marie-Anne Lachaux, Thibaut Lavril, Jenya Lee, Diana Liskovich, Yinghai Lu, Yuning Mao, Xavier Martinet, Todor Mihaylov, Pushkar Mishra, Igor Molybog, Yixin Nie, Andrew Poulton, Jeremy Reizenstein, Rashi Rungta, Kalyan Saladi, Alan Schelten, Ruan Silva, Eric~Michael Smith, Ranjan Subramanian, Xiaoqing~Ellen Tan, Binh Tang, Ross Taylor, Adina Williams, Jian~Xiang Kuan, Puxin Xu, Zheng Yan, Iliyan Zarov, Yuchen Zhang, Angela Fan, Melanie Kambadur, Sharan Narang, Aurelien Rodriguez, Robert Stojnic, Sergey Edunov, and Thomas
  Scialom.
\newblock Llama 2: Open foundation and fine-tuned chat models, 2023.

\bibitem[Tu et~al.(2019)Tu, Wang, Huang, Tang, He, and Zhou]{tu2019multihop}
Ming Tu, Guangtao Wang, Jing Huang, Yun Tang, Xiaodong He, and Bowen Zhou.
\newblock Multi-hop reading comprehension across multiple documents by reasoning over heterogeneous graphs, 2019.

\bibitem[TurkuNLP(2023)]{TurkuNLPturkuParaphraseCorpus}
TurkuNLP.
\newblock {Turku Paraphrase Corpus}.
\newblock \url{https://huggingface.co/datasets/TurkuNLP/turku_paraphrase_corpus}, 2023.
\newblock Accessed: 2023-11-28.

\bibitem[{Universal NER}(2023)]{uner_llm_inst}
{Universal NER}.
\newblock {UNER LLM Instructions}.
\newblock \url{https://huggingface.co/datasets/universalner/uner_llm_instructions}, 2023.
\newblock Accessed: 2023-11-28.

\bibitem[Urbizu et~al.(2023)Urbizu, San~Vicente, Saralegi, and Corral]{urbizu-etal-2023-enough}
Gorka Urbizu, I{\~n}aki San~Vicente, Xabier Saralegi, and Ander Corral.
\newblock Not enough data to pre-train your language model? {MT} to the rescue!
\newblock In \emph{Findings of the Association for Computational Linguistics: ACL 2023}, pp.\  3826--3836, Toronto, Canada, July 2023. Association for Computational Linguistics.
\newblock URL \url{https://aclanthology.org/2023.findings-acl.235}.

\bibitem[Vanmassenhove et~al.(2021)Vanmassenhove, Shterionov, and Gwilliam]{vanmassenhove-etal-2021-machine}
Eva Vanmassenhove, Dimitar Shterionov, and Matthew Gwilliam.
\newblock Machine translationese: Effects of algorithmic bias on linguistic complexity in machine translation.
\newblock In \emph{Proceedings of the 16th Conference of the European Chapter of the Association for Computational Linguistics: Main Volume}, pp.\  2203--2213, Online, April 2021. Association for Computational Linguistics.
\newblock \doi{10.18653/v1/2021.eacl-main.188}.
\newblock URL \url{https://aclanthology.org/2021.eacl-main.188}.

\bibitem[Vashishtha et~al.(2023)Vashishtha, Ahuja, and Sitaram]{vashishtha2023evaluating}
Aniket Vashishtha, Kabir Ahuja, and Sunayana Sitaram.
\newblock On evaluating and mitigating gender biases in multilingual settings, 2023.

\bibitem[Vigouroux(2013)]{doi:10.1146/annurev-anthro-092611-145804}
C\'{e}cile~B. Vigouroux.
\newblock Francophonie.
\newblock \emph{Annual Review of Anthropology}, 42\penalty0 (1):\penalty0 379--397, 2013.
\newblock \doi{10.1146/annurev-anthro-092611-145804}.
\newblock URL \url{https://doi.org/10.1146/annurev-anthro-092611-145804}.

\bibitem[Wan et~al.(2023)Wan, Wallace, Shen, and Klein]{wan2023poisoning}
Alexander Wan, Eric Wallace, Sheng Shen, and Dan Klein.
\newblock Poisoning language models during instruction tuning, 2023.

\bibitem[Wang et~al.(2018)Wang, Singh, Michael, Hill, Levy, and Bowman]{wang2019glue}
Alex Wang, Amanpreet Singh, Julian Michael, Felix Hill, Omer Levy, and Samuel~R Bowman.
\newblock Glue: A multi-task benchmark and analysis platform for natural language understanding.
\newblock \emph{arXiv preprint arXiv:1804.07461}, 2018.

\bibitem[Wang et~al.(2019)Wang, Pruksachatkun, Nangia, Singh, Michael, Hill, Levy, and Bowman]{wang2019superglue}
Alex Wang, Yada Pruksachatkun, Nikita Nangia, Amanpreet Singh, Julian Michael, Felix Hill, Omer Levy, and Samuel~R Bowman.
\newblock Superglue: A stickier benchmark for general-purpose language understanding systems.
\newblock \emph{arXiv preprint arXiv:1905.00537}, 2019.

\bibitem[Wang et~al.(2023)Wang, Yang, Du, Fan, and Li]{wang2023clinicalgpt}
Guangyu Wang, Guoxing Yang, Zongxin Du, Longjun Fan, and Xiaohu Li.
\newblock Clinicalgpt: Large language models finetuned with diverse medical data and comprehensive evaluation, 2023.

\bibitem[Wang et~al.(2022{\natexlab{a}})Wang, Rubinstein, and Cohn]{wang-etal-2022-measuring}
Jun Wang, Benjamin Rubinstein, and Trevor Cohn.
\newblock Measuring and mitigating name biases in neural machine translation.
\newblock In \emph{Proceedings of the 60th Annual Meeting of the Association for Computational Linguistics (Volume 1: Long Papers)}, pp.\  2576--2590, Dublin, Ireland, May 2022{\natexlab{a}}. Association for Computational Linguistics.
\newblock \doi{10.18653/v1/2022.acl-long.184}.
\newblock URL \url{https://aclanthology.org/2022.acl-long.184}.

\bibitem[Wang et~al.(2022{\natexlab{b}})Wang, Kordi, Mishra, Liu, Smith, Khashabi, and Hajishirzi]{wang2022self}
Yizhong Wang, Yeganeh Kordi, Swaroop Mishra, Alisa Liu, Noah~A Smith, Daniel Khashabi, and Hannaneh Hajishirzi.
\newblock Self-instruct: Aligning language model with self generated instructions.
\newblock \emph{ArXiv preprint}, abs/2212.10560, 2022{\natexlab{b}}.
\newblock URL \url{https://arxiv.org/abs/2212.10560}.

\bibitem[Wang et~al.(2022{\natexlab{c}})Wang, Mishra, Alipoormolabashi, Kordi, Mirzaei, Arunkumar, Ashok, Dhanasekaran, Naik, Stap, Pathak, Karamanolakis, Lai, Purohit, Mondal, Anderson, Kuznia, Doshi, Patel, Pal, Moradshahi, Parmar, Purohit, Varshney, Kaza, Verma, Puri, Karia, Sampat, Doshi, Mishra, Reddy, Patro, Dixit, Shen, Baral, Choi, Smith, Hajishirzi, and Khashabi]{wang2022supernaturalinstructions}
Yizhong Wang, Swaroop Mishra, Pegah Alipoormolabashi, Yeganeh Kordi, Amirreza Mirzaei, Anjana Arunkumar, Arjun Ashok, Arut~Selvan Dhanasekaran, Atharva Naik, David Stap, Eshaan Pathak, Giannis Karamanolakis, Haizhi~Gary Lai, Ishan Purohit, Ishani Mondal, Jacob Anderson, Kirby Kuznia, Krima Doshi, Maitreya Patel, Kuntal~Kumar Pal, Mehrad Moradshahi, Mihir Parmar, Mirali Purohit, Neeraj Varshney, Phani~Rohitha Kaza, Pulkit Verma, Ravsehaj~Singh Puri, Rushang Karia, Shailaja~Keyur Sampat, Savan Doshi, Siddhartha Mishra, Sujan Reddy, Sumanta Patro, Tanay Dixit, Xudong Shen, Chitta Baral, Yejin Choi, Noah~A. Smith, Hannaneh Hajishirzi, and Daniel Khashabi.
\newblock Super-naturalinstructions: Generalization via declarative instructions on 1600+ nlp tasks, 2022{\natexlab{c}}.

\bibitem[Wang et~al.(2022{\natexlab{d}})Wang, Mishra, Alipoormolabashi, Kordi, Mirzaei, Arunkumar, Ashok, Dhanasekaran, Naik, Stap, et~al.]{wang2022super}
Yizhong Wang, Swaroop Mishra, Pegah Alipoormolabashi, Yeganeh Kordi, Amirreza Mirzaei, Anjana Arunkumar, Arjun Ashok, Arut~Selvan Dhanasekaran, Atharva Naik, David Stap, et~al.
\newblock Super-naturalinstructions: Generalization via declarative instructions on 1600+ nlp tasks.
\newblock \emph{arXiv preprint arXiv:2204.07705}, 2022{\natexlab{d}}.

\bibitem[Wang et~al.(2022{\natexlab{e}})Wang, Mishra, Alipoormolabashi, Kordi, Mirzaei, Naik, Ashok, Dhanasekaran, Arunkumar, Stap, Pathak, Karamanolakis, Lai, Purohit, Mondal, Anderson, Kuznia, Doshi, Pal, Patel, Moradshahi, Parmar, Purohit, Varshney, Kaza, Verma, Puri, Karia, Doshi, Sampat, Mishra, Reddy~A, Patro, Dixit, and Shen]{wang-etal-2022-super}
Yizhong Wang, Swaroop Mishra, Pegah Alipoormolabashi, Yeganeh Kordi, Amirreza Mirzaei, Atharva Naik, Arjun Ashok, Arut~Selvan Dhanasekaran, Anjana Arunkumar, David Stap, Eshaan Pathak, Giannis Karamanolakis, Haizhi Lai, Ishan Purohit, Ishani Mondal, Jacob Anderson, Kirby Kuznia, Krima Doshi, Kuntal~Kumar Pal, Maitreya Patel, Mehrad Moradshahi, Mihir Parmar, Mirali Purohit, Neeraj Varshney, Phani~Rohitha Kaza, Pulkit Verma, Ravsehaj~Singh Puri, Rushang Karia, Savan Doshi, Shailaja~Keyur Sampat, Siddhartha Mishra, Sujan Reddy~A, Sumanta Patro, Tanay Dixit, and Xudong Shen.
\newblock Super-{N}atural{I}nstructions: Generalization via declarative instructions on 1600+ {NLP} tasks.
\newblock In \emph{Proceedings of the 2022 Conference on Empirical Methods in Natural Language Processing}, pp.\  5085--5109, Abu Dhabi, United Arab Emirates, December 2022{\natexlab{e}}. Association for Computational Linguistics.
\newblock URL \url{https://aclanthology.org/2022.emnlp-main.340}.

\bibitem[Warstadt et~al.(2018)Warstadt, Singh, and Bowman]{warstadt2018neural}
Alex Warstadt, Amanpreet Singh, and Samuel~R Bowman.
\newblock Neural network acceptability judgments.
\newblock \emph{arXiv preprint arXiv:1805.12471}, 2018.

\bibitem[Wei et~al.(2022{\natexlab{a}})Wei, Bosma, Zhao, Guu, Yu, Lester, Du, Dai, and Le]{wei2022finetuned}
Jason Wei, Maarten Bosma, Vincent Zhao, Kelvin Guu, Adams~Wei Yu, Brian Lester, Nan Du, Andrew~M. Dai, and Quoc~V Le.
\newblock Finetuned language models are zero-shot learners.
\newblock In \emph{International Conference on Learning Representations}, 2022{\natexlab{a}}.
\newblock URL \url{https://openreview.net/forum?id=gEZrGCozdqR}.

\bibitem[Wei et~al.(2022{\natexlab{b}})Wei, Wang, Schuurmans, Bosma, Xia, Chi, Le, Zhou, et~al.]{wei2022chain}
Jason Wei, Xuezhi Wang, Dale Schuurmans, Maarten Bosma, Fei Xia, Ed~Chi, Quoc~V Le, Denny Zhou, et~al.
\newblock Chain-of-thought prompting elicits reasoning in large language models.
\newblock \emph{Advances in Neural Information Processing Systems}, 35:\penalty0 24824--24837, 2022{\natexlab{b}}.

\bibitem[Wei et~al.(2023)Wei, Wei, Lin, Li, Zhang, Ren, Li, Wan, Cao, Xie, et~al.]{wei2023polylm}
Xiangpeng Wei, Haoran Wei, Huan Lin, Tianhao Li, Pei Zhang, Xingzhang Ren, Mei Li, Yu~Wan, Zhiwei Cao, Binbin Xie, et~al.
\newblock Polylm: {A}n open source polyglot large language model.
\newblock \emph{arXiv preprint arXiv:2307.06018}, 2023.

\bibitem[Welbl et~al.(2017)Welbl, Liu, and Gardner]{SciQ}
Johannes Welbl, Nelson~F Liu, and Matt Gardner.
\newblock Crowdsourcing multiple choice science questions.
\newblock \emph{arXiv preprint arXiv:1707.06209}, 2017.

\bibitem[West et~al.(2020)West, Newby, Cheng, Erickson, and Clements]{West2020}
Richard~E West, Timothy Newby, Zui Cheng, Alyssa Erickson, and Kyle Clements.
\newblock Acknowledging all learning: Alternative, micro, and open credentials.
\newblock \emph{Handbook of Research in Educational Communications and Technology: Learning Design}, pp.\  593--613, 2020.

\bibitem[Whitehouse et~al.(2023)Whitehouse, Choudhury, and Aji]{Whitehouse2023LLM}
Chenxi Whitehouse, Monojit Choudhury, and Alham~Fikri Aji.
\newblock Llm-powered data augmentation for enhanced crosslingual performance.
\newblock \emph{arXiv preprint arXiv:2305.14288}, 2023.

\bibitem[Willis(1998)]{willis1998adinkra}
W~Bruce Willis.
\newblock \emph{The Adinkra dictionary: A visual primer on the language of Adinkra}.
\newblock Pyramid Complex, 1998.

\bibitem[Wilson(2023)]{Wilson2023}
Joseph Wilson.
\newblock Voicing an algorithm: trials of strength in artificial intelligence research.
\newblock \emph{Anthropology News}, 2023.
\newblock URL \url{https://www.anthropology-news.org/articles/combat-by-algorithm-trials-of-strength-in-artificial-intelligence-research/}.

\bibitem[Winata et~al.(2023{\natexlab{a}})Winata, Aji, Yong, and Solorio]{winata-etal-2023-decades}
Genta Winata, Alham~Fikri Aji, Zheng~Xin Yong, and Thamar Solorio.
\newblock The decades progress on code-switching research in {NLP}: A systematic survey on trends and challenges.
\newblock In \emph{Findings of the Association for Computational Linguistics: ACL 2023}, pp.\  2936--2978, Toronto, Canada, July 2023{\natexlab{a}}. Association for Computational Linguistics.
\newblock URL \url{https://aclanthology.org/2023.findings-acl.185}.

\bibitem[Winata et~al.(2023{\natexlab{b}})Winata, Aji, Cahyawijaya, Mahendra, Koto, Romadhony, Kurniawan, Moeljadi, Prasojo, Fung, Baldwin, Lau, Sennrich, and Ruder]{winata-etal-2023-nusax}
Genta~Indra Winata, Alham~Fikri Aji, Samuel Cahyawijaya, Rahmad Mahendra, Fajri Koto, Ade Romadhony, Kemal Kurniawan, David Moeljadi, Radityo~Eko Prasojo, Pascale Fung, Timothy Baldwin, Jey~Han Lau, Rico Sennrich, and Sebastian Ruder.
\newblock {N}usa{X}: Multilingual parallel sentiment dataset for 10 {I}ndonesian local languages.
\newblock In \emph{Proceedings of the 17th Conference of the European Chapter of the Association for Computational Linguistics}, pp.\  815--834, Dubrovnik, Croatia, May 2023{\natexlab{b}}. Association for Computational Linguistics.
\newblock URL \url{https://aclanthology.org/2023.eacl-main.57}.

\bibitem[Wittenburg(2021)]{WittOS}
Peter Wittenburg.
\newblock {Open Science and Data Science}.
\newblock \emph{Data Intelligence}, 3\penalty0 (1):\penalty0 95--105, 02 2021.

\bibitem[Witteveen \& Andrews(2019)Witteveen and Andrews]{witteveen2019paraphrasing}
Sam Witteveen and Martin Andrews.
\newblock Paraphrasing with large language models.
\newblock \emph{arXiv preprint arXiv:1911.09661}, 2019.

\bibitem[Wolfram(1997)]{wolfram1997issues}
Walt Wolfram.
\newblock Issues in dialect obsolescence: An introduction.
\newblock \emph{American speech}, 72\penalty0 (1):\penalty0 3--11, 1997.

\bibitem[Wu et~al.(2007)Wu, Gerlach, and Young]{wu2007empirical}
Chorng-Guang Wu, James~H Gerlach, and Clifford~E Young.
\newblock An empirical analysis of open source software developers’ motivations and continuance intentions.
\newblock \emph{Information \& Management}, 44\penalty0 (3):\penalty0 253--262, 2007.

\bibitem[Wu et~al.(2021)Wu, Ouyang, Ziegler, Stiennon, Lowe, Leike, and Christiano]{wu2021recursively}
Jeff Wu, Long Ouyang, Daniel~M Ziegler, Nisan Stiennon, Ryan Lowe, Jan Leike, and Paul Christiano.
\newblock Recursively summarizing books with human feedback.
\newblock \emph{arXiv preprint arXiv:2109.10862}, 2021.

\bibitem[Xie et~al.(2022)Xie, Wu, Shi, Zhong, Scholak, Yasunaga, Wu, Zhong, Yin, Wang, et~al.]{xie2022unifiedskg}
Tianbao Xie, Chen~Henry Wu, Peng Shi, Ruiqi Zhong, Torsten Scholak, Michihiro Yasunaga, Chien-Sheng Wu, Ming Zhong, Pengcheng Yin, Sida~I Wang, et~al.
\newblock Unifiedskg: Unifying and multi-tasking structured knowledge grounding with text-to-text language models.
\newblock \emph{arXiv preprint arXiv:2201.05966}, 2022.

\bibitem[Xu et~al.(2023{\natexlab{a}})Xu, Sun, Zheng, Geng, Zhao, Feng, Tao, and Jiang]{xu2023wizardlm}
Can Xu, Qingfeng Sun, Kai Zheng, Xiubo Geng, Pu~Zhao, Jiazhan Feng, Chongyang Tao, and Daxin Jiang.
\newblock Wizardlm: Empowering large language models to follow complex instructions, 2023{\natexlab{a}}.

\bibitem[Xu et~al.(2023{\natexlab{b}})Xu, Ma, Wang, Xiao, and Chen]{xu2023instructions}
Jiashu Xu, Mingyu~Derek Ma, Fei Wang, Chaowei Xiao, and Muhao Chen.
\newblock Instructions as backdoors: Backdoor vulnerabilities of instruction tuning for large language models, 2023{\natexlab{b}}.

\bibitem[Xu et~al.(2020)Xu, Hu, Zhang, Li, Cao, Li, Xu, Sun, Yu, Yu, Tian, Dong, Liu, Shi, Cui, Li, Zeng, Wang, Xie, Li, Patterson, Tian, Zhang, Zhou, Liu, Zhao, Zhao, Yue, Zhang, Yang, Richardson, and Lan]{xu-etal-2020-clue}
Liang Xu, Hai Hu, Xuanwei Zhang, Lu~Li, Chenjie Cao, Yudong Li, Yechen Xu, Kai Sun, Dian Yu, Cong Yu, Yin Tian, Qianqian Dong, Weitang Liu, Bo~Shi, Yiming Cui, Junyi Li, Jun Zeng, Rongzhao Wang, Weijian Xie, Yanting Li, Yina Patterson, Zuoyu Tian, Yiwen Zhang, He~Zhou, Shaoweihua Liu, Zhe Zhao, Qipeng Zhao, Cong Yue, Xinrui Zhang, Zhengliang Yang, Kyle Richardson, and Zhenzhong Lan.
\newblock {CLUE}: A {C}hinese language understanding evaluation benchmark.
\newblock In \emph{Proceedings of the 28th International Conference on Computational Linguistics}, pp.\  4762--4772, Barcelona, Spain (Online), December 2020. International Committee on Computational Linguistics.
\newblock \doi{10.18653/v1/2020.coling-main.419}.
\newblock URL \url{https://aclanthology.org/2020.coling-main.419}.

\bibitem[Xue et~al.(2021)Xue, Constant, Roberts, Kale, Al-Rfou, Siddhant, Barua, and Raffel]{mt5-2020}
Linting Xue, Noah Constant, Adam Roberts, Mihir Kale, Rami Al-Rfou, Aditya Siddhant, Aditya Barua, and Colin Raffel.
\newblock m{T}5: A massively multilingual pre-trained text-to-text transformer.
\newblock In Kristina Toutanova, Anna Rumshisky, Luke Zettlemoyer, Dilek Hakkani-Tur, Iz~Beltagy, Steven Bethard, Ryan Cotterell, Tanmoy Chakraborty, and Yichao Zhou (eds.), \emph{Proceedings of the 2021 Conference of the North American Chapter of the Association for Computational Linguistics: Human Language Technologies}, pp.\  483--498, Online, June 2021. Association for Computational Linguistics.
\newblock \doi{10.18653/v1/2021.naacl-main.41}.
\newblock URL \url{https://aclanthology.org/2021.naacl-main.41}.

\bibitem[Yang et~al.(2023)Yang, Xie, Peng, Xu, Sun, and Li]{yang2023dataset}
Shuo Yang, Zeke Xie, Hanyu Peng, Min Xu, Mingming Sun, and Ping Li.
\newblock Dataset pruning: Reducing training data by examining generalization influence, 2023.

\bibitem[Yang et~al.(2015)Yang, Yih, and Meek]{yang-etal-2015-wikiqa}
Yi~Yang, Wen-tau Yih, and Christopher Meek.
\newblock {W}iki{QA}: A challenge dataset for open-domain question answering.
\newblock In \emph{Proceedings of the 2015 Conference on Empirical Methods in Natural Language Processing}, pp.\  2013--2018, Lisbon, Portugal, September 2015. Association for Computational Linguistics.
\newblock \doi{10.18653/v1/D15-1237}.
\newblock URL \url{https://aclanthology.org/D15-1237}.

\bibitem[Yang et~al.(2019)Yang, Zhang, Tar, and Baldridge]{pawsx2019emnlp}
Yinfei Yang, Yuan Zhang, Chris Tar, and Jason Baldridge.
\newblock {PAWS-X: A Cross-lingual Adversarial Dataset for Paraphrase Identification}.
\newblock In \emph{Proc. of EMNLP}, 2019.

\bibitem[Yang et~al.(2018)Yang, Qi, Zhang, Bengio, Cohen, Salakhutdinov, and Manning]{yang2018hotpotqa}
Zhilin Yang, Peng Qi, Saizheng Zhang, Yoshua Bengio, William~W. Cohen, Ruslan Salakhutdinov, and Christopher~D. Manning.
\newblock {HotpotQA}: A dataset for diverse, explainable multi-hop question answering.
\newblock In \emph{Conference on Empirical Methods in Natural Language Processing ({EMNLP})}, 2018.

\bibitem[Ye et~al.(2021)Ye, Lin, and Ren]{ye2021crossfit}
Qinyuan Ye, Bill~Yuchen Lin, and Xiang Ren.
\newblock {C}ross{F}it: A few-shot learning challenge for cross-task generalization in {NLP}.
\newblock In \emph{Proceedings of the 2021 Conference on Empirical Methods in Natural Language Processing}, pp.\  7163--7189, Online and Punta Cana, Dominican Republic, 2021. Association for Computational Linguistics.
\newblock \doi{10.18653/v1/2021.emnlp-main.572}.
\newblock URL \url{https://aclanthology.org/2021.emnlp-main.572}.

\bibitem[Yong et~al.(2023{\natexlab{a}})Yong, Menghini, and Bach]{yong2023lowresource}
Zheng~Xin Yong, Cristina Menghini, and Stephen Bach.
\newblock Low-resource languages jailbreak {GPT}-4.
\newblock In \emph{Socially Responsible Language Modelling Research}, 2023{\natexlab{a}}.
\newblock URL \url{https://openreview.net/forum?id=pn83r8V2sv}.

\bibitem[Yong et~al.(2023{\natexlab{b}})Yong, Schoelkopf, Muennighoff, Aji, Adelani, Almubarak, Bari, Sutawika, Kasai, Baruwa, Winata, Biderman, Raff, Radev, and Nikoulina]{yong-etal-2023-bloom}
Zheng~Xin Yong, Hailey Schoelkopf, Niklas Muennighoff, Alham~Fikri Aji, David~Ifeoluwa Adelani, Khalid Almubarak, M~Saiful Bari, Lintang Sutawika, Jungo Kasai, Ahmed Baruwa, Genta Winata, Stella Biderman, Edward Raff, Dragomir Radev, and Vassilina Nikoulina.
\newblock {BLOOM}+1: Adding language support to {BLOOM} for zero-shot prompting.
\newblock In \emph{Proceedings of the 61st Annual Meeting of the Association for Computational Linguistics (Volume 1: Long Papers)}, pp.\  11682--11703, Toronto, Canada, July 2023{\natexlab{b}}. Association for Computational Linguistics.
\newblock URL \url{https://aclanthology.org/2023.acl-long.653}.

\bibitem[Yong et~al.(2023{\natexlab{c}})Yong, Zhang, Forde, Wang, Subramonian, Lovenia, Cahyawijaya, Winata, Sutawika, Cruz, et~al.]{yong2023prompting}
Zheng~Xin Yong, Ruochen Zhang, Jessica~Zosa Forde, Skyler Wang, Arjun Subramonian, Holy Lovenia, Samuel Cahyawijaya, Genta~Indra Winata, Lintang Sutawika, Jan Christian~Blaise Cruz, et~al.
\newblock Prompting multilingual large language models to generate code-mixed texts: The case of south east asian languages.
\newblock In \emph{Sixth Workshop on Computational Approaches to Linguistic Code-Switching}, 2023{\natexlab{c}}.

\bibitem[Yu et~al.(2023)Yu, Zhuang, Zhang, Meng, Ratner, Krishna, Shen, and Zhang]{yu2023large}
Yue Yu, Yuchen Zhuang, Jieyu Zhang, Yu~Meng, Alexander Ratner, Ranjay Krishna, Jiaming Shen, and Chao Zhang.
\newblock Large language model as attributed training data generator: A tale of diversity and bias, 2023.

\bibitem[Yuan et~al.(2021)Yuan, Ippolito, Nikolaev, Callison{-}Burch, Coenen, and Gehrmann]{DBLP:journals/corr/abs-2111-06467}
Ann Yuan, Daphne Ippolito, Vitaly Nikolaev, Chris Callison{-}Burch, Andy Coenen, and Sebastian Gehrmann.
\newblock Synthbio: {A} case study in human-ai collaborative curation of text datasets.
\newblock \emph{CoRR}, abs/2111.06467, 2021.
\newblock URL \url{https://arxiv.org/abs/2111.06467}.

\bibitem[Yusuf(2022)]{Yusuf2022}
Tajudeen Yusuf.
\newblock Politeness in arabic and yoruba: Personal pronouns as a case study.
\newblock \emph{Asian Journal of Language, Literature and Culture Studies}, 5\penalty0 (2):\penalty0 82--88, 2022.

\bibitem[Zampieri et~al.(2017)Zampieri, Malmasi, Ljube{\v{s}}i{\'c}, Nakov, Ali, Tiedemann, Scherrer, and Aepli]{zampieri-etal-2017-findings}
Marcos Zampieri, Shervin Malmasi, Nikola Ljube{\v{s}}i{\'c}, Preslav Nakov, Ahmed Ali, J{\"o}rg Tiedemann, Yves Scherrer, and No{\"e}mi Aepli.
\newblock Findings of the {V}ar{D}ial evaluation campaign 2017.
\newblock In \emph{Proceedings of the Fourth Workshop on {NLP} for Similar Languages, Varieties and Dialects ({V}ar{D}ial)}, pp.\  1--15, Valencia, Spain, April 2017. Association for Computational Linguistics.
\newblock \doi{10.18653/v1/W17-1201}.
\newblock URL \url{https://aclanthology.org/W17-1201}.

\bibitem[Zampieri et~al.(2020)Zampieri, Nakov, and Scherrer]{zampieri2020natural}
Marcos Zampieri, Preslav Nakov, and Yves Scherrer.
\newblock Natural language processing for similar languages, varieties, and dialects: A survey.
\newblock \emph{Natural Language Engineering}, 26\penalty0 (6):\penalty0 595--612, 2020.

\bibitem[Zhang et~al.(2022)Zhang, Maslej, Brynjolfsson, Etchemendy, Lyons, Manyika, Ngo, Niebles, Sellitto, Sakhaee, Shoham, Clark, and Perrault]{zhang2022ai}
Daniel Zhang, Nestor Maslej, Erik Brynjolfsson, John Etchemendy, Terah Lyons, James Manyika, Helen Ngo, Juan~Carlos Niebles, Michael Sellitto, Ellie Sakhaee, Yoav Shoham, Jack Clark, and Raymond Perrault.
\newblock The ai index 2022 annual report, 2022.

\bibitem[Zhang et~al.(2023{\natexlab{a}})Zhang, Shi, Liu, Yuan, Li, Dong, Shu, Li, Wang, Lin, Huang, and Fu]{zhang2023chinese}
Ge~Zhang, Yemin Shi, Ruibo Liu, Ruibin Yuan, Yizhi Li, Siwei Dong, Yu~Shu, Zhaoqun Li, Zekun Wang, Chenghua Lin, Wenhao Huang, and Jie Fu.
\newblock Chinese open instruction generalist: A preliminary release, 2023{\natexlab{a}}.

\bibitem[Zhang et~al.(2023{\natexlab{b}})Zhang, Fang, Zhang, Ma, Zhou, Huang, Bu, Gui, Chen, Chen, and Feng]{zhang2023bayling}
Shaolei Zhang, Qingkai Fang, Zhuocheng Zhang, Zhengrui Ma, Yan Zhou, Langlin Huang, Mengyu Bu, Shangtong Gui, Yunji Chen, Xilin Chen, and Yang Feng.
\newblock Bayling: Bridging cross-lingual alignment and instruction following through interactive translation for large language models, 2023{\natexlab{b}}.

\bibitem[Zhang et~al.(2018{\natexlab{a}})Zhang, Liu, Liu, Gao, Duh, and Durme]{zhang2018record}
Sheng Zhang, Xiaodong Liu, Jingjing Liu, Jianfeng Gao, Kevin Duh, and Benjamin~Van Durme.
\newblock Record: Bridging the gap between human and machine commonsense reading comprehension, 2018{\natexlab{a}}.

\bibitem[Zhang et~al.(2023{\natexlab{c}})Zhang, Aljunied, Gao, Chia, and Bing]{zhang2023m3exam}
Wenxuan Zhang, Sharifah~Mahani Aljunied, Chang Gao, Yew~Ken Chia, and Lidong Bing.
\newblock M3exam: A multilingual, multimodal, multilevel benchmark for examining large language models, 2023{\natexlab{c}}.

\bibitem[Zhang et~al.(2015)Zhang, Zhao, and LeCun]{NIPS2015_250cf8b5}
Xiang Zhang, Junbo Zhao, and Yann LeCun.
\newblock Character-level convolutional networks for text classification.
\newblock In C.~Cortes, N.~Lawrence, D.~Lee, M.~Sugiyama, and R.~Garnett (eds.), \emph{Advances in Neural Information Processing Systems}, volume~28. Curran Associates, Inc., 2015.
\newblock URL \url{https://proceedings.neurips.cc/paper_files/paper/2015/file/250cf8b51c773f3f8dc8b4be867a9a02-Paper.pdf}.

\bibitem[Zhang et~al.(2019)Zhang, Baldridge, and He]{zhang2019paws}
Yuan Zhang, Jason Baldridge, and Luheng He.
\newblock Paws: Paraphrase adversaries from word scrambling, 2019.

\bibitem[Zhang et~al.(2018{\natexlab{b}})Zhang, Liu, Li, Zhou, and Chen]{ZhiruiZhang18}
Zhirui Zhang, Shujie Liu, Mu~Li, Ming Zhou, and Enhong Chen.
\newblock Joint training for neural machine translation models with monolingual data.
\newblock \emph{CoRR}, abs/1803.00353, 2018{\natexlab{b}}.
\newblock URL \url{http://arxiv.org/abs/1803.00353}.

\bibitem[Zhou et~al.(2023)Zhou, Liu, Xu, Iyer, Sun, Mao, Ma, Efrat, Yu, Yu, et~al.]{zhou2023lima}
Chunting Zhou, Pengfei Liu, Puxin Xu, Srini Iyer, Jiao Sun, Yuning Mao, Xuezhe Ma, Avia Efrat, Ping Yu, Lili Yu, et~al.
\newblock Lima: Less is more for alignment.
\newblock \emph{arXiv preprint arXiv:2305.11206}, 2023.

\bibitem[Zhuo et~al.(2024)Zhuo, Zebaze, Suppattarachai, von Werra, de~Vries, Liu, and Muennighoff]{zhuo2024astraios}
Terry~Yue Zhuo, Armel Zebaze, Nitchakarn Suppattarachai, Leandro von Werra, Harm de~Vries, Qian Liu, and Niklas Muennighoff.
\newblock Astraios: Parameter-efficient instruction tuning code large language models.
\newblock \emph{arXiv preprint arXiv:2401.00788}, 2024.

\bibitem[Ziegler et~al.(2020)Ziegler, Stiennon, Wu, Brown, Radford, Amodei, Christiano, and Irving]{ziegler2020finetuning}
Daniel~M. Ziegler, Nisan Stiennon, Jeffrey Wu, Tom~B. Brown, Alec Radford, Dario Amodei, Paul Christiano, and Geoffrey Irving.
\newblock Fine-tuning language models from human preferences, 2020.

\bibitem[Zoph et~al.(2016)Zoph, Yuret, May, and Knight]{zoph-etal-2016-transfer}
Barret Zoph, Deniz Yuret, Jonathan May, and Kevin Knight.
\newblock Transfer learning for low-resource neural machine translation.
\newblock In \emph{Proceedings of the 2016 Conference on Empirical Methods in Natural Language Processing}, pp.\  1568--1575, Austin, Texas, November 2016. Association for Computational Linguistics.
\newblock \doi{10.18653/v1/D16-1163}.
\newblock URL \url{https://aclanthology.org/D16-1163}.

\end{thebibliography}

\newpage

\appendix

\section{\aya Annotation Platform}
\label{apx:aya_platform_design}

In this section, we discuss the detailed design and development of the \aya Annotation Platform and the gamification strategy employed. Together, these attempts aimed to ensure high-quality curation of the \aya Dataset.

\begin{figure}[h!]
  \centering
  \includegraphics[scale=.38]{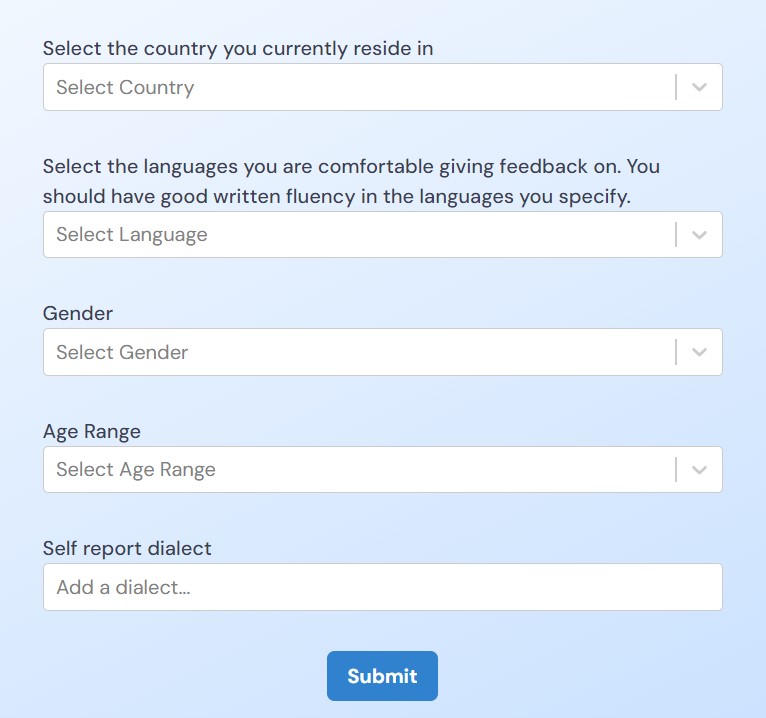}
\caption{The onboarding page for a new user. We collect some demographic information and ask them to specify the languages they are comfortable annotating data in. These are the language options they are presented with later on in the UI.}
\label{fig:onboarding}
\end{figure}

\subsection{Engagement Strategies}

We decided to employ gamification methods to enhance annotator engagement \citep{deFranga2015, Bastanfard2023, Morschheuser2017}. Our strategy involved using a points-based rewards system, motivating contributors through social media recognition, and fostering friendly competition with leaderboards. Regular mini-challenges and sprints helped to create collective achievement goals and fostered a sense of community \citep{Bastanfard2023}. Real-time feedback reinforced positive behavior and customization options, such as avatars, provided a personalized experience. Overall, these gamification strategies aimed to boost engagement, improve data quality, and created a more enjoyable experience for the crowd-sourcing participants. \citep{Morschheuser2017}.

To recognize and incentivize the efforts of our contributors, we established a tiered reward system based on contribution milestones: 500, 1,000, and 5,000 contributions. Contributors who achieved these goals on the project leaderboard were rewarded with certificates and specially designed \aya project apparel. The attire varied according to the contribution level, with different packages for each milestone. Additionally, the most active contributors were prominently acknowledged in the project's dataset documentation, highlighting their significant role as key contributors to the project. This system not only motivated contributors but also served as a token of appreciation for their dedication and hard work.

\begin{figure}[t]
  \centering
    \includegraphics[width=0.4\textwidth]{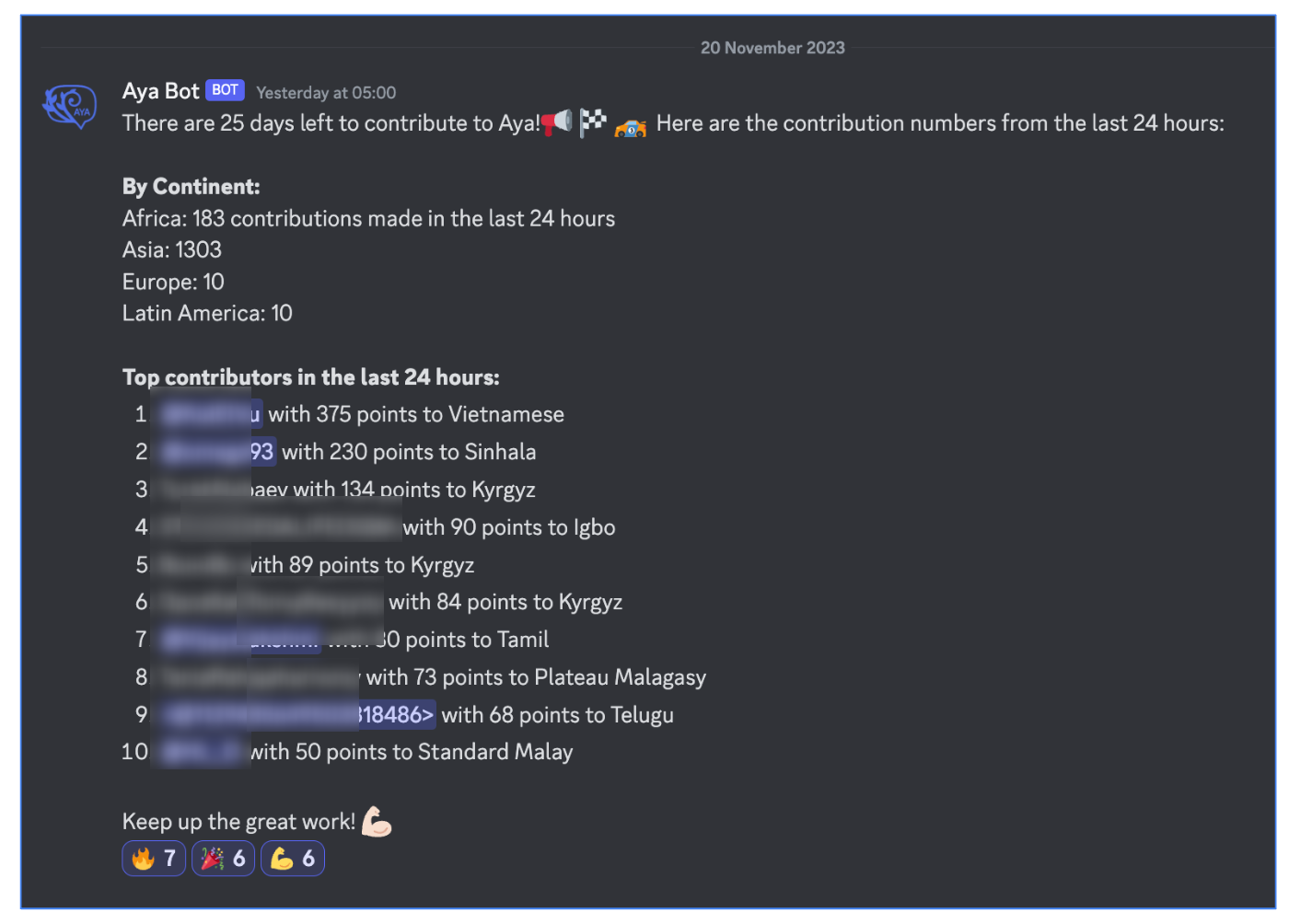}
		\caption{Aya Discord Bot message aggregating daily statistics and top 10 annotators}
		\label{fig:aya_bot}
\end{figure}

In addition to the leaderboard, the \aya Discord Bot was developed to recognize contributors with a high number of points. This bot recognized the daily top 10 contributors by tagging them in a message that was posted on the \aya Discord server. It also aggregated daily total contributions per region and specified how many days were left until the data collection phase ended. As shown in Figure~\ref{fig:aya_bot}, these messages provided a regular snapshot of progression that allowed annotators to see the dataset grow across all languages.

\begin{figure}[htb!]
  \centering
  \begin{minipage}[b]{.80\textwidth}
    \includegraphics[width=\textwidth]{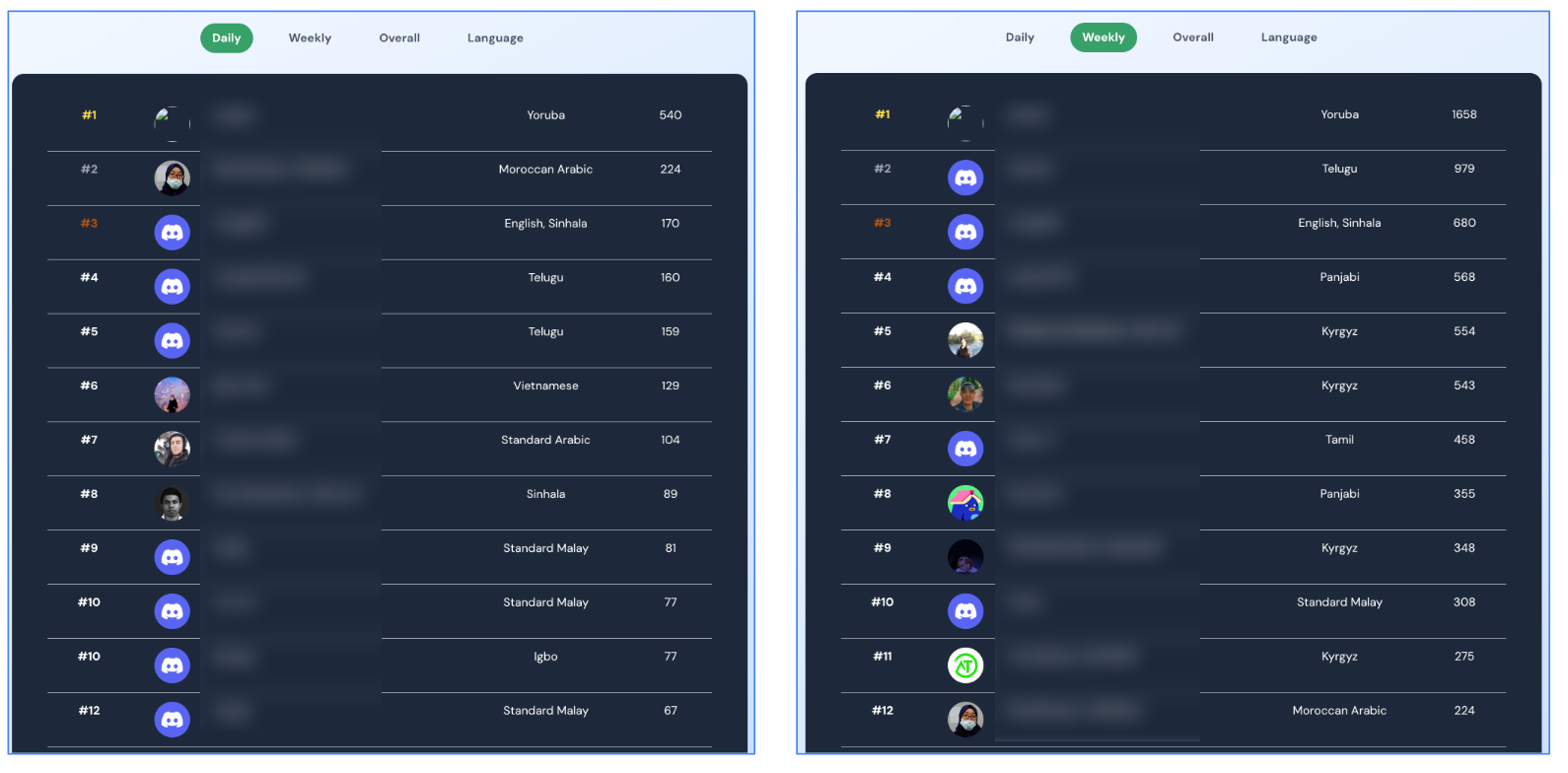}
  \end{minipage}
  \hfill
  \begin{minipage}[b]{.80\textwidth}
    \includegraphics[width=\textwidth]{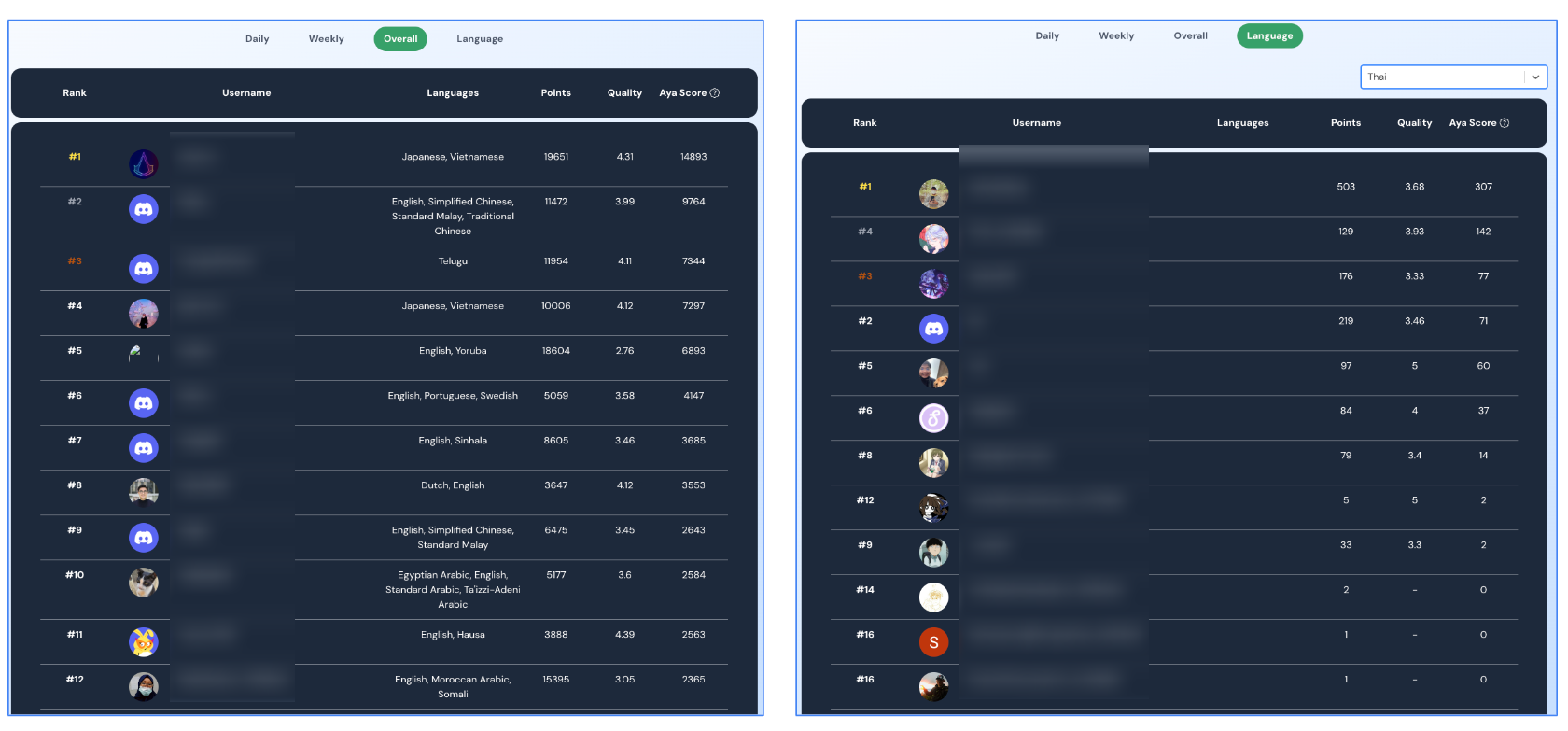}
  \end{minipage}
  \caption{\aya Leaderboard. Daily and weekly leaderboards are shown on the top left and right respectively. At the bottom are language-specific and overall leader boards, where annotators are ranked based on their \aya score.}
\label{fig:leaderboards}
\end{figure}

\FloatBarrier

\subsection{Accessibility of Registration Tools}

\begin{figure}[htb!]
     \centering
     \begin{subfigure}[T]{0.48\textwidth}
         \centering
         \includegraphics[scale=0.05]{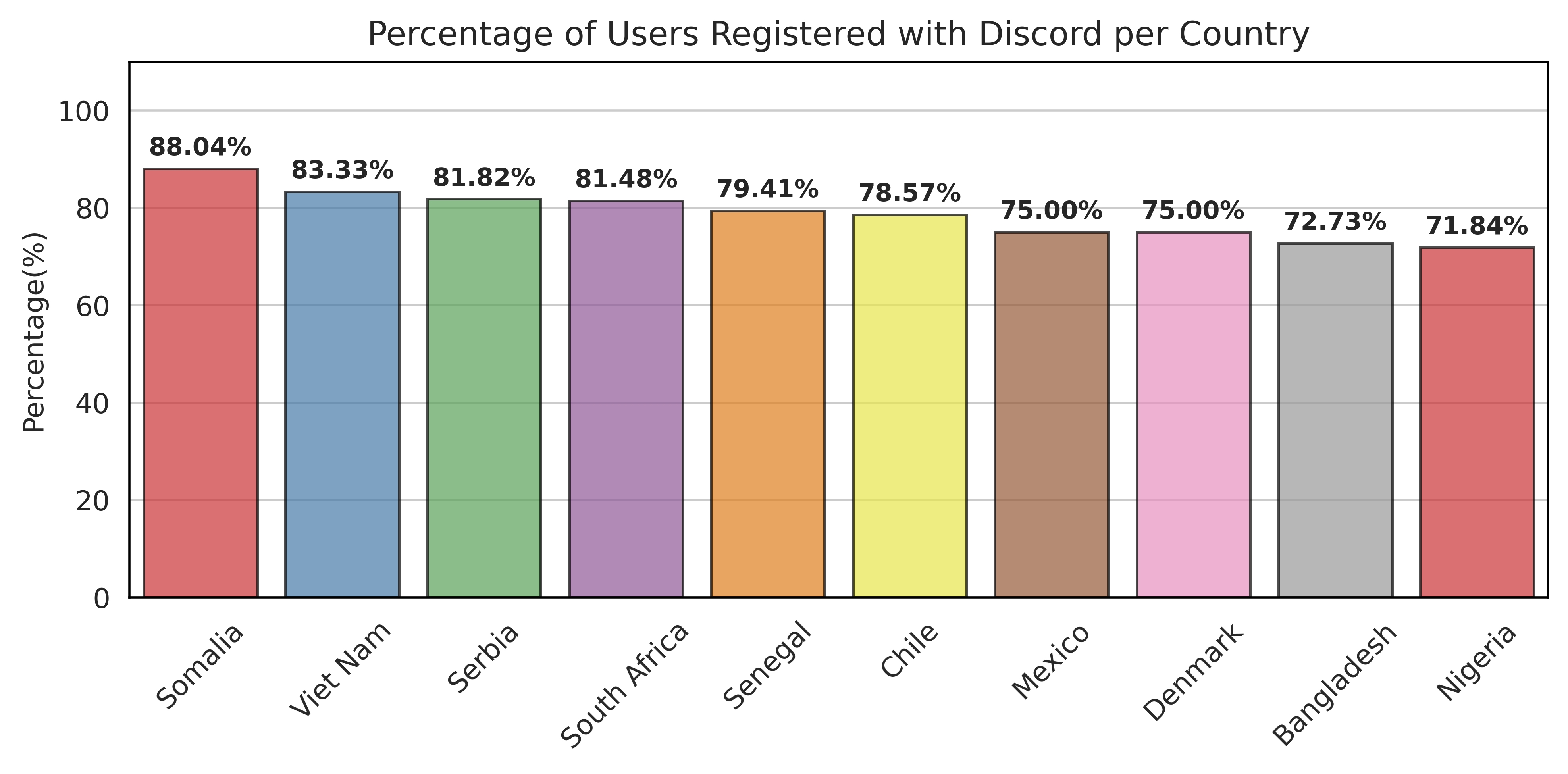}
     \end{subfigure}
     \hfill
     \begin{subfigure}[T]{0.48\textwidth}
         \centering
         \includegraphics[scale=0.35]{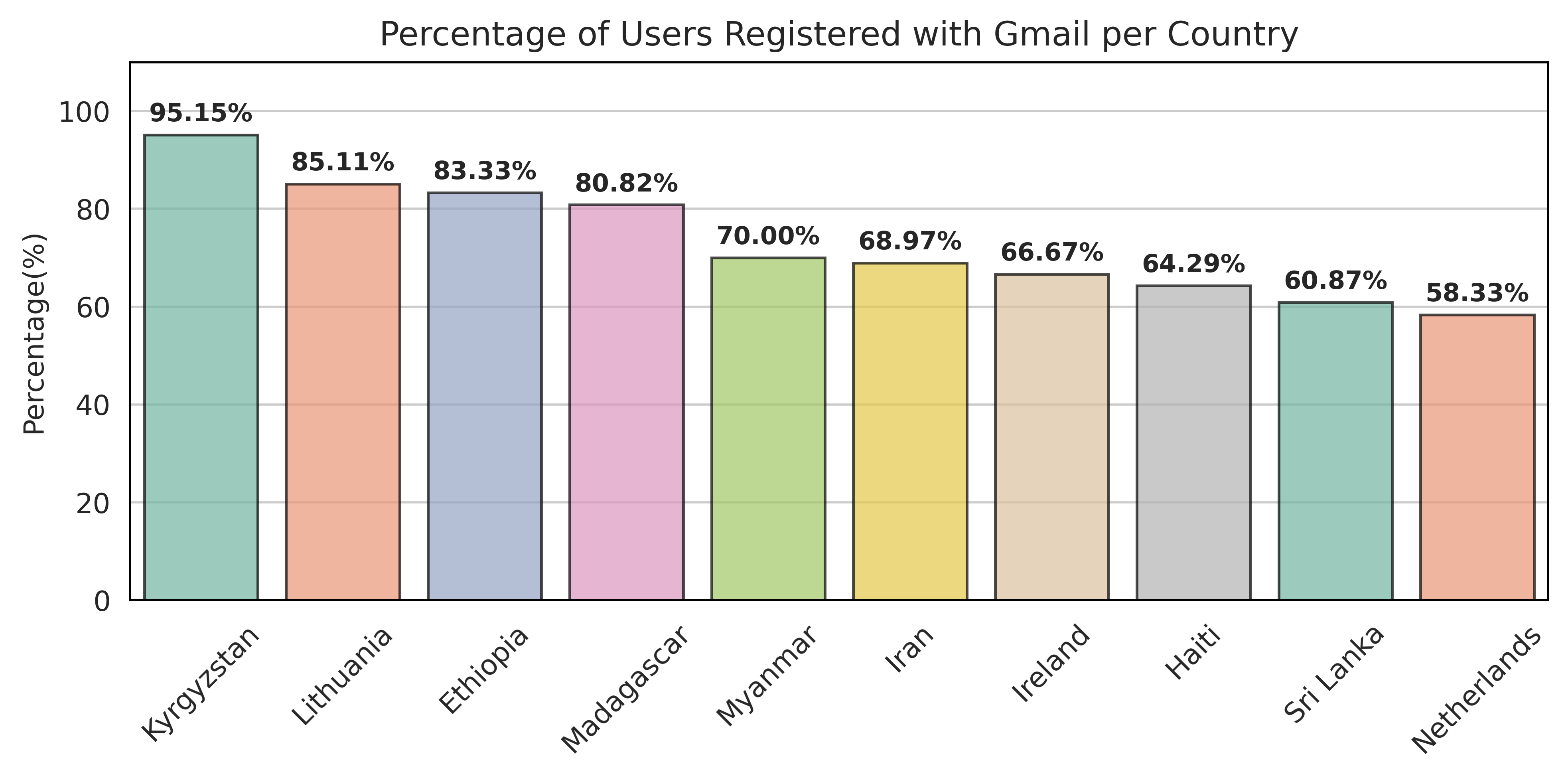}
     \end{subfigure}
  \caption{Percentage of \aya UI users that registered via \textbf{Left:} Discord and \textbf{Right:} Gmail.}
\label{fig:percentage_user_by_sso}
\end{figure}

\begin{figure}[htb!]
  \centering
  \includegraphics[scale=.6]{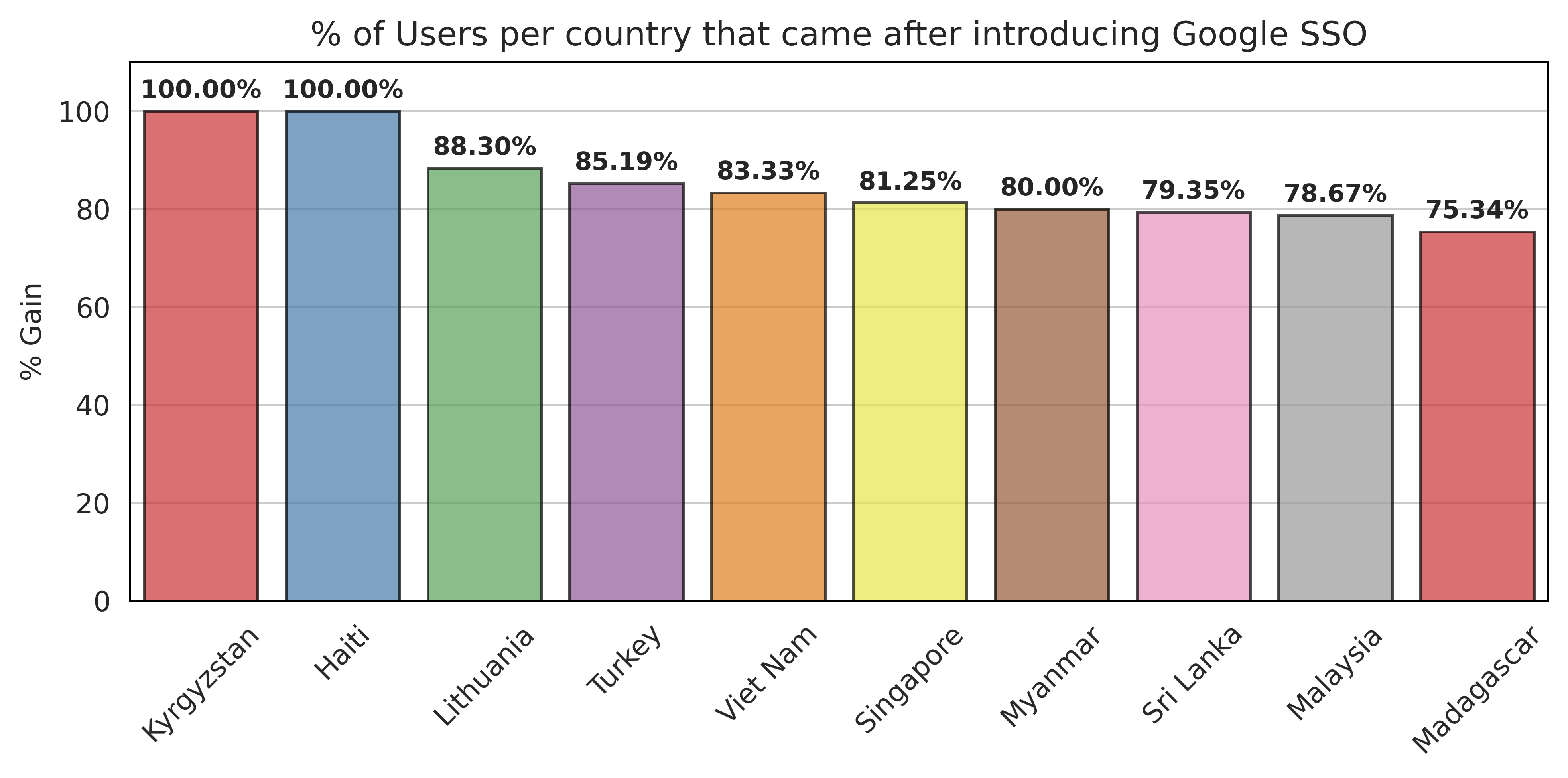}
  \caption{Percentage of the \aya UI users that registered after the introduction of Google SSO in the platform.}\label{fig:percentage_user_after_google_sso}
\end{figure}

The accessibility and popularity of registration tools differed from country to country and this had an impact on where the \aya UI users joined from.
Figure~\ref{fig:percentage_user_by_sso} compares the percentage of registered users using Discord and Gmail to sign up in the top 10 countries. After introducing Google SSO, we observed a significant jump in the number of registered users from several new countries (Figure~\ref{fig:percentage_user_after_google_sso}).

\section{UI Tasks}

\subsection{Reviewing Annotators}\label{sec:reviewing-annotators}

In the \aya UI, we display the original and re-annotated versions of both prompt and completion along with the name of the annotator who did the re-annotation. Reviewers are tasked with rating the re-annotated prompt and completion on a scale from 1 to 5. The ratings are defined as follows:

\begin{enumerate}[nosep, left=0pt, labelsep=0.5em, label=\textbf{(\arabic*)}]
    \item Much worse than the original annotation;
    \item Worse than the original annotation;
    \item No noticeable difference compared to the original annotation;
    \item Better than the original annotation;
    \item Much better than the original annotation.
\end{enumerate}

If any of the prompt/completion pairs receive a rating below 5 (i.e., anything lower than ``Much better than the original annotation''), the reviewer is provided with an option to modify the provided prompt and completion pair to improve its quality. An ``Additional Feedback'' text box is also available for reviewers to explain the reasoning behind their chosen rating.

\subsection{Aya Score}
\label{sec:highquality}
To encourage annotators to make high-quality submissions, we designed a ranking score favoring contributions that received high peer ratings. 
The \aya Score for an individual annotator is:

\begin{equation}
    \text{Score}_\aya = w_1 \times (\mathcal{E}) + w_2 \times \mathcal{C},
    \label{eq:aya_score}
\end{equation}

with weights $w_1$, $w_2$ defined as: 

\begin{equation}
    w_1 = \max\left(0, (\hat{Q} - 3)\right),
    w_2 = \left( \frac{\mathcal{T_{+}}}{\mathcal{T}} \right)
    \label{eq:aya_score_weights}
\end{equation}

where:
\begin{itemize}
    \item $\mathcal{E}$ represents the total number of re-annotations in which the annotator edited prompts and completions 
    \item $\mathcal{C}$ represents the total number of original annotations submitted by an annotator
    \item $\hat{Q}$ represents the average quality rating derived for an annotator through peer review via Annotation Feedback task
    \item$\mathcal{T_{+}}$ represents the total number of original annotations made by an annotator that received positive feedback in Re-annotations task
    \item $\mathcal{T}$ represents the total number original annotations made by an annotator that feedback received in Re-annotations task. 
\end{itemize}


The rationale behind introducing the $\text{Score}_\aya$ was to boost a competitive environment among annotators and encourage them to focus on submitting high-quality data, consequently improving the overall quality of the \aya Dataset. 

\subsection{Annotation Guidelines}
\label{sec:annotation-guidelines}
The annotators were provided with the following evaluation criteria for what a good prompt and completion pair must look like. Re-Annotations were then performed if they determined that the prompts or completions needed editing.
\begin{enumerate}[noitemsep,topsep=0pt,parsep=0pt,partopsep=0pt]
    \item No grammatical or spelling mistakes in both the prompt and completion text.
    \item The prompt provides clear instruction on what the task is. 
    \item The completion answers the prompt correctly. Both the prompts and completions should be in full sentences and coherent, with reasonable length.
    \item For original annotations, the prompts and completions should not be generated from other language models.
\end{enumerate}

\textbf{Re-Annotations $\:$}
Before editing, annotators rated the quality of existing prompt and completion pairs by choosing either the thumbs-up or thumbs-down option. If the provided prompt and completion pair were already of good quality according to the criteria above, then annotators rated them with thumbs up and moved ahead without editing.
Overall, annotators were encouraged to re-annotate the completions, in particular by adding more details and context to them since many of them were often short one-word answers.

\section{Language Representation via Community}
\label{sec:repr_communities}

\subsection{Division by Regions} 
\label{apx:division_by_region}

We chose to divide languages into four primary regions: \textit{Africa, Asia, Europe,} and \textit{Latin America}. These four regions were established in order to facilitate the administration of user contributions and were not intended to prescribe boundaries within which certain languages are exclusively spoken. 

The language statistics by region are as follows: \textit{Africa} (14 languages), \textit{Asia} (41 languages), \textit{Europe} (42 languages), and \textit{Latin America} (4 languages). Almost all the languages were assigned to a region but there are some exceptions, \texttt{Maori} and \texttt{Samoan} were unassigned to any specific region as they didn't align with the predefined regions. \texttt{English} was left unassigned, serving as a common language across all regions. Additionally, contributions in \texttt{Spanish} and \texttt{Portuguese} were distributed between \textit{Europe} and \textit{Latin America} based on contributors' countries. Similarly, \texttt{Arabic} contributions were shared between \textit{Africa} and \textit{Asia} depending on the contributors' country of origin.
Additional dialects of \texttt{Arabic} were included in regions separate from that of their parent language because we had a significant number of speakers from these regions eager to contribute to their respective dialects.  Each region had at least one ``Regional Lead'' responsible for coordinating ``Language Ambassadors,'' and for recruiting fluent speakers for the languages within their area.

\subsection{Language Ambassadors} 

The Language Ambassador's role was pivotal in bridging the gap between the data collected in a language and its speakers. An essential criterion for selection was native fluency in the specific language. The Language Ambassador's expertise in specific languages and familiarity with the cultures of the language speakers was invaluable. They assisted not only in spreading awareness among participants but also in identifying and addressing potential data issues specific to each language, such as languages incorrectly assigned to their region. Their cultural and linguistic insights enabled them to make informed decisions, like choosing suitable data sources for collection in their respective languages.
Not every language had a designated Language Ambassador, and some had more than one. In total, we had 84 Language Ambassadors over the course of the initiative. Their combined efforts played a vital role in broadening the contributor base for each language. Support for the Language Ambassadors' progress and trouble-shooting challenges they faced was coordinated asynchronously and through weekly online meetups, discussed in Sec.~\ref{apx:platforms} and Sec.~\ref{apx:meetings}.

\subsection{Regional Leads}

There were a total of six Regional Leads: two for Latin America, one for Africa, one for Asia, and two for Europe. The selection for Regional Lead roles was on a voluntary basis, with the only requirement being that they must originate from the regions they intended to lead. The invitation for this role was specifically extended to individuals who were already actively participating in Cohere For AI projects or engaged in other open science projects. Regional Leads had several key roles throughout the project, such as selecting Language Ambassadors and aiding their efforts in attracting more annotators and maintaining their engagement.

\section{Communication}\label{apx:communication}

\subsection{Platforms} 
\label{apx:platforms}

We established a Discord server for coordination between Regional Leads, Language Ambassadors and annotators. The server provided basic channels for internal communications: introductions, inquiries, and announcements, as well as specific channels for Language Ambassadors, for each region, and for each language, along with any other channels that proved useful for the particular region. For external communications, we used social media platforms (e.g., X, LinkedIn, WhatsApp, Facebook), recognizing that the choice of communication platform varied based on cultural and regional preferences. Using multiple platforms not only facilitated internal organization but also broadened our project's outreach by providing flexible and inclusive means of outreach to diverse communities and audiences.

\subsection{Meetings} 
\label{apx:meetings}

In addition to asynchronous communication through Discord, we conducted meetings to maintain team collaboration and cohesion:
\begin{enumerate}

\item \textbf{Regional Leads and Language Ambassadors Meeting}: A weekly meeting in which Regional Leads and Language Ambassadors shared project updates, exchanged ideas, and addressed questions from Language Ambassadors. It served as an excellent platform for gathering ideas from Language Ambassadors and brainstorming new strategies to engage annotators effectively.

\item \textbf{New Contributor Introduction Meeting}: Held weekly, this meeting aimed to introduce new contributors to the project's specifics. It included explanations about the motivations behind the project, a walk-through of the \aya UI, and a sharing of regional statistics. Additionally, this meeting provided examples of both good and bad annotations and edits to guide new annotators in their work. It concluded with a synchronous challenge for the annotators to submit a few initial annotations in real time to familiarize them with the process and allow them to ask questions if they got stuck.

\item \textbf{Regional Leads Meeting}: Held bi-weekly, this meeting brought together Regional Leads to assess progress, discuss upcoming steps, and provide advice on how to engage and sustain contributions for their respective regions. Furthermore, this meeting facilitated collaborative troubleshooting efforts and helped make important decisions for the following week.

\item \textbf{Technical Update}: This meeting was dedicated to sharing technical updates, with a focus on recent UI progress, data, and benchmarking. The purpose of this monthly update was to ensure all team members and annotators were well-informed about the project's current status and upcoming priorities. It was a place for open discussion to hear feedback from everyone interested in the project.

\item \textbf{Language Specific Meeting}: Held weekly or biweekly, these meetings were co-working sessions or datathons led by the language ambassadors with their respective annotators to submit annotations synchronously. It also acted as an onboarding session to welcome new contributors from regions that could not join the New Contributor Introduction Meeting due to conflicting time zones. Demonstrations on using the UI, as well as brainstorming sessions, were conducted to improve the representation of specific languages in the project.  
\end{enumerate}

\section{Language Groupings}\label{sec:lang-groups} In this work we will refer to groups of languages to be ``lower-'', ``mid-'' or ``higher''-resourced according to their recorded, written, and catalogued NLP resources~\citep{joshi-etal-2020-state}. \cite{joshi-etal-2020-state} group languages into 5 distinct clusters based on the amount of data from a combined range of source (LDC catalog\footnote{\url{https://catalog.ldc.upenn.edu/}}, ELRA Map\footnote{\url{https://catalog.elra.info/en-us/}}, Wikipedia \footnote{\url{https://wikipedia.org/}}), which we interpret as a proxy for data availability for pretraining and IFT training of LLMs. 
We group these 5 distinct clusters into a rough taxonomy of \textbf{lower-resourced (LR)}, \textbf{mid-resourced (MR)} and \textbf{higher-resourced (HR)} (See Table~\ref{tab:lang_group}). See Table~\ref{tab:language_codes} for full mapping of languages to categories. 
We note that this grouping is inevitably imperfect; languages and their varieties cannot absolutely nor universally be classified based on this single dimension~\citep{H_m_l_inen_2021,lignos-etal-2022-toward, bird-2022-local}. The categorization in our case serves the purpose of aggregation in our analysis of the data distribution.

\begin{table}[htb!]
    \centering
    \small
    \begin{tabular}{lccl}
        \toprule
        Group & Category & Languages & Examples \\
        \midrule
        \multirow{2}{*}{Higher-Resourced} & 5 & 7 & Arabic, Chinese, English, French, Spanish \\
         & 4 & 18 & Hindi, Italian, Portuguese, Russian, Turkish \\
         \noalign{\smallskip} 
        \midrule
        \noalign{\smallskip}
        Mid-Resourced& 3 & 25 & Afrikaans, Indonesian, Kazakh, Malay, Latvian  \\
         \noalign{\smallskip} 
        \midrule 
        \noalign{\smallskip}
        \multirow{3}{*}{Lower-Resourced} & 2 & 13 & Hausa, Icelandic, Irish, Lao, Maltese\\
         & 1 & 39 & Albanian, Gujarati, Igbo, Luxembourgish \\
         & 0 & 12$^*$& Kurdish, Kyrgyz, Sinhala, Yiddish \\
        \bottomrule
    \end{tabular}
    \caption{Language grouping for the \aya Collection. We assign categories to languages based on~\citep{joshi-etal-2020-state}. (*) We assign label 0 to two languages not found in \citet{joshi-etal-2020-state}'s taxonomy (\texttt{manipuri} and \texttt{ngaju}). 
    }
    \label{tab:lang_group}
\end{table}

\section{Post-Editing the \textsc{dolly-machine-translated} Test Set}\label{apx:postedit_setup}
\subsection{Annotators}
\textbf{Annotator Selection} The primary demographic make-up of the participants in the evaluations was recruited based on their proficiency in the language groups. The proficiency was self-reported, and our requirements were natively proficient or professionally proficient in the specific languages needed for the project. Outside of this, the participants come from diverse social backgrounds comprised of students and individuals with full-time or part-time jobs that do annotation as a ``side gig''. 

\textbf{Socio-Demographics} 
The annotator pool is comprised of people from diverse backgrounds, and this spans across socioeconomic backgrounds, careers, levels of education, and self-reported gender and sexual identities. We do not ask any annotators to share or report any of these statistical pieces of information in a formal way; any insights into this are gathered organically and through self-reporting by the annotators. 

\textbf{Quality Considerations} We do not believe that any socio-demographic characteristics have led to any impact on the data that has been annotated. Through every part of the project we have reiterated the importance of this work and the fact that this is helping to support a global-scale research project. We are confident in the trust we have built with the annotators in this project, and they care greatly about the overall outcome and therefore have been diligent in completing the task with a high degree of accuracy. 
Where possible, we have done our best to have annotators work on this project and be representatives of the communities that the project aims to support.

\textbf{Compensation}
The annotators were paid 30 CAD per hour. No special consideration was made to the hourly rate as that is the standard rate offered to Cohere’s annotators who work on highly complex tasks. 

\subsection{Annotation Process}
\textbf{Communication} Annotators were briefed by one of the authors in a virtual introduction session, and were able to ask questions and raise issues throughout the annotation task in a Slack channel. They were also encouraged to share frequent error patterns, artifacts, or hard decisions that they encountered throughout the task with the authors and other annotators.

\textbf{Schedule} There was no fixed time schedule for the annotations and annotators contributed a varying amount of hours and ratings, depending on their availabilities and speed. Each translation was post-edited by one annotator, and there were 3--4 annotators involved in each task. After post-edits were completed, a second annotator (not the original post-editor) assessed the post-edit for quality and proposed new final edits if necessary. 

\textbf{Interface} Post-edits were collected on Google Sheets with an interface built in Google Apps Script. 

\subsection{Instructions}
The instructions given to professional annotators for the \textsc{dolly-machine-translated}
 test set post-edits were the following:
``As an annotator, you have the task to improve the quality of the prompts for our multilingual model! 
The prompts are originally machine-translated from English, and sometimes the translation introduces errors in the prompts that make them hard to follow. \\

We need your help to identify these cases, and to edit these translations so that they...
\begin{enumerate}
    \item Convey the same instruction/task/request as the English original --- not more and not less.
    \item Are grammatically correct.
    \item Are free from phrases too literally translated from English (we call this “Translationese”).
\end{enumerate}
This is how:\\
For each pair of English prompt and translated prompt shown, decide whether the prompt is okay as it is (according to the above criteria), or needs an edit. 
\begin{itemize}
    \item If it needs an edit, edit the prompt until the quality is satisfactory (in the field “Edited Prompt”). Try to keep your edits minimal. Then confirm that the edited prompt fulfills the above three criteria.
    \item 
If it’s okay as is, just proceed (without editing the “Edited Prompt” field) to confirm that it fulfills the above three criteria.''
\end{itemize}

Annotations were done through an interface built on top of Google Sheets. One annotator edited each prompt, and another verified the edit, if necessary had a discussion and edited the original edit. Three to four editors collaborated on each language.

\subsection{Post-Editing Effort}
For each prompt, we measure the post-editing effort with Human-targeted Translation Error Rate (HTER) ~\citep{specia-farzindar-2010-estimating}, an edit-distance metric that compares the original machine translation with the post-edited version in terms of edit operations on units of words. 
This also gives us an estimate of how severe the errors in the original translations were, and how critically the post-editors assessed the original translations.
Analogously, we estimate with a Human-targeted Character F-Score (HChrF) score how much the original translation overlaps with the final post-edited translation. This metric is based on the ChrF score~\citep{popovic-2015-chrf} and operates on character-level matches. Computations of HTER and HChrF are based on the \texttt{sacrebleu} implementation~\citep{post-2018-call}. 

Table~\ref{tab:pe_effort} reports these statistics for the six languages of the \textsc{dolly-machine-translated} test set. We find that editors edited at least 41\% of prompts in all languages, a surprisingly high number. This indicates that translation errors in the \textsc{dolly-machine-translated} test set are quite common.
For Russian, the post-editing effort was overall largest, with an average of 37.43 HTER, which means that 37.43\% of words in the final post-edit had to be edited from the original.
This stands in contrast with the post-edits for French, where a similar ratio of original prompts was edited (84.5\% compared to 86.5\% for Russian), but to a much lesser extent (5.56 HTER).

\begin{table}[]
    \centering
    \begin{tabular}{lccc}
    \toprule
        Language & \% of Prompts Edited & HTER & HChrF\\
        \midrule
        Arabic & 41.0\% & 10.78 & 92.74\\
        French &84.5\% & 5.56 & 96.81\\
        Hindi & 60.0\% & 6.16 & 95.00\\
        Russian & 86.5\% & 37.43 & 75.92\\
        Serbian & 72.5\% & 9.06 & 92.79\\
        Spanish & 75.5\% & 9.13 & 93.25\\
        \bottomrule
    \end{tabular}
    \caption{Post-editing effort measured by the overall percentage of edited dolly test prompts,  HTER (Human-targeted Translation Error Rate: the higher, the more effort), and HChrF (Human-Targeted Character F-Score: the lower, the more effort). }
    \label{tab:pe_effort}
\end{table}

\section{Translation using NLLB}\label{app:translation_nllb}

\textbf{Additional Generation details} One caveat with using NLLB is that since the model was trained on single sentence pairs, the translations tend to cut off abruptly when full paragraphs are translated. To get around this, we sentence-tokenize the paragraphs using the sentence-splitter Python package (similar to ~\citep{nllbteam2022language}) and concatenate them post-translation. To avoid known translation introduced artefacts, We also filter any samples which have \texttt{<unk>} tokens introduced by the NLLB tokenizer or model. 



\subsection{Translation Quality of NLLB}\label{app:nllb_quality}
\begin{figure}
    \centering
    \begin{subfigure}[b]{\textwidth}
          \centering
          \includegraphics[width=\textwidth]{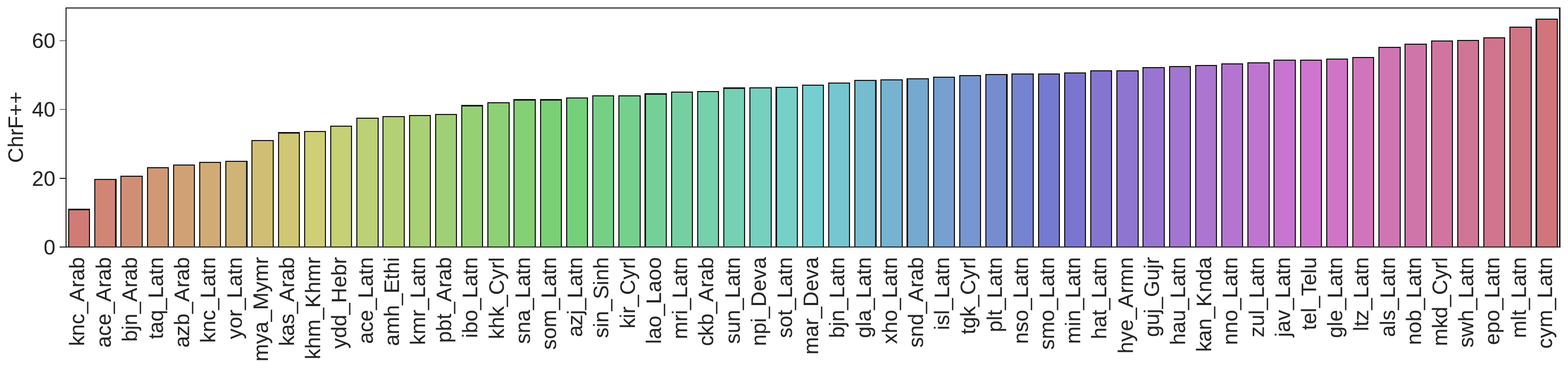}
         \caption{Low-resource Languages:~\citep{joshi-etal-2020-state} classes 0, 1, 2}
    \end{subfigure}
    \begin{subfigure}[b]{\textwidth}
          \centering
          \includegraphics[width=\textwidth]{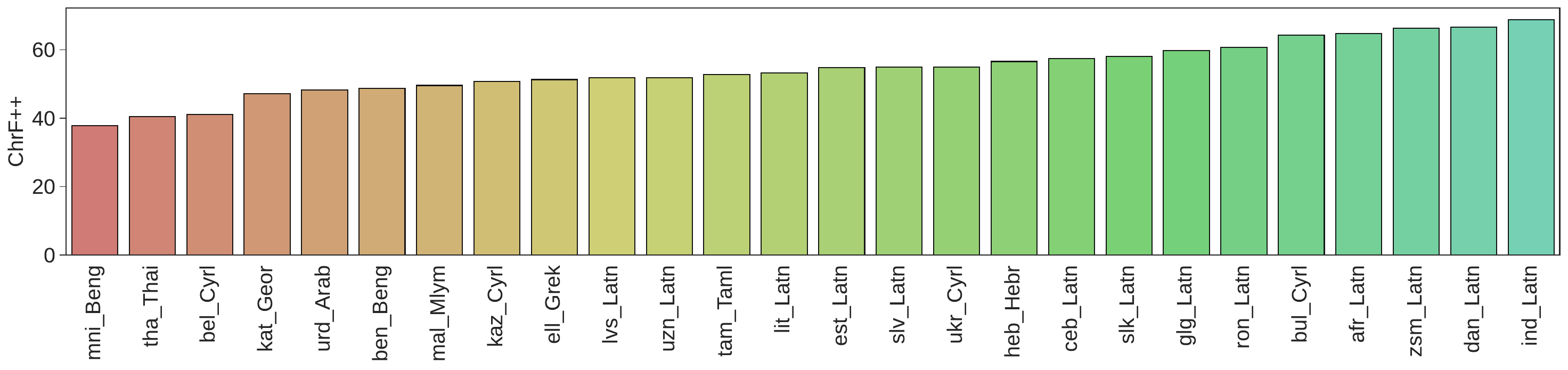}
         \caption{Mid-resource Languages: ~\citep{joshi-etal-2020-state} class 3}
    \end{subfigure}
    \begin{subfigure}[b]{\textwidth}
          \centering
          \includegraphics[width=\textwidth]{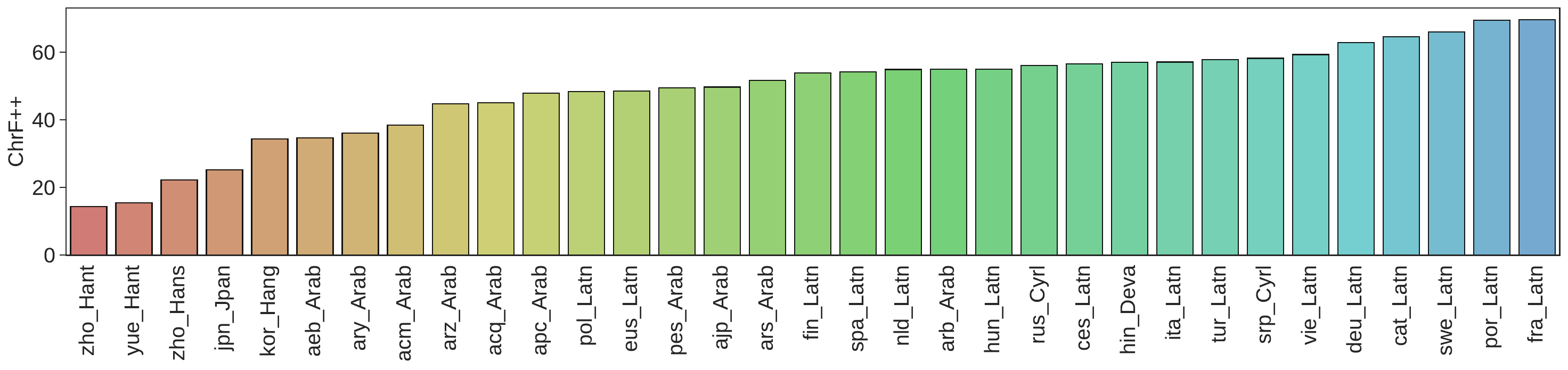}
         \caption{High-resource Languages: ~\citep{joshi-etal-2020-state} classes 4, 5}
    \end{subfigure}
   
    \caption{NLLB Translation Quality: ChrF++ scores on FLORES for translations from English into the \aya target languages that are covered by NLLB, grouped by their resourcefulness according to~\citep{joshi-etal-2020-state}.}
    \label{fig:nllb_quality}
\end{figure}

Figure~\ref{fig:nllb_quality} illustrates NLLB translation quality as measured by ChrF++ on the FLORES benchmark for the languages of interest for \aya, grouped by their resourcefulness according to~\citep{joshi-etal-2020-state}. The scores were extracted from \url{https://github.com/facebookresearch/fairseq/blob/nllb/README.md} for the dense 3.3B model.

\newpage
\section{Additional Figures}

\begin{figure}[ht!]
  \centering
  \includegraphics[width=\textwidth]{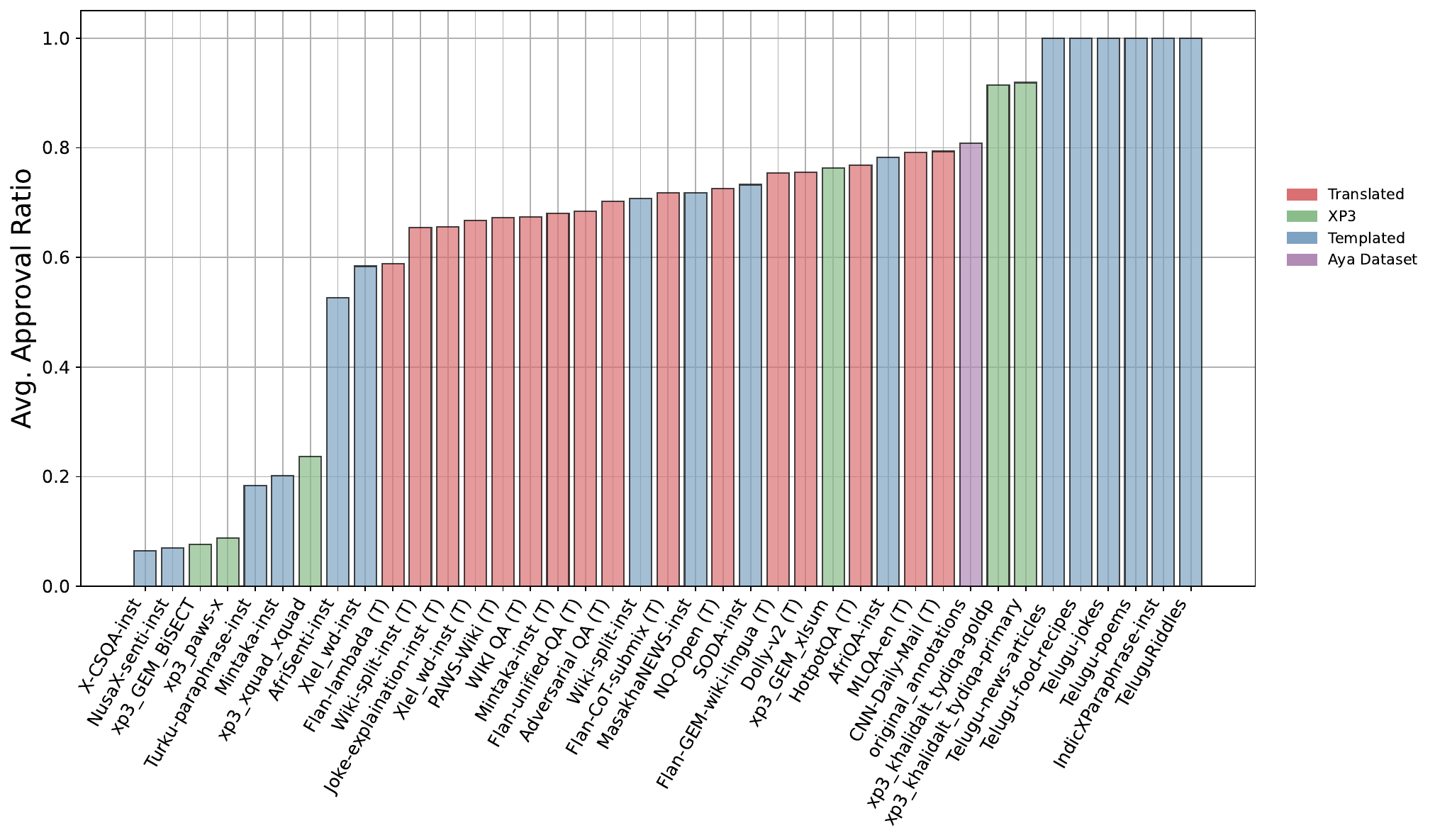}
  \caption{Average Approval Ratio per dataset,
constrained to datasets receiving at least 20 votes.}
  \label{avg-approval-ratio-per-dataset}
\end{figure}

\begin{figure}[ht!]
    \centering
    \begin{subfigure}[b]{.45\textwidth}
    \centering
    {\includegraphics[width=\linewidth]{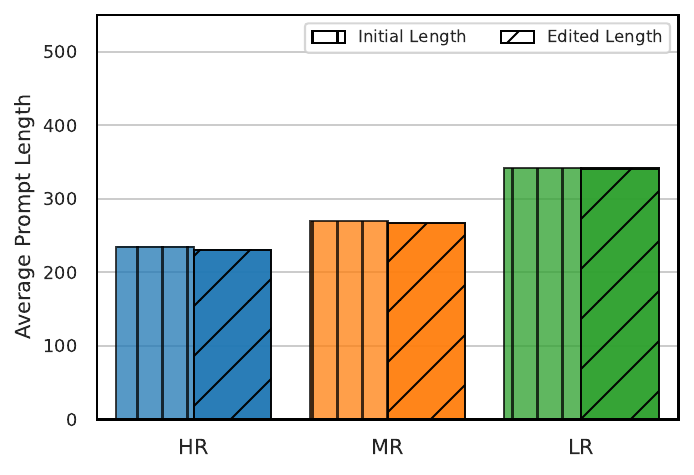}}
    \caption{Average prompt length}
    \end{subfigure}
    \begin{subfigure}[b]{.45\textwidth}
    \centering
    {\includegraphics[width=\linewidth]{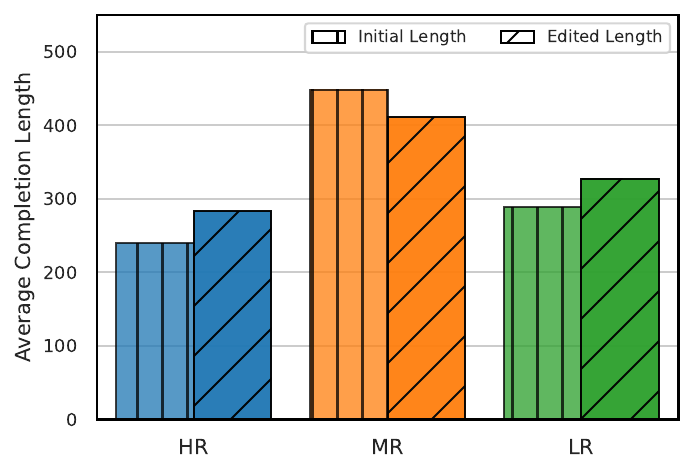}}
    \caption{Average completion length}
    \end{subfigure}
    \caption{Average prompt and completion length of instances in the \aya Dataset before and after re-annotation across different language categories.
    }
    \label{fig:Avg_prompt_completion_length_task_1_lang_categories}
\end{figure} 

\begin{figure}
    \centering
    \begin{subfigure}[b]{\textwidth}
          \centering
          \includegraphics[width=\textwidth]{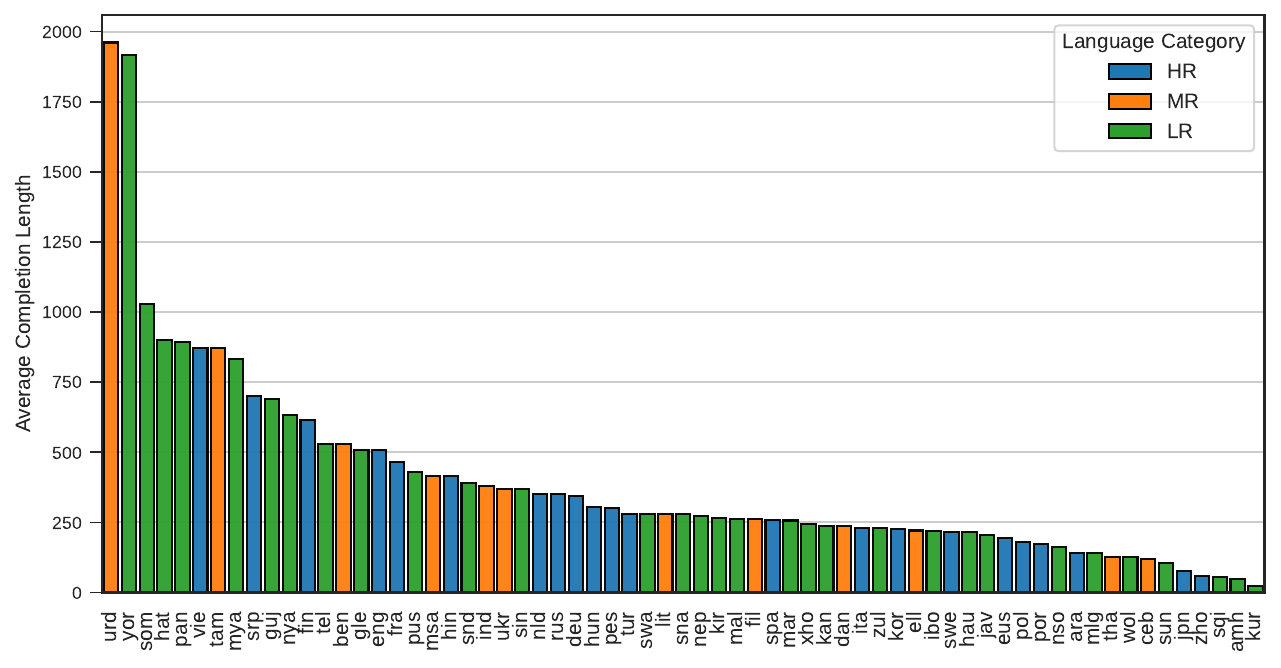}
         \caption{Average completion length for every language in the \aya Dataset.}
    \end{subfigure}
    \begin{subfigure}[b]{\textwidth}
          \centering
          \includegraphics[width=\textwidth]{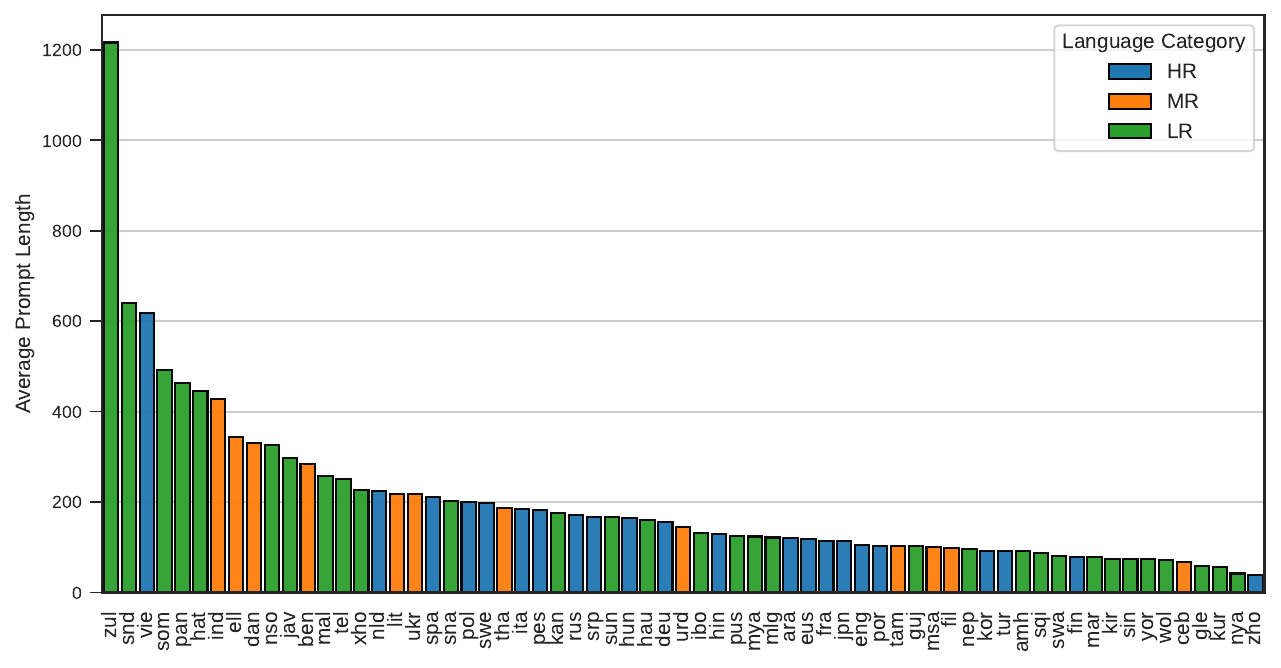}
         \caption{Average prompt length for every language in the \aya Dataset}
    \end{subfigure}
    \caption{Average prompt and completion length for every language in the \aya Dataset}
    \label{fig:per_language_prompt_comp_length}
\end{figure}

\begin{figure}[ht!]
  \centering
  \includegraphics[width=\textwidth]{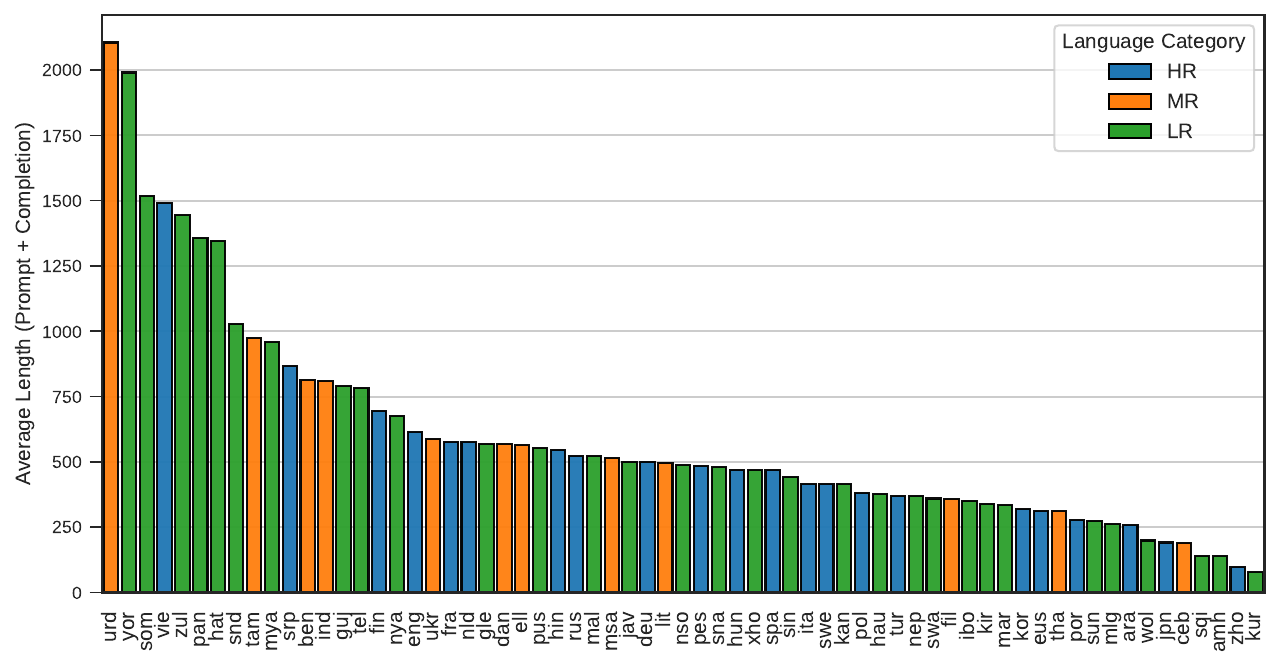}
  \caption{Average prompt and completion length across different languages in \textbf{\aya Dataset}.}
  \label{fig:Avg_prompt_comletion_length-Aya_dataset}
\end{figure}

\newpage
\section{Additional Tables}
\begin{center}
\scriptsize
\newcolumntype{L}[1]{>{\raggedright\arraybackslash}m{#1}} 
\begin{longtable}{L{3cm}L{1cm}L{2cm}L{2cm}L{1cm}L{1cm}L{1.5cm}L{2cm}}
\toprule
Dataset & \#Langs & Template lang & Dataset lang & $\bar{L}_{prompt}$ & $\bar{L}_{compl.}$ & License & Task \\
\midrule
\endfirsthead
\rowcolor{Gray}
AfriQA-inst \citep{ogundepo-etal-2023-cross} & 12 & & \multicolumn{1}{m{2cm}}{bem, fon, hau, ibo, kin, swh, twi, wol, yor, zul, eng, fra} & 46 & 15 & CC BY 4.0 & Question Answering \\
AfriSenti-inst \citep{muhammad-etal-2023-afrisenti} & 9 & & \multicolumn{1}{m{2cm}}{amh, arq, hau, ibo, kin, ary, por, swh, twi} & 168 & 44 & CC BY 4.0 & Sentiment Analysis \\
\rowcolor{Gray}
Amharic QA \citep{abedissa2023amqa} & 1 & amh & amh & 1114 & 33 & MIT license & Question Answering \\
News-summary-instruct \citep{TahmidHAnnotatedNewsSummary} & 1 & ben & ben & 174 & 67 & CC0 1.0 & Summarization\\
\rowcolor{Gray}
Arpa-instruct \citep{syntaxshillArpaAya} & 1 & hye & hye & 165 & 118 & Artistic-2.0 & Paraphrasing\\
Telugu-food-recipes \citep{SuryaKrishna02AyaTeluguFoodRecipes} & 1 & tel & tel & 70 & 870 & Apache 2.0 & Generation \\
\rowcolor{Gray}
Telugu-jokes \citep{SuryaKrishna02AyaTeluguJokes} & 1 & tel & tel & 80 & 276 & Apache 2.0 & Generation \\
Telugu-news-articles  \citep{SuryaKrishna02AyaTeluguNewsArticle} & 1 & tel & tel & 448 & 426 & Apache 2.0 & Generation \\
\rowcolor{Gray}
Telugu-poems \citep{SuryaKrishna02AyaTeluguPoems} & 1 & tel & tel & 357 & 198 & Apache 2.0 & Generation \\
FarsTail-Instruct \citep{amirkhani2023farstail,hghader1FarsTailInstructLLM} & 1 & pes & pes & 224 & 112 & Apache 2.0 & Natural Language Inference \\
\rowcolor{Gray}
Hindi-article-summarization \citep{hindi-article-summarization} & 1 & hin & hin & 3813 & 175 & CC BY-SA 4.0 & Summarization\\
Hindi-article-generation \citep{hindi-headline-article-generation} & 1 & hin & hin & 102 & 3683 & CC BY-SA 4.0 & Generation\\
\rowcolor{Gray}
IMDB-Dutch-instruct \citep{maas-EtAl:2011:ACL-HLT2011,imdb-dutch-instruct} & 1 & nld & nld & 1470 & 31 & Apache 2.0 & Sentiment Analysis\\
IndicSentiment-inst \citep{doddapaneni-etal-2023-towards,aya-indicsentiment} & 11 & \multicolumn{1}{m{2cm}}{eng} & \multicolumn{1}{m{2cm}}{ben, guj, hni, kan, mal, mar, pan, tam, tel, urd, eng} & 174 & 141 & MIT & Translation\\
\rowcolor{Gray}
IndicXParaphrase-inst \citep{doddapaneni-etal-2023-towards,aya-parapharse,SuryaKrishna02AyaTeluguParaphrase} & 7 & \multicolumn{1}{m{2cm}}{ben, guj, hin, mar, pan, mal, tel} & \multicolumn{1}{m{2cm}}{ben, guj, hin, mar, pan, mal, tel} & 132 & 93 & MIT & Paraphrase Identification\\
Indo-stories-instruct \citep{indonesian_instruct_stories,javanese_instruct_stories,sundanese_instruct_stories} & 3 & ind, sun, jav & ind, sun, jav & 345 & 322 & CC BY 4.0 & Translation\\
\rowcolor{Gray}
Joke-explaination-inst \citep{theblackcat102JokeExplaination} & 1 & & eng & 118 & 548 & MIT & Generation \\
Lijnews-instruct \citep{lijnews-instruct-ita-lij,lijnews-instruct-lij-ita} & 2 & ita, lij & it, lij & 893 & 898 & CC BY 4.0 & Translation\\
\rowcolor{Gray}
LLM-Japanese-Vanilla-inst \citep{Suzuki2023-llmvanilla,llm-japanese-dataset-vanilla-aya-format} & 1 & jpn & jpn & 60 & 97 & CC BY-SA 4.0 & Question Answering \\
MasakhaNEWS-inst \citep{adelani2023masakhanews} & 16 & & \multicolumn{1}{m{2cm}}{amh, eng, fra, hau, ibo, lin, cgg, orm, pcm, run, sna, som, swh, tir, xho, yor} & 1483 & 1459 & AFL-3.0 & Text Classification\\
\rowcolor{Gray}
Mintaka-inst \citep{sen-etal-2022-mintaka} & 9 & eng & \multicolumn{1}{m{2cm}}{arb, deu, spa, fra, jpn, por, hin, ita, eng} & 102 & 49 & CC BY 4.0 & Question Answering\\
NTX-LLM-inst \citep{chen2023dataset,ntx_llm_inst}, & 13 & \multicolumn{1}{m{2cm}}{arb, zho, nld, eng, fra, deu, hin, ita, jpn, kor, por, spa, tur} & \multicolumn{1}{m{2cm}}{arb, zho, nld, eng, fra, deu, hin, ita, jpn, kor, por, spa, tur} & 917 & 493 & CC BY-SA 4.0 & Information Extraction\\
\rowcolor{Gray}
NusaX-senti-inst \citep{winata-etal-2023-nusax} & 12 & & \multicolumn{1}{m{2cm}}{ace, ban, bjn, bug, eng, ind, jav, mad, min, nij, sun, bbc} & 219 & 22 & Apache 2.0 & Sentiment Analysis\\
Persian-instruct-pn \citep{pnSummary,ShafaghAyaPersianInstructionPnSummary, ShafaghAyaPersianInstructionPnSummaryTitle} & 1 & pes & pes & 1713 & 128 & MIT & Summarization\\
\rowcolor{Gray}
SCB-MT-2020-prompt \citep{lowphansirikul2020scb,pythainlpScbMt2020en2thprompt,pythainlpScbMt2020th2enprompt} & 2 & tha, eng & tha, eng & 181 & 127 & CC BY-SA 4.0 & Translation\\
Scirepeval-biomimicry-inst \citep{Singh2022SciRepEvalAM} & 1 & & eng & 996 & 523 & ODC-BY  & Scientific Document Representation  \\
\rowcolor{Gray}
Seed-instruct-lij \citep{seed-23,seed-instruct-eng-lij,seed-instruct-lij-eng} & 2 & lij, eng & lij, eng & 184 & 186 & CC BY-SA 4.0 & Translation\\
SODA-inst \citep{kim2022soda} & 1 & & eng & 412 & 328 & CC BY 4.0 & Dialogue\\
\rowcolor{Gray}
TamilStories \citep{aitamilnaduTamilStories} & 1 & tam & tam & 2266 & 2172 & Apache 2.0 & Generation \\
TeluguRiddles \citep{desik98TeluguRiddles} & 1 & tel & tel & 74 & 44 & Apache 2.0 & Question Answering\\
\rowcolor{Gray}
Thai-USEmbassy-prompt \citep{pythainlpThaiUsembassyEn2thPrompt,pythainlpThaiUsembassyth2enPrompt} & 2 & tha, eng & tha, eng & 2131 & 2077 & CC0 1.0 & Translation\\
Thai-POS-inst \citep{pythainlpThaiPosPrompt} & 1 & tha & tha & 72 & 36 & CC BY-SA 3.0 & Generation\\
\rowcolor{Gray}
Thai-Wiktionary-inst \citep{thai-wiktionary-prompt} & 1 & tha & tha & 35 & 147 & CC BY-SA 3.0 & Generation \\
Thirukkural-instruct \citep{aitamilnaduThirukkuralInstuct} & 1 & tam & tam & 133 & 542 & Apache 2.0 & Generation\\
\rowcolor{Gray}
Turku-paraphrase-inst \citep{kanerva-etal-2021-finnish,TurkuNLPturkuParaphraseCorpus} & 1 & fin & fin & 108 & 59 & CC BY-SA 4.0 & Paraphrase Identification\\
UA-Gec-inst \citep{syvokon-etal-2023-ua,ua_gec_instruction_tuning} & 1 & ukr & ukr & 192 & 148 & CC BY 4.0 & Generation\\
\rowcolor{Gray}
UNER-LLM-inst \citep{mayhew2023universal,uner_llm_inst} & 11 & \multicolumn{1}{m{2cm}}{zho, hrv, dan, eng, deu, por, rus, srp, slk, swe, tgl} & \multicolumn{1}{m{2cm}}{zho, hrv, dan, eng, deu, por, rus, srp, slk, swe, tgl} & 768 & 109 & CC BY-SA 4.0 & Named Entity Recognition\\
Urdu-News-Gen-Article \citep{hussain-etal-2021-urdunews,Urdu-Instruct-News-Article-Generation} & 1 & urd & urd & 109 & 1313 & CC BY 4.0 & Generation\\
\rowcolor{Gray}
Urdu-News-Category-Class \citep{hussain-etal-2021-urdunews,Urdu-Instruct-News-Category-Classification} & 1 & urd  & urd & 1407 & 43 & CC BY 4.0 & Text Classification\\
Urdu-News-Gen-Headline \citep{hussain-etal-2021-urdunews,Urdu-Instruct-News-Headline-Generation} & 1 & urd  & urd & 1314 & 94 & CC BY 4.0 & Generation\\
\rowcolor{Gray}
Wiki-split-inst \citep{botha-etal-2018-learning} & 1 & & eng & 200 & 166 & CC BY 4.0 & Text Simplification\\
X-CSQA-inst \citep{lin-etal-2021-common} & 16 & & \multicolumn{1}{m{2cm}}{eng, zho, deu, spa, fra, ita, jpn, nld, pol, por, rus, arb, vie, hin, swa, urd} & 197 & 21 & MIT & Question Answering \\
\rowcolor{Gray}
Xlel_wd-inst \citep{pratapa-etal-2022-multilingual} & 44 & & & 379 & 190 & CC BY 4.0 & Event Linking\\
XWikis-inst \citep{clads-emnlp} & 4 & & \multicolumn{1}{m{2cm}}{ces, fra, eng, deu} & 5662 & 346 & MIT & Summarization \\
\bottomrule
\caption{List of datasets in \aya Collection (templated datasets).}\\
\label{tab:list_of_templated_datasets_aya_collection}\\
\end{longtable}
\end{center}

\begin{center}
\scriptsize
\newcolumntype{L}[1]{>{\raggedright\arraybackslash}m{#1}}
\begin{longtable}{L{6cm}L{1cm}L{1.5cm}L{1.5cm}L{2cm}L{2.2cm}}
\toprule
Dataset & \#Langs & $\bar{L}_{prompt}$ & $\bar{L}_{compl.}$ & License & Task \\
\midrule
\rowcolor{Gray}
Adversarial QA (T) \citep{bartolo2020beat} & 101 & 159 & 721 & CC BY-SA 3.0  & Question Answering \\
CNN-Daily-Mail (T) \citep{DBLP:journals/corr/SeeLM17} \citep{hermann2015teaching} & 101 & 1980 & 305 & Apache 2.0 & Summarization\\
\rowcolor{Gray}
Flan-Coqa (T) \citep{wei2022finetuned, reddy-etal-2019-coqa} & 101 & 2143 & 364 & Multiple* & Question Answering\\
Flan-CoT-submix (T) \citep{wei2022finetuned} & 101 & 239 & 160 & Unknown & Generation\\
\rowcolor{Gray}
Flan-GEM-wiki-lingua (T) \citep{wei2022finetuned,ladhak-etal-2020-wikilingua} & 101 & 1732 & 572 & CC BY-NC-SA 3.0 &  Summarization\\

Flan-lambada (T) \citep{wei2022finetuned,paperno2016lambada}& 101 & 232 & 7 & CC BY 4.0 & Generation\\
\rowcolor{Gray}
Flan-unified-QA (T) \citep{wei2022finetuned,khashabi2020unifiedqa} & 101 & 281 & 13 & Apache 2.0& Question Answering\\

HotpotQA (T) \citep{yang2018hotpotqa} & 101 & 129 & 15 & CC BY-SA 4.0 & Question Answering\\
\rowcolor{Gray}
Joke-explaination-inst (T) \citep{theblackcat102JokeExplaination}  & 101 & 111 & 545 & MIT & Generation \\

Mintaka-inst (T) \citep{sen-etal-2022-mintaka} & 101 & 54 & 67 & CC BY 4.0 & Question Answering\\
\rowcolor{Gray}
MLQA-en (T) \citep{lewis2020mlqa} & 101 & 819 & 20 & CC BY-SA 3.0 & Question Answering\\

NQ-Open (T) \citep{kwiatkowski-etal-2019-natural} & 101 & 68 & 14  & CC BY-SA 3.0 & Question Answering\\
\rowcolor{Gray}
PAWS-Wiki (T) \citep{zhang2019paws} & 101 & 308 & 6 & Custom license, attribution & Paraphrase Identification\\

PIQA (T) \citep{Bisk2020} & 101 & 304 & 100 & Unknown & Question Answering\\
\rowcolor{Gray}
SODA-inst (T) ~\citep{kim2022soda} & 101 & 86 & 208 & CC BY 4.0 & Dialogue\\
WIKI QA (T) \citep{yang-etal-2015-wikiqa} & 101 & 205 & 36 & MSR DLA* & Question Answering\\
\rowcolor{Gray}
Wiki-split-inst (T) \citep{botha-etal-2018-learning} & 101 & 126 & 220 & CC BY-SA 4.0 & Text Simplification \\
Xlel_wd-inst (T) \citep{pratapa-etal-2022-multilingual} & 101 & 300 & 274  & CC BY 4.0 & Event Linking \\
\rowcolor{Gray}
Dolly-v2 (T) ~\citep{DatabricksBlog2023DollyV2}  & 101 & 427 & 357 & CC BY-SA 3.0 & Generation \\
\bottomrule
\caption{List of datasets in \Aya Collection (translated datasets).}
\label{tab:aya_collection_translated}
\end{longtable}
\end{center}

\begin{center}
\scriptsize
\begin{longtable}{lll}
\toprule
Main Task Type    &   Fine-grained Task Type   & Dataset \\
\midrule
Question Answering  & & AfriQA-inst \citep{ogundepo-etal-2023-cross} \\
& & Amharic QA \citep{abedissa2023amqa} \\
& & LLM-Japanese-Vanilla-inst \citep{llm-japanese-dataset-vanilla-aya-format} \\
& & Mintaka-inst \citep{sen-etal-2022-mintaka} \\
& & X-CSQA-inst \citep{lin-etal-2021-common} \\
& & TeluguRiddles \citep{desik98TeluguRiddles} \\
\midrule
Natural Language    & Summarization           & News-summary-instruct \citep{TahmidHAnnotatedNewsSummary} \\
Generation          & & Persian-instruct-pn \citep{ShafaghAyaPersianInstructionPnSummary} \\
& & Hindi-article-summarization \citep{hindi-article-summarization} \\
& & XWikis-inst \citep{clads-emnlp} \\
\cmidrule{2-3}
& Translation             & IndicSentiment-inst \citep{aya-indicsentiment} \\
& & Indo-stories-instruct \citep{indonesian_instruct_stories,javanese_instruct_stories,sundanese_instruct_stories} \\
& & Lijnews-instruct \citep{lijnews-instruct-ita-lij,lijnews-instruct-lij-ita} \\
& & SCB-MT-2020-prompt \citep{pythainlpScbMt2020en2thprompt,pythainlpScbMt2020th2enprompt}    \\
& & Thai-USEmbassy-prompt \citep{pythainlpThaiUsembassyEn2thPrompt,pythainlpThaiUsembassyth2enPrompt} \\
& & SEED-instruct-lij \citep{seed-instruct-eng-lij,seed-instruct-lij-eng} \\
\cmidrule{2-3}
& Paraphrasing            & Arpa-instruct \citep{syntaxshillArpaAya} \\
& & IndicXParaphrase-inst \citep{aya-parapharse,SuryaKrishna02AyaTeluguParaphrase} \\
& & Turku-paraphrase-inst \citep{TurkuNLPturkuParaphraseCorpus} \\
\cmidrule{2-3}
& Text Simplification & Wiki-split-inst \citep{botha-etal-2018-learning} \\
\cmidrule{2-3}
& Dialogue & SODA-inst \citep{kim2022soda} \\
\cmidrule{2-3}
& NL Generation & Telugu-food-recipes \citep{SuryaKrishna02AyaTeluguFoodRecipes} \\ 
& & Telugu-jokes \citep{SuryaKrishna02AyaTeluguJokes} \\
& & Telugu-news-articles  \citep{SuryaKrishna02AyaTeluguNewsArticle} \\ 
& &Telugu-poems \citep{SuryaKrishna02AyaTeluguPoems} \\
& & TamilStories \citep{aitamilnaduTamilStories} \\
& & Joke-explaination-inst \citep{theblackcat102JokeExplaination} \\
& & Thirukkural-instruct \citep{aitamilnaduThirukkuralInstuct} \\
& & Hindi-article-generation \citep{hindi-headline-article-generation}  \\
& & Thai-Wiktionary-inst \citep{thai-wiktionary-prompt}\\
& & UA-Gec-inst \citep{ua_gec_instruction_tuning}                   \\
& & Urdu-News-Gen-Article \citep{Urdu-Instruct-News-Article-Generation}  \\
& & Urdu-News-Gen-Headline \citep{Urdu-Instruct-News-Headline-Generation}  \\
& & Thai-POS-inst \citep{pythainlpThaiPosPrompt} \\
\midrule
Text Classification  & Sentiment Analysis      & AfriSenti-inst \citep{muhammad-etal-2023-afrisenti}  \\
& & IMDB-Dutch-instruct \citep{imdb-dutch-instruct} \\
& & NusaX-senti-inst \citep{winata-etal-2023-nusax} \\
\cmidrule{2-3}
& Information Extraction  & NTX-LLM-inst \citep{ntx_llm_inst} \\
\cmidrule{2-3}
& Named Entity Recognition & UNER-LLM-inst \citep{uner_llm_inst} \\
\cmidrule{2-3}
&  Natural Language Inference & FarsTail-Instruct \citep{hghader1FarsTailInstructLLM} \\
\cmidrule{2-3}
&  Event Linking  & Xlel_wd-inst \citep{pratapa-etal-2022-multilingual} \\
\cmidrule{2-3}
&  Sci.\ Doc.\ Representation & Scirepeval-biomimicry-inst \citep{Singh2022SciRepEvalAM} \\
\cmidrule{2-3}
& Text Classification & Urdu-News-Category-Class \citep{Urdu-Instruct-News-Category-Classification} \\
& & MasakhaNEWS-inst \citep{adelani2023masakhanews} \\
\bottomrule
\caption{Task Taxonomy of Templated Datasets (\aya Collection). We classify the templated datasets with a standard task taxonomy of three main tasks: Question Answering, Natural Language Generation, and Text Classification (Table~\ref{tab:task_taxonomy}). We then have a fine-grained task taxonomy within each task, such as Summarization, Translation, Paraphrasing, Sentiment Analysis, Information Extraction, and Named Entity Recognition. If there is not a recognized fine-grained task taxonomy for a specific dataset, we put it in the main task type category.}
\label{tab:templated_taxonomy}
\end{longtable}
\end{center}

\begin{center}
\scriptsize
\newcolumntype{L}[1]{>{\raggedright\arraybackslash}m{#1}}
\begin{longtable}{L{3cm}L{4cm}L{7cm}}
\toprule
Main Task Type    &   Fine-grained Task Type   & Dataset \\
\midrule
Question Answering  & & 
Adversarial QA (T) \citep{bartolo2020beat} \\& & Flan-Coqa (T) \citep{wei2022finetuned, reddy-etal-2019-coqa} \\
& & Flan-unified-QA (T) \citep{wei2022finetuned,khashabi2020unifiedqa} \\
& & HotpotQA (T) \citep{yang2018hotpotqa} \\
& & Mintaka-inst (T) \citep{sen-etal-2022-mintaka} \\
& & MLQA-en (T) \citep{lewis2020mlqa} \\ 
& & NQ-Open (T) \citep{kwiatkowski-etal-2019-natural} \\ 
& & PIQA (T) \citep{Bisk2020} \\
& & WIKI QA (T) \citep{yang-etal-2015-wikiqa} \\
\midrule
Natural Language    & Summarization           & CNN-Daily-Mail (T) \citep{DBLP:journals/corr/SeeLM17} \citep{hermann2015teaching} \\
Generation          & & Flan-GEM-wiki-lingua (T)\citep{wei2022finetuned,ladhak-etal-2020-wikilingua} \\
\cmidrule{2-3}
& Text Simplification & Wiki-split-inst (T) \citep{botha-etal-2018-learning} \\
\cmidrule{2-3}
& Dialogue & SODA-inst (T) \citep{kim2022soda} \\
\cmidrule{2-3}
& NL Generation &  Joke-explaination-inst (T) \citep{theblackcat102JokeExplaination} \\
& & 
 Flan-CoT-submix (T)\citep{wei2022finetuned}  \\
& & Flan-lambada (T) \citep{wei2022finetuned,paperno2016lambada}  \\
& & Dolly-v2 (T)~\citep{DatabricksBlog2023DollyV2}
 \\
\midrule
Text Classification  
&  Event Linking  & Xlel_wd-inst (T) \citep{pratapa-etal-2022-multilingual} \\
\cmidrule{2-3}
&  Paraphrase
Identification & PAWS-Wiki (T) \citep{zhang2019paws} \\
\bottomrule
\caption{Task Taxonomy of Translated Datasets (Aya Collection). We classify the translated datasets similar to templated datasets (Table~\ref{tab:templated_taxonomy}). If there is not a recognized fine-grained task taxonomy for a specific dataset, we put it in the main task type category.}
\label{tab:translated_taxonomy}
\end{longtable}
\end{center}

\begin{center}
\scriptsize
\newcolumntype{L}[1]{>{\raggedright\arraybackslash}m{#1}}
\begin{longtable}{L{5cm}L{1cm}L{2cm}L{1cm}L{1cm}L{1.5cm}L{2cm}}
\toprule
Dataset & \#Langs & Dataset Language & $\bar{L}_{prompt}$ & $\bar{L}_{compl.}$ & License & Task \\
\midrule
\rowcolor{Gray}
adversarial_qa dbert \citep{bartolo2020beat, adversarial_qa_dbert}          & 1 & eng & 655 & 263 & CC BY-SA 3.0 & Question Answering \\
adversarial_qa dbidaf \citep{bartolo2020beat, adversarial_qa_dbidaf}        & 1 & eng & 669 & 256 & CC BY-SA 4.0 & Question Answering\\
\rowcolor{Gray}
adversarial_qa droberta \citep{bartolo2020beat, adversarial_qa_droberta}    & 1 & eng & 742 & 243 & CC BY-SA 4.0 & Question Answering\\
ag_news \citep{gulli2005ag, ag_news}                          & 1 & eng & 292 & 40 & BSD-3-Clause & Text Classification\\
\rowcolor{Gray}
ai2_arc ARC-Challenge \citep{allenai:arc}                                   & 1 & eng & 351 & 33 & GPL-3 & Question Answering\\
ai2_arc ARC-Easy \citep{allenai:arc}                                        & 1 & eng & 307 & 26 & GPL-3 & Question Answering\\
\rowcolor{Gray}
amazon_polarity \citep{NIPS2015_250cf8b5}                           & 1 & eng & 454 & 83 & BSD-3-Clause & Sentiment Analysis\\
app_reviews \citep{ZurichOpenRepositoryandArchive:dataset}                  & 1 & eng & 159 & 28 & Unknown & Sentiment Analysis\\
\rowcolor{Gray}
clue c3 \citep{xu-etal-2020-clue}                                           & 1 & zho & 338 & 7 & Apache 2.0 & Question Answering\\
clue cmrc2018 \citep{cui-etal-2019-span}                                    & 1 & zho & 426 & 178 & CC BY-SA 4.0 & Question Answering\\
\rowcolor{Gray}
clue csl \citep{li2022csl}                                                  & 1 & zho & 315 & 64 & Apache 2.0 & Question Answering\\
clue drcd \citep{shao2019drcd}                                              & 1 & zho & 436 & 128 & CC BY-SA 3.0 & Question Answering\\
\rowcolor{Gray}
clue tnews \citep{xu-etal-2020-clue}                                               & 1 & zho & 235 & 7 & Apache 2.0 & Question Answering\\
cnn_dailymail\_3.0.0 \citep{nallapati2016abstractive}                       & 1 & eng & 1699 & 646 & Unknown & Summarization\\
\rowcolor{Gray}
common_gen \citep{lin-etal-2020-commongen}                                  & 1 & eng & 96 & 49 & MIT & Generation\\
cos_e\_v1.11 \citep{rajani2019explain}                                      & 1 & eng & 208 & 19 & BSD-3-Clause & Generation\\
\rowcolor{Gray}
cosmos_qa \citep{huang-etal-2019-cosmos}                                    & 1 & eng & 547 & 51 & Unknown & Question Answering\\
dbpedia\_14 \citep{dbpedia}                                                 & 1 & eng & 378 & 64 & Apache 2.0 & Topic Classification\\
\rowcolor{Gray}
dream \citep{gu-etal-2022-dream}                                            & 1 & eng & 511 & 152 & Apache 2.0 & Question Answering\\
duorc ParaphraseRC \citep{DuoRC}                                            & 1 & eng & 1438 & 663 & MIT & Question Answering\\
\rowcolor{Gray}
duorc SelfRC \citep{DuoRC}                                                  & 1 & eng & 1411 & 645 & MIT & Question Answering\\
GEM/BiSECT \citep{kim-etal-2021-bisect}                                     & 3 & eng, spa, fra & 346 & 251 & Unknown & Text Simplification\\
\rowcolor{Gray}
GEM/xlsum \citep{hasan-etal-2021-xl}                                        & 2 & eng, ben & 1156 & 636 & CC BY-NC-SA 4.0 & Summarization\\
gigaword \citep{Rush_2015, graff2003english}                                & 1 & eng & 181 & 80 & Unknown & Summarization\\
\rowcolor{Gray}
glue mrpc \citep{warstadt2018neural, wang2019glue, dolan2005automatically}  & 1 & eng & 270 & 38 & MIT & Text Classification\\
glue qqp \citep{warstadt2018neural, wang2019glue, qqp}                      & 1 & eng & 199 & 4 & Unknown & Text Classification\\
\rowcolor{Gray}
imdb \citep{maas-EtAl:2011:ACL-HLT2011}                                     & 1 & eng & 1089 & 106 & Unknown & Sentiment Analysis\\
tydiqa-goldp \citep{tydiqa}                                                 & 6 & \multicolumn{1}{m{2cm}}{arb, ben, eng, ind, swh, tel} & 526 & 115 & Apache 2.0 & Question Answering\\
\rowcolor{Gray}
tydiqa-primary \citep{tydiqa}                                               & 6 & \multicolumn{1}{m{2cm}}{arb, ben, eng, ind, swa, tel} & 1110 & 332 & Apache 2.0 & Question Answering\\
kilt_tasks hotpotqa \citep{kilt_tasks}                                      & 1 & eng & 137 & 15 & MIT & Question Answering\\
\rowcolor{Gray}
multi_news \citep{alex2019multinews}                                        & 1 & eng & 3466 & 1442 & Custom license & Summarization\\
openbookqa main \citep{OpenBookQA2018}                                      & 1 & eng & 163 & 16  & Apache 2.0 & Question Answering \\
\rowcolor{Gray}
xlwic xlwic \citep{raganato-etal-2020-xl}                                   & 1 & eng & 225 & 3 & CC BY-NC 4.0 & Text Classification\\
paws labeled_final \citep{zhang2019paws}                                    & 1 & eng & 285 & 12 & Custom license & Paraphrase Identification\\
\rowcolor{Gray}
paws-x \citep{pawsx2019emnlp}                                               & 4 & \multicolumn{1}{m{2cm}}{eng, spa, fra, zho} & 255 & 11 & Custom license & Paraphrase Identification\\
piqa \citep{Bisk2020}                                                       & 1 & eng & 256 & 72 & AFL 3.0 & Question Answering\\
\rowcolor{Gray}
qasc \citep{allenai:qasc}                                                   & 1 & eng & 314	& 38 & Apache 2.0 & Question Answering\\
quail \citep{DBLP:conf/aaai/RogersKDR20}                                    & 1 & eng & 1752 & 18 & CC BY-NC-SA 4.0 & Question Answering\\
\rowcolor{Gray}
quarel \citep{quarel_v1}                                                    & 1 & eng & 289 & 10 & CC BY 4.0 & Question Answering\\
quartz \citep{quartz}                                                       & 1 & eng & 307 & 9 & CC BY 4.0 & Question Answering\\
\rowcolor{Gray}
quoref \citep{allenai:quoref}                                               & 1 & eng & 1556 & 388 & CC BY 4.0 & Question Answering\\
race high \citep{lai-etal-2017-race}                                        & 1 & eng & 1723 & 229 & Custom license & Question Answering\\
\rowcolor{Gray}
race middle \citep{lai-etal-2017-race}                                      & 1 & eng & 1141 & 144 & Custom license & Question Answering\\
ropes \citep{Lin2019ReasoningOP}                                            & 1 & eng & 886 & 97 & CC BY 4.0 & Question Answering\\
\rowcolor{Gray}
rotten_tomatoes \citep{Pang+Lee:05a}                                        & 1 & eng & 152 & 18 & Unknown  & Sentiment Analysis\\
samsum \citep{gliwa-etal-2019-samsum}                                       & 1 & eng & 473 & 170 & CC BY-NC-ND 4.0 & Summarization\\
\rowcolor{Gray}
sciq \citep{SciQ}                                                           & 1 & eng & 346 & 139 & CC BY-NC 3.0 & Question Answering\\
social_i_qa \citep{sap2019socialiqa}                                        & 1 & eng & 182 & 15 & CC BY 4.0 & Question Answering\\
\rowcolor{Gray}
squad_v2 \citep{2016arXiv160605250R}                                        & 1 & eng & 689 & 82 & CC BY-SA 4.0 & Question Answering\\
super_glue boolq \citep{clark2019boolq, wang2019superglue}                  & 1 & eng & 653 & 76 & CC BY-SA 3.0 & Question Answering \\
\rowcolor{Gray}
super_glue multirc \citep{MultiRC2018}                                      & 1 & eng & 1509 & 120 & Custom license & Question Answering\\
super_glue record \citep{zhang2018record}                                   & 1 & eng & 1175 & 70 & Apache 2.0 & Question Answering\\
\rowcolor{Gray}
super_glue wic \citep{pilehvar2019wic}                                      & 1 & eng & 170 & 3 & CC BY-NC 4.0 & Text Classification\\
trec \citep{li-roth-2002-learning, hovy-etal-2001-toward}                   & 1 & eng & 144 & 9 & Unknown & Text Classification\\
\rowcolor{Gray}
trivia_qa unfiltered \citep{2017arXivtriviaqa}                              & 1 & eng & 148 & 92 & Unknown & Question Answering\\
web_questions \citep{berant-etal-2013-semantic}                             & 1 & eng & 70 & 17 & Unknown & Question Answering\\
\rowcolor{Gray}
wiki_bio \citep{DBLP:journals/corr/LebretGA16}                              & 1 & eng & 586 & 328 & CC BY-SA 3.0 & Generation\\
wiki_hop original \citep{tu2019multihop}                                    & 1 & eng & 6363 & 748 & CC BY-SA 3.0 & Question Answering\\
\rowcolor{Gray}
wiki_qa \citep{yang-etal-2015-wikiqa}                                       & 1 & eng & 224 & 26 & Custom license & Question Answering\\
wiqa \citep{wiqa}                                                           & 1 & eng & 408 & 44 & Apache-2.0 & Question Answering\\
\rowcolor{Gray}
xquad \citep{Artetxe:etal:2019}                                       & 2 & zho, vie & 652 & 173 & CC BY-SA 4.0 & Question Answering\\
xsum \citep{Narayan2018DontGM}                                              & 1 & eng & 1412 & 250 & MIT & Summarization\\
\rowcolor{Gray}
yelp_review_full \citep{NIPS2015_250cf8b5}                                  & 1 & eng & 620 & 91 & Custom license & Sentiment Analysis\\
\caption{List of  xP3 datasets \citep{muennighoff-etal-2023-crosslingual}.}
\label{tab:templated_xp3}
\end{longtable}
\end{center}

\vspace{-25pt}
\section{Data Cards}

Following \citet{Pushkarna2022} and the HuggingFace data card template\footnote{https://huggingface.co/docs/datasets/v2.15.0/en/dataset_card}, we present the data card for the \aya Dataset.

\setlength{\columnseprule}{0.7pt}
\setlength{\columnsep}{20pt}

\begin{card}{\color{white} \textsf{Data Card for the \aya Dataset}}{ayadsymbol}{bgblue}
\small 

\label{apx:aya_dataset_datacard}

    \begin{mybox}{}
    
    {The \aya Dataset is a multilingual instruction fine-tuning dataset curated by an open-science community. The dataset contains a total of 204,114 annotated prompt-completion pairs.
    }

    \begin{itemize}
            \setlength\itemsep{0em}
            \item Curated by: 2,007 contributors from 110 countries
            \item Language(s): 65 languages
            \item License: Apache 2.0
            \item Repository: \url{https://huggingface.co/datasets/CohereForAI/aya_dataset}
    \end{itemize}
    \end{mybox}

    \begin{mybox}{\textsf{Authorship}}
    \begin{multicols}{3}
        \textsf{\textbf{Publishing Organization:}}\\
        Cohere For AI

        \columnbreak 

        \textsf{\textbf{Industry Type:}}\\
        Not-for-profit - Tech

        \columnbreak 

        \textsf{\textbf{Contact Details:}}\\
        {\color{red}\url{https://aya.for.ai/}}
    \end{multicols}
    \end{mybox}

    \begin{mybox}{\textsf{Example of Data Points}}
    
    The dataset contains multilingual prompts and completions in the following format:
    {\texttt{\{prompt: "What day is followed by Saturday?", 
               completion : "Saturday is followed by Sunday.",
               language: "English"
             \}  
            }}
    \end{mybox}

    \begin{mybox}{\textsf{Motivations \& Intentions}}
        \textsf{\textbf{Curation Rationale}}: The curation effort employed an open-science approach to create a diverse instruction-style dataset through annotators across the globe that ensures comprehensive representation across all languages. The success of the curation effort, led by volunteers across diverse backgrounds, was significantly influenced by their hope to meaningfully bring NLP advancements to their languages.
    \end{mybox}

        \begin{mybox}{\textsf{Provenance}}
    {
    \begin{multicols}{2}
            \textsf{\textbf{Methods Used}}\\
            crowd-sourced through volunteer annotations, followed by a quality assessment phase in which samples from the dataset were checked.
            
            \columnbreak %

            \textsf{\textbf{Methodology Details}}\\
            \textsf{Source:} Original annotations and edits of open-source NLP datasets \\
            \textsf{Platform:} \aya Annotation Platform\\
            \textsf{Dates of Collection:} Jun 2023 - Dec 2023
            
            \columnbreak 
        
    \end{multicols}
    }
    \end{mybox}

       \begin{mybox}{\textsf{Dataset Version and Maintenance}}

    \begin{multicols}{3}
            \textsf{\textbf{Maintenance Status}}\\
            Actively Maintained
            
            \columnbreak 

            \textsf{\textbf{Version Details}}\\
            Current version: 1.0\\
            Last Update: 12/2023\\
            First Release: 02/2024

            \columnbreak 

            \textsf{\textbf{Maintenance Plan}}\\
            Updates will be periodically made available based on volunteer contributions
        
    \end{multicols}

    \end{mybox}
\end{card}

\newpage
\begin{card}{\color{black} \textsf{Data Card for the \aya Collection}}{ayac}{bgyellow}
\small

\label{apx:aya_collection_datacard}

    \begin{mybox2}{}
    The \aya Collection incorporates instruction-style templates from fluent speakers and applies them to a curated list of 44 datasets. It also includes translations of 19 instruction-style datasets into 101 languages. This collection provides 513,579,625 instances of prompts and completions covering a wide range of tasks.. 

    \begin{itemize}
            \setlength\itemsep{0em}
            \item Curated by: 2007 contributors from 110 countries
            \item Language(s): 114 languages
            \item License: Apache 2.0
            \item Repository: \url{https://huggingface.co/datasets/CohereForAI/aya_collection}
    \end{itemize}
    \end{mybox2}

    \begin{mybox2}{\textsf{Authorship}}
    \begin{multicols}{3}
        \textsf{Publishing Organization:}\\
        Cohere For AI

        \columnbreak 

        \textsf{Industry Type:}\\
        Not-for-profit - Tech

        \columnbreak 

        \textsf{Contact Details:}\\
        \url{https://aya.for.ai}
    \end{multicols}
    \end{mybox2}

    \begin{mybox2}{\textsf{Example of Data Points}}
    
    The dataset contains multilingual prompts and completions in the following format:
    {\texttt{\{`prompt': "Generate an article for the given headline: \{\{headline\}\}", 
               `completion': "\{\{news_article\}\}",
               `lang': "English"
             \}  
            }}
    \end{mybox2}

    \begin{mybox2}{\textsf{Motivations \& Intentions}}
        \textsf{\textbf{Curation Rationale}}: 
        {Automatic augmentation of existing datasets serves to enhance the available linguistic resources for multiple languages.
 List of languages were established from mT5 and aligned with annotators' language list and NLLB translation model. The datasets were translated directly from English for all languages.}
    \end{mybox2}

        \begin{mybox2}{\textsf{Provenance}}
    {
    \begin{multicols}{2}
            \textsf{\textbf{Methods Used}}\\
            combination of crowd-sourced templating and automatic translation.
            
            \columnbreak 

            \textsf{\textbf{Methodology Details}}\\
            \textsf{Source:} Existing NLP datasets \\
            \textsf{Platform:} \aya Annotation Platform\\
            \textsf{Dates of Collection:} Jun 2023 - Dec 2023

            \columnbreak %
        
    \end{multicols}
    }
    \end{mybox2}

        \begin{mybox2}{\textsf{Dataset Version and Maintenance}}

    \begin{multicols}{3}
            \textsf{\textbf{Maintenance Status}}\\
            Actively Maintained
            
            \columnbreak 

            \textsf{\textbf{Version Details}}\\
            Current version: 1.0\\
            Last updated: 12/2023\\
            Release date: 02/2024

            \columnbreak 

            \textsf{\textbf{Maintenance Plan}}\\
            {No updates planned.}
        
    \end{multicols}

    \end{mybox2}
\end{card}

\pagebreak
\begin{card}{\color{black} \textsf{Data Card for the \aya Evaluation Suite}}{ayae}{ayaebackground} 
\small
\label{apx:aya_evaluation_datacard}

    \begin{mybox3}{}
    The \aya Evaluation Suite contains a total of 25,750 open-ended conversation-style prompts covering 101 languages of three subsets: \\
    \textbf{\textsc{aya-human-annotated}}: 250 original human-written prompts in 7 languages each.\\
    \textbf{\textsc{dolly-machine-translated}}: 200 human-selected prompts from~\citet{DatabricksBlog2023DollyV2}, automatically translated with the NLLB model~\citep{nllbteam2022language} from English into 101 languages. \\
    \textbf{\textsc{dolly-human-edited}}: 200 dolly-machine-translated prompts post-edited by fluent speakers for 6 languages.

    \begin{itemize}
            \setlength\itemsep{0em}
            \item Curated by: contributors, professional annotators, and synthetic generation 
            \item Language(s): 101 languages
            \item License: Apache 2.0
            \item Repository: \url{https://huggingface.co/datasets/CohereForAI/aya_evaluation_suite}
    \end{itemize}
    \end{mybox3}

    \begin{mybox3}{\textsf{Authorship}}
    \begin{multicols}{3}
        \textsf{Publishing Organization:}\\
        Cohere For AI

        \columnbreak 

        \textsf{Industry Type:}\\
        Not-for-profit - Tech

        \columnbreak 

        \textsf{Contact Details:}\\
        \url{https://aya.for.ai}
    \end{multicols}
    \end{mybox3}

    \begin{mybox3}{\textsf{Example of Data Points}}
    
    The dataset contains multilingual prompts in the following format:
    {\texttt{\{`prompt': "Which is a species of fish? Bleak or Weary", 
               `lang': "English"
             \}  
            }}
    \end{mybox3}

    \begin{mybox3}{\textsf{Motivations \& Intentions}}
        \textsf{\textbf{Curation Rationale}}: 
        {This evaluation suite is tailored for testing the generation quality of multilingual models, with the aim to balance language coverage and human-sourced quality. It covers prompts originally written in each language, as well as English-centric translated and manually curated or edited prompts for a linguistically broad but rich testbed.
        The list of languages was established from mT5 and aligned with annotators' language list and the NLLB translation model. 
        }

    \end{mybox3}

        \begin{mybox3}{\textsf{Provenance}}
    {
    \begin{multicols}{2}
            \textsf{\textbf{Methods Used}}\\
            combination of original annotations by volunteers, automatic translation, and post-editing of translations by professional annotators.
            
            \columnbreak

            \textsf{\textbf{Methodology Details}}\\
            \textsf{Source:} Original annotations and translations and post-edits of Dolly \\
            \textsf{Platform:} \aya Annotation Platform\\
            \textsf{Dates of Collection:} Jun 2023 - Dec 2023

            \columnbreak
        
    \end{multicols}
    }
    \end{mybox3}

        \begin{mybox3}{\textsf{Dataset Version and Maintenance}}

    \begin{multicols}{3}
            \textsf{\textbf{Maintenance Status}}\\
            Actively Maintained
            
            \columnbreak

            \textsf{\textbf{Version Details}}\\
            Current version: 1.0\\
            Last updated: 02/2024\\
            Release date: 02/2024

            \columnbreak

            \textsf{\textbf{Maintenance Plan}}\\
            No updates planned.
        
    \end{multicols}

    \end{mybox3}
    
\end{card}

\invisiblesection{\aya Collection Templates}
\label{tab:aya_collection_templates}
\includepdf[pages=-]{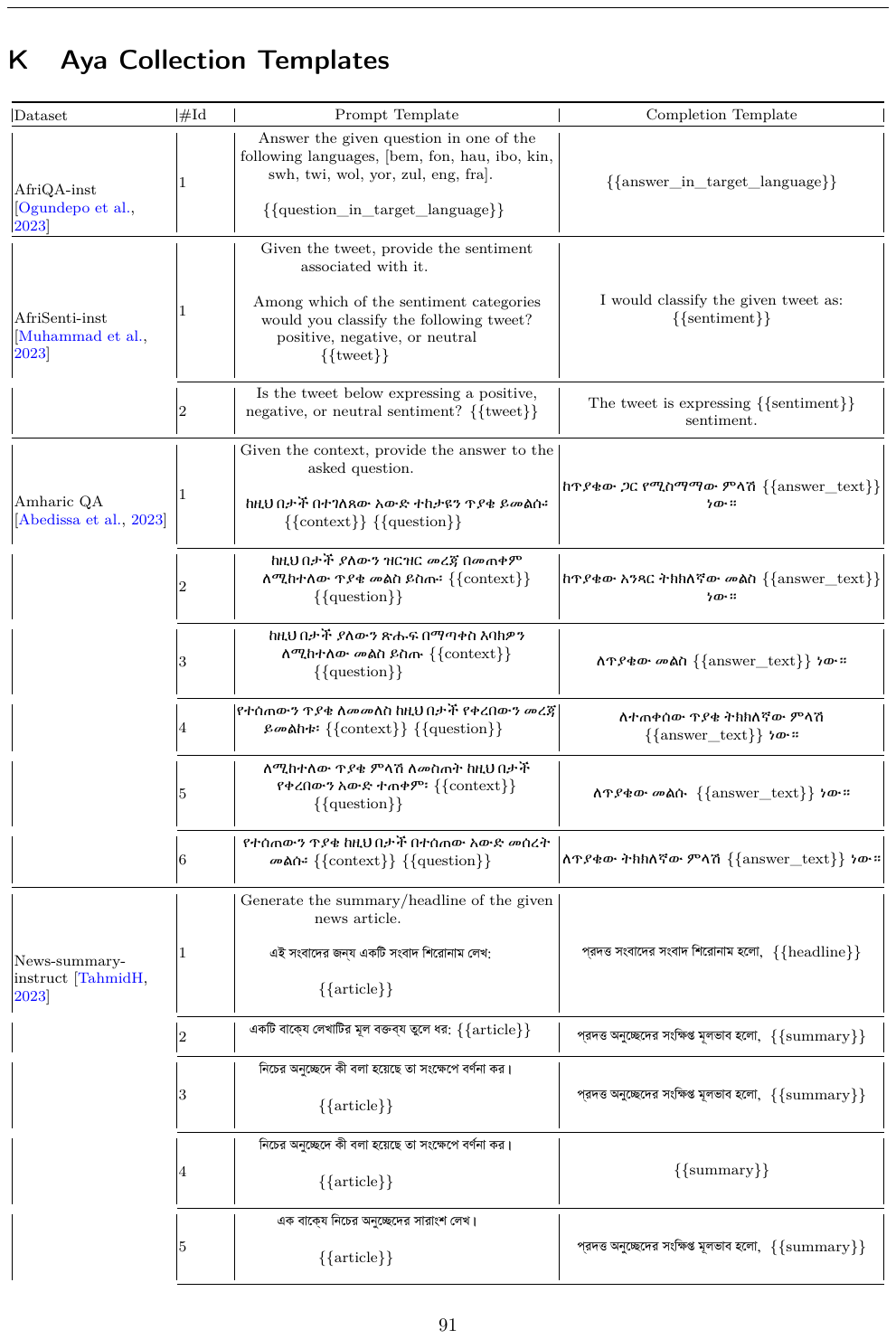}

\end{document}